\documentclass{article}

\PassOptionsToPackage{numbers, compress}{natbib}

\usepackage[final]{neurips_data_2022}

\usepackage[utf8]{inputenc} 
\usepackage[T1]{fontenc}    
\usepackage[breaklinks=true]{hyperref}       
\usepackage{url}            
\usepackage{booktabs}       
\usepackage{amsfonts}       
\usepackage{nicefrac}       
\usepackage{microtype}      
\usepackage{xcolor}         

\usepackage[pdftex]{graphicx}
\usepackage{subfig}
\usepackage{multirow}
\usepackage{todonotes}
\usepackage{siunitx}
\usepackage[export]{adjustbox}
\usepackage{duckuments} 

\usepackage{listings}
\usepackage{datasheet}
\usepackage{amssymb}
\usepackage{pifont}
\usepackage{array,multirow,graphicx}
\usepackage{stackengine}
\usepackage{rotating}

\newcommand{\cmark}{\ding{51}}%
\newcommand{\xmark}{\ding{55}}%

\newcommand*{\headformat}[1]{\textbf{#1}}
\newlength{\maxlen}
\newcommand*{\head}[1]{%
	\begin{sideways}
		\makebox[\maxlen][l]{\headformat{#1}}
\end{sideways}}

\title{CARLANE: A Lane Detection Benchmark for Unsupervised Domain Adaptation from Simulation to multiple Real-World Domains}

\author{%
	Julian Gebele\thanks{Equal contribution} \textsuperscript{ ,1}\, 
	Bonifaz Stuhr\footnotemark[1] \textsuperscript{ ,1,2}\, 
	Johann Haselberger\footnotemark[1] \textsuperscript{ ,1,3}\\
	\textsuperscript{1}University of Applied Science Kempten\\
	\textsuperscript{2}Autonomous University of Barcelona\\
	\textsuperscript{3}Technische Universität Berlin\\
	\texttt{carlane.benchmark@gmail.com}  
}

\begin{document}
	
	\maketitle
	
	\begin{abstract}
		Unsupervised Domain Adaptation demonstrates great potential to mitigate domain shifts by transferring models from labeled source domains to unlabeled target domains. While Unsupervised Domain Adaptation has been applied to a wide variety of complex vision tasks, only few works focus on lane detection for autonomous driving. This can be attributed to the lack of publicly available datasets. To facilitate research in these directions, we propose CARLANE, a 3-way sim-to-real domain adaptation benchmark for 2D lane detection. CARLANE encompasses the single-target datasets MoLane and TuLane and the multi-target dataset MuLane. These datasets are built from three different domains, which cover diverse scenes and contain a total of 163K unique images, 118K of which are annotated. In addition we evaluate and report systematic baselines, including our own method, which builds upon Prototypical Cross-domain Self-supervised Learning. We find that false positive and false negative rates of the evaluated domain adaptation methods are high compared to those of fully supervised baselines. This affirms the need for benchmarks such as CARLANE to further strengthen research in Unsupervised Domain Adaptation for lane detection. CARLANE, all evaluated models and the corresponding implementations are publicly available at \href{https://carlanebenchmark.github.io}{\textbf{https://carlanebenchmark.github.io}}.
	\end{abstract}
	
	\section{Introduction}
	Vision-based deep learning systems for autonomous driving have made significant progress in the past years \cite{Munir2021, qin2020ultra, pan2018SCNN, Garnett2019, SimuLanes2022}. Recent state-of-the-art methods achieve remarkable results on public, real-world benchmarks but require labeled, large-scale datasets. Annotations for these datasets are often hard to acquire, mainly due to the high expenses of labeling in terms of cost, time, and difficulty.
	Instead, simulation environments for autonomous driving, such as CARLA \cite{Carla2017}, can be utilized to generate abundant labeled images automatically. 
	However, models trained on data from simulation often experience a significant performance drop in a different domain, i.e., the real world, mainly due to the domain shift \cite{saenko2010adapting}. Unsupervised Domain Adaptation (UDA) methods \cite{ganin2015unsupervised, wilson2020survey, Long2015DAN, ZhuDSAN2020, Ganin2016, Tzeng2017ADDA, Xu2019, Sun2019SSL} try to mitigate the domain shift by transferring models from a fully-labeled source domain to an unlabeled target domain. 
	This eliminates the need for annotating images but assumes that the target domain is accessible at training time. 
	While UDA has been applied to complex tasks for autonomous driving such as object detection \cite{Munir2021, Xu2021} and semantic segmentation \cite{WU_2021_CVPR, Zhao2019SemSeg}, only few works focus on lane detection \cite{Garnett2020, SimuLanes2022}. This can be attributed to the lack of public UDA datasets for lane detection.
	
	\begin{figure}[t]
		\centering
		\small
		\begin{tabular}{rc@{}c@{\hskip 0.2cm}c@{}c}
			~ & \multicolumn{2}{c}{\textbf{MoLane}} & \multicolumn{2}{c}{\textbf{TuLane}} \\
			%
			\textbf{Source} & 
			\includegraphics[width=.2\linewidth,valign=m]{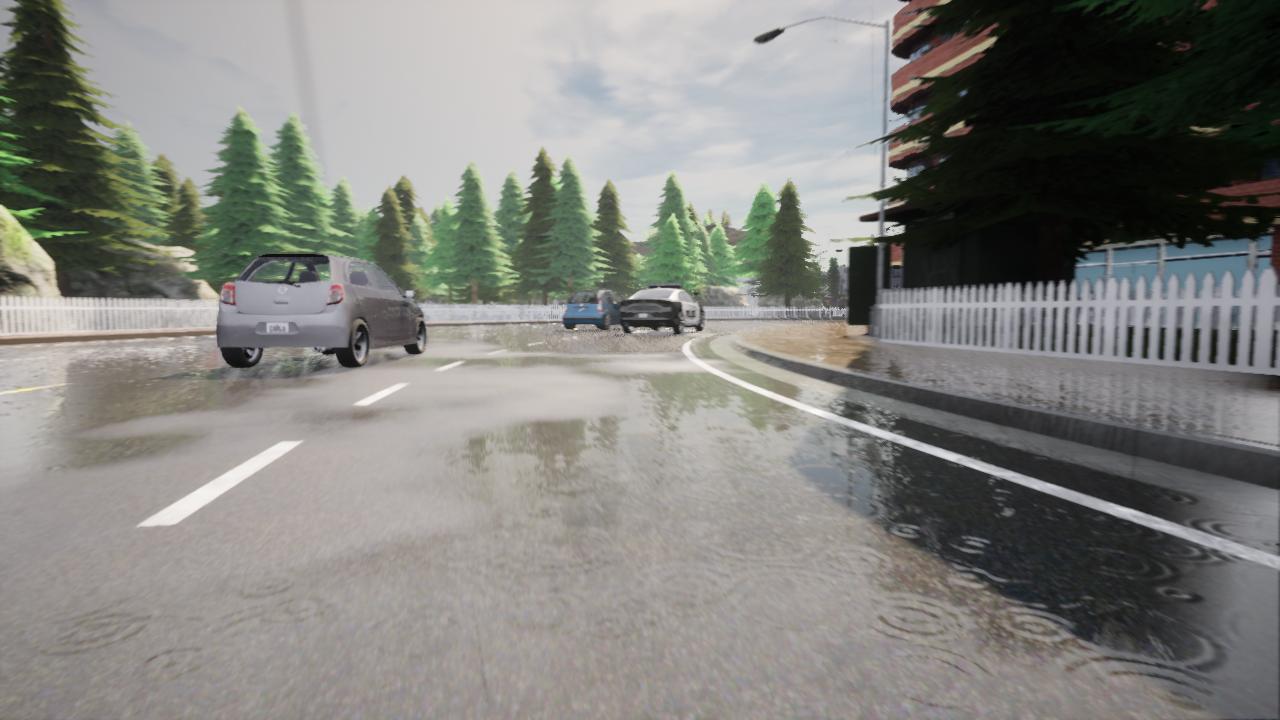} & 
			\includegraphics[width=.2\linewidth,valign=m]{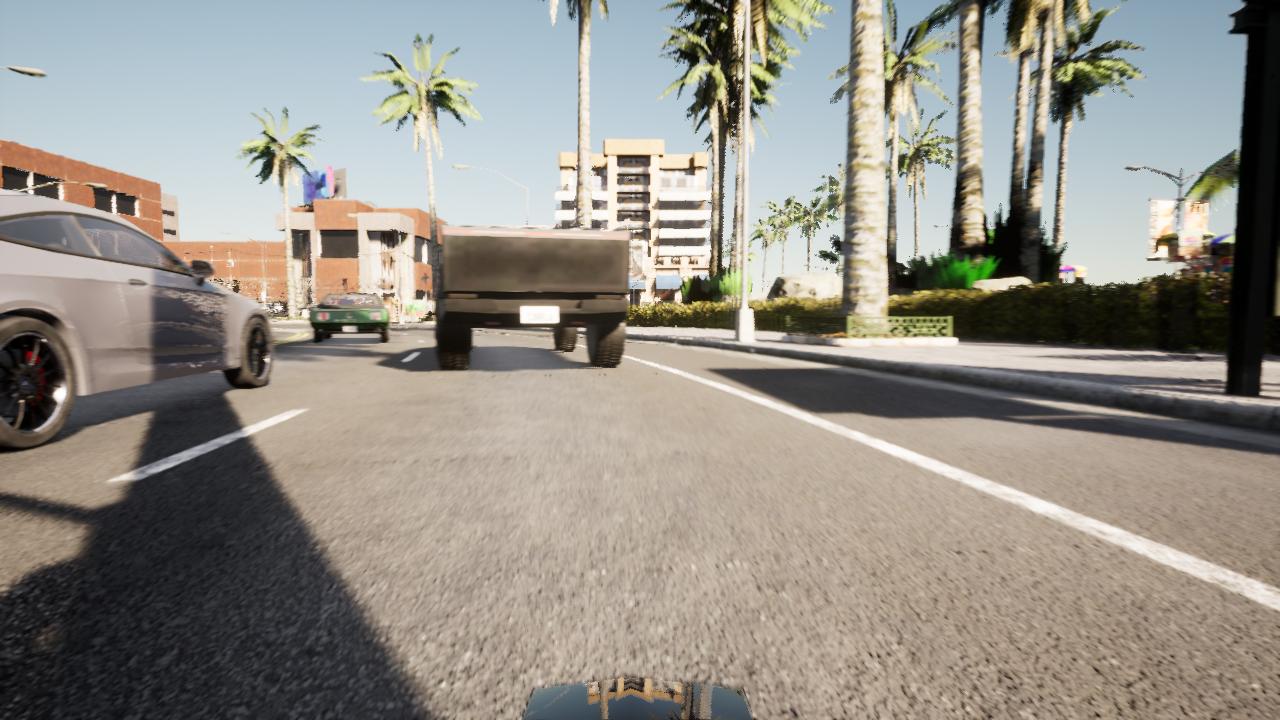} &
			\includegraphics[width=.2\linewidth,valign=m]{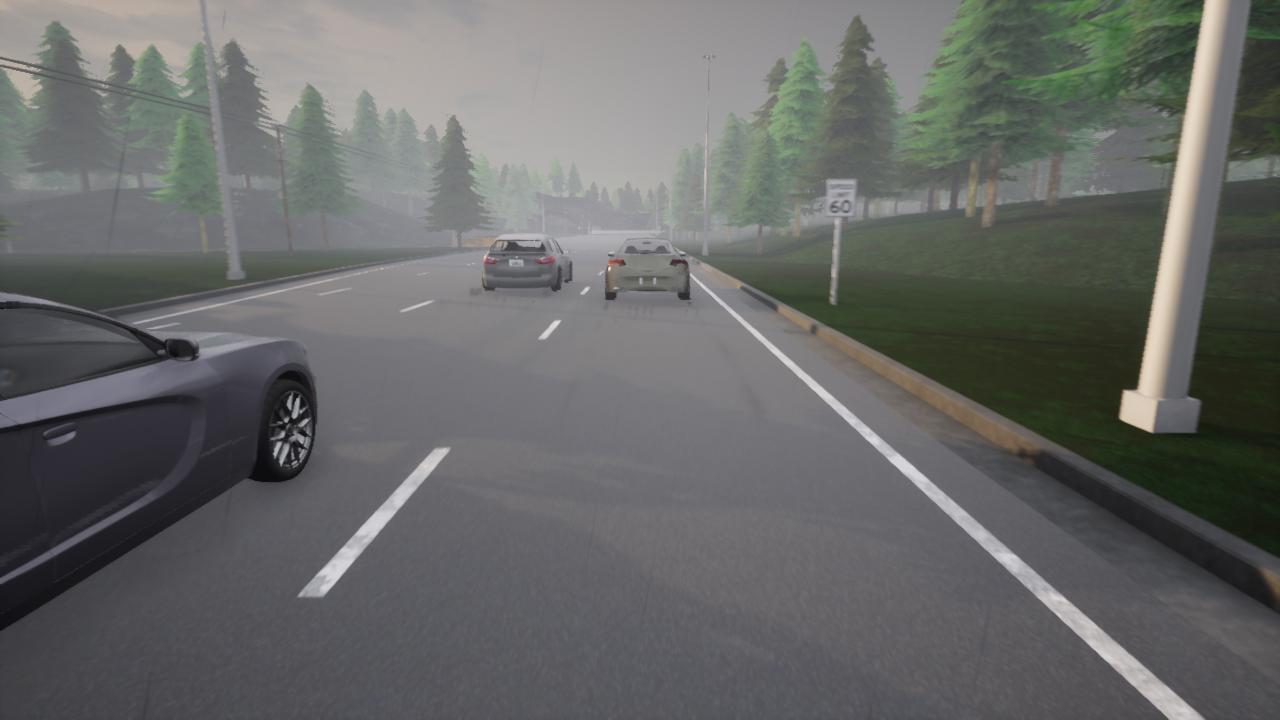} & 
			\includegraphics[width=.2\linewidth,valign=m]{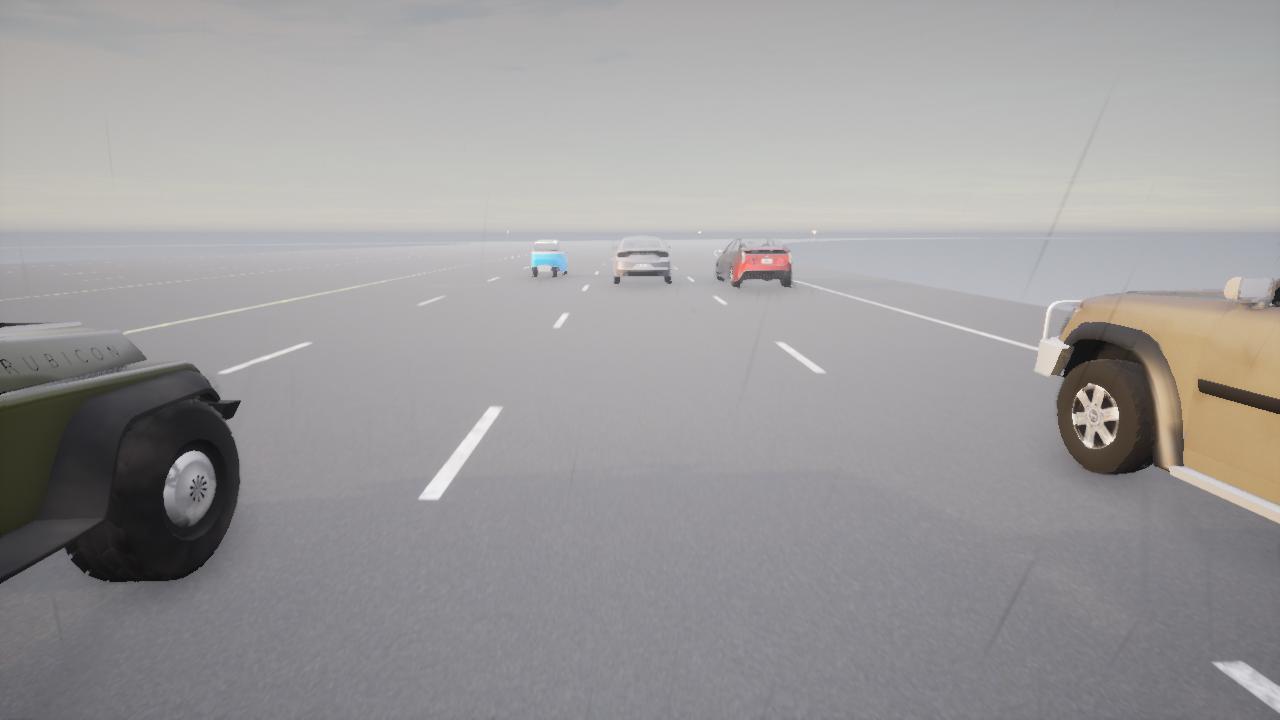}\\
			%
			\textbf{Target} & 
			\includegraphics[width=.2\linewidth,valign=m]{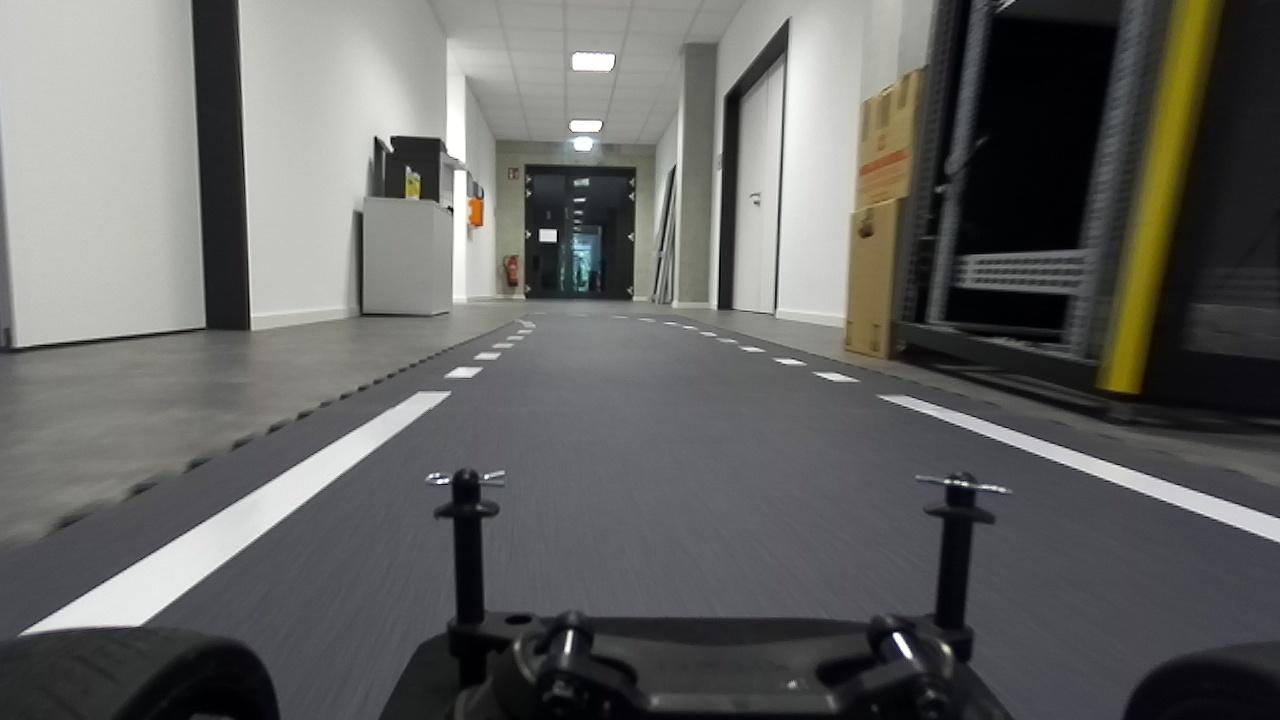} & 
			\includegraphics[width=.2\linewidth,valign=m]{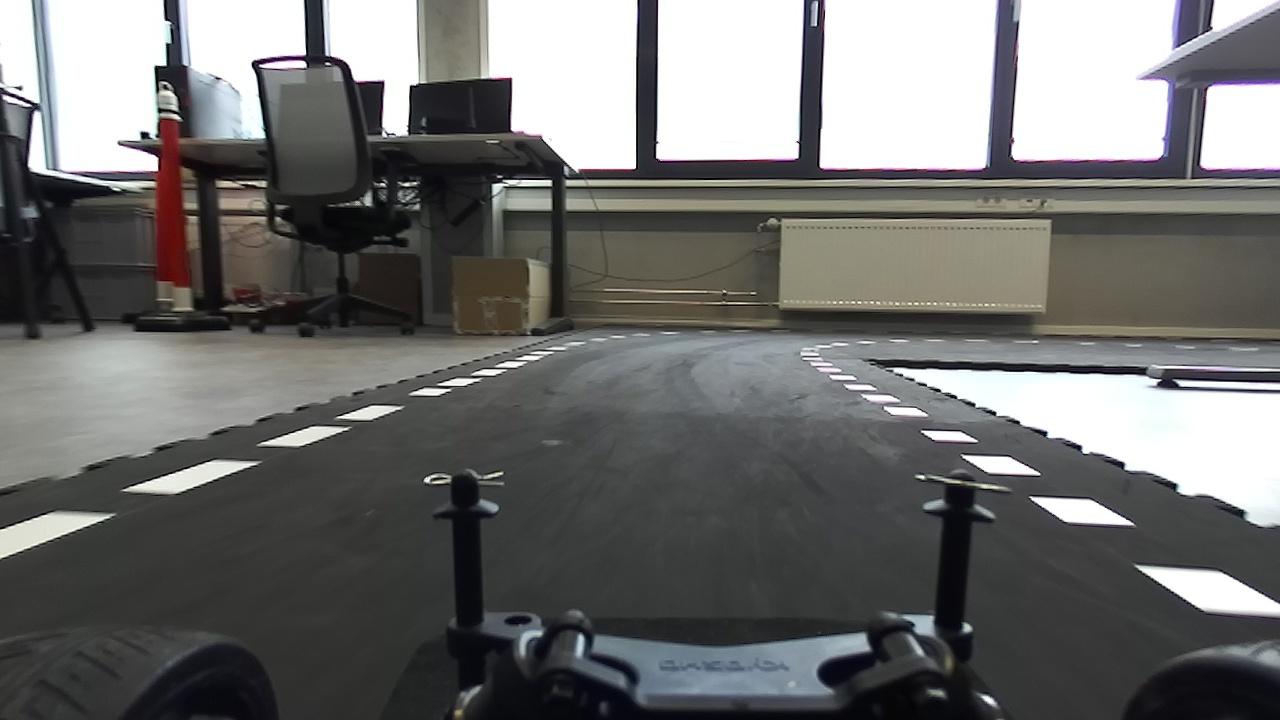} &
			\includegraphics[width=.2\linewidth,valign=m]{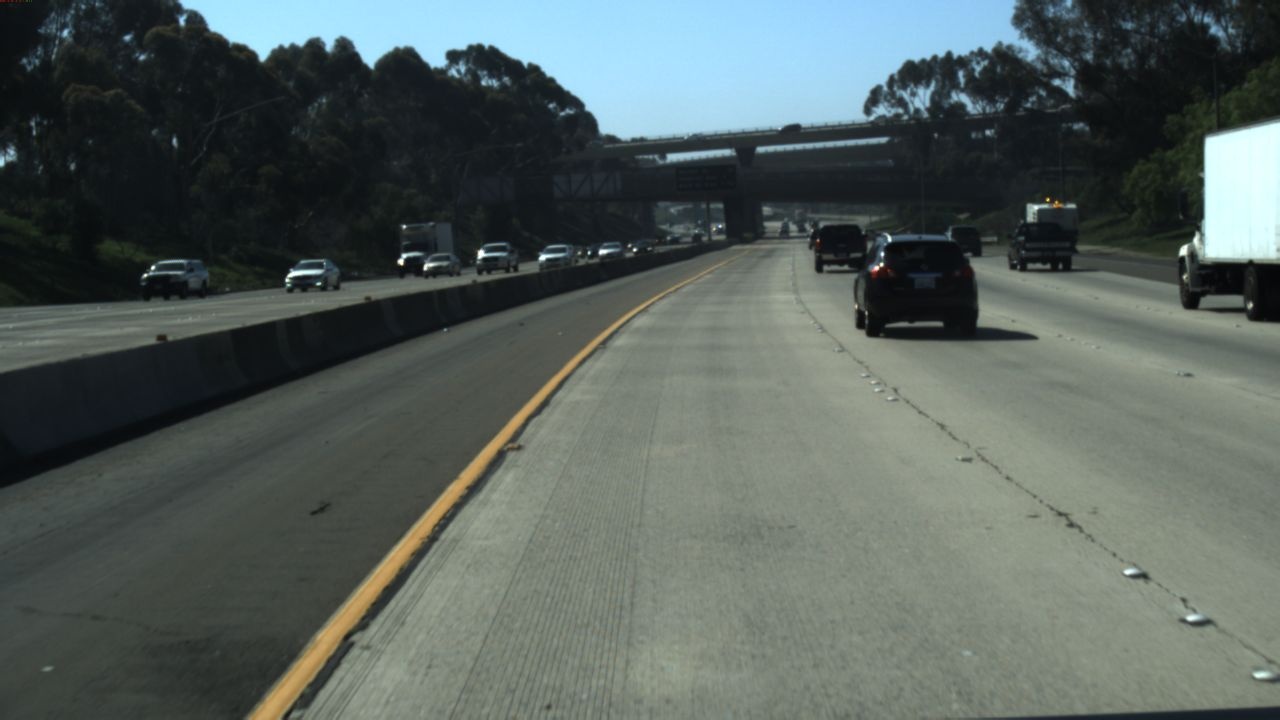} & 
			\includegraphics[width=.2\linewidth,valign=m]{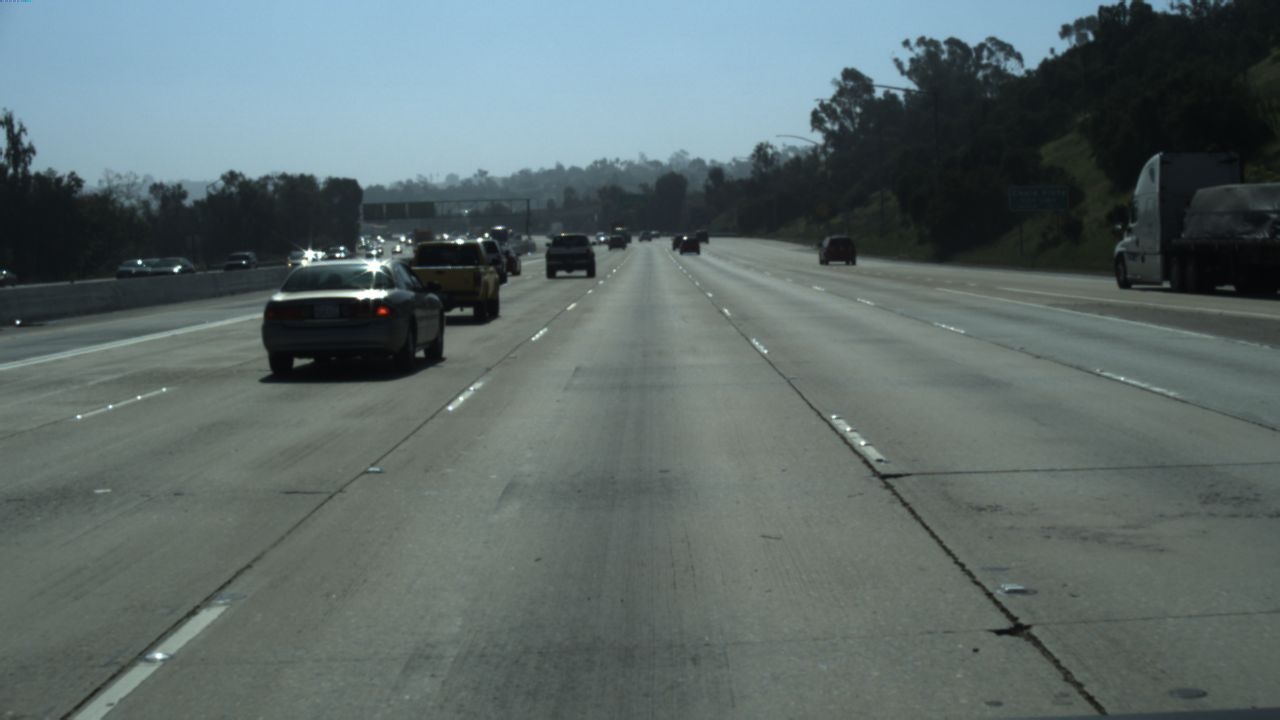}\\
		\end{tabular}
		\caption{Images sampled from our CARLANE Benchmark.}
		\label{fig:intro}
	\end{figure}
	
	To compensate for this data scarcity and encourage future research, we introduce CARLANE, a sim-to-real domain adaptation benchmark for lane detection. 
	We use the CARLA simulator for data collection in the source domain with a free-roaming waypoint-based agent and data from two distinct real-world domains as target domains. This enables us to construct a benchmark that consists of three datasets:
	\vspace{-2pt}
	
	\textit{(1) MoLane} focuses on abstract lane markings in the domain of a 1/8th \textit{Mo}del vehicle. We collect 80K labeled images from simulation as the source domain and 44K unlabeled real-world images from several tracks with two lane markings as the target domain. Further, we apply domain randomization as well as data balancing. For evaluation, we annotate 2,000 validation and 1,000 test images with our labeling tool.
	
	\vspace{-2pt}
	\textit{(2) TuLane} incorporates 24K balanced and domain-randomized images from simulation as the source domain and the well-known \textit{Tu}Simple \cite{TuSimple2017} dataset with 3,268 real-world images from U.S. highways with up to four labeled lanes as the target domain. The target domain of MoLane is a real-world abstraction from the target domain of TuLane, which may result in interesting insights about UDA.
	
	\vspace{-2pt}
	\textit{(3) MuLane} is a balanced combination of \textit{M}oLane and T\textit{u}Lane with two target domains. For the source domain, we randomly sample 24K images from MoLane and combine them with TuLane's synthetic images. For the target domains, we randomly sample 3,268 images from MoLane and combine them with TuSimple. This allows us to investigate multi-target UDA for lane detection.
	\vspace{-2pt}
	
	To establish baselines and investigate UDA on our benchmark, we evaluate several adversarial discriminative methods, such as DANN \cite{Ganin2016}, ADDA \cite{Tzeng2017ADDA} and SGADA \cite{sgada2021}. Additionally, we propose SGPCS, which builds upon PCS \cite{yue2021prototypical} with a pseudo labeling approach to achieve state-of-the-art performance. 
	
	Our contributions are three-fold: (1) We introduce CARLANE, a 3-way sim-to-real benchmark, allowing single- and multi-target UDA. (2) We provide several dataset tools, i.e., an agent to collect images with lane annotations in CARLA and a labeling tool to annotate the real-world images manually. (3) We evaluate several well-known UDA methods to establish baselines and discuss results on both single- and multi-target UDA. To the best of our knowledge, we are the first to adapt a lane detection model from simulation to multiple real-world domains.

	\section{Related Work}
	
	\subsection{Data Generation for Sim-to-Real Lane Detection}
	
	In recent years, much attention has been paid to lane detection benchmarks in the real world, such as CULane \cite{pan2018SCNN}, TuSimple \cite{TuSimple2017}, LLAMAS \cite{llamas2019}, and BDD100K \cite{BDD100k}. Despite the popularity of these benchmarks, there is few research that focuses on sim-to-real lane detection datasets. Garnett et al. \cite{Garnett2019} propose a method for generating synthetic images with 3D lane annotations in the open-source engine blender. Their \textit{synthetic-3D-lanes} dataset contains 300K train, 1,000 validation and 5,000 test images, while their real-world \textit{3D-lanes} dataset consists of 85K images, which are annotated in a semi-manual manner. Utilizing the data generation method from \cite{Garnett2019}, Garnett et al. \cite{Garnett2020} collect 50K labeled synthetic images to perform sim-to-real domain adaptation for 3D lane detection. At this point, the source domain of the dataset is not publicly available. 
	
	Recently, Hu et al. \cite{SimuLanes2022} investigated UDA techniques for 2D lane detection. Their proposed data generation method relies on CARLA's built-in agent to automatically collect 16K synthetic images. However, the dataset is not publicly available at this point. In comparison, our method leverages an efficient and configurable waypoint-based agent. Furthermore, in contrast to the aforementioned works, considering only single-source single-target UDA, we additionally focus on multi-target UDA.
	
	\subsection{Unsupervised Domain Adaptation}
	Unsupervised Domain Adaptation has been extensively studied in recent years \cite{wilson2020survey}. 
	In an early work, Ganin et al. \cite{ganin2015unsupervised} propose a gradient reversal layer between the features extractor and a domain classifier to learn similar feature distributions for distinct domains. 
	Early discrepancy-based methods employ a distance metric to measure the discrepancy of the source and target domain \cite{Long2015DAN, SunCORAL}. A prominent example is DAN \cite{Long2015DAN} which uses maximum mean discrepancies (MMD) \cite{Gretton2007, Gretton2012} to match embeddings of different domain distributions. Recently, DSAN \cite{ZhuDSAN2020} builds upon DAN with local MMD and exploits fine-grained features to align subdomains accurately. 
	
	Domain alignment can also be achieved through adversarial learning \cite{Goodfellow2014}. Adversarial discriminative methods such as DANN \cite{Ganin2016} or ADDA \cite{Tzeng2017ADDA} employ a domain classifier or discriminator, encouraging the feature extractor to produce domain-invariant representations. While these methods mainly rely on feature-level alignment, adversarial generative methods \cite{Hoffman_cycada2017, Bousmalis_pixelDA} operate on pixel-level. 
	
	In a recent trend, self-supervised learning methods are leveraged as auxiliary tasks to improve domain adaptation effectiveness and to capture in-domain semantic structures \cite{Xu2019, Sun2019SSL, Kim2020CDS, yue2021prototypical, xie2020self}. Furthermore, self-supervised learning is utilized for cross-domain alignment as well, by matching class-discriminative features \cite{Kim2020CDS, wang2020classes}, task-discriminative features \cite{wei2021toalign}, class prototypes \cite{tanwisuth2021prototype, yue2021prototypical} or equivalent samples in the domains \cite{zhao2021reducing}.
	
	Furthermore, other recent works mitigate optimization inconsistencies by minimizing the gradients discrepancy of the source samples and target samples \cite{du2021cross} or by applying a meta-learning scheme between the domain alignment and the targeted classification task \cite{wei2021metaalign}.
	
	\section{Data Generation}
	To construct our benchmark, we gather image data from a real 1/8th model vehicle, and the CARLA simulator \cite{Carla2017}. Ensuring the verification of results and transferability to real driving scenarios, we extend our benchmark with the TuSimple dataset \cite{TuSimple2017}. This enables gradual testing, starting from simulation, followed by model cars, and ending with full-scale real word experiments. Data variety is achieved through domain randomization in all domains. However, naively performing domain randomization might lead to an imbalanced dataset. Therefore, similar driving scenarios are sampled across all domains, and a bagging approach is utilized to uniformly collect lanes by their curvature with respect to the camera position. We strictly follow TuSimple's data format \cite{TuSimple2017} to maintain consistency across all our datasets. 
	
	\subsection{Real-World Environment}
\begin{figure}
		\centering
		\subfloat[]{\includegraphics[trim={-1.5cm 0 -1.5cm 0}, clip, scale=0.08]{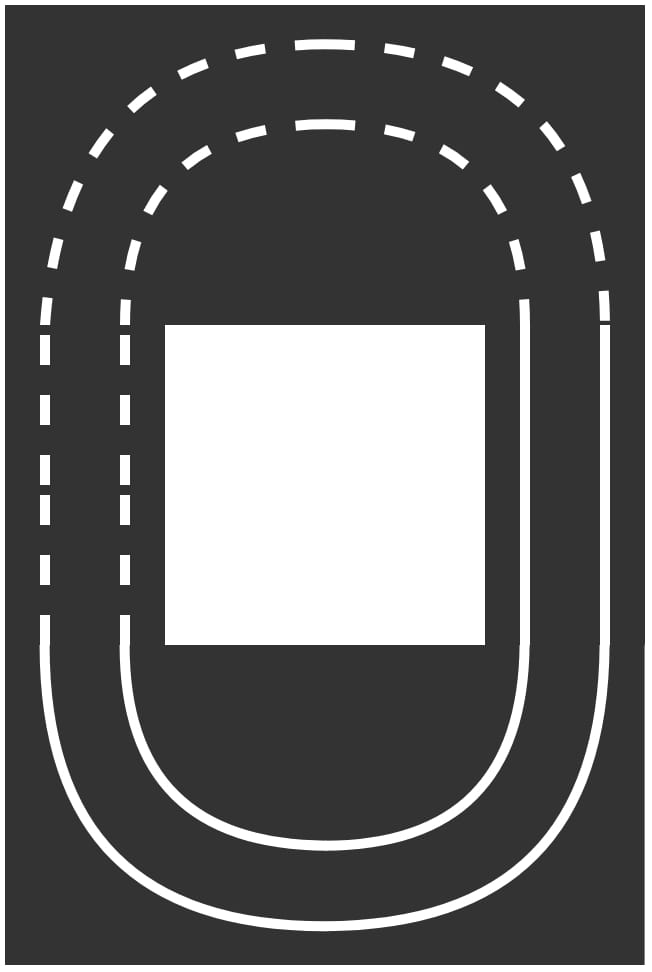}}
		\subfloat[]{\includegraphics[trim={0 0 0 0}, clip, scale=0.08]{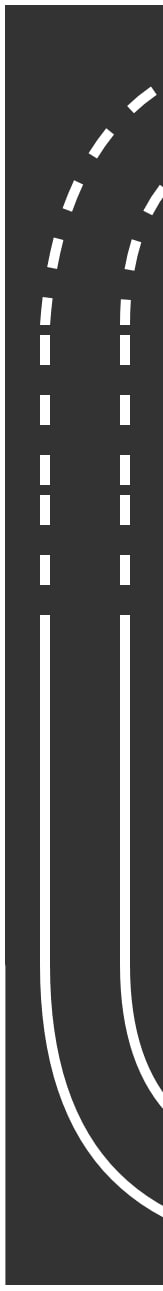}}
		\hspace*{0.1cm}
		\subfloat[]{\includegraphics[trim={0 0 0 0}, clip, scale=0.08]{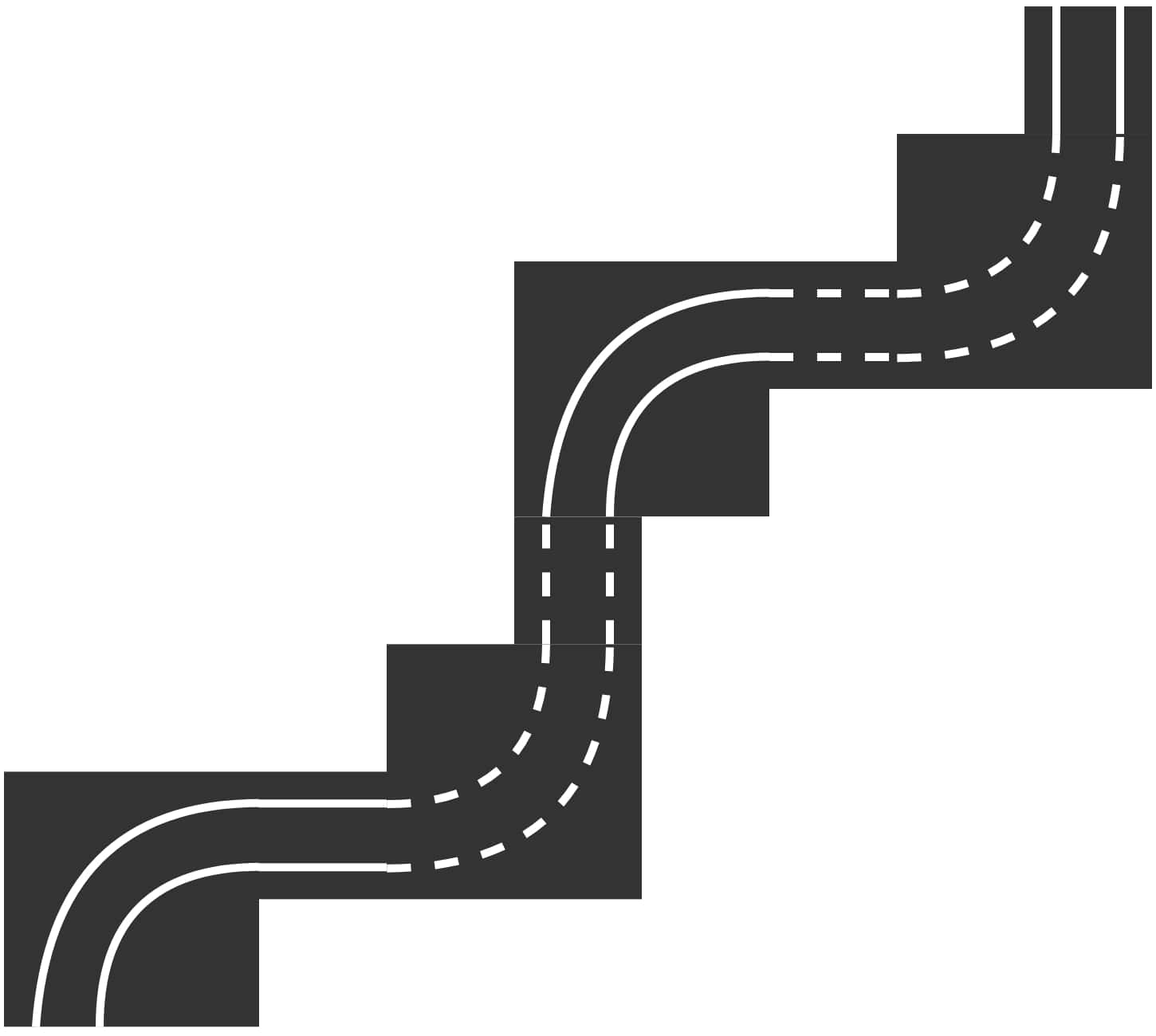}}
		\hspace*{-1.7cm}
		\subfloat[]{\includegraphics[scale=0.08]{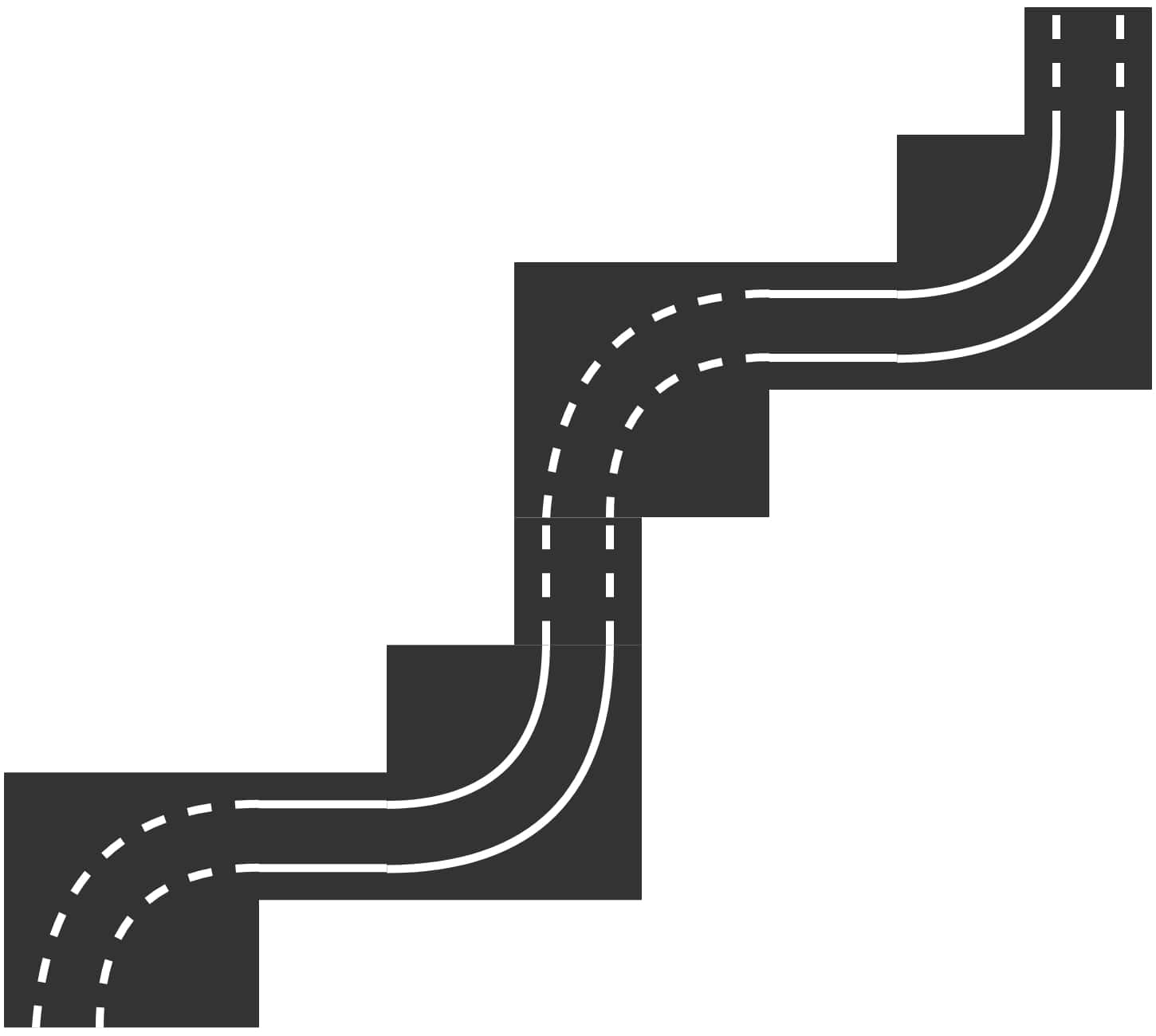}}
		\hspace*{0.1cm}
		\subfloat[]{\includegraphics[scale=0.08]{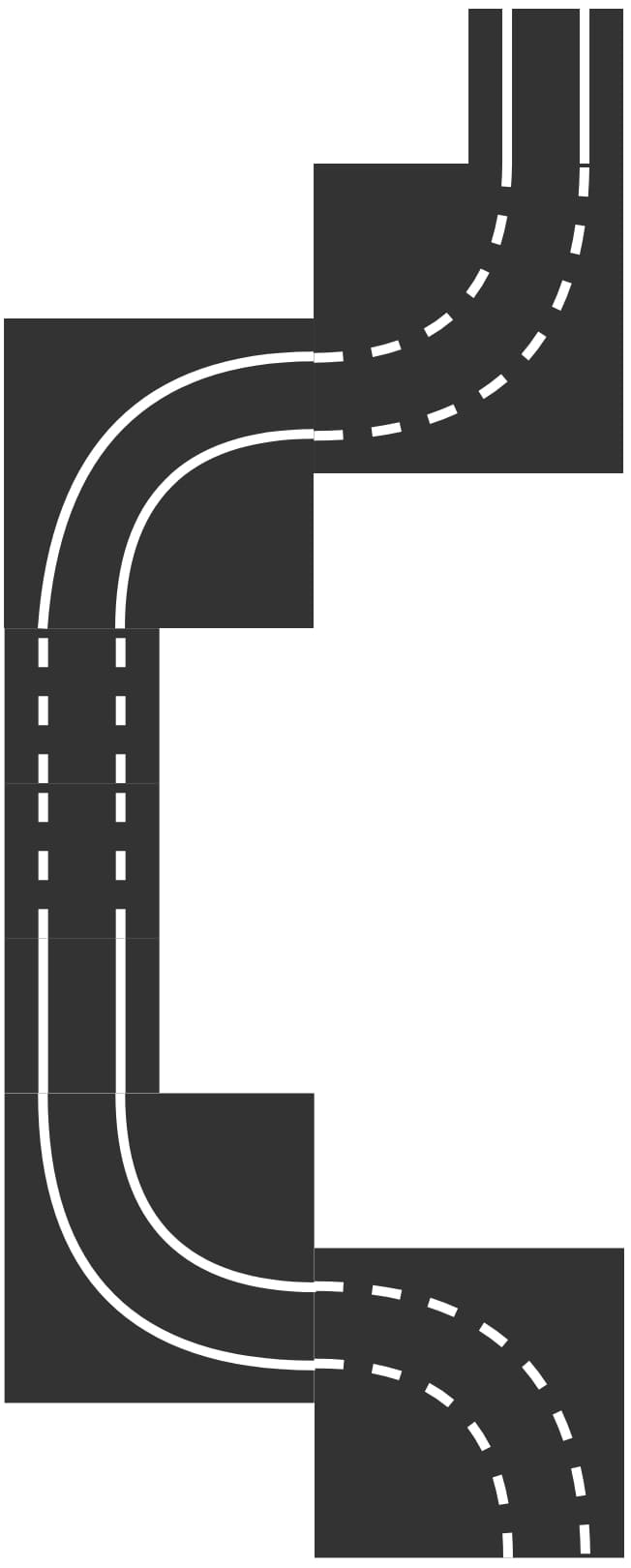}}
		\hspace*{0.1cm}
		\subfloat[]{\includegraphics[scale=0.08]{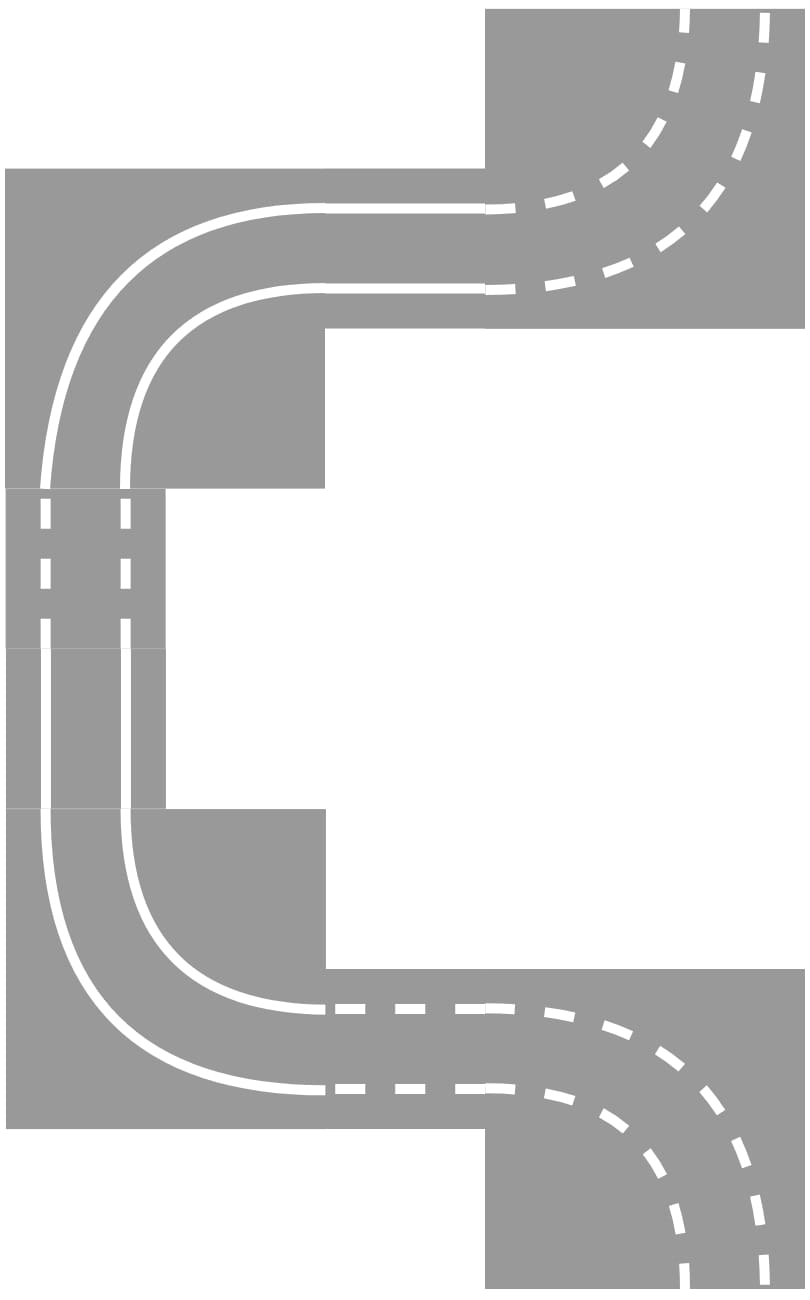}}
		\caption{Overview of our track types for MoLane. (a) - (d) show the black version of the training and validation tracks. These tracks are also constructed using a light gray surface material. (e) and (f) depict our test tracks.}
		\label{fig:tracks_dark}
	\end{figure}

	As shown in \autoref{fig:tracks_dark}, we build six different 1/8th race tracks, where each track is available in two different surface materials (dark and light gray). We vary between dotted and solid lane markings, which are pure white and \SI{50}{\milli\metre} thick. The lanes are constantly \SI{750}{\milli\metre} wide, and the smallest inner radius is \SI{250}{\milli\metre}. The track layouts are designed to roughly contain the same proportion of straight and curved segments to obtain a balanced label distribution. We construct these tracks in four locations with alternating backgrounds and lighting conditions. 
	
	\subsection{Real-World Data Collection}
	Raw image data is recorded from a front-facing Stereolabs ZEDM camera with 30 FPS and a resolution of $1280 \times 720$ pixels. A detailed description of the 1/8th car can be found in the Appendix. The vehicle is moved with a quasi-constant velocity clockwise and counter-clockwise to cover both directions of each track. All collected images from tracks (e) and (f) are used for the test subset. In addition, we annotate lane markings with our labeling tool for validation and testing, which is made publicly available.
	
	\subsection{Simulation Environment}
	We utilize the built-in APIs from CARLA to randomize multiple aspects of the agent and environment, such as weather, daytime, ego vehicle position, camera position, distractor vehicles, and world objects (i.a., walls, buildings, and plants). Weather and daytime are varied systematically by adapting parameters for cloud density, rain intensity, puddles, wetness, wind strength, fog density, sun azimuth, and sun altitude. For further details, we refer to our implementation. To occlude the lanes similar to real-world scenarios, up to five neighbor vehicles are spawned randomly in the vicinity of the agent. We consider five different CARLA maps in urban and highway environments (Town03, Town04, Town05, Town06, and Town10) to collect our dataset, as the other towns' characteristics are not suitable for our task (i.a., mostly straight lanes). In addition, we collect data from the same towns without world objects to strengthen the focus on lane detection, similar to our model vehicle target domain.
	
	\subsection{Simulation Data Agent}
	We implement an efficient agent based on waypoint navigation, which roams randomly and reliably in the aforementioned map environments and collects $1280\times720$ images. In each step, the waypoint navigation stochastically traverses the CARLA road map with a fixed lookahead distance of one meter. 
	In addition, we sample offset values $\Delta y_k$ from the center lane within the range \SI{\pm 1.20}{\metre}.
	To avoid saturation at the lane borders, which would occur with a sinusoidal function, we use the triangle wave function:
	\begin{equation}
		\Delta y_k =\frac{2m}{\pi}\arcsin(\sin(i_k))
	\end{equation}
	where $m$ is the maximal offset and $i_k$ is incremented by $0.08$ for each simulation step $k$. Per frame, our agent moves to the next waypoint with an increment of one meter, enabling the collection of highly diverse data in a fast manner. We use a bagging approach for balancing, which allows us to define lane classes based on their curvature.
	
	\section{The CARLANE Benchmark}
	\label{sec:CARLANE}
	The CARLANE Benchmark consists of three distinct sim-to-real datasets, which we build from our three different domains. The details of the individual subsets can be found in \autoref{table:Dataset overview}.
	
	\vspace{-2pt}
	\textit{MoLane} consists of images from CARLA and the real 1/8th model vehicle. For the abstract real-world domain, we collect 46,843 images with our model vehicle, of which 2,000 validation and 1,000 test images are labeled. For the source domain, we use our simulation agent to gather 84,000 labeled images. To match the label distributions between both domains, we define five lane classes based on the relative angle $\beta$ of the agent to the center lane for our bagging approach: strong left curve ($\beta\leq$\ang{-45}), soft left curve (\ang{-45} $ < \beta \leq $ \ang{-15}), straight (\ang{-15} $ < \beta <$ \ang{15}), soft right curve (\ang{15} $ \leq \beta < $ \ang{45}) and strong right curve (\ang{45}$\leq \beta$). In total, MoLane encompasses 130,843 images. 
	
	\vspace{-2pt}
	\textit{TuLane} consists of images from CARLA, and a cleaned version of the TuSimple dataset \cite{TuSimple2017}, which is licensed under the Apache License, Version 2.0. To clean test set annotations, we utilize our labeling tool to ensure that the up to four lanes closest to the car are correctly labeled. We adapt the bagging classes to align the source dataset with TuSimple's lane distribution: left curve (\ang{-12} $ < \beta \leq$ \ang{5}), straight (\ang{-5} $ < \beta <$ \ang{5}) and right curve (\ang{5} $ \leq \beta < $ \ang{12}). 
	
	\vspace{-2pt}
	\textit{MuLane} is a multi-target UDA dataset and is a balanced mixture of images from MoLane and TuLane. For MuLane's entire training set and its source domain validation and test set, we use all available images from TuLane and sample the same amount of images from MoLane. We adopt the 1,000 test images from MoLane's target domain and sample 1,000 test images from TuSimple to form MuLane's test set. For the validation set, we use the 2,000 validation images from MoLane and 2,000 of the remaining validation and test images of TuLane's target domain. In total, MuLane consists of 65,336 images.
	
	To further analyze CARLANE, we visualize the ground truth lane distributions in Figure \ref{fig:dataset_distribution}. We observe that the lane distributions of source and target data from our datasets are well aligned. 
	
	MoLane, TuLane, and MuLane are publicly available at \href{https://carlanebenchmark.github.io}{https://carlanebenchmark.github.io} and licensed under the Apache License, Version 2.0.
	
	\begin{table}
		\caption{Dataset overview. Unlabeled images denoted by *, partially labeled images denoted by ** }
		\label{table:Dataset overview}
		\small
		\centering
		\begin{tabular}{lcccccc}
			\toprule
			Dataset                  & domain             & total images & train   & validation  & test  & lanes       \\ \midrule
			\multirow{2}{*}{MoLane}  & CARLA simulation   & 84,000       & 80,000  & 4,000       & -     & \(\leq\) 2  \\ 
			& model vehicle      & 46,843       & 43,843* & 2,000       & 1,000 & \(\leq\) 2  \\ \midrule
			\multirow{2}{*}{TuLane}  & CARLA simulation   & 26,400       & 24,000  & 2,400       & -     & \(\leq\) 4  \\ 
			& TuSimple \cite{TuSimple2017} & 6,408        & 3,268   & 358         & 2,782 & \(\leq\) 4  \\ \midrule
			\multirow{2}{*}{MuLane}  & CARLA simulation   & 52,800       & 48,000  & 4,800       & -     & \(\leq\) 4  \\ 
			& model vehicle + TuSimple \cite{TuSimple2017} & 12,536      & 6,536** & 4,000       & 2,000 & \(\leq\) 4  \\ 
			\bottomrule
		\end{tabular}
	\end{table}
	
	\subsection{Dataset Format}
	For each dataset, we split training, validation, and test samples into source and target subsets. Lane annotations are stored within a \emph{.json} file containing the lanes' y-values discretized by raw anchors, the lanes' x-values, and the image file path following the data format of TuSimple\cite{TuSimple2017}. Additionally, we adopt the method from \cite{qin2020ultra} to generate \emph{.png} lane segmentations and a \emph{.txt} file containing the linkage between the raw images and their segmentation as well as the presence and absence of a lane.

	\subsection{Dataset Tasks}
	The main task of our datasets is UDA for lane detection, where the goal is to predict lane annotations $Y_{t} \in \mathbb{R}^{R \times G \times N}$ given the input image $X_{t} \in \mathbb{R}^{H \times W \times 3}$ from the unlabeled target domain $\mathcal{D}_{\mathcal{T}} = \{(X_{t})\}_{t\in \mathcal{T}}$. $R$ defines the number of row anchors, $G$ the number of griding cells, and $N$ the number of lane annotations available in the dataset, where the definition of $Y_{t}$ follows \cite{TuSimple2017}. During training time, the images $X_{s} \in \mathbb{R}^{H \times W \times 3}$, corresponding labels $Y_{s} \in \mathbb{R}^{H \times W \times C}$ from the source domain $\mathcal{D}_{\mathcal{S}} = \{(X_{s},Y_{s})\}_{s\in \mathcal{S}}$, and the unlabeled target images $X_{t}$ are available. Additionally, MuLane focuses on multi-target UDA, where $\mathcal{D}_{\mathcal{T}} = \{(X_{t_1})\cup(X_{t_2})\}_{t_1\in \mathcal{T}_1, t_2\in \mathcal{T}_2}$.
	
	Although we focus on sim-to-real UDA, our datasets can be used for unsupervised and semi-supervised tasks and partially for supervised learning tasks. Furthermore, a real-to-real transfer can be performed between the target domains of our datasets.
	
	\begin{figure}
		\centering
		\small
		\begin{tabular}{rccc}
			~ & \textbf{MoLane} & \textbf{TuLane} & \textbf{MuLane} \\
			%
			\textbf{Source} & 
			\includegraphics[width=.22\linewidth,valign=m,trim={6.1cm 3.9cm 12.3cm 4cm},clip]{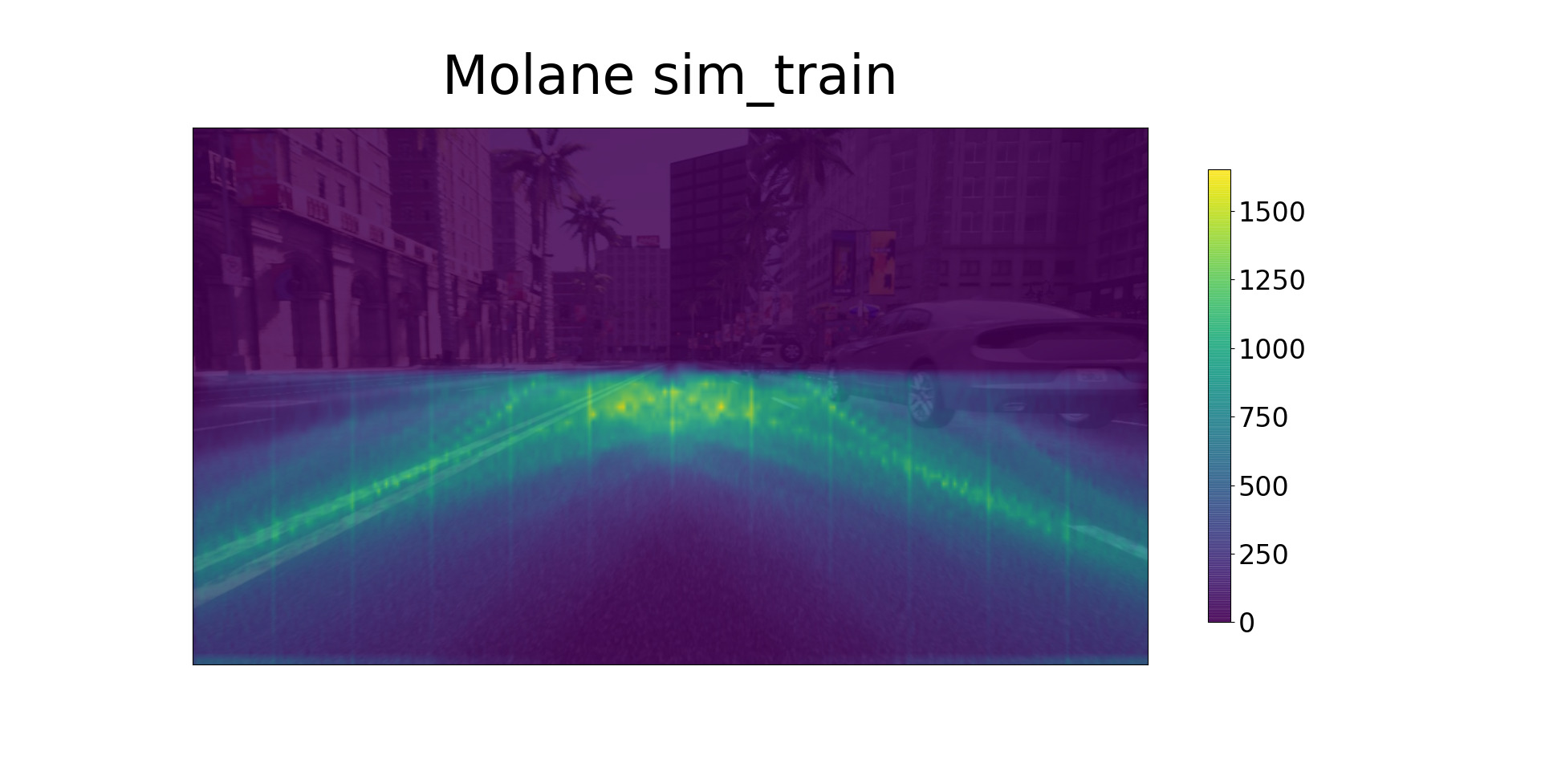} & \includegraphics[width=.22\linewidth,valign=m,trim={6.1cm 3.9cm 12.3cm 4cm},clip]{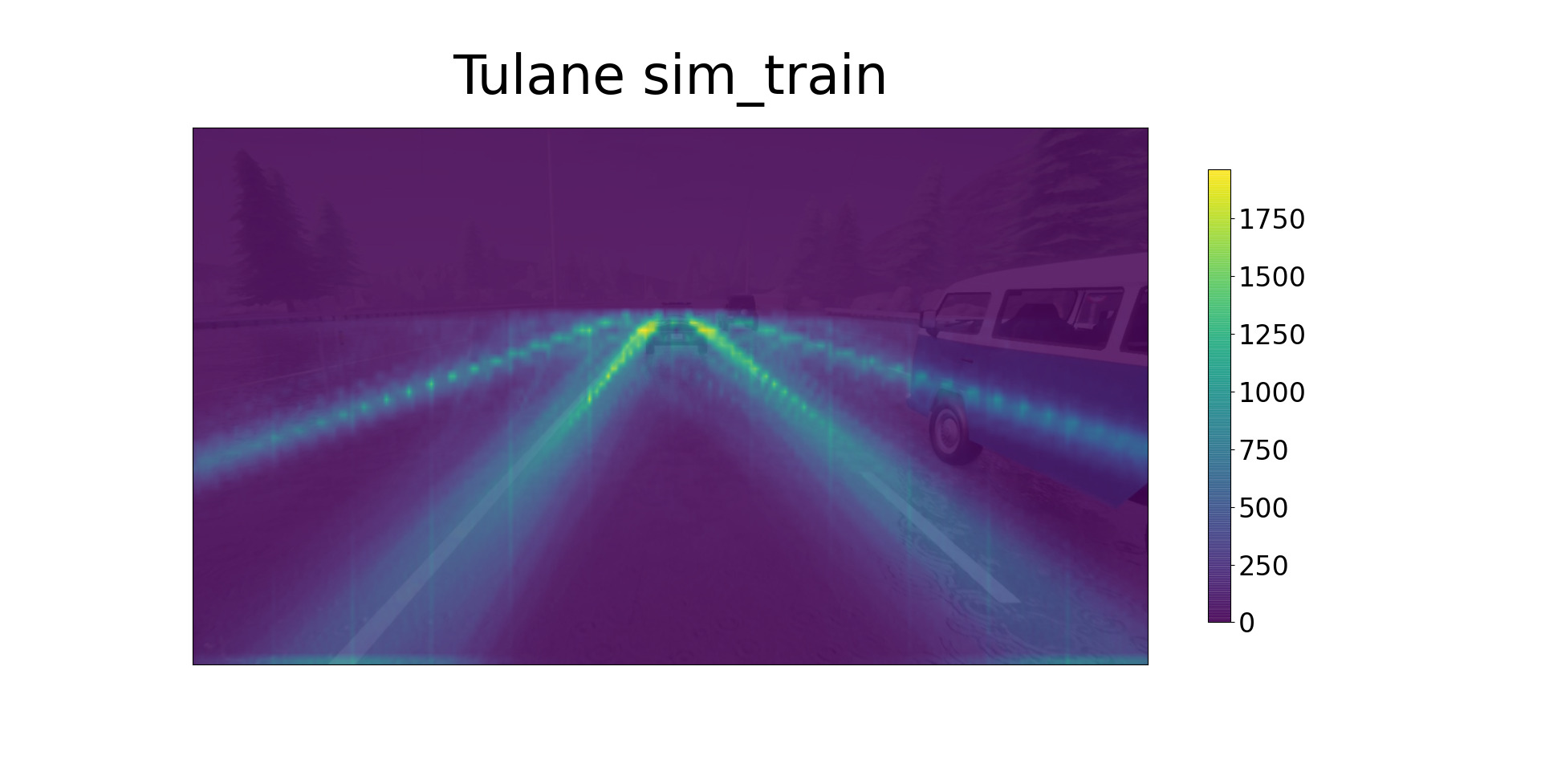} & \includegraphics[width=.22\linewidth,valign=m,trim={6.1cm 3.9cm 12.3cm 4cm},clip]{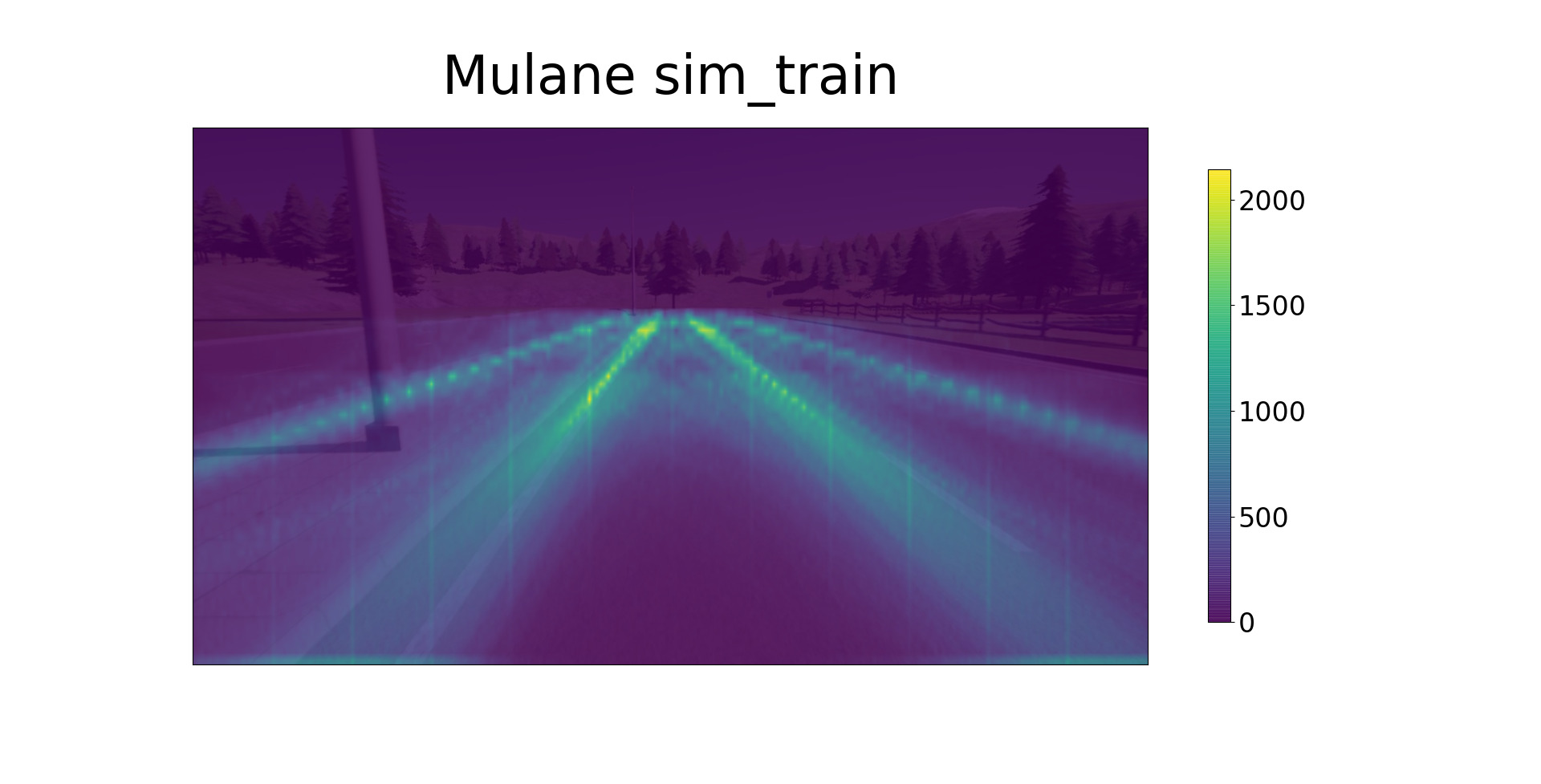}\\
			~ & ~ & ~ \\
			\textbf{Target} & 
			\includegraphics[width=.22\linewidth,valign=m,trim={6.1cm 3.9cm 12.3cm 4cm},clip]{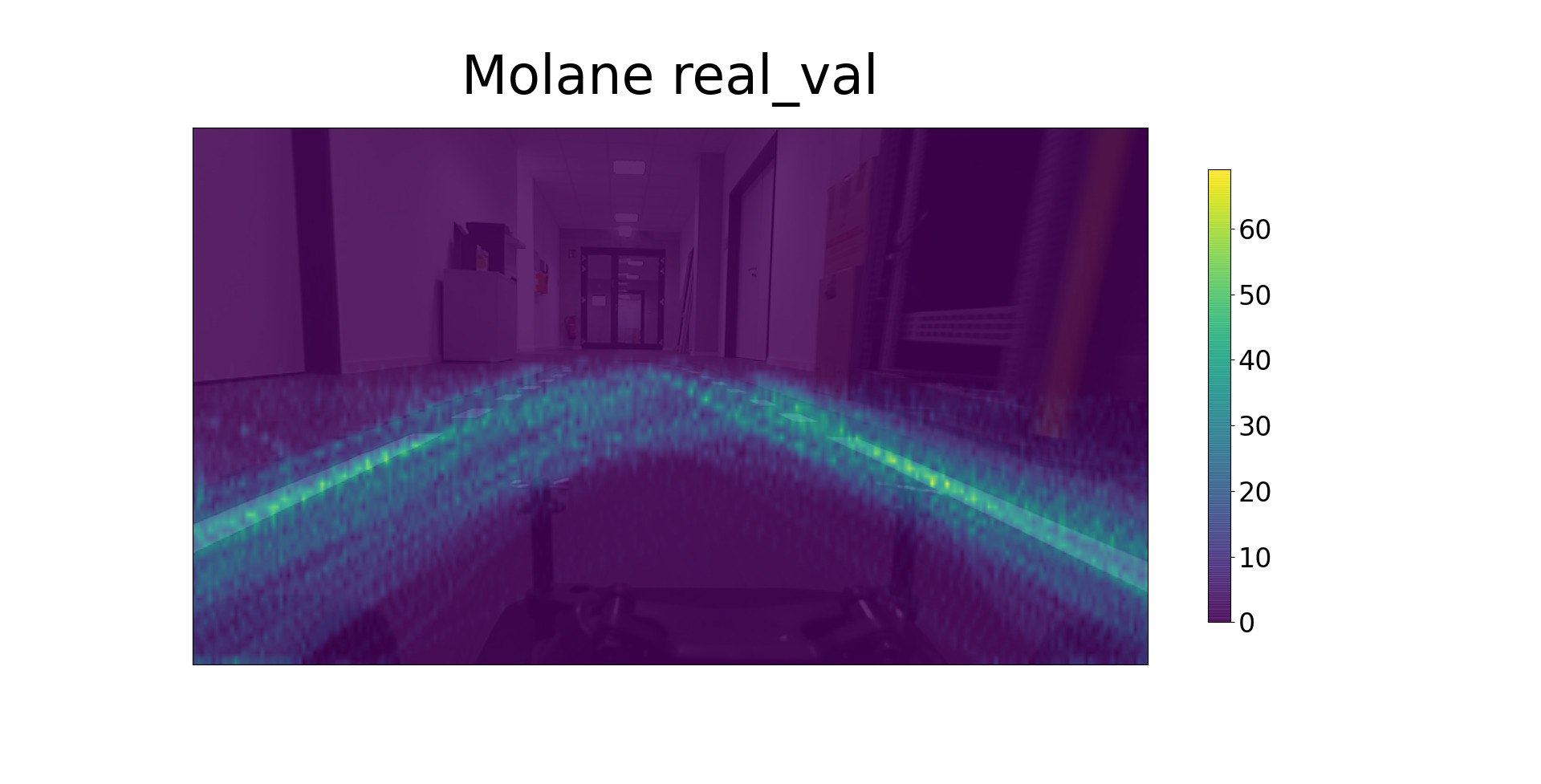} & \includegraphics[width=.22\linewidth,valign=m,trim={6.1cm 3.9cm 12.3cm 4cm},clip]{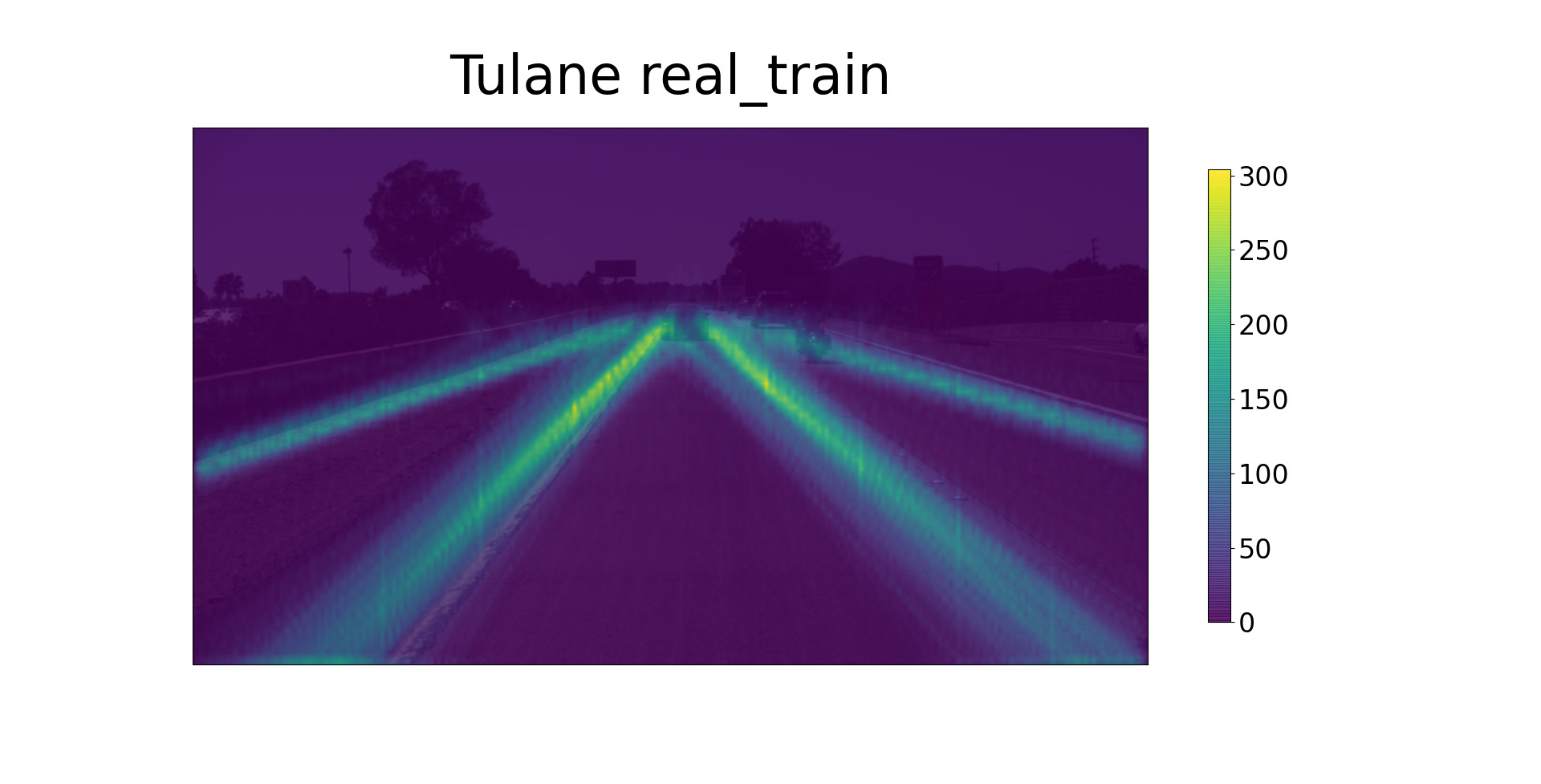} & \includegraphics[width=.22\linewidth,valign=m,trim={6.1cm 3.9cm 12.3cm 4cm},clip]{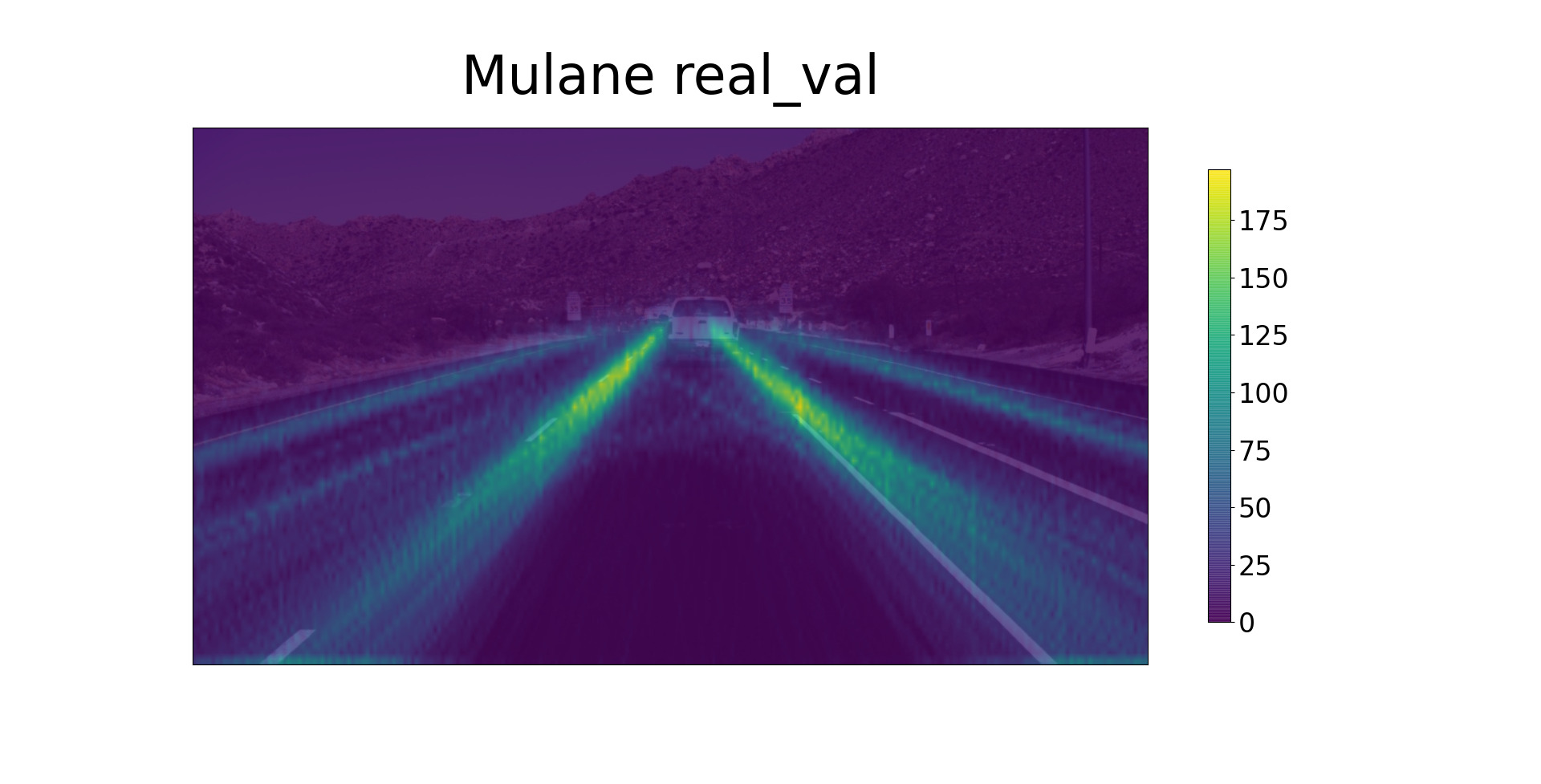}\\
		\end{tabular}
		\caption{Lane annotation distributions of the three subsets of CARLANE. Since the real-world training data of MoLane and MuLane is unlabeled, we utilize their validation data for visualization.}
		\label{fig:dataset_distribution}
	\end{figure}
	
	\section{Benchmark Experiments}
	\label{sec:Experiments}
	We conduct experiments on our CARLANE Benchmark for several UDA methods from the literature and our proposed method. Additionally, we train fully supervised baselines on all domains.
	
	\subsection{Metrics}
	For evaluation, we use the following metrics:
	
	\textit{(1) Lane Accuracy} (LA) \cite{qin2020ultra} is defined by $\textrm{LA} = \frac{p_{c}}{p_{y}}$, where $p_c$ is the number of correctly predicted lane points and $p_{y}$ is the number of ground truth lane points. Lane points are considered as correct if their $L_1$ distance is smaller than the given threshold $t_{pc}=\frac{20}{\cos(a_{yl})}$, where $a_{yl}$ is the angle of the corresponding ground truth lane.
	
	\textit{(2) False Positives} (FP) and \textit{False Negatives} (FN) \cite{qin2020ultra}: To further determine the error rate and to draw more emphasis on mispredicted or missing lanes, we measure false positives with $\textrm{FP}=\frac{l_{f}}{l_p}$ and false negatives with $\textrm{FN}=\frac{l_m}{l_{y}}$, where $l_{f}$ is the number of mispredicted lanes, $l_p$ is the number of predicted lanes, $l_{m}$ is the number of missing lanes and $l_{y}$ is the number of ground truth lanes. Following \cite{qin2020ultra}, we classify lanes as mispredicted, if the $LA < 85\%$. 
	
	\subsection{Baselines}
	We use Ultra Fast Structure-aware Deep Lane Detection (UFLD) \cite{qin2020ultra} as baseline and strictly adopt its training scheme and hyperparameters. UFLD treats lane detection as a row-based classification problem and utilizes the row anchors defined by TuSimple \cite{TuSimple2017}.
	To achieve a lower bound for the evaluated UDA methods, we train UFLD as a supervised baseline on the source simulation data (UFLD-SO). Furthermore, we train our baseline on the labeled real-world training data for a surpassable fully-supervised performance in the target domain (UFLD-TO). Since the training images from MoLane and MuLane have no annotations, we train UFLD-TO in these cases on the labeled validation images and validate our model on the entire test set.
	
	\subsection{Compared UDA Methods}
	We evaluate the following feature-level UDA methods on the CARLANE Benchmark by adopting their default hyperparameters and tuning them accordingly. Each model is initialized with the pre-trained feature encoder of our baseline model (UFLD-SO). The optimized hyperparameters can be found in \autoref{table:Hyperpparameters}. 
	\vspace{-2pt}
	
	\textit{(1) DANN} \cite{Ganin2016} is an adversarial discriminative method that utilizes a shared feature encoder and a dense domain classifier connected via a gradient reversal layer.
	\vspace{-2pt}
	
	\textit{(2) ADDA} \cite{Tzeng2017ADDA} employs a feature encoder for each domain and a dense domain discriminator. Following ADDA, we freeze the weights of the pre-trained classifier of UFLD-SO to obtain final predictions.
	
	\textit{(3) SGADA} \cite{sgada2021} builds upon ADDA and utilizes its predictions as pseudo labels for the target training images. Since UFLD treats lane detection as a row-based classification problem, we reformulate the pseudo label selection mechanism. For each lane, we select the highest confidence value from the griding cells of each row anchor. Based on their griding cell position, the confidence values are divided into two cases: absent lane points and present lane points. Thereby, the last griding cell represents absent lane points as in \cite{qin2020ultra}. For each case, we calculate the mean confidence over the corresponding lanes. We then use the thresholds defined by SGADA
	to decide whether the prediction is treated as a pseudo label.
	
	\textit{(4) SGPCS} (ours) builds upon PCS \cite{yue2021prototypical} and performs in-domain contrastive learning and cross-domain self-supervised learning via cluster prototypes. Our overall objective function comprises the in-domain and cross-domain loss from PCS, the losses defined by UFLD, and our adopted pseudo loss from SGADA. We adjust the momentum for memory bank feature updates to $0.5$ and use spherical K-means \cite{johnson2019billion} with $K=2,500$ to cluster them into prototypes.
	
	\begin{table}
		\caption{Optimized hyperparameters to achieve the reported results. $C$ denotes domain classifier parameters, $D$ denotes domain discriminator parameters, adv the adversarial loss from \cite{Tzeng2017ADDA} and cls the classifier loss, sim the similarity loss and aux the auxiliary loss from \cite{qin2020ultra}. Loss weights are set to $1.0$ unless stated otherwise.}
		\label{table:Hyperpparameters}
		\tiny
		\centering
		\begin{tabular}{cccccll}
			\toprule
			Method  	& Initial Learning Rate    	             & Scheduler     & Batch Size  & Epochs    &Losses    & Other Changes \\   
			\midrule	
			UFLD-SO                &$4e^{-4}$                 & Cosine Annealing           & 4           &150  & cls, sim, aux & - \\       
			DANN                   & $1e^{-5}$, $C$: $1e^{-3}$   & $\frac{1e^{-5}}{(1 + 10p)^{0.75}}$   & 4  & 30 & cls, sim, aux, adv \cite{Ganin2016}  & $C$: 3 fc layers (1024-1024-2)  \\   
			ADDA                    & $1e^{-6}$, $D$: $1e^{-3}$  & Constant           & 16          & 30 & map \cite{Tzeng2017ADDA}, adv \cite{Tzeng2017ADDA}& $D$: 3 fc layers (500-500-2) \\    
			SGADA                   & $1e^{-6}$, $D$: $1e^{-3}$  & Constant           & 16          & 15 &  map \cite{Tzeng2017ADDA}, adv \cite{Tzeng2017ADDA}, pseudo: 0.25  & Pseudo label selection \\
			\multirow{2}{*}{SGPCS}  &\multirow{2}{*}{$4e^{-4}$} & \multirow{2}{*}{Cosine Annealing} & \multirow{2}{*}{16} & \multirow{2}{*}{10} & in-domain \cite{yue2021prototypical}, cross-domain \cite{yue2021prototypical} & \multirow{2}{*}{-} \\
			&&&&& cls, sim, aux, pseudo: 0.25 & \\
			UFLD-TO                   &$4e^{-4}$  & Cosine Annealing           & 4          & 300 & cls, sim, aux & - \\	   
			\bottomrule	
		\end{tabular}
	\end{table}
	
	\subsection{Implementation Details}
	\label{sec:ImplementationDetails}
	We implement all methods in PyTorch 1.8.1 and train them on a single machine with four RTX 2080 Ti GPUs. Tuning all methods took a total amount of compute of approximately 3.5 petaflop/s-days. The training times for each model range from 4-13 days for UFLD baselines and 6-44 hours for domain adaption methods. In addition, we found that applying output scaling on the last linear layer of the model yields slightly better results. Therefore, we divide the models' output by 0.5. Our implementation is publicly available at \href{https://carlanebenchmark.github.io}{https://carlanebenchmark.github.io}.
	
	\subsection{Evaluation}
	\textbf{Quantitative Evaluation.}
	In \autoref{table:Results} we report the results on MoLane, TuLane, and MuLane across five different runs. We observe that UFLD-SO is able to generalize to a certain extent to the target domain. This is mainly due to the alignment of semantic structure from both domains. ADDA, SGADA, and our proposed SGPCS manage to adapt the model to the target domain slightly and consistently. However, DANN suffers from negative transfer \cite{wang2019characterizing} when trained on MoLane and MuLane. The negative transfer of DANN for complex domain adaptation tasks is also observed in other works \cite{ tanwisuth2021prototype,fan2022self, wang2019characterizing, kim2020cross} and can be explained by the source domain's data distribution and the model complexity \cite{wang2019characterizing}. In our case, the source domain contains labels not present in the target domain, as shown in Figure \ref{fig:dataset_distribution}, which is more pronounced in MoLane and MuLane.
	
	We want to emphasize that with an accuracy gain of a maximum of 5.14\% (SGPCS) and high false positive and false negative rates, the domain adaptation methods are not able to achieve comparable results to the supervised baselines (UFLD-TO). Furthermore, we observe that false positive and false negative rates increase significantly on MuLane, indicating that the multi-target dataset forms the most challenging task. False positives and false negatives represent wrongly detected and missing lanes which can lead to crucial impacts on autonomous driving functions. These results affirm the need for the proposed CARLANE Benchmark to further strengthen the research in UDA for lane detection.
	
	\begin{table}
		\caption{Performance on the test set. Lane accuracy (LA), false positives (FP), and false negatives (FN) are reported in \%.}
		\label{table:Results}
		\tiny
		\centering
		\setlength{\tabcolsep}{4.6pt}
		\begin{tabular}{c|ccc|ccc|ccc}
			\toprule
			\multirow{2}{*}{ResNet-18} & \multicolumn{3}{c|}{MoLane}       & \multicolumn{3}{c|}{TuLane}                       & \multicolumn{3}{c}{MuLane} \\
			& LA           & FP & FN         & LA            & FP    &  FN                      & LA       & FP    &  FN   \\
			\midrule
			UFLD-SO                   & 88.15          & 34.35  &  28.45        & 87.43           & 34.21 & 23.48                    & 79.61      & 44.78 & 33.36 \\
			DANN \cite{Ganin2016}     & 85.25$\pm$0.49 & 39.07$\pm$1.21  & 36.18 $\pm$1.50 & 88.74$\pm$0.32 & 32.71$\pm$0.52  & 21.64$\pm$0.65  & 78.25$\pm$0.62 & 48.67$\pm$1.17 & 41.69$\pm$1.80  \\
			ADDA \cite{Tzeng2017ADDA} & 92.10$\pm$0.21 & 19.71$\pm$0.77  & 11.46$\pm$0.92 & 90.72$\pm$0.15 & 29.73$\pm$0.36  & 17.67$\pm$0.42  & 81.25$\pm$0.33 & 41.69$\pm$0.40 & 28.58$\pm$0.57   \\
			SGADA \cite{sgada2021}    & 93.15$\pm$0.12 & 16.23$\pm$0.22 & 7.68$\pm$0.26 & \textbf{91.70}$\pm$0.13 & \textbf{28.42}$\pm$0.34 & \textbf{16.10}$\pm$0.43 & 82.05$\pm$0.10 & \textbf{40.37}$\pm$0.33 & \textbf{25.95}$\pm$0.36\\
			SGPCS (ours)                & \textbf{93.29}$\pm$0.07 & \textbf{16.22}$\pm$0.17 & \textbf{7.29}$\pm$0.13  & 91.55$\pm$0.13 & 28.52$\pm$0.21 & 16.16$\pm$0.26 & \textbf{82.91}$\pm$0.19 & 40.76$\pm$0.57 & 26.14$\pm$0.75\\
			\midrule
			UFLD-TO                   & 97.15           & 0.96  &    0.05       &  94.97          & 18.05           & 3.84          & 87.64  & 29.48 & 11.52 \\ 
			
			\midrule
			\midrule
			
			ResNet-34                & LA            & FP & FN        & LA            & FP     & FN                       & LA        & FP    & FN    \\ 
			\midrule
			UFLD-SO                  & 88.76          & 31.30     &   26.70    & 89.42           & 32.35  & 21.19                    & 80.70      & 43.63 & 31.40 \\
			DANN \cite{Ganin2016}    & 89.58$\pm$0.63  & 28.75$\pm$1.53 & 23.24$\pm$2.16  & 91.06$\pm$0.14 & 30.17$\pm$0.20 & 18.54$\pm$0.25    & 80.40$\pm$0.15 & 43.52$\pm$0.37 & 31.53$\pm$0.66\\
			ADDA \cite{Tzeng2017ADDA}& 91.76$\pm$0.29  & 21.04$\pm$0.81 & 13.06$\pm$1.02  & 91.39$\pm$0.16 & 28.76$\pm$0.30 & 16.63$\pm$0.36    & 81.64$\pm$0.34 & 40.74$\pm$0.48 & 27.50$\pm$0.78\\
			SGADA \cite{sgada2021}   & 92.59$\pm$0.18  & 18.32$\pm$0.24& 9.88$\pm$0.29   & 92.04$\pm$0.09 & 28.18$\pm$0.20 & 15.99$\pm$0.24    & \textbf{82.93}$\pm$0.03 & \textbf{39.45}$\pm$0.11 & \textbf{24.98}$\pm$0.13\\
			SGPCS (ours)               & \textbf{92.82}$\pm$0.27 & \textbf{17.10}$\pm$0.73 & \textbf{8.77}$\pm$0.95 & \textbf{93.29}$\pm$0.18 & \textbf{25.68}$\pm$0.48 & \textbf{12.73}$\pm$0.59 & 82.87$\pm$0.17 & 40.13$\pm$0.28 & 25.38$\pm$0.45\\
			\midrule
			UFLD-TO                  & 96.92           & 0.94   &   0.03     & 94.43          & 20.74      & 7.20  & 87.62  &  29.19  & 11.08 \\
			\bottomrule
			
		\end{tabular}
	\end{table}
	
	\begin{figure}
		\centering
		\small
		\begin{tabular}{c@{}c@{}c@{}c@{}c}
			UFLD-SO & DANN & ADDA & SGADA & SGPCS \\
			\includegraphics[width=.2\linewidth,valign=m]{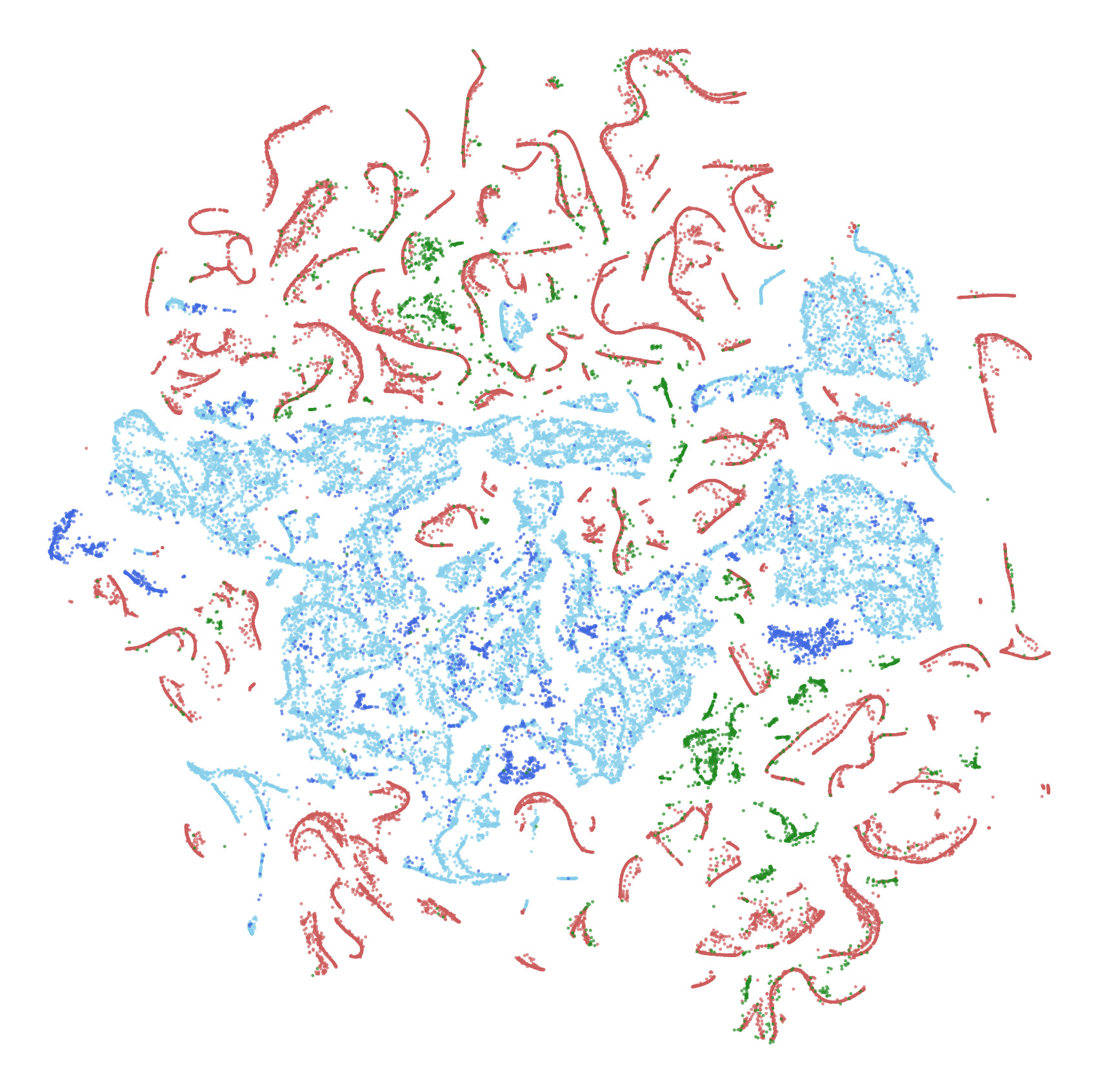} & \includegraphics[width=.2\linewidth,valign=m]{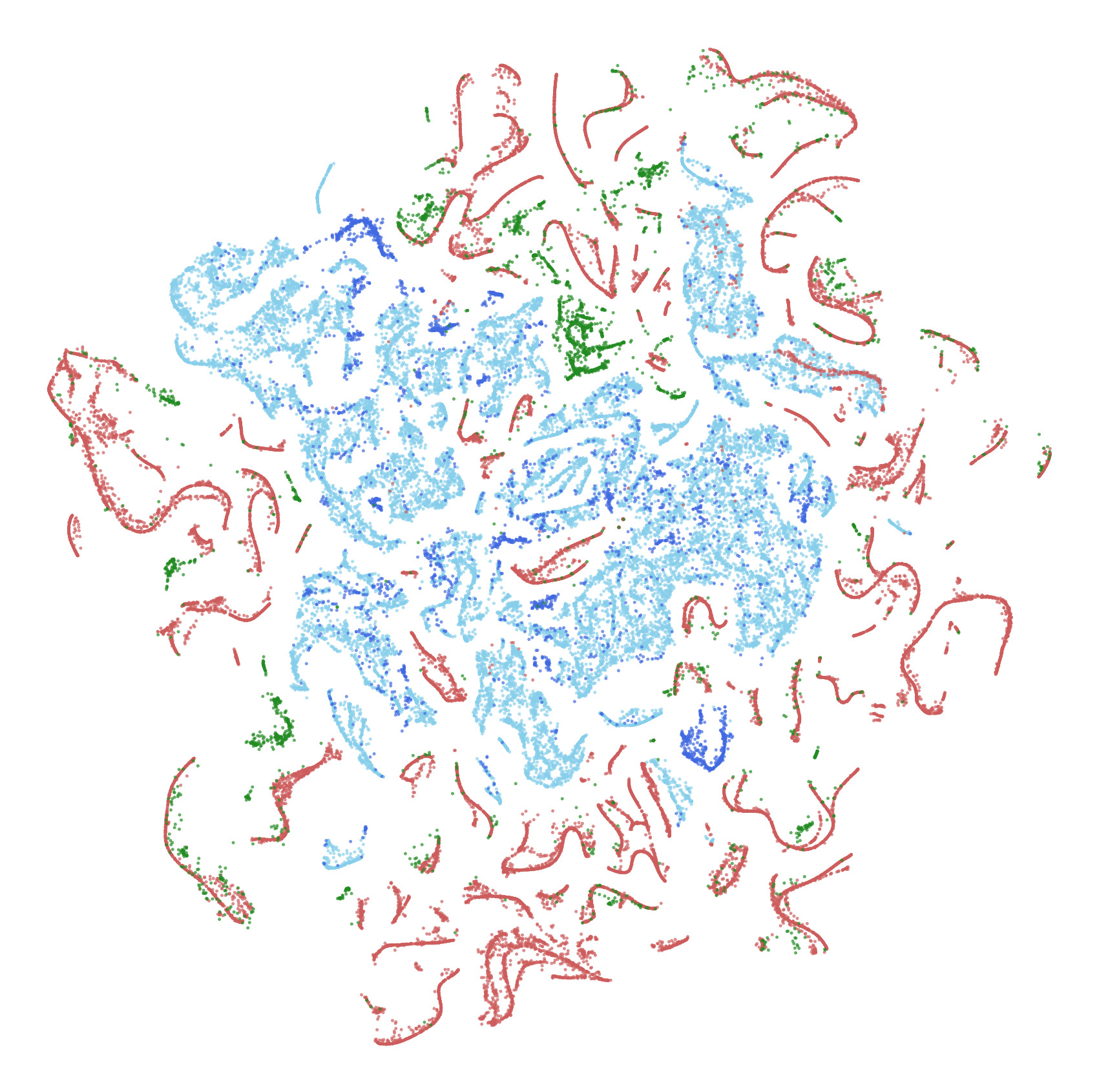} &
			\includegraphics[width=.2\linewidth,valign=m]{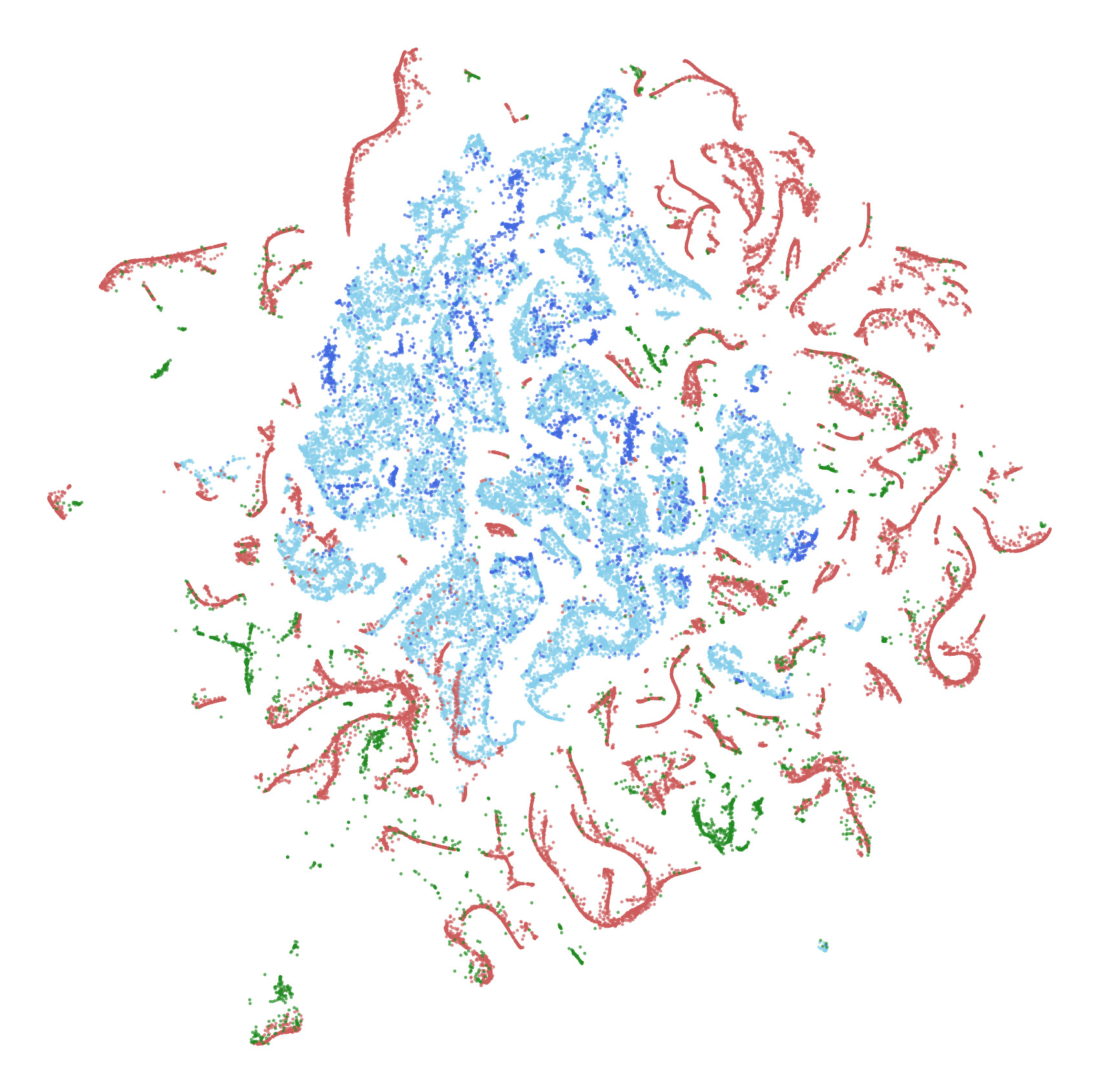} &
			\includegraphics[width=.2\linewidth,valign=m]{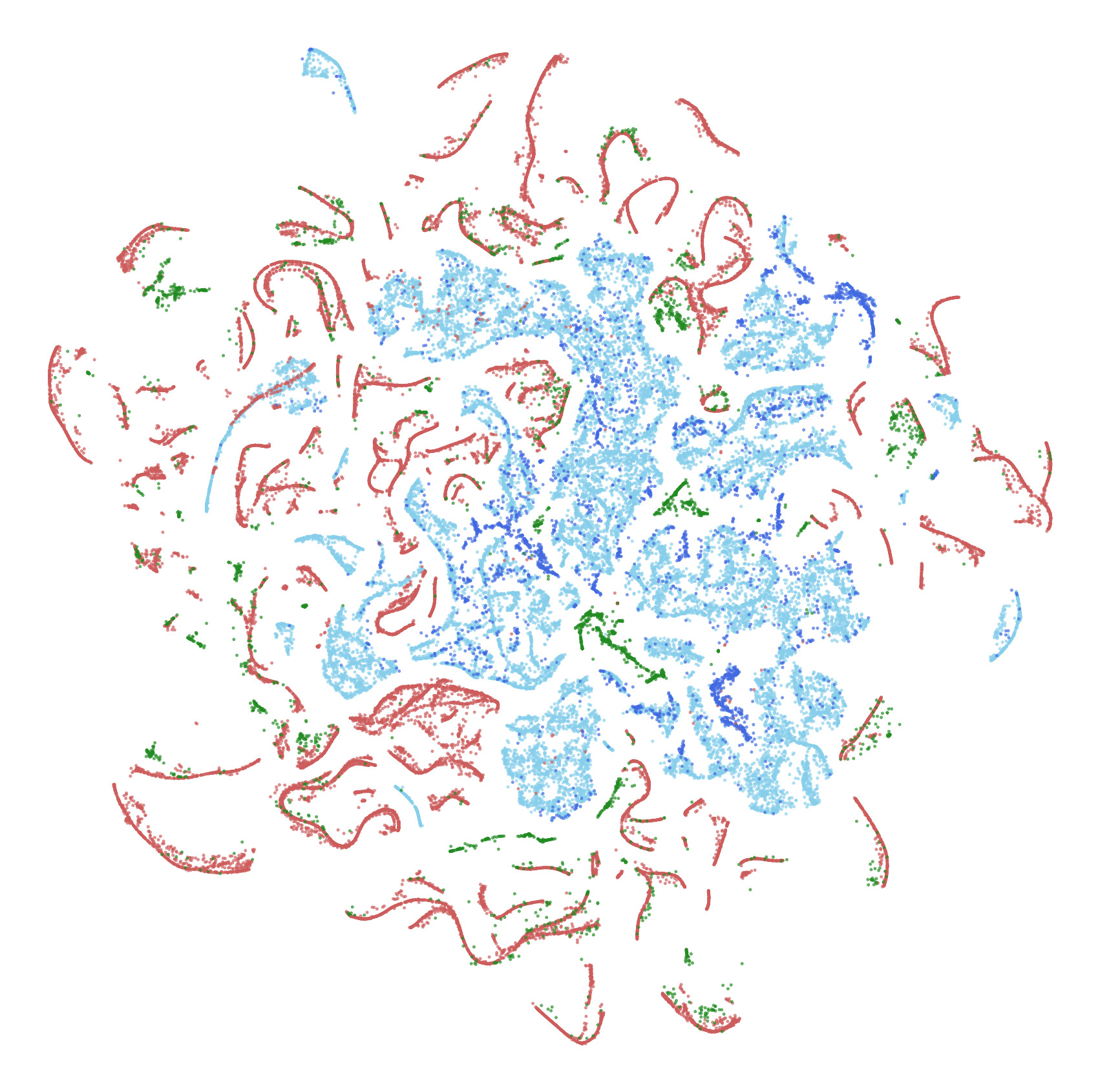} & \includegraphics[width=.2\linewidth,valign=m]{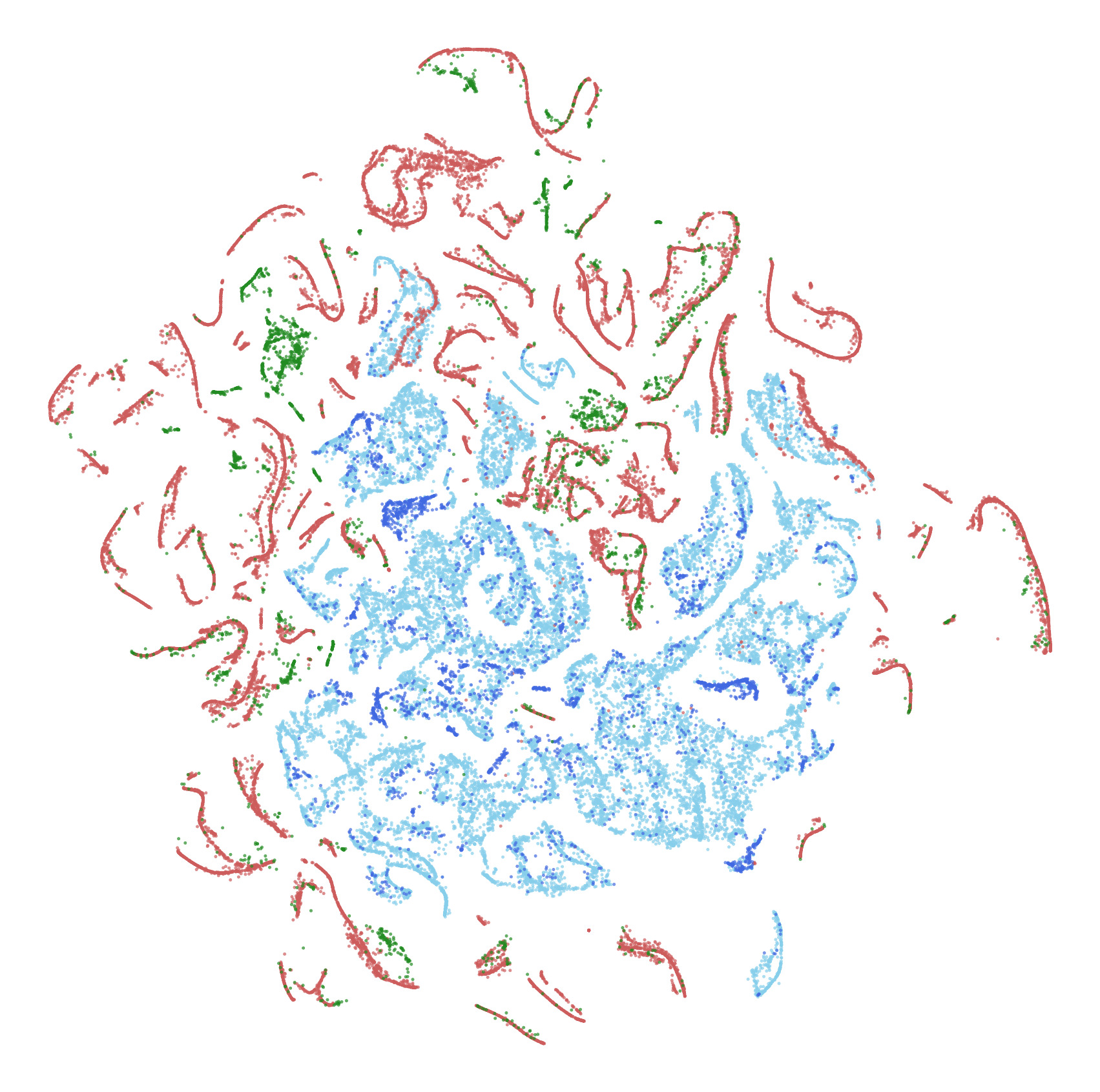}\\
		\end{tabular}
		\caption{t-SNE visualization of MuLane dataset. The source domain is marked in blue, the real-world model vehicle target domain in red, and the TuSimple domain in green.}
		\label{fig:TSNE_plot_mulane}
	\end{figure}

\begin{figure}
		\centering
		\small
		\begin{tabular}{rc@{}c@{}c@{}c}
			~ & \textbf{MoLane} & \textbf{TuLane} & \multicolumn{2}{c}{\textbf{MuLane}} \\
			%
			\textbf{UFLD-SO} & 
			\includegraphics[width=.18\linewidth,valign=m]{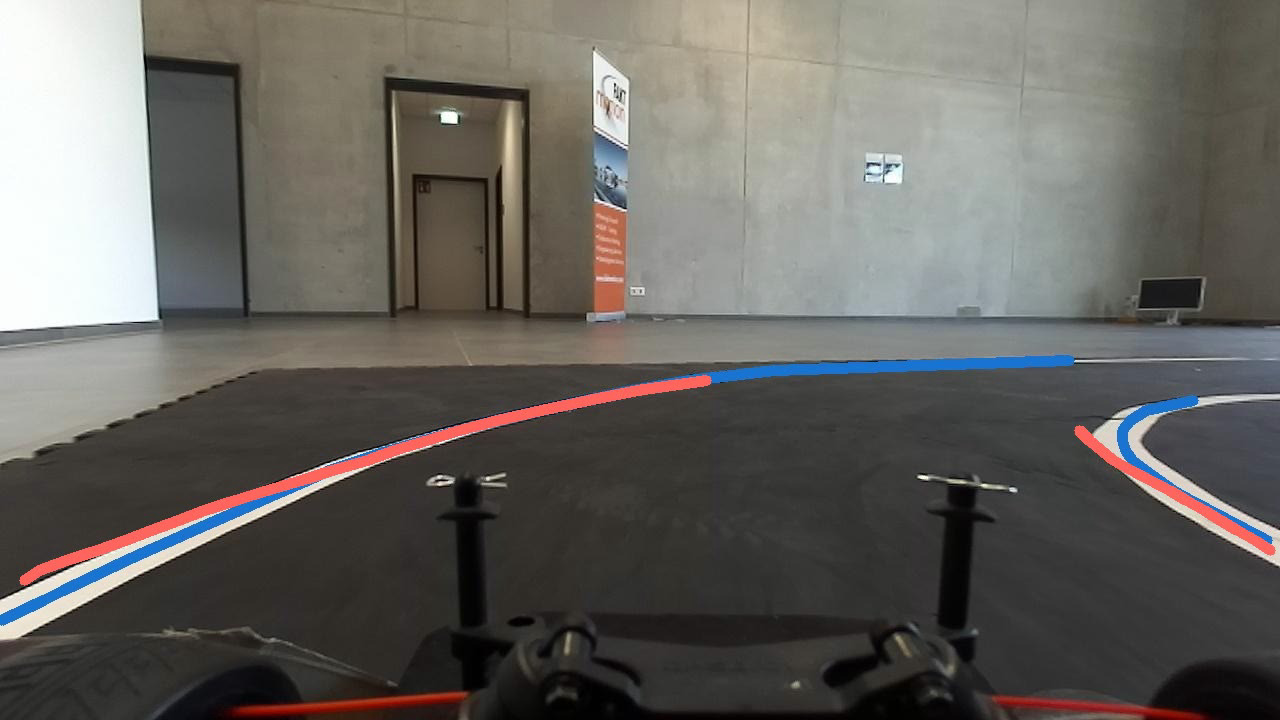} & \includegraphics[width=.18\linewidth,valign=m]{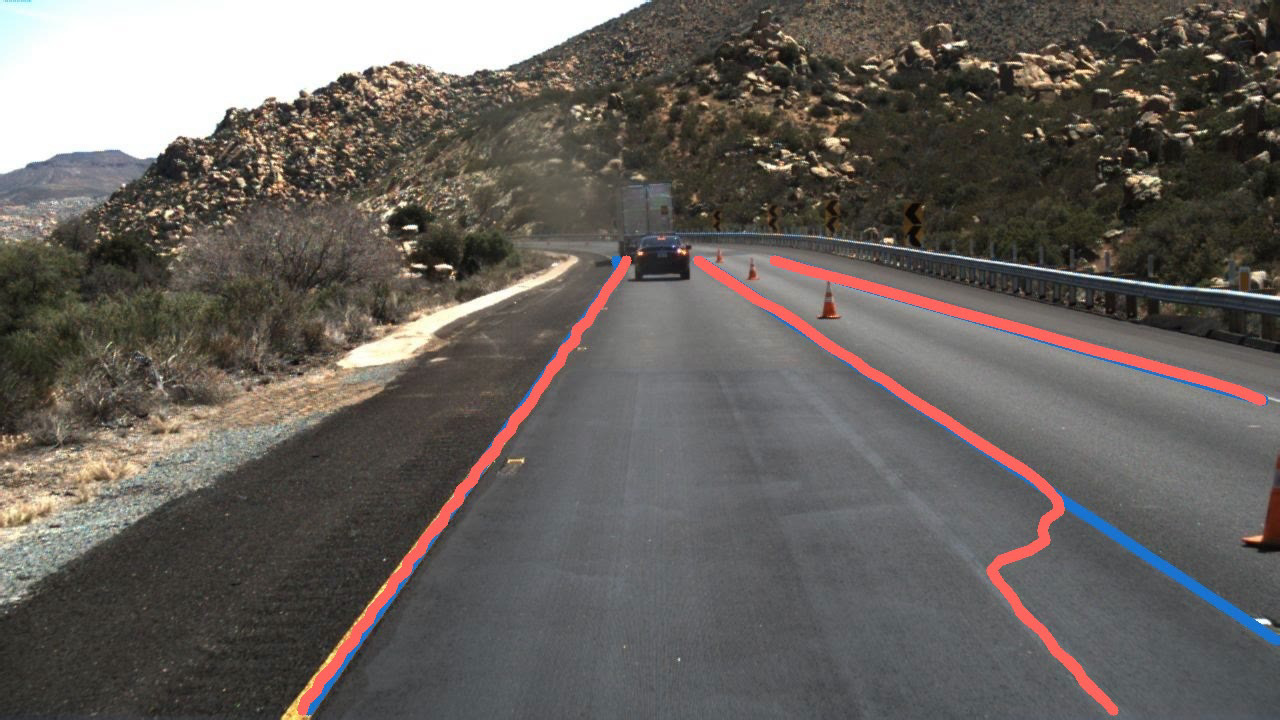} &
			\includegraphics[width=.18\linewidth,valign=m]{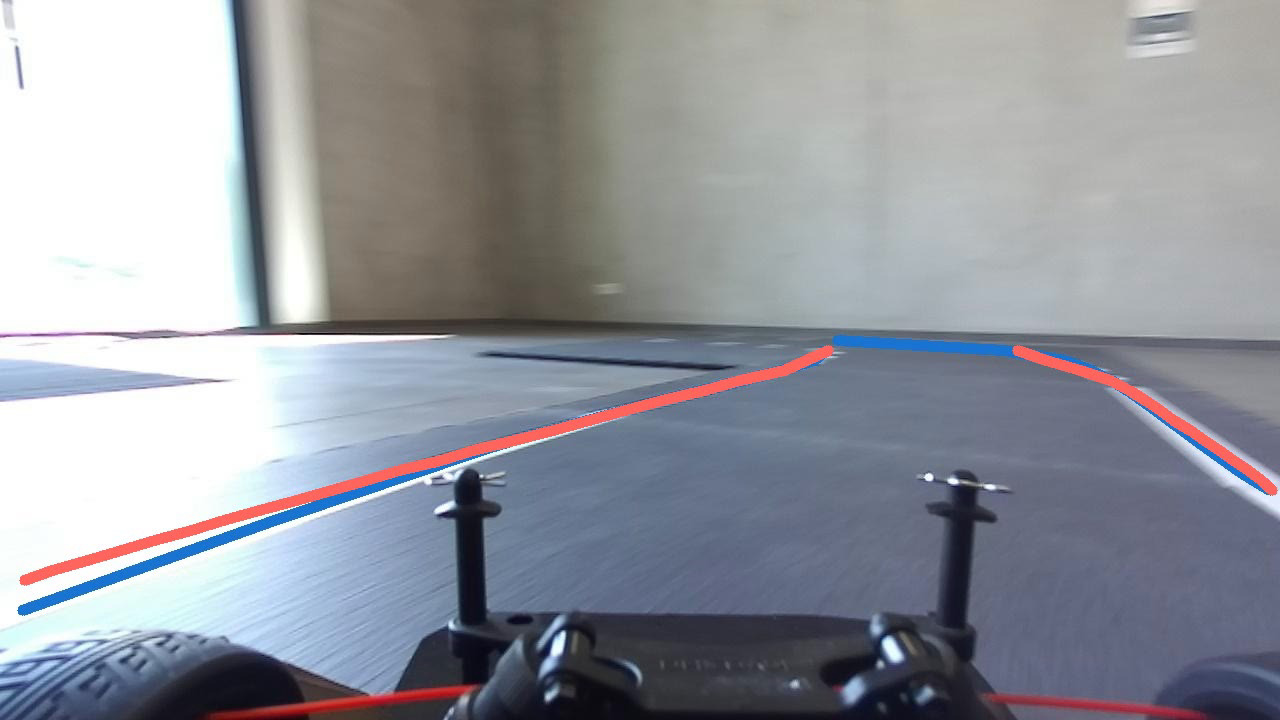} & \includegraphics[width=.18\linewidth,valign=m]{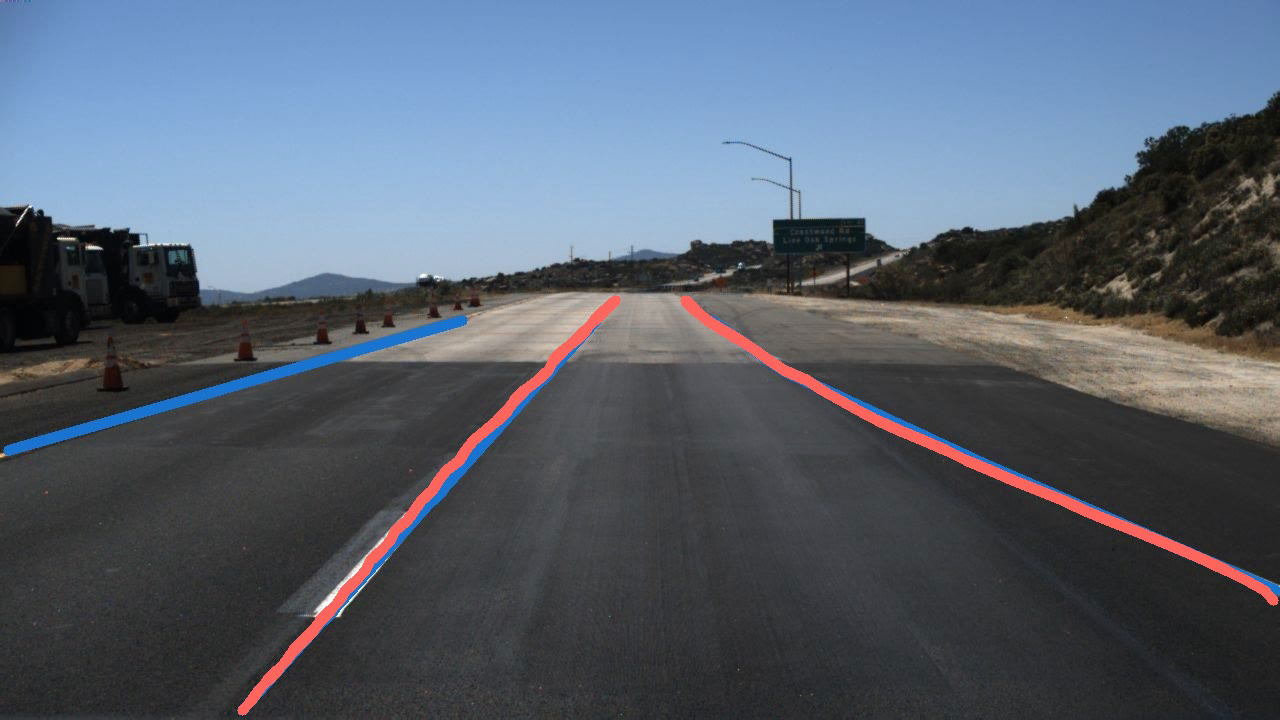}\\
			%
			\textbf{DANN} & 
			\includegraphics[width=.18\linewidth,valign=m]{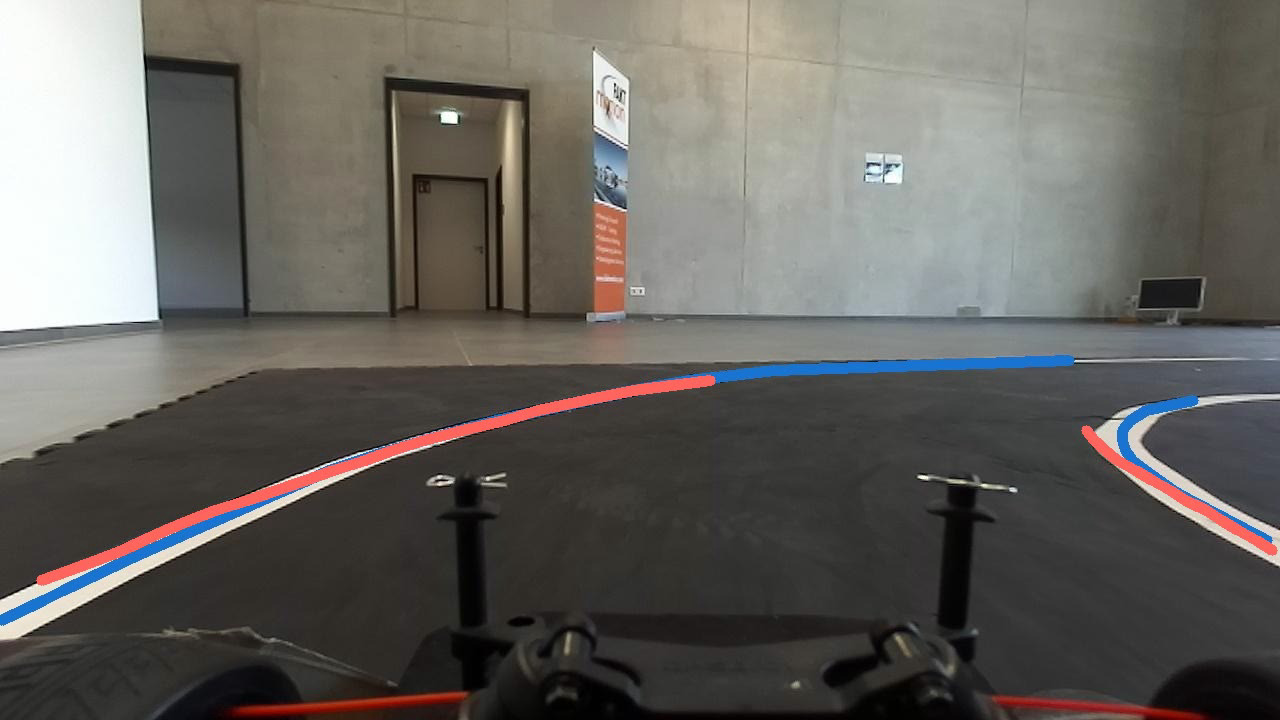} & 
			\includegraphics[width=.18\linewidth,valign=m]{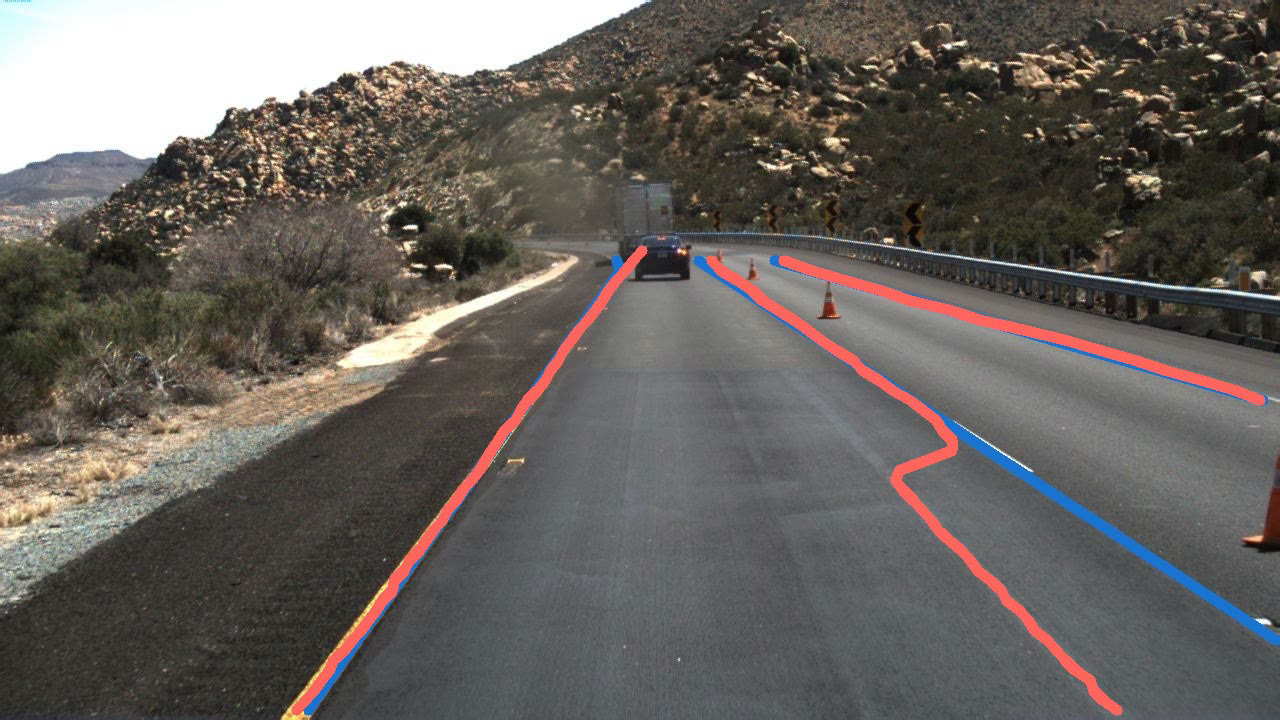} &
			\includegraphics[width=.18\linewidth,valign=m]{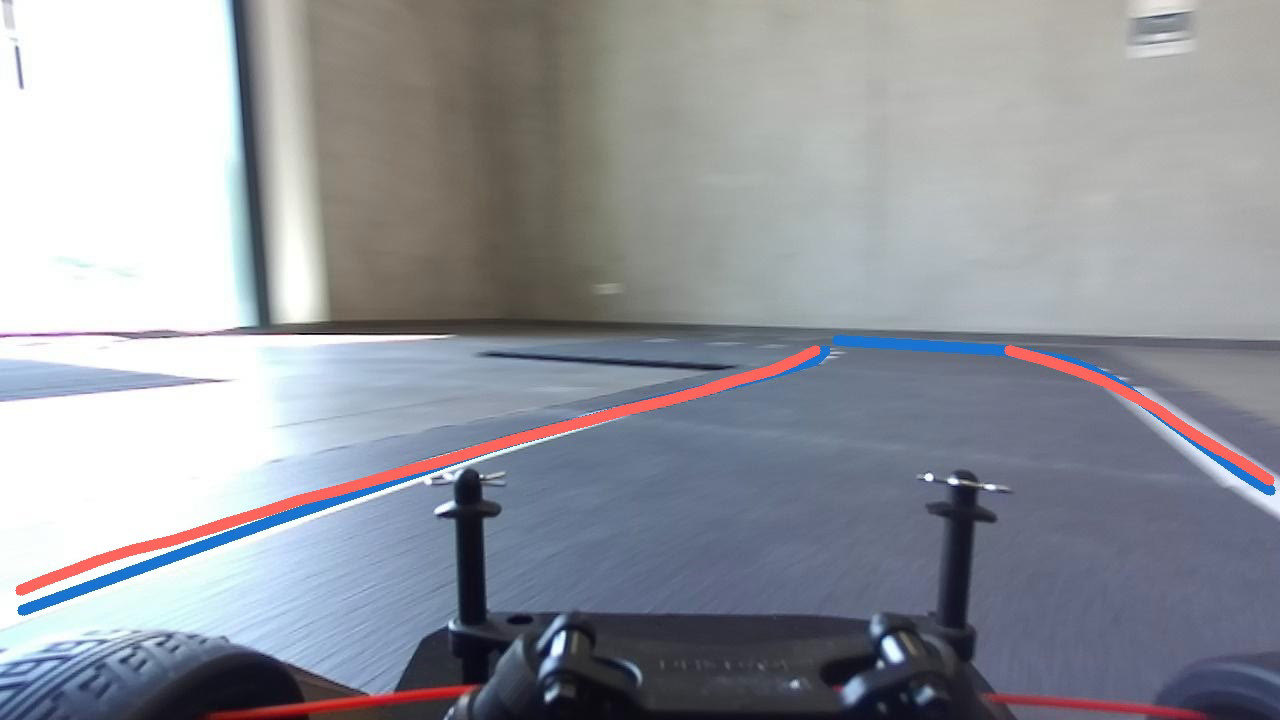} & \includegraphics[width=.18\linewidth,valign=m]{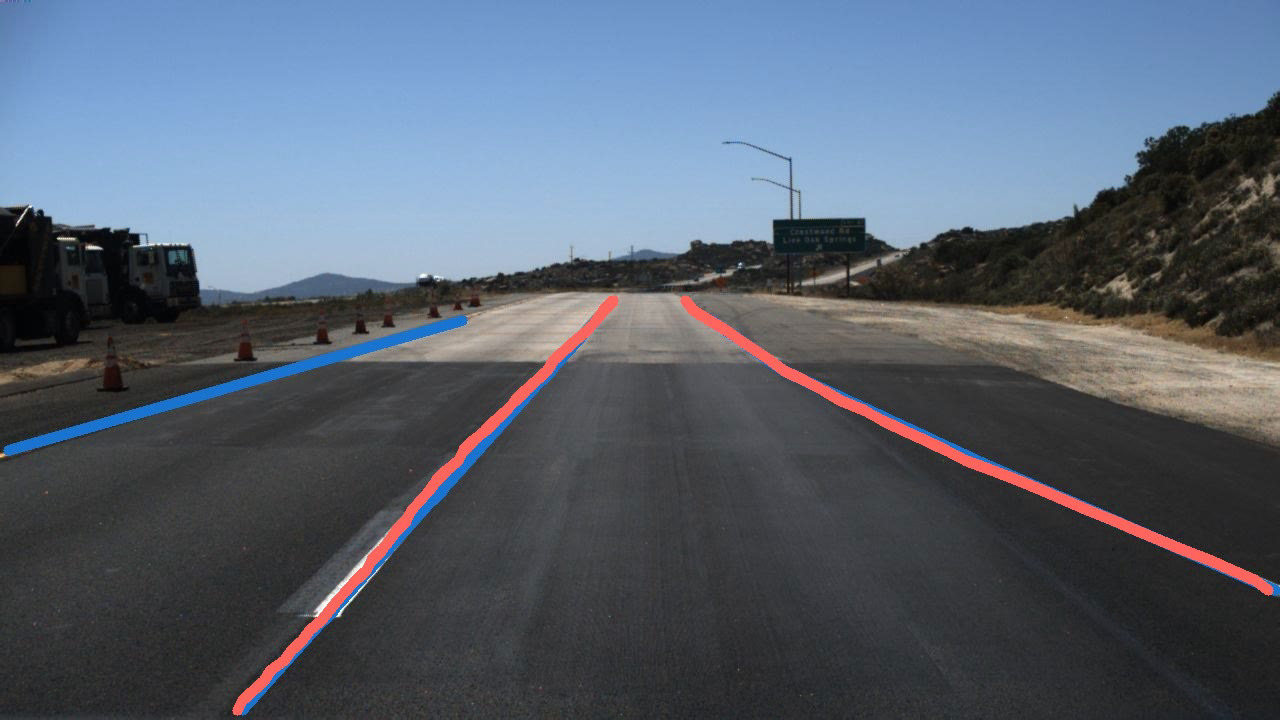}\\
			%
			\textbf{ADDA} & 
			\includegraphics[width=.18\linewidth,valign=m]{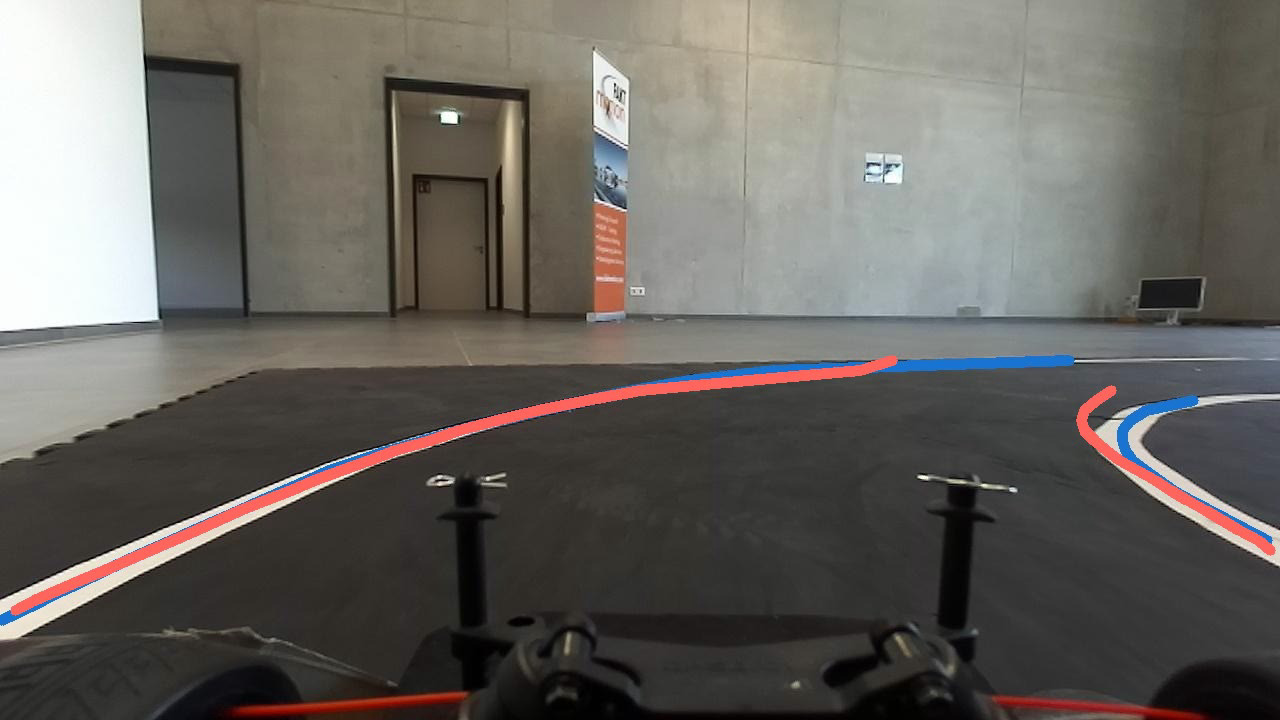} & 
			\includegraphics[width=.18\linewidth,valign=m]{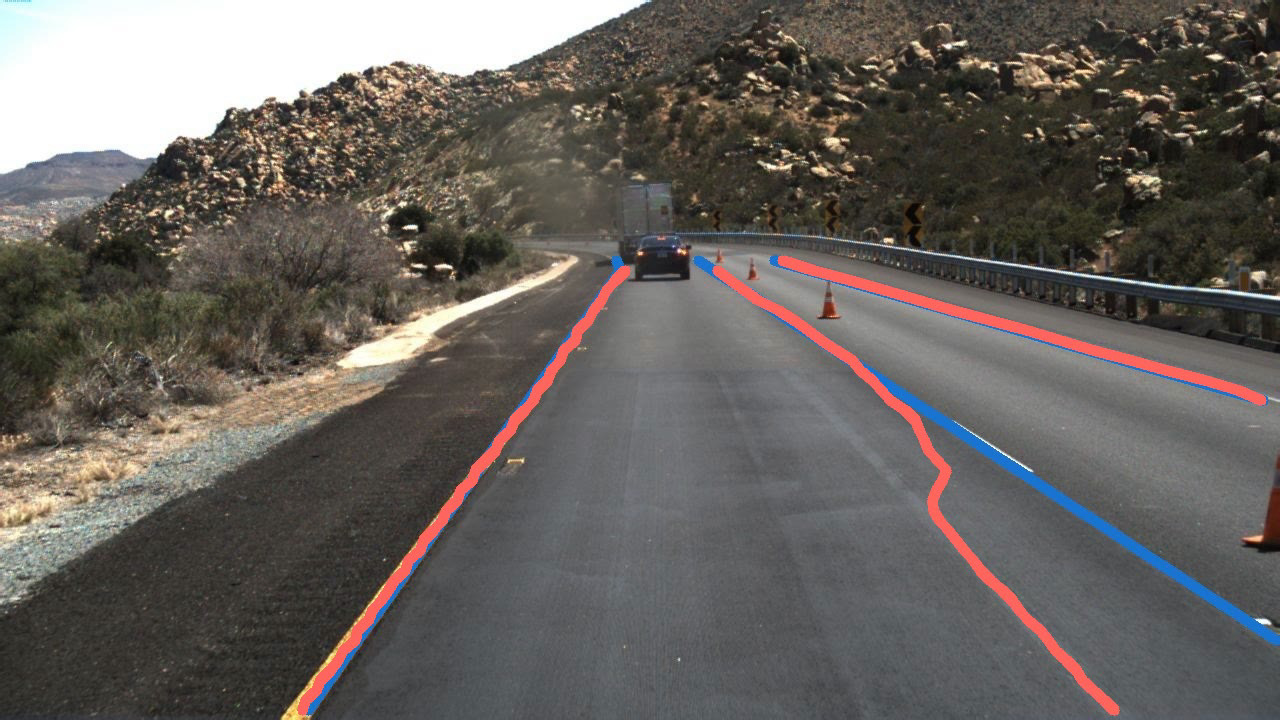} &
			\includegraphics[width=.18\linewidth,valign=m]{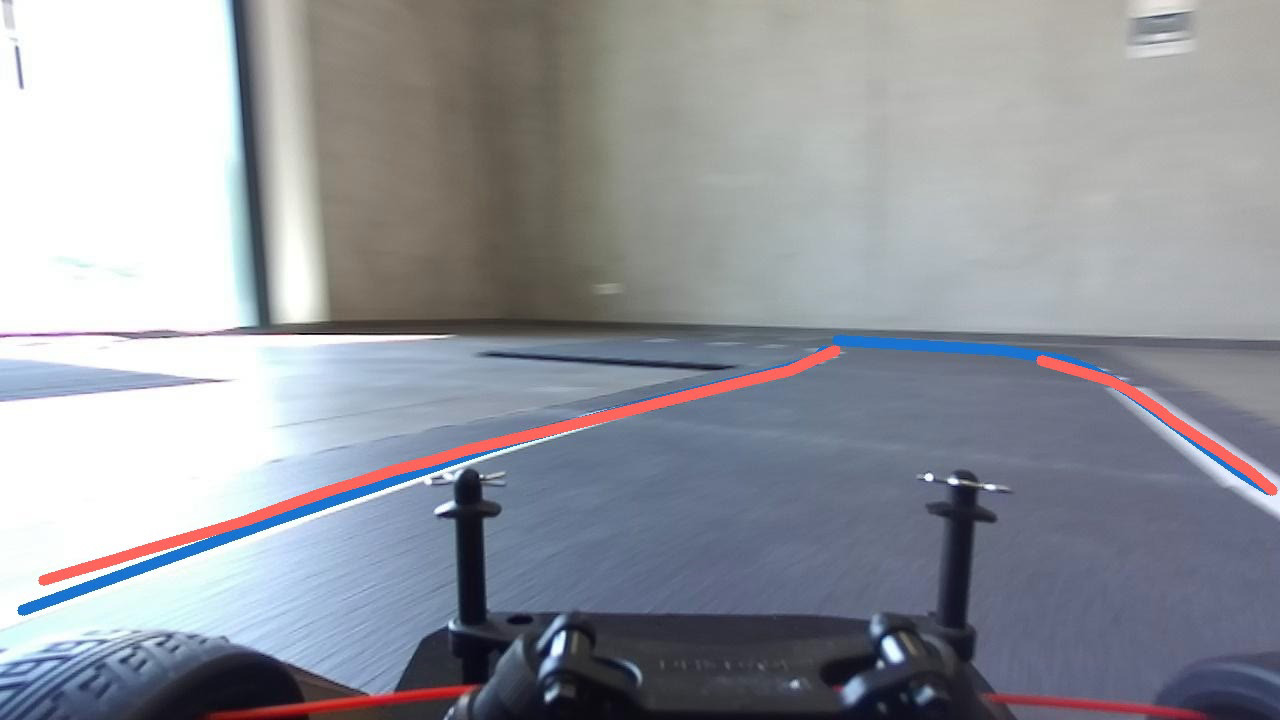} & \includegraphics[width=.18\linewidth,valign=m]{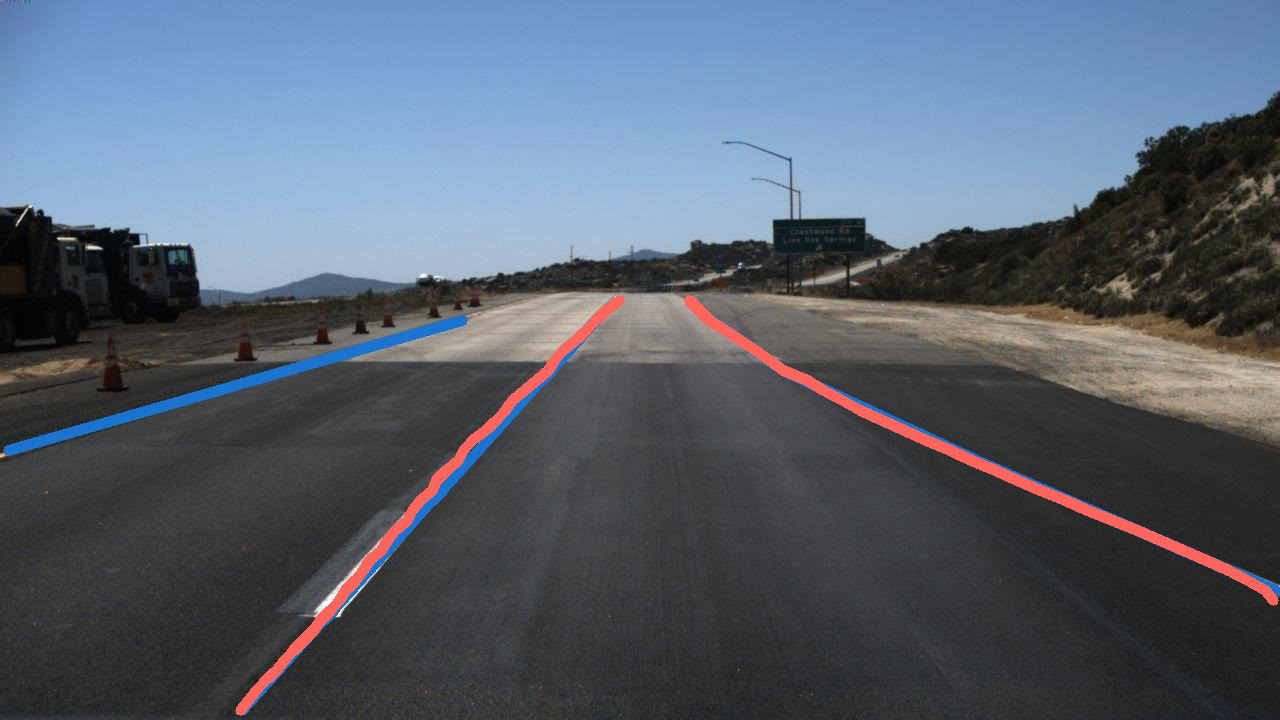}\\
			%
			\textbf{SGADA} & 
			\includegraphics[width=.18\linewidth,valign=m]{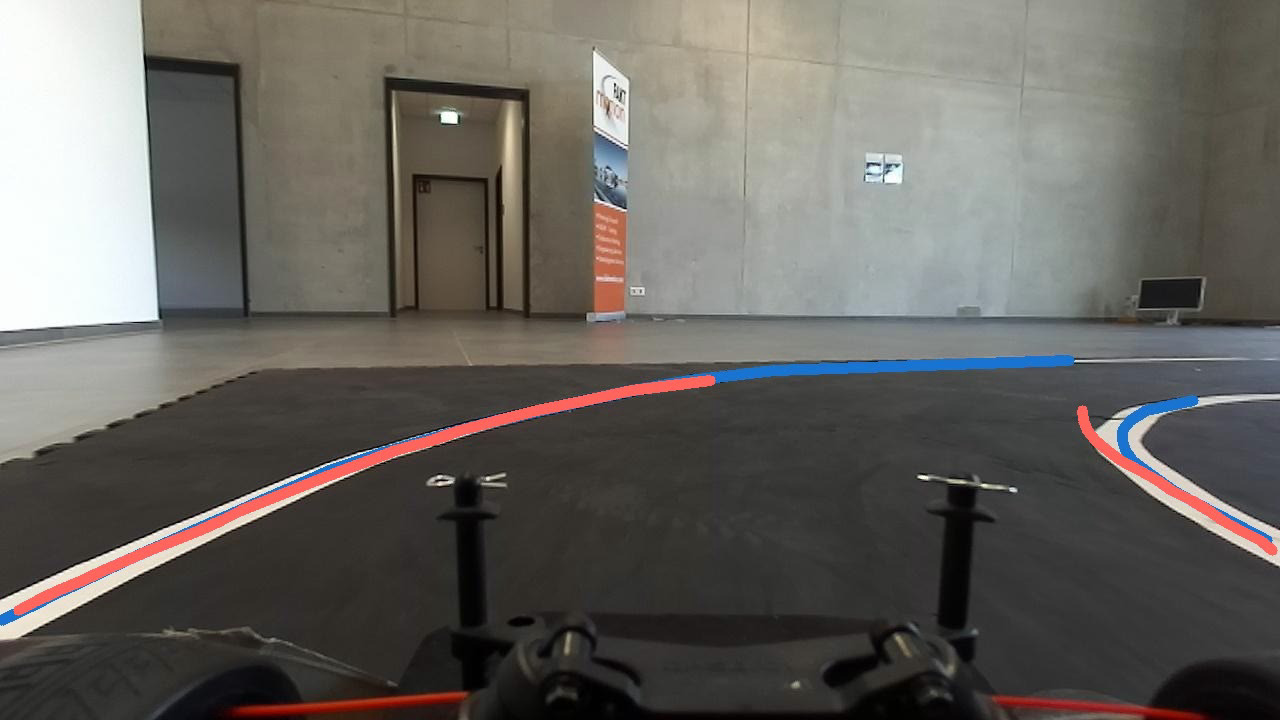} & 
			\includegraphics[width=.18\linewidth,valign=m]{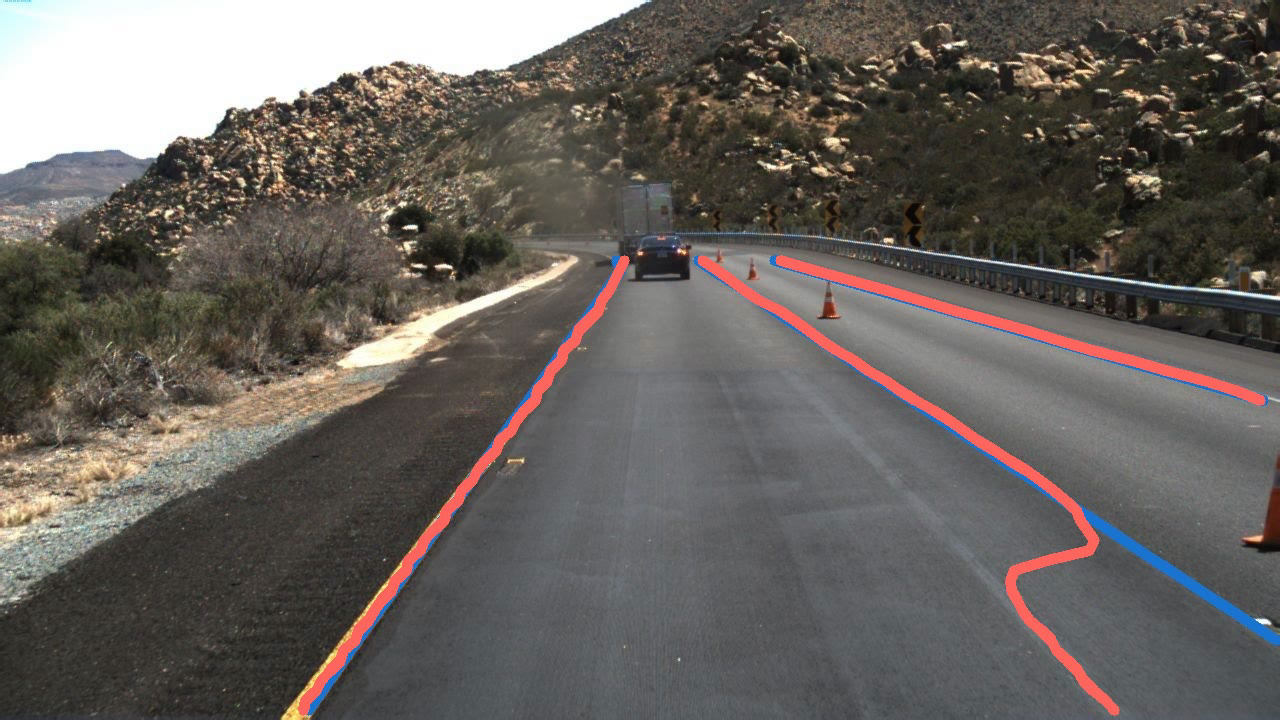} &
			\includegraphics[width=.18\linewidth,valign=m]{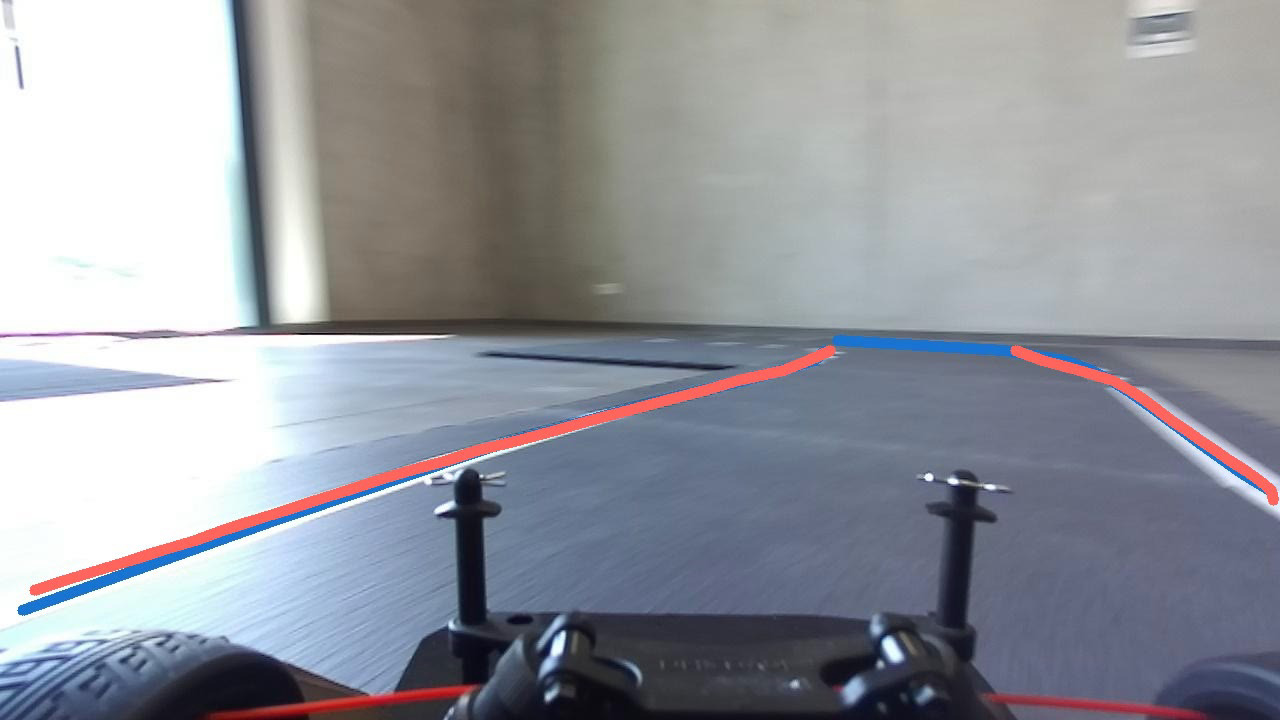} & \includegraphics[width=.18\linewidth,valign=m]{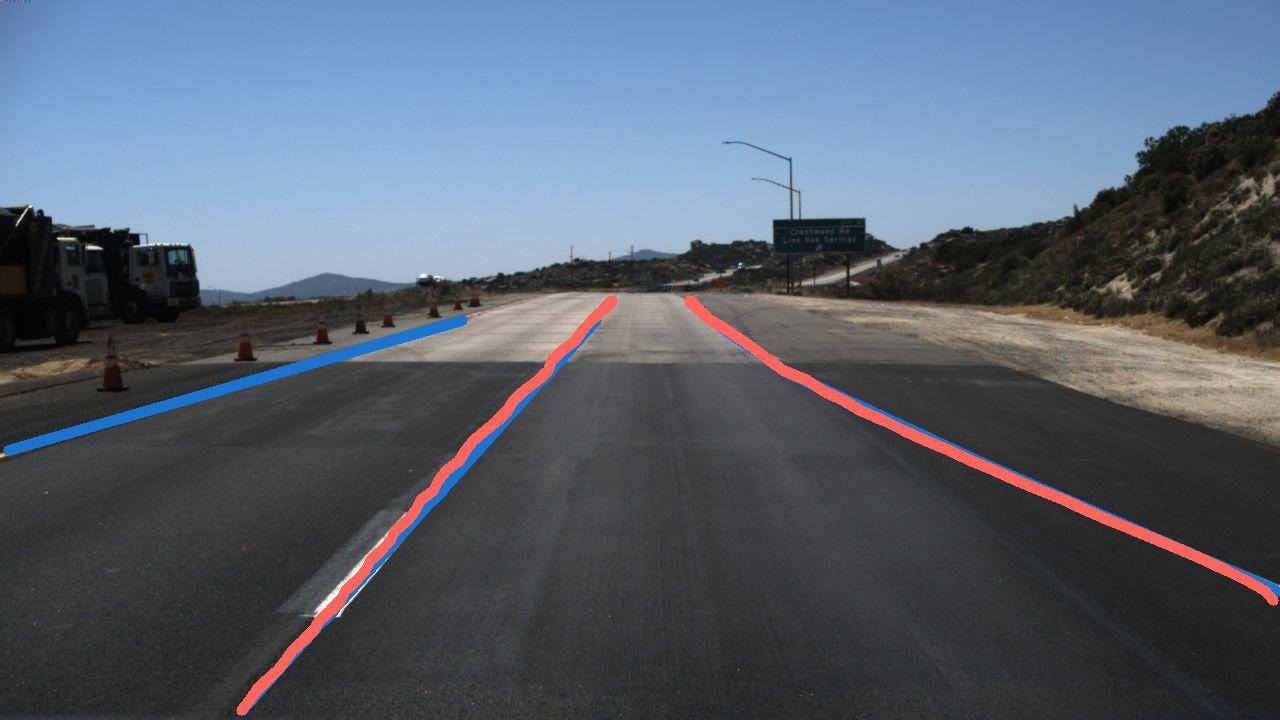}\\
			%
			\textbf{SGPCS} & 
			\includegraphics[width=.18\linewidth,valign=m]{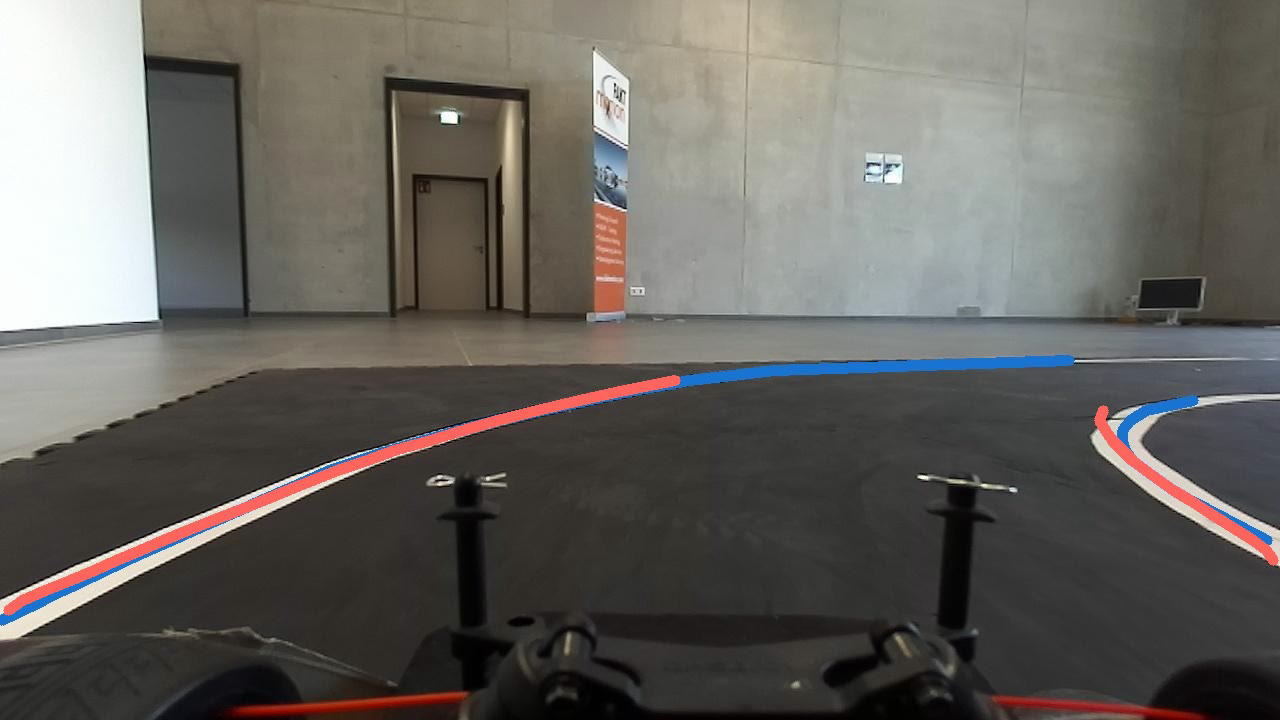} & \includegraphics[width=.18\linewidth,valign=m]{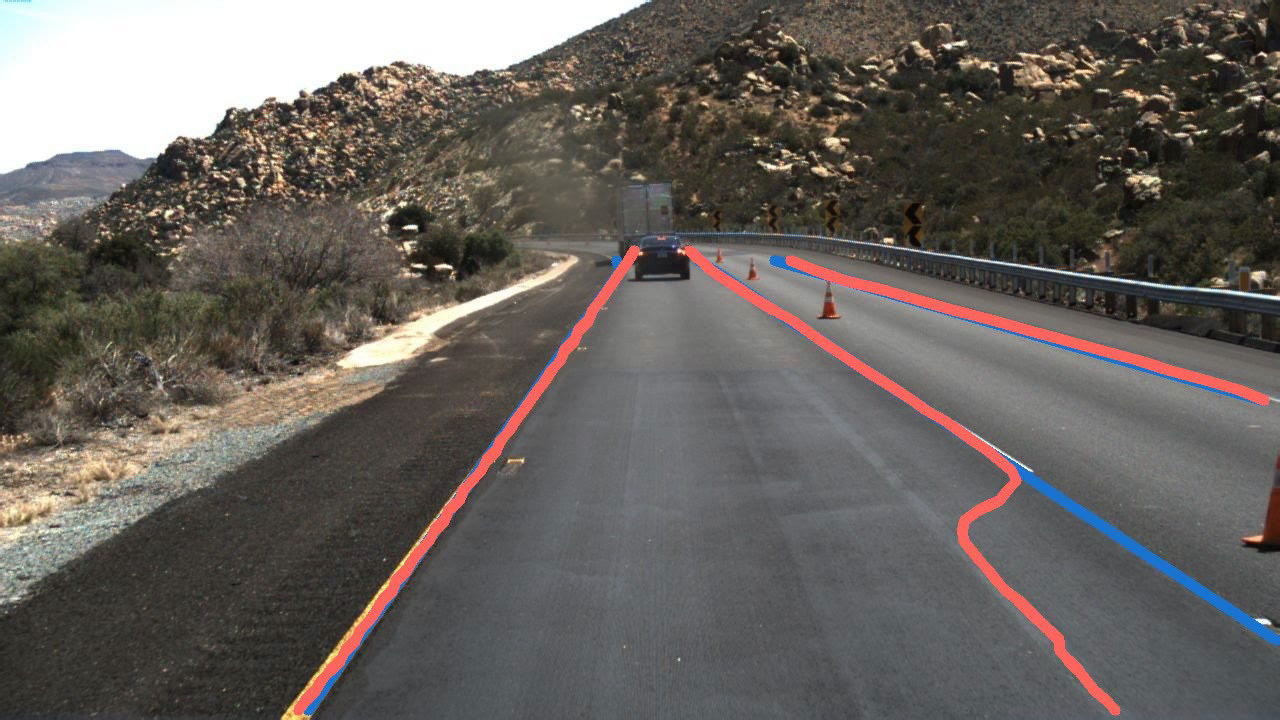} &
			\includegraphics[width=.18\linewidth,valign=m]{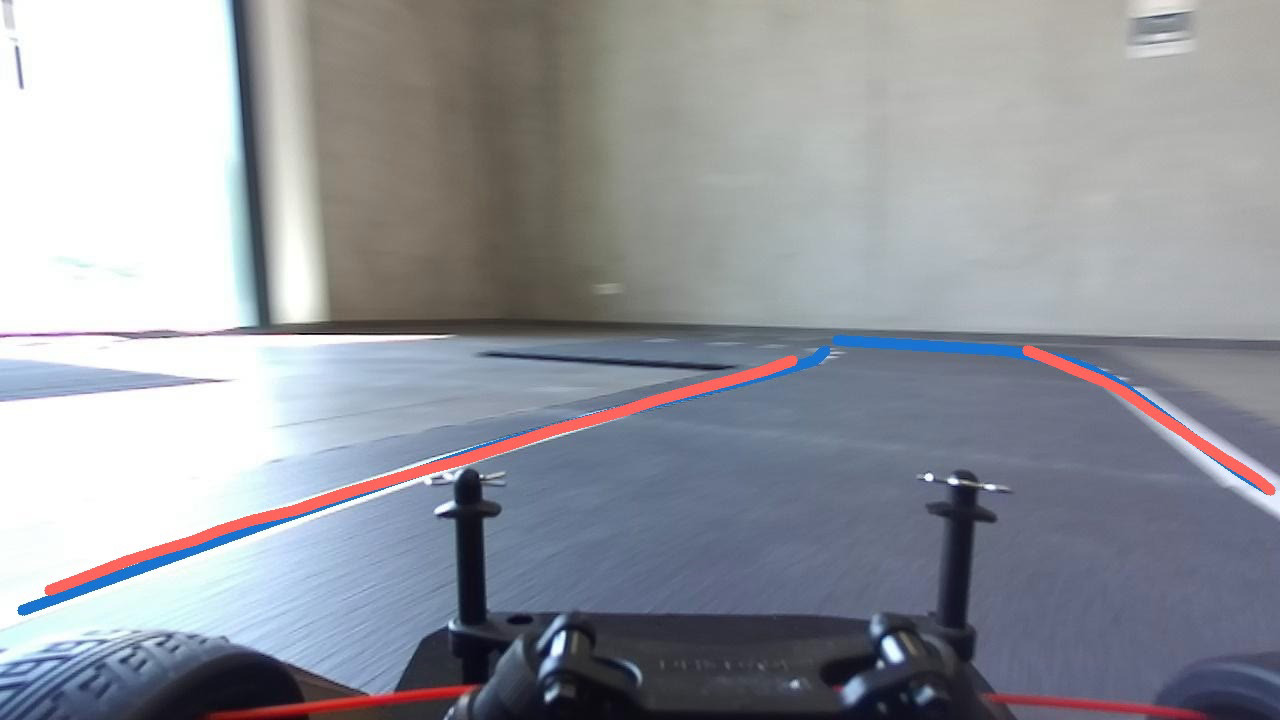} & \includegraphics[width=.18\linewidth,valign=m]{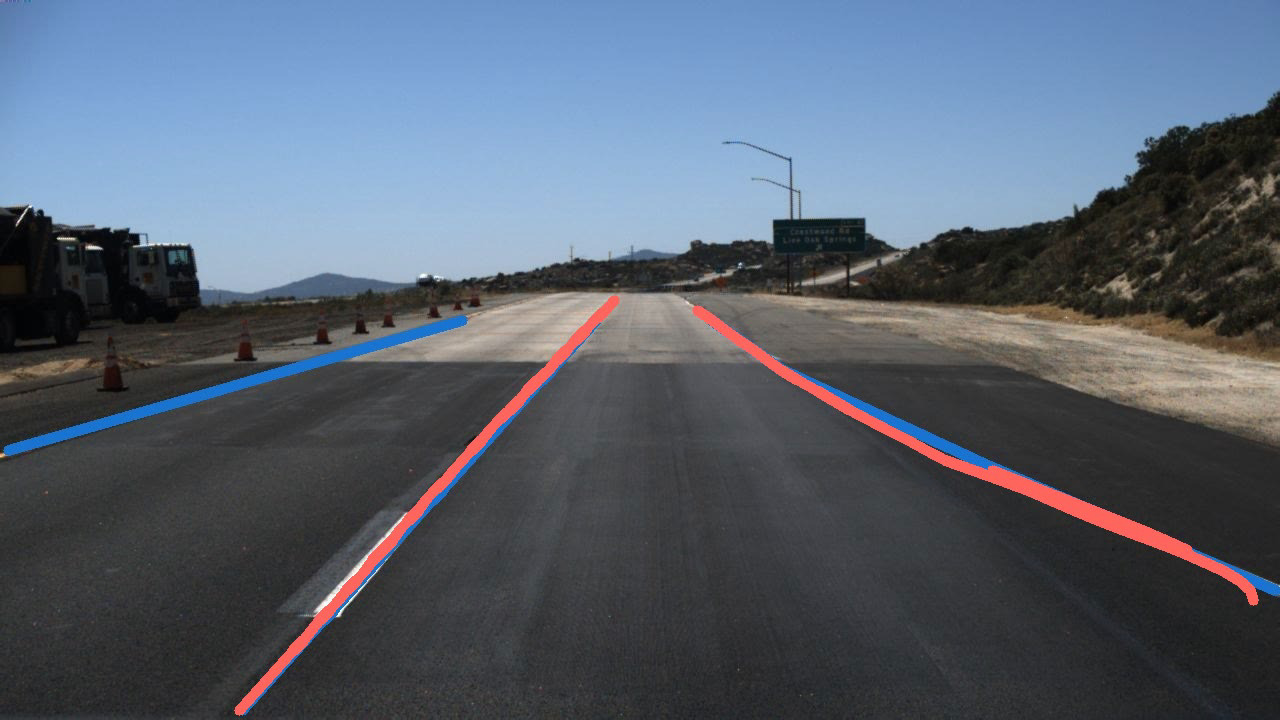}\\
			%
			\textbf{UFLD-TO} & 
			\includegraphics[width=.18\linewidth,valign=m]{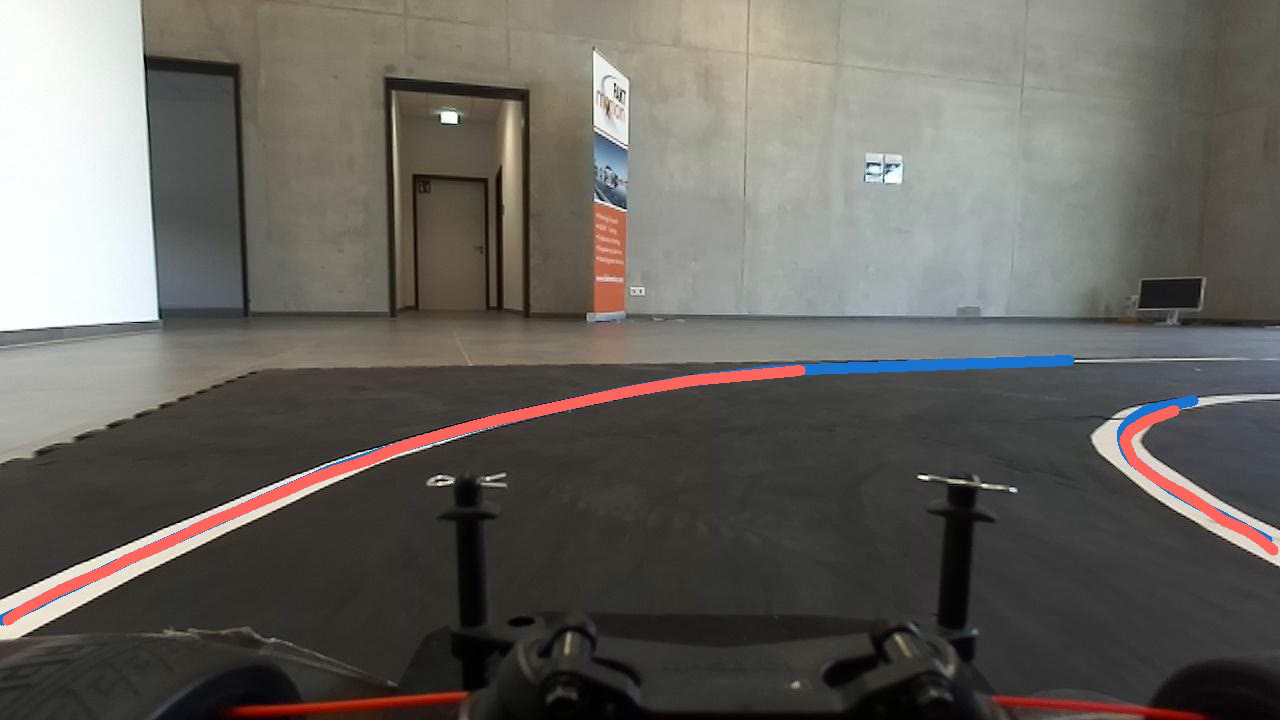} & \includegraphics[width=.18\linewidth,valign=m]{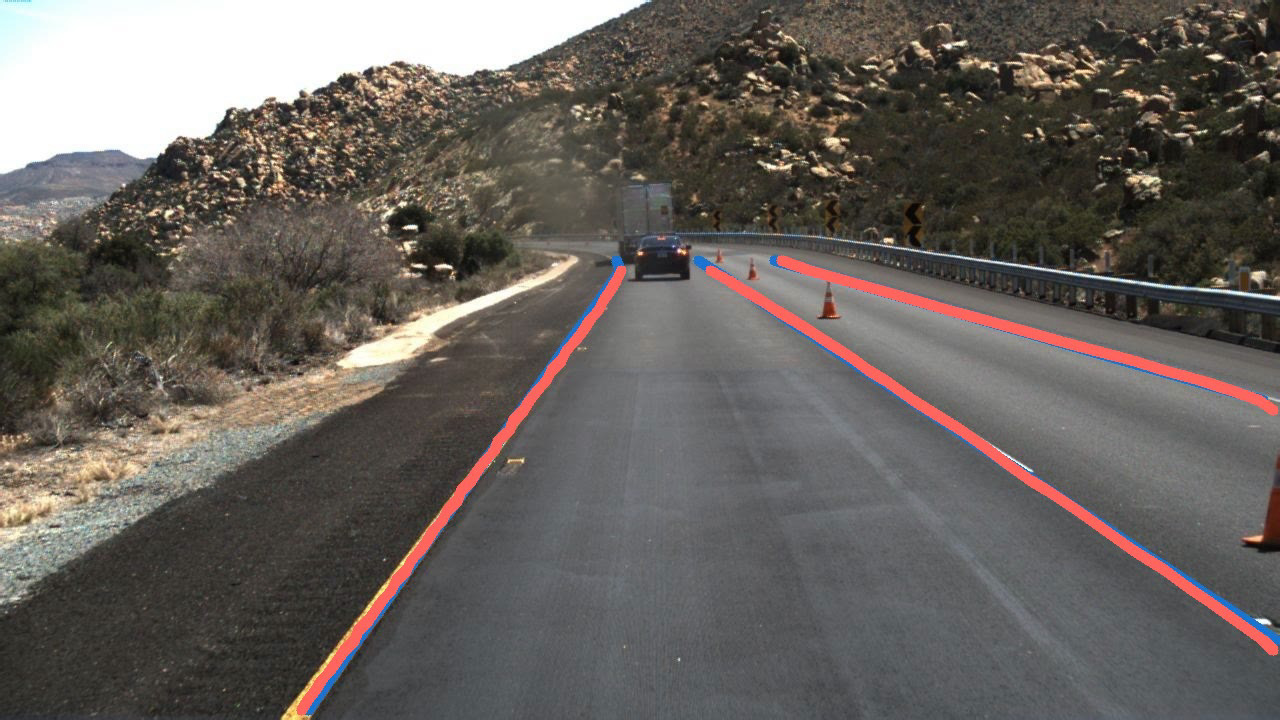} &
			\includegraphics[width=.18\linewidth,valign=m]{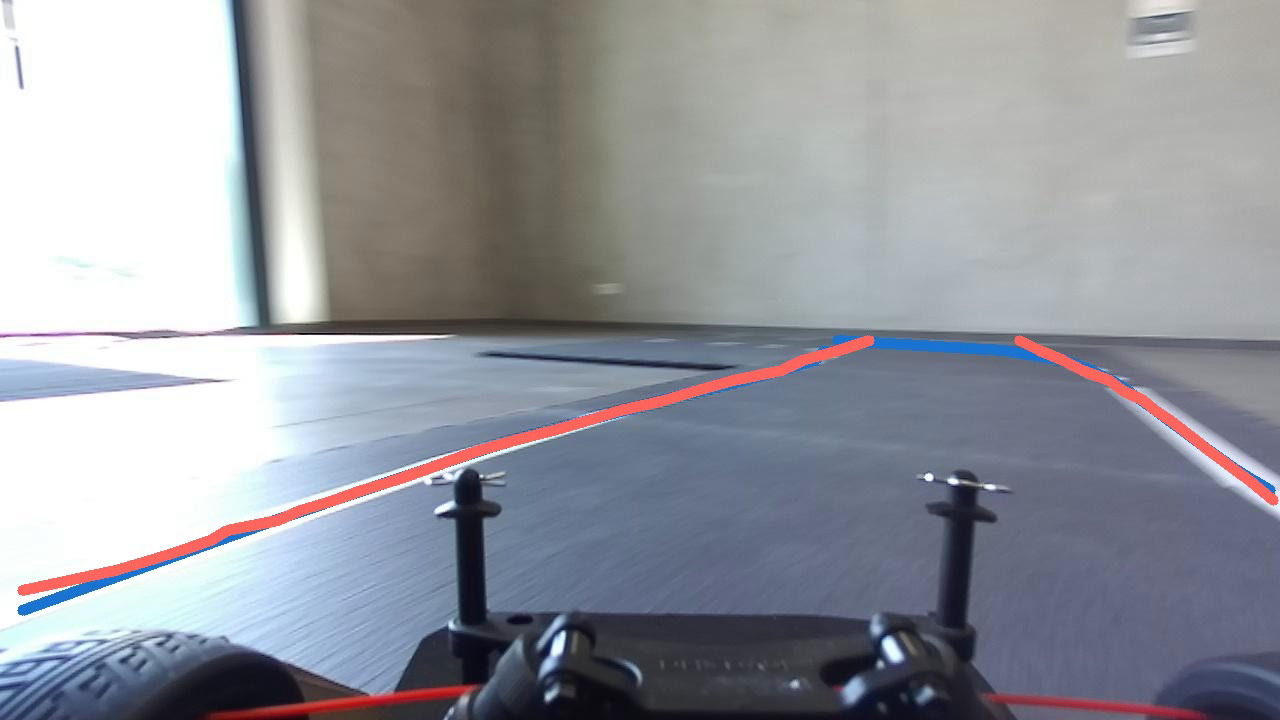} & \includegraphics[width=.18\linewidth,valign=m]{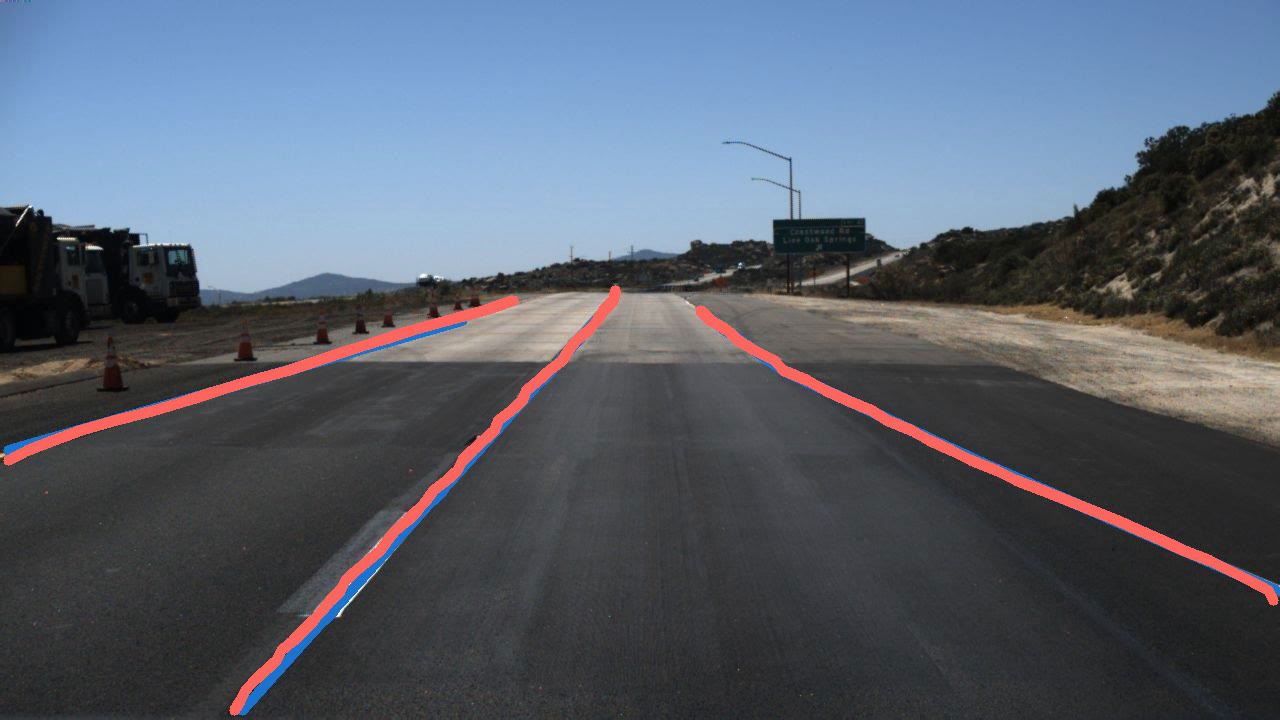}\\
		\end{tabular}
		\caption{Qualitative results of target domain predictions. Ground truth lane annotations are marked in blue, predictions in red.}
		\label{fig:qualitative_resuls}
	\end{figure}

	\textbf{Qualitative Evaluation.}
	We use t-SNE \cite{van2008visualizing} to visualize the features of the features encoders for the source and target domains of MuLane in \autoref{fig:TSNE_plot_mulane}. t-SNE visualizations of MoLane and TuLane can be found in the Appendix. In accordance with the quantitative results, we observe only a slight adaptation of the source and target domains features for ADDA, SGADA, and SGPCS compared to the supervised baseline UFLD-SO. Consequently, the examined well-known domain adaptation methods have no significant effect on feature alignment. In addition, we show results from the evaluated methods in \autoref{fig:qualitative_resuls} and observe that the models are able to predict target domain lane annotations in many cases but are not able to achieve comparable results to the supervised baseline (UFLD-TO).
	
	In summary, we find quantitatively and qualitatively that the examined domain adaptation methods do not significantly improve the performance of lane detection and feature adaptation. For this reason, we believe that the proposed benchmark could facilitate the exploration of new domain adaptation methods to overcome these problems. 
	
	\section{Conclusion}
	\label{sec:Conclusion}
	We present CARLANE, the first UDA benchmark for lane detection. CARLANE was recorded in three domains and consists of three datasets: the single-target datasets MoLane and TuLane and the multi-target dataset MuLane, which is a balanced combination of both. Based on the UFLD model, we conducted experiments with different UDA methods on CARLANE and found that the selected methods are able to adapt the model to target domains slightly and consistently. However, none of the methods achieve comparable results to the supervised baselines. The most significant performance differences are noticeable in the high false positive and false negative rates of the UDA methods compared to the target-only baselines, which is even more pronounced in the MuLane multi-target task. These false-positive and false-negative rates can negatively impact autonomous driving functions since they represent misidentified and missing lanes. Furthermore, as shown in the t-SNE plots of \autoref{fig:TSNE_plot_mulane}, the examined well-known domain adaptation methods have no significant effect on feature alignment. The current difficulties of the examined UDA methods to adequately align the source and target domains confirm the need for the proposed CARLANE benchmark. We believe that CARLANE eases the development and comparison of UDA methods for lane detection. In addition, we open-source all tools for dataset creation and labeling and hope that CARLANE facilitates future research in these directions. 
	
	\textbf{Limitations.} One limitation of our work is that we only use a fixed set of track elements within our 1/8th scaled environment. These track elements represent only a limited number of distinct curve radii. Furthermore, neither buildings nor traffic signs exist in MoLane's model vehicle target domain.
	Moreover, the full-scale real-world target domain of TuLane is derived from TuSimple. TuSimple's data was predominantly collected under good and medium conditions and lacks variation in weather and time of day. In addition, we want to emphasize that collecting data for autonomous driving is still an ongoing effort and that datasets such as TuSimple do not cover all possible real-world driving scenarios to ensure safe, practical use. For the synthetically generated data, we limited ourselves to using existing CARLA maps without defining new simulation environments. Despite these limitations, CARLANE serves as a supportive dataset for further research in the field of UDA.
	\vspace{-2pt}
	
	\textbf{Ethical and Responsible Use.} Considering the limitations of our work, UDA methods trained on TuLane and MuLane should be tested with care and under the right conditions on a full-scale car. However, real-world testing with MoLane in the model vehicle domain can be carried out in a safe and controlled environment. Additionally, TuLane contains open-source images with unblurred license plates and people. This data should be treated with respect and in accordance with privacy policies. In general, our work contributes to the research in the field of autonomous driving, in which a lot of unresolved ethical and legal questions are still being discussed. The step-by-step testing possibility across three domains makes it possible for our benchmark to include an additional safety mechanism for real-world testing.
	
	\bibliographystyle{ieeetr}
	\bibliography{refs}
	
	\newpage
	
	\section*{Checklist}
	\begin{enumerate}
		
		\item For all authors...
		\begin{enumerate}
			\item Do the main claims made in the abstract and introduction accurately reflect the paper's contributions and scope?
			\answerYes{See Section~\ref{sec:Conclusion}.}
			\item Did you describe the limitations of your work?
			\answerYes{See Section~\ref{sec:Conclusion}.}
			\item Did you discuss any potential negative societal impacts of your work?
			\answerYes{See Section~\ref{sec:Conclusion}.}
			\item Have you read the ethics review guidelines and ensured that your paper conforms to them?
			\answerYes{}
		\end{enumerate}
		
		\item If you are including theoretical results...
		\begin{enumerate}
			\item Did you state the full set of assumptions of all theoretical results?
			\answerNA{}
			\item Did you include complete proofs of all theoretical results?
			\answerNA{}
		\end{enumerate}
		
		\item If you ran experiments (e.g. for benchmarks)...
		\begin{enumerate}
			\item Did you include the code, data, and instructions needed to reproduce the main experimental results (either in the supplemental material or as a URL)?
			\answerYes{See Abstract, Section~\ref{sec:ImplementationDetails} and the supplemental material.}
			\item Did you specify all the training details (e.g., data splits, hyperparameters, how they were chosen)?
			\answerYes{See Section~\ref{sec:CARLANE}, Section~\ref{sec:Experiments}, Section~\ref{sec:ImplementationDetails}, Table~\ref{table:Dataset overview}, Table~\ref{table:Hyperpparameters} and the supplemental material.}
			\item Did you report error bars (e.g., with respect to the random seed after running experiments multiple times)?
			\answerYes{Table~\ref{table:Results}.}
			\item Did you include the total amount of compute and the type of resources used (e.g., type of GPUs, internal cluster, or cloud provider)?
			\answerYes{Section~\ref{sec:ImplementationDetails}.}
		\end{enumerate}
		
		\item If you are using existing assets (e.g., code, data, models) or curating/releasing new assets...
		\begin{enumerate}
			\item If your work uses existing assets, did you cite the creators?
			\answerYes{We cited the TuSimple dataset \cite{TuSimple2017}.}
			\item Did you mention the license of the assets?
			\answerYes{See Section~\ref{sec:CARLANE}.}
			\item Did you include any new assets either in the supplemental material or as a URL?
			\answerYes{}
			\item Did you discuss whether and how consent was obtained from people whose data you're using/curating?
			\answerYes{TuSimple is open-source and licensed under the Apache License, Version 2.0 (January 2004).}
			\item Did you discuss whether the data you are using/curating contains personally identifiable information or offensive content?
			\answerYes{See Section~\ref{sec:Conclusion}}
		\end{enumerate}
		
		\item If you used crowdsourcing or conducted research with human subjects...
		\begin{enumerate}
			\item Did you include the full text of instructions given to participants and screenshots, if applicable?
			\answerNA{}
			\item Did you describe any potential participant risks, with links to Institutional Review Board (IRB) approvals, if applicable?
			\answerNA{}
			\item Did you include the estimated hourly wage paid to participants and the total amount spent on participant compensation?
			\answerNA{}
		\end{enumerate}
		
	\end{enumerate}

	\newpage
	\appendix
	
	\section{Appendix}
	
\subsection{Example Usage of the CARLANE Benchmark}
A Jupyter Notebook with a tutorial to read the datasets for usage in PyTorch can be found at \href{https://carlanebenchmark.github.io}{https://carlanebenchmark.github.io}. 

\subsection{Model Vehicle Description}
\begin{figure}[ht]
	\centering
	\includegraphics[width=.3\linewidth]{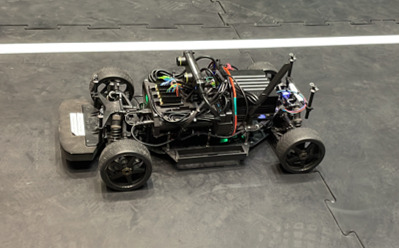}
	\caption{Picture of the 1/8th model vehicle we built to capture images in our 1/8th target domain.} 
	\label{fig:appendix_model_vehicle}
\end{figure}

In \autoref{fig:appendix_model_vehicle}, the self-built 1/8th model vehicle is shown, which we used to gather the images for the 1/8th scaled target domain. A NVIDIA Jetson AGX is the central computation unit powered by a separate Litionite Tanker Mini 25000mAh battery. For image collection, we utilize the software framework ROS Melodic and a Stereolabs ZEDM stereo camera with an integrated IMU. The camera is directly connected to the AGX and captures images with a resolution of $1280 \times 720$ pixels and a rate of $30$ FPS.

\subsection{Reproducibility of the Baselines}
To ensure reproducibility, we strictly follow UFLD \cite{qin2020ultra} and the corresponding UDA method for model architecture and hyperparameters. Thereby, we utilize UFLD as an encoder for the UDA method. We provide a detailed table of the tuned hyperparameters, architecture changes, and objectives in the main text. In addition, the trained weights of our baselines, their entire implementation, and the configuration files of our baselines are made publicly available at \href{https://carlanebenchmark.github.io}{https://carlanebenchmark.github.io}. 

\textbf{Initialization}. We initialize convolutional layer weights with kaiming normal and their biases with 0.0. Linear layer weights are initialized with normal (mean = 0.0, std = 0.01), batch normalization weights and biases are initialized with 1.0.

\subsection{Additional Results}
\begin{figure}[hb]
	\centering
	\small
	\begin{tabular}{c@{}c@{}c@{}c@{}c}
		UFLD-SO & DANN & ADDA & SGADA & SGPCS \\
		\includegraphics[width=.2\linewidth,valign=m]{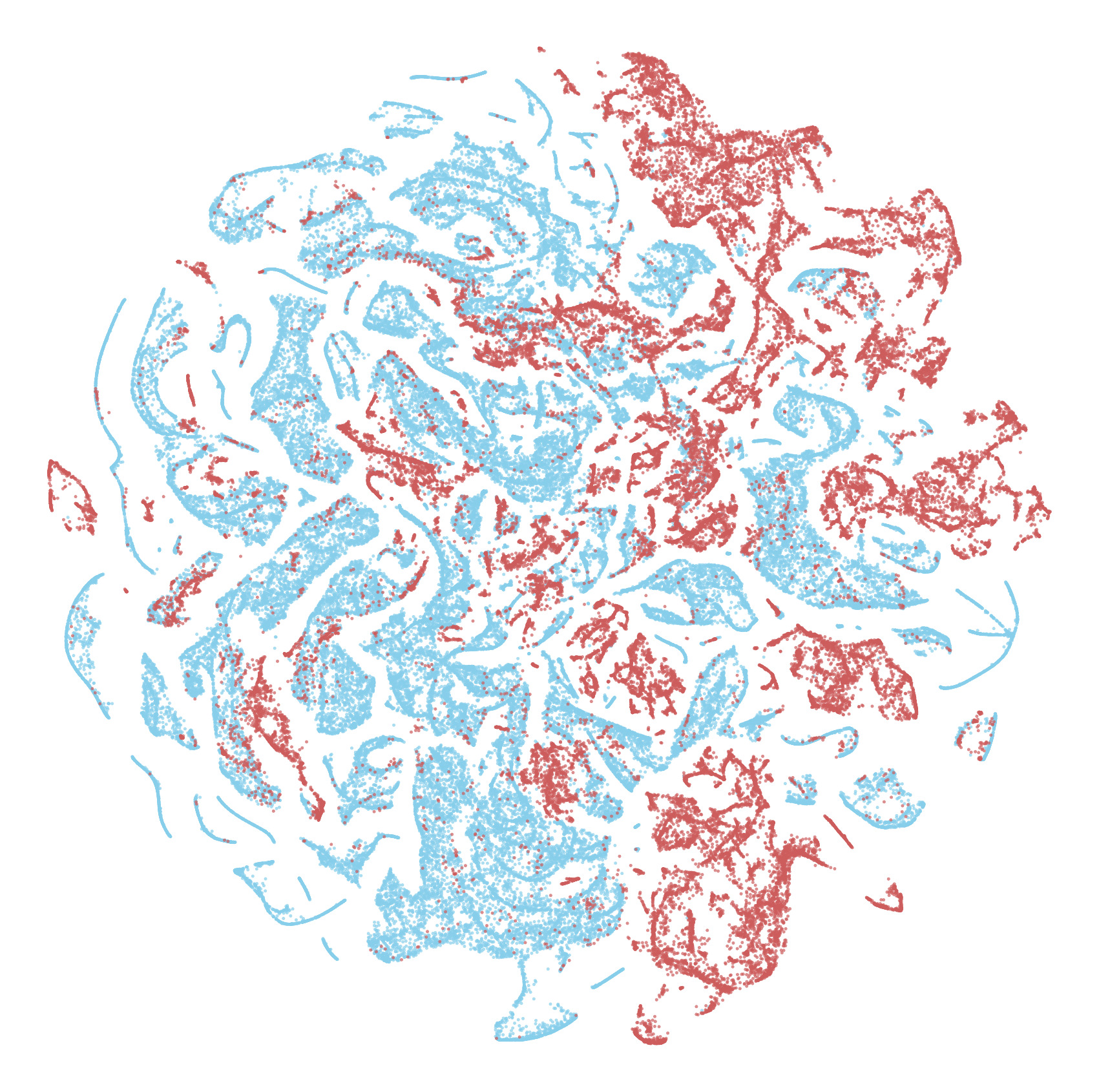} & \includegraphics[width=.2\linewidth,valign=m]{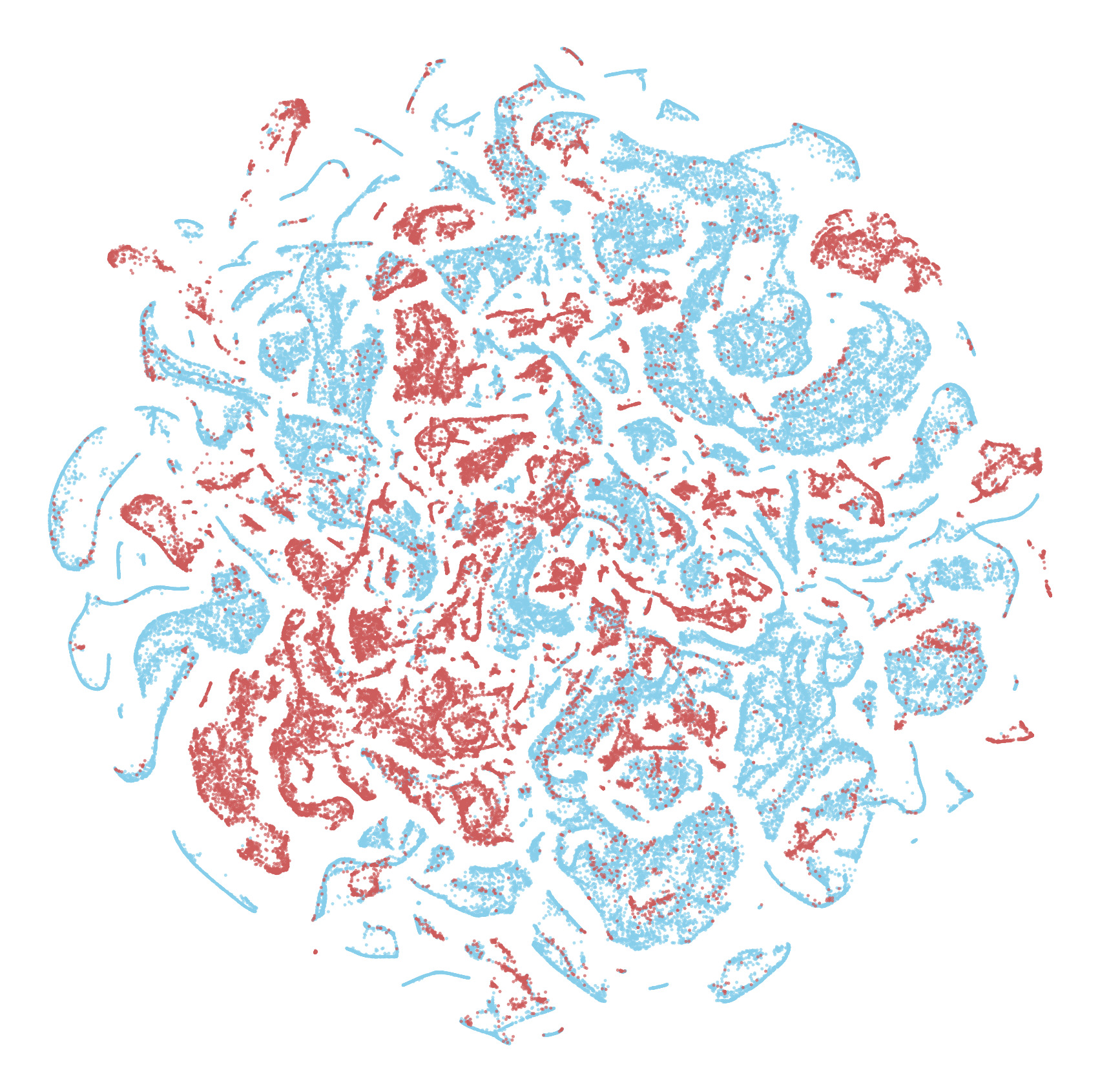} &
		\includegraphics[width=.2\linewidth,valign=m]{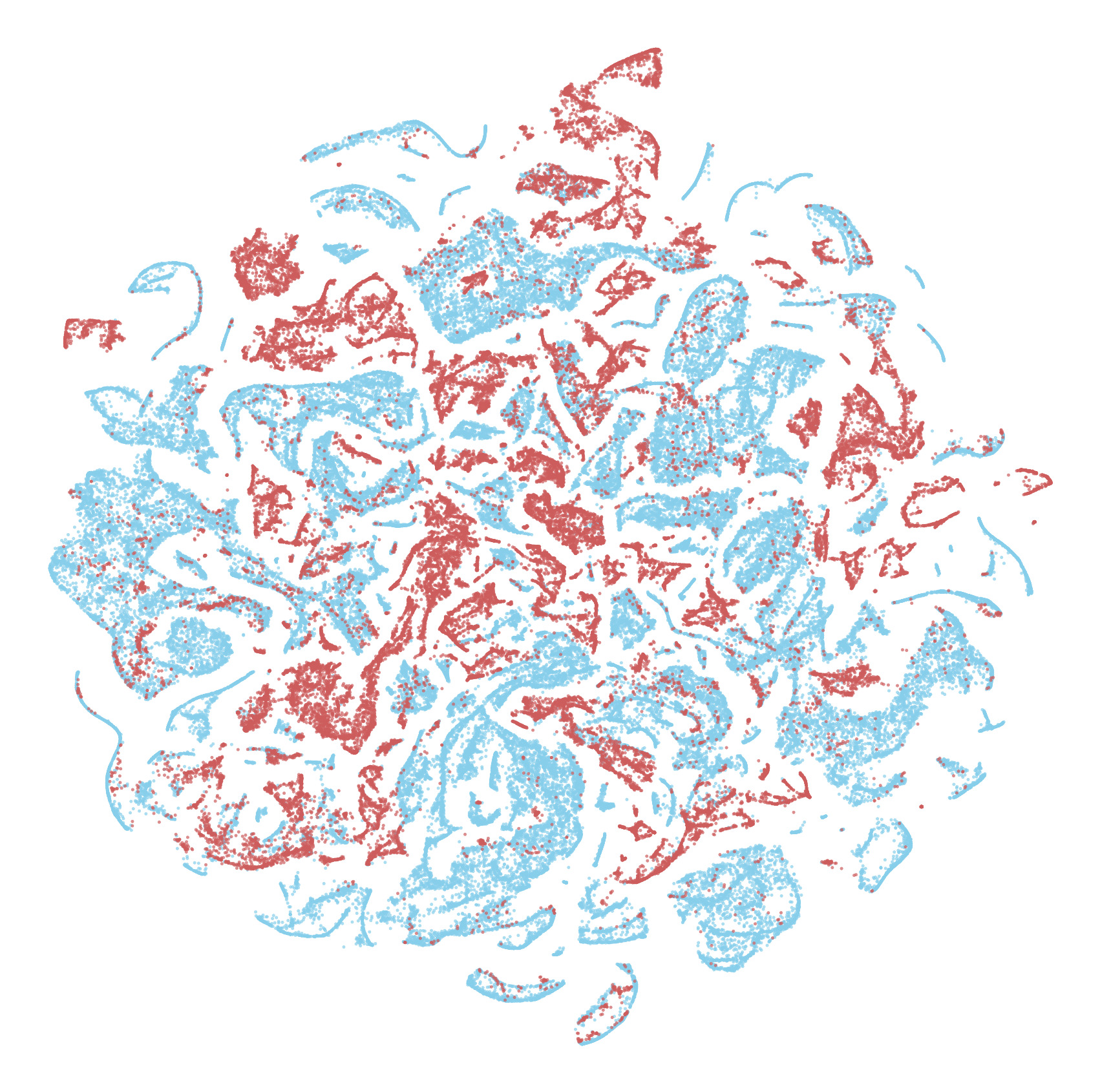} &
		\includegraphics[width=.2\linewidth,valign=m]{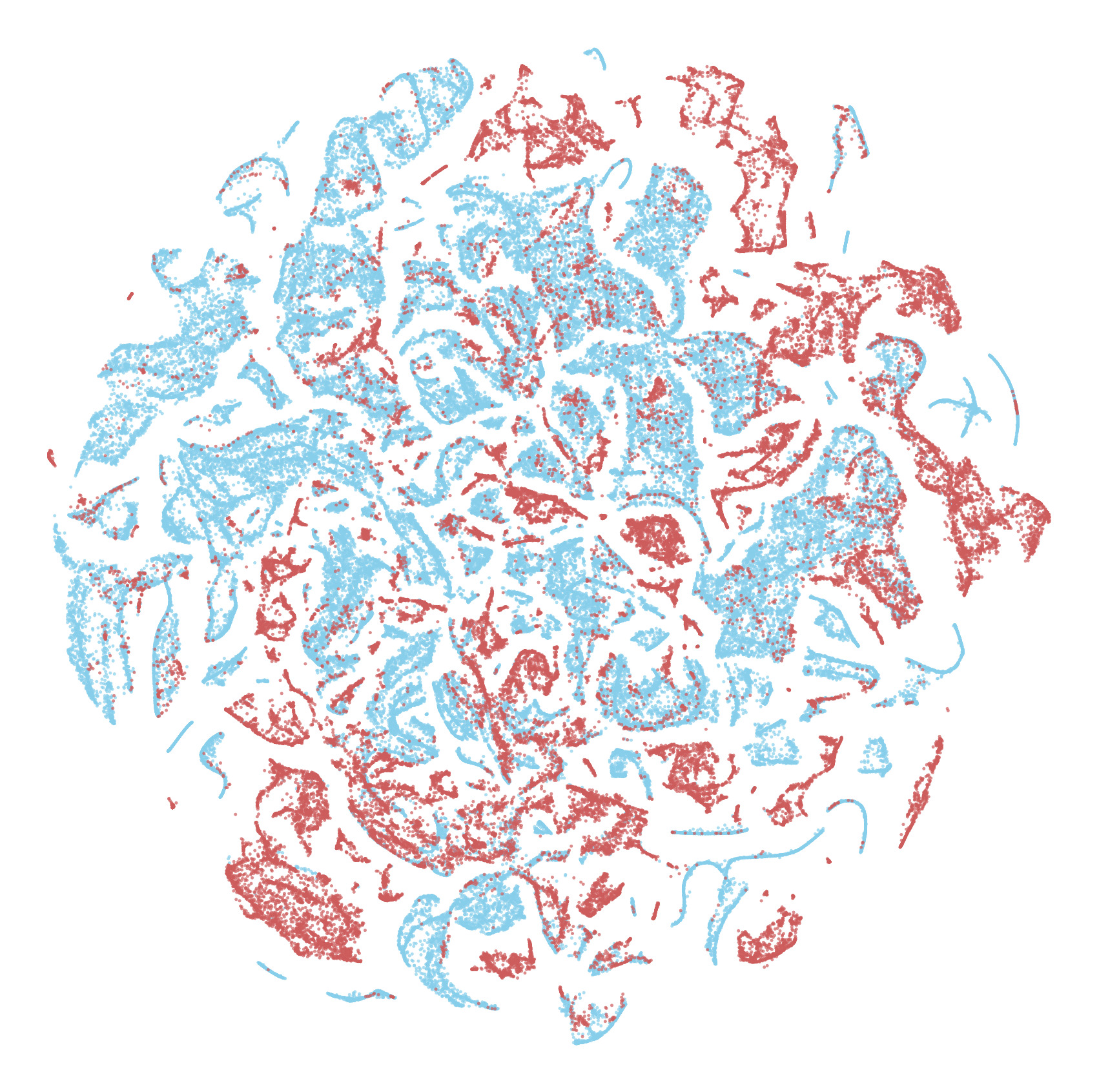} & \includegraphics[width=.2\linewidth,valign=m]{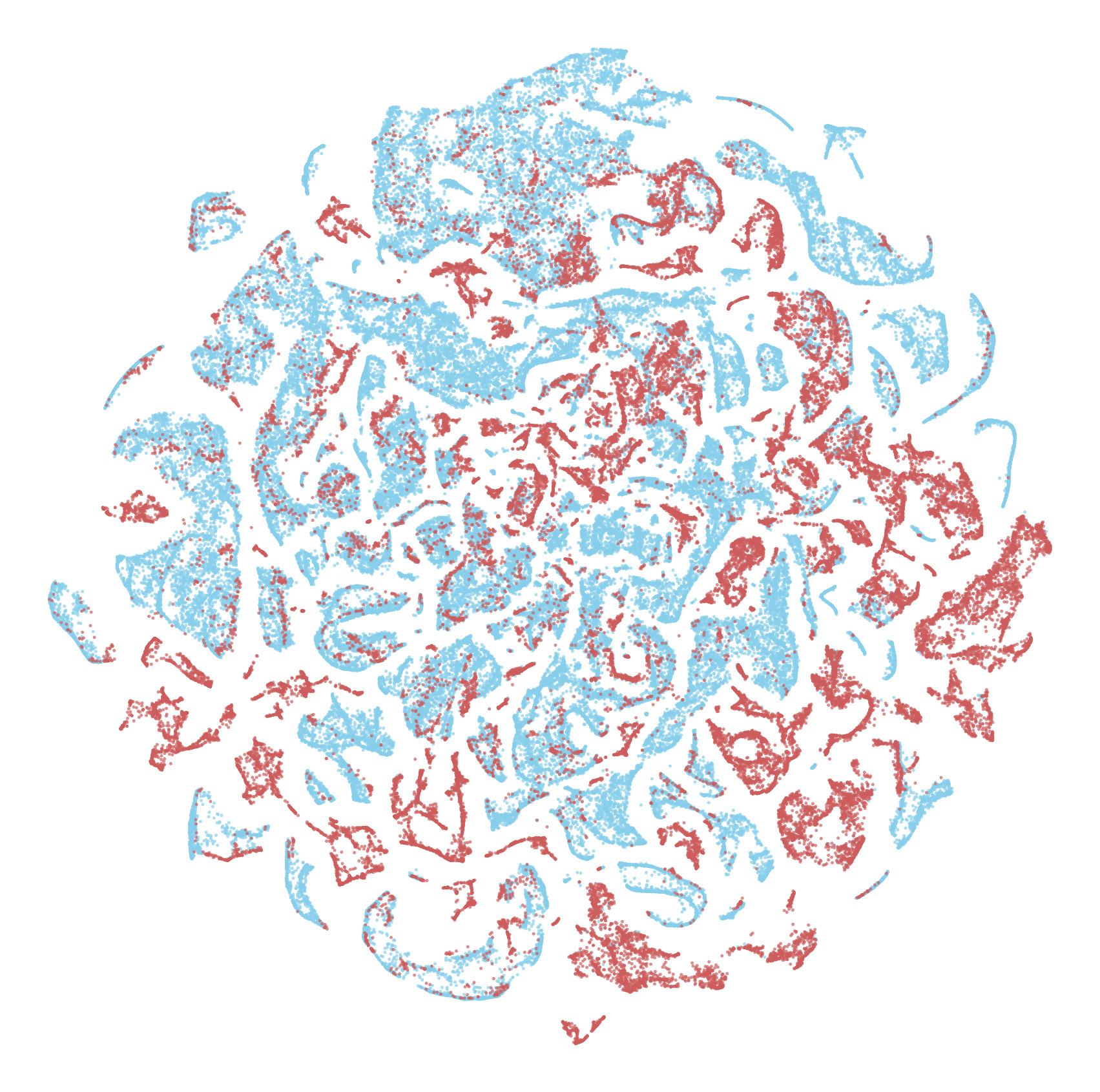}\\
		\includegraphics[width=.2\linewidth,valign=m]{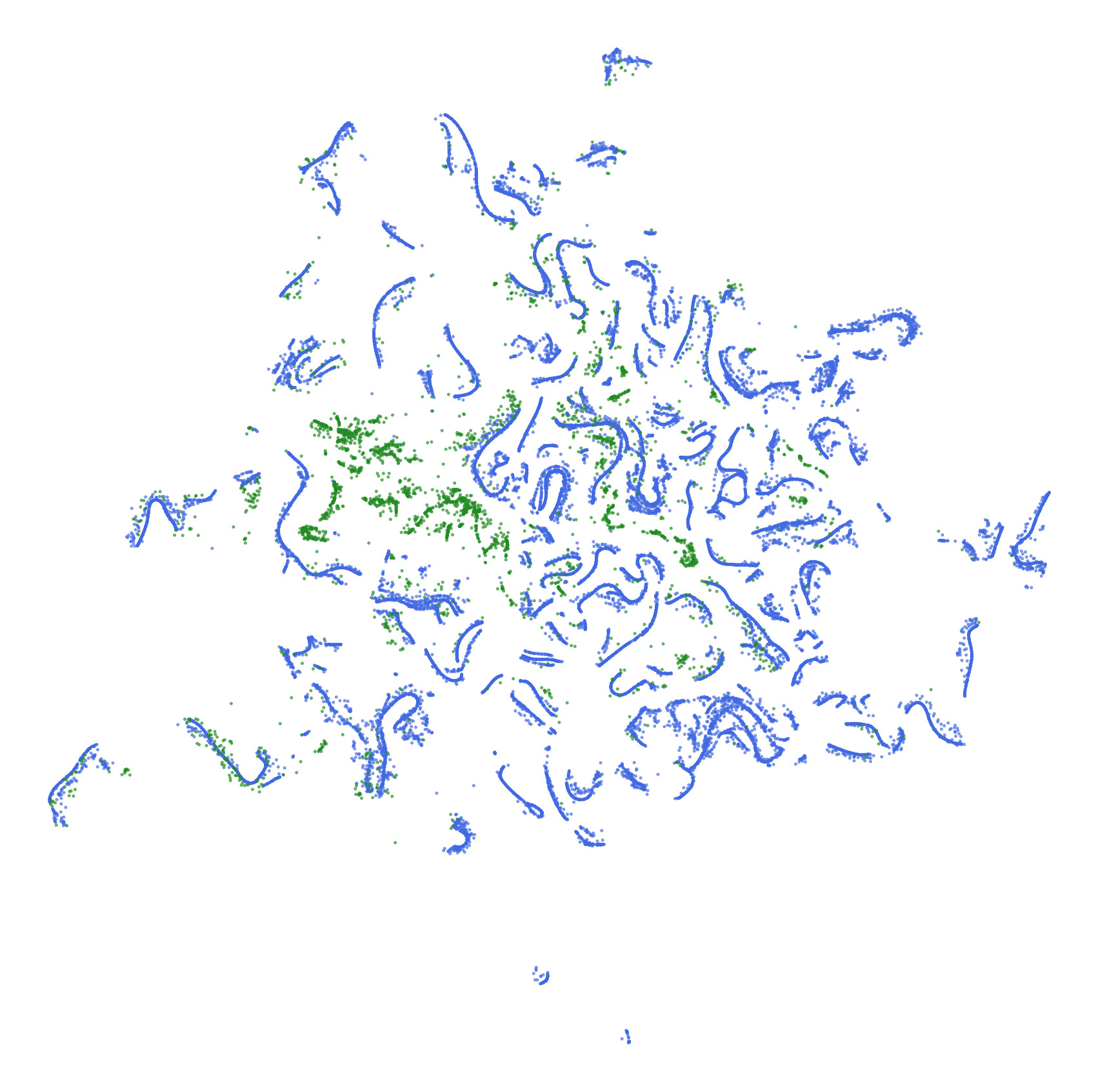} & \includegraphics[width=.2\linewidth,valign=m]{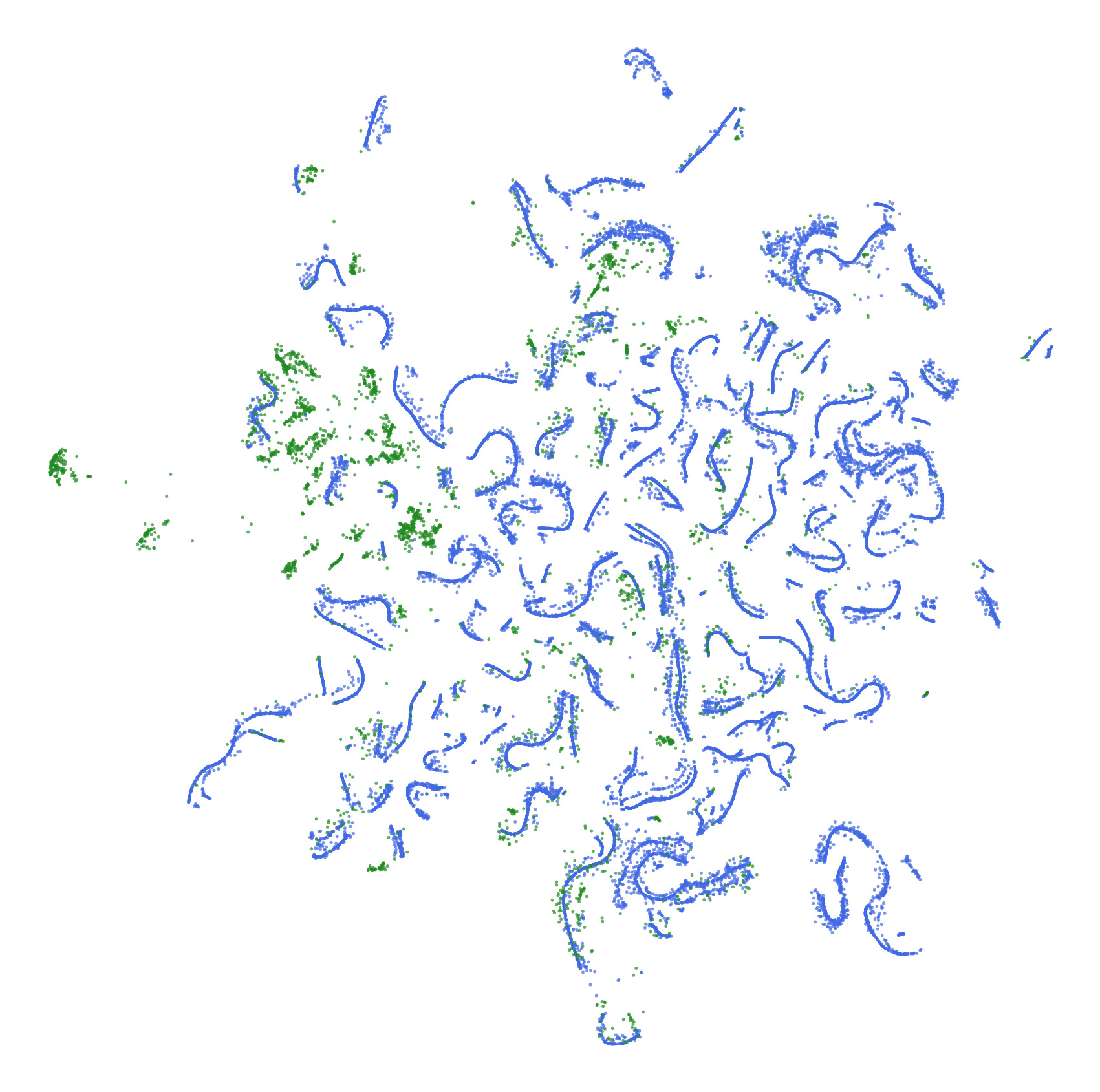} &
		\includegraphics[width=.2\linewidth,valign=m]{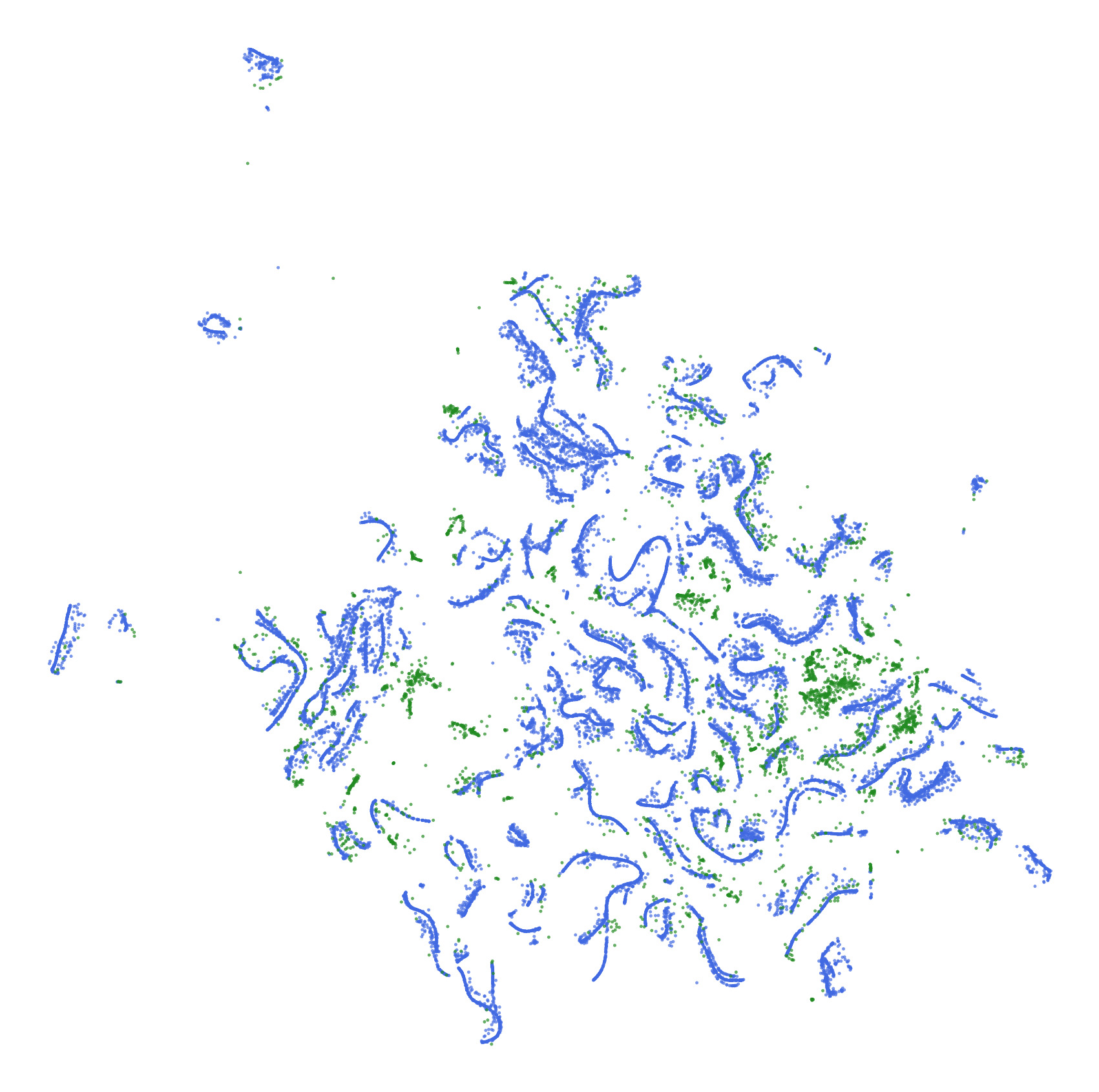} &
		\includegraphics[width=.2\linewidth,valign=m]{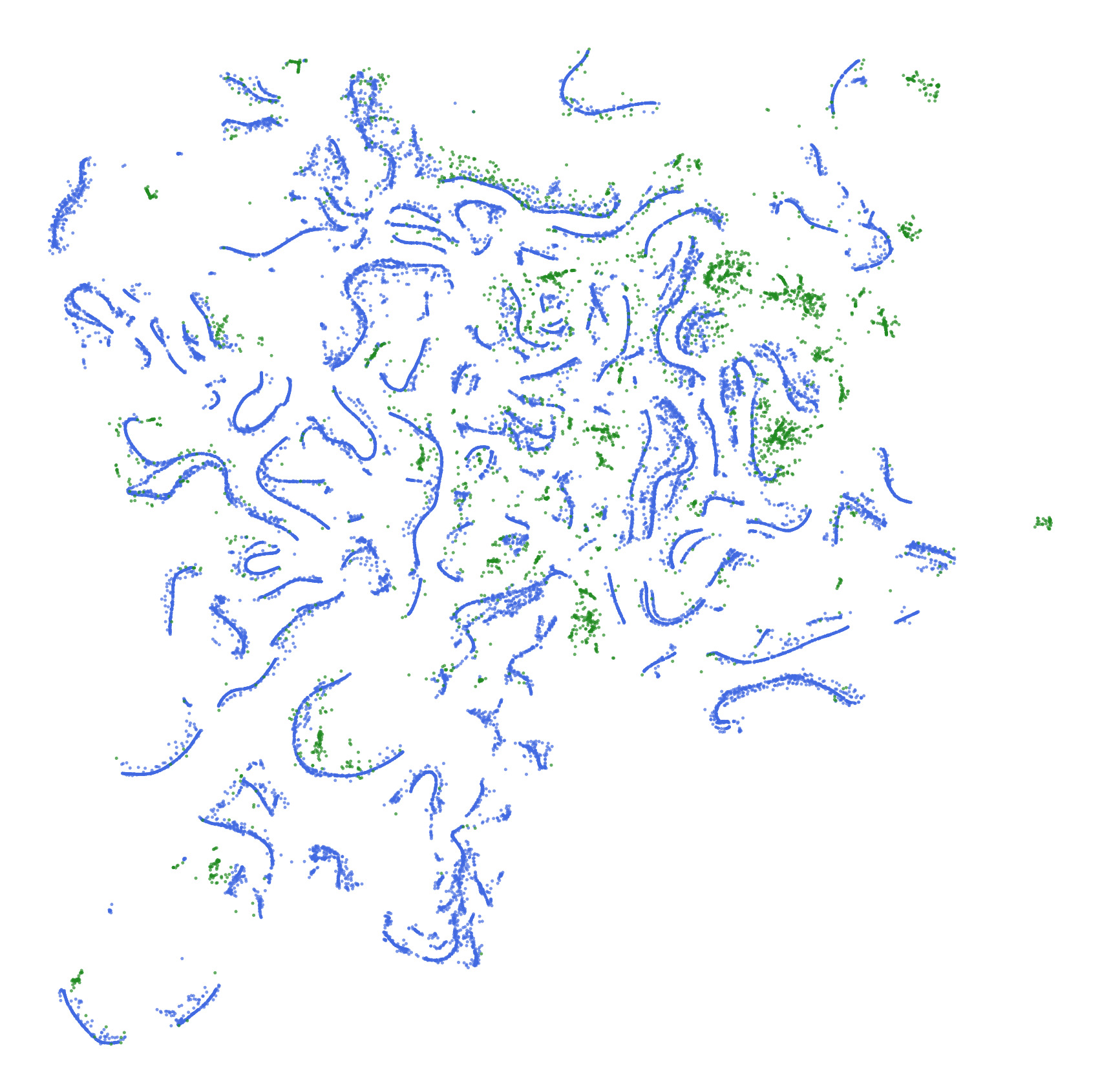} & \includegraphics[width=.2\linewidth,valign=m]{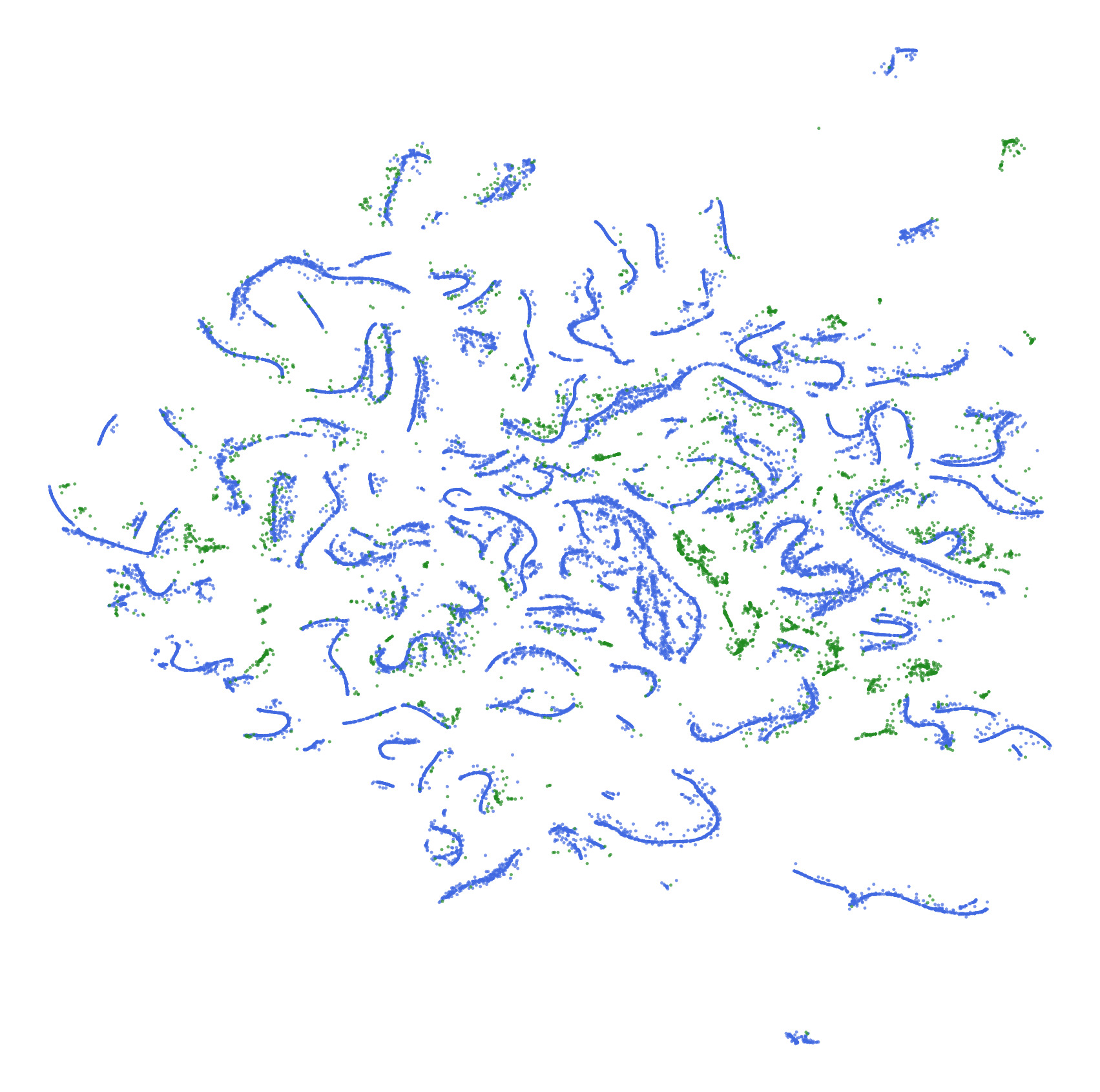}\\
	\end{tabular}
	\caption{t-SNE visualizations of the MoLane dataset (top) and the TuLane dataset (bottom). The source domain is marked in blue,  the real-world model vehicle target domain in red, and TuLane's target domain in green.}
	\label{fig:TSNE_plot_molane_tulane}
\end{figure}

\textbf{t-SNE feature clustering.} \autoref{fig:TSNE_plot_molane_tulane} shows the t-SNE feature clustering of the trained baselines for the MoLane and TuLane dataset, respectively. We observe that few features of both domains spread over the entire plot for higher-performing UDA methods. However, there are still large clusters of features from one domain, indicating that the domain adaptation only occurred slightly.

\begin{figure}
	\centering
	\small
	\begin{tabular}{rc@{}c@{}c@{}c}
		~ & \textbf{MoLane} & \textbf{TuLane} & \multicolumn{2}{c}{\textbf{MuLane}} \\
		%
		\textbf{UFLD-SO} & 
		\includegraphics[width=.18\linewidth,valign=m]{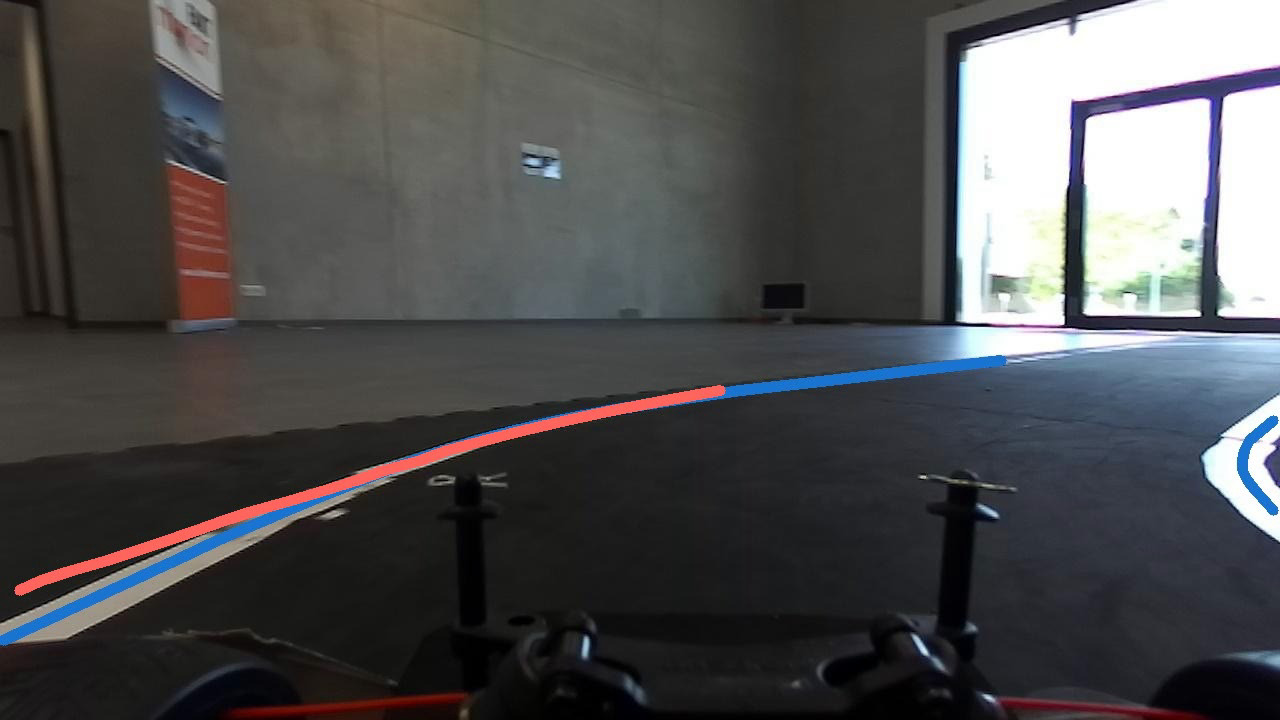} & \includegraphics[width=.18\linewidth,valign=m]{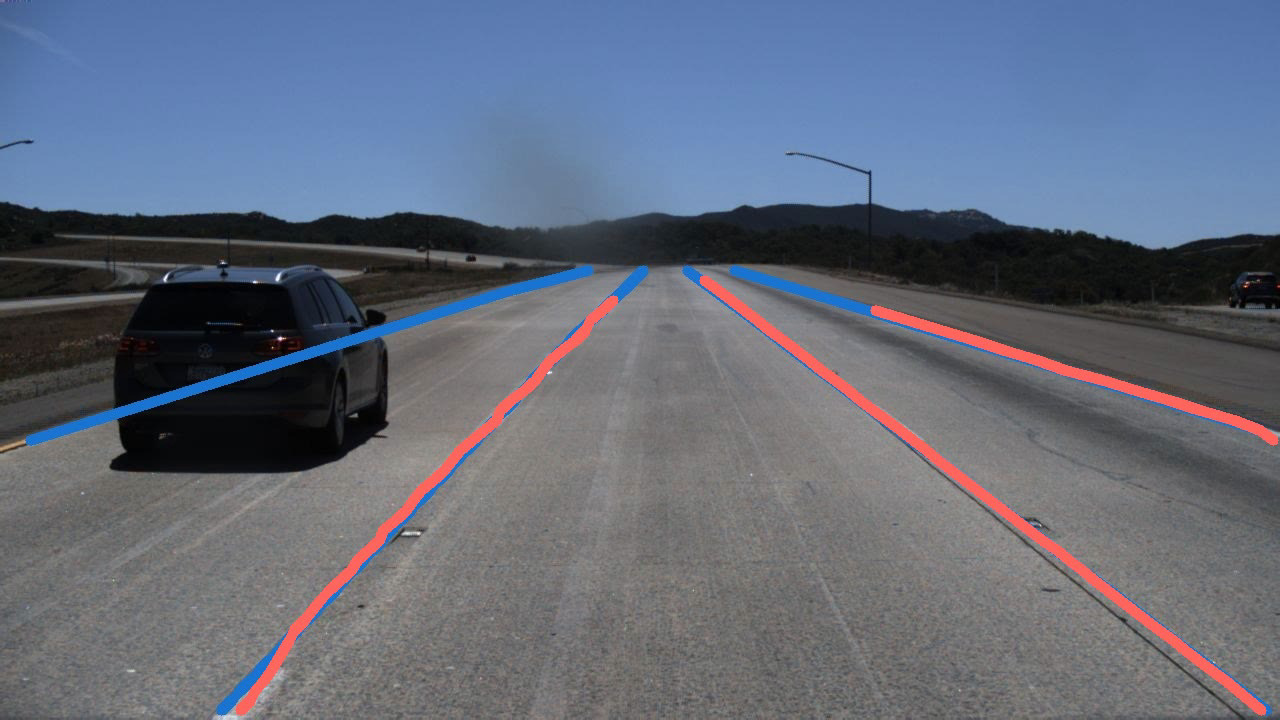} &
		\includegraphics[width=.18\linewidth,valign=m]{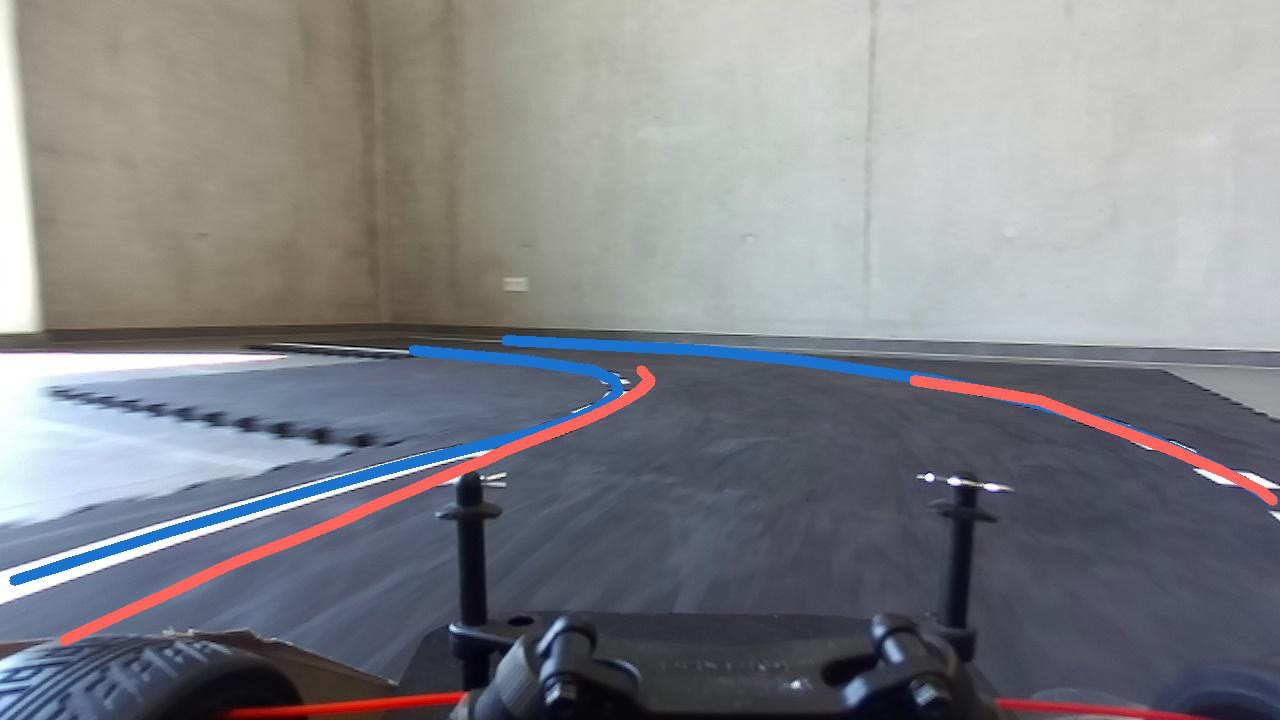} & \includegraphics[width=.18\linewidth,valign=m]{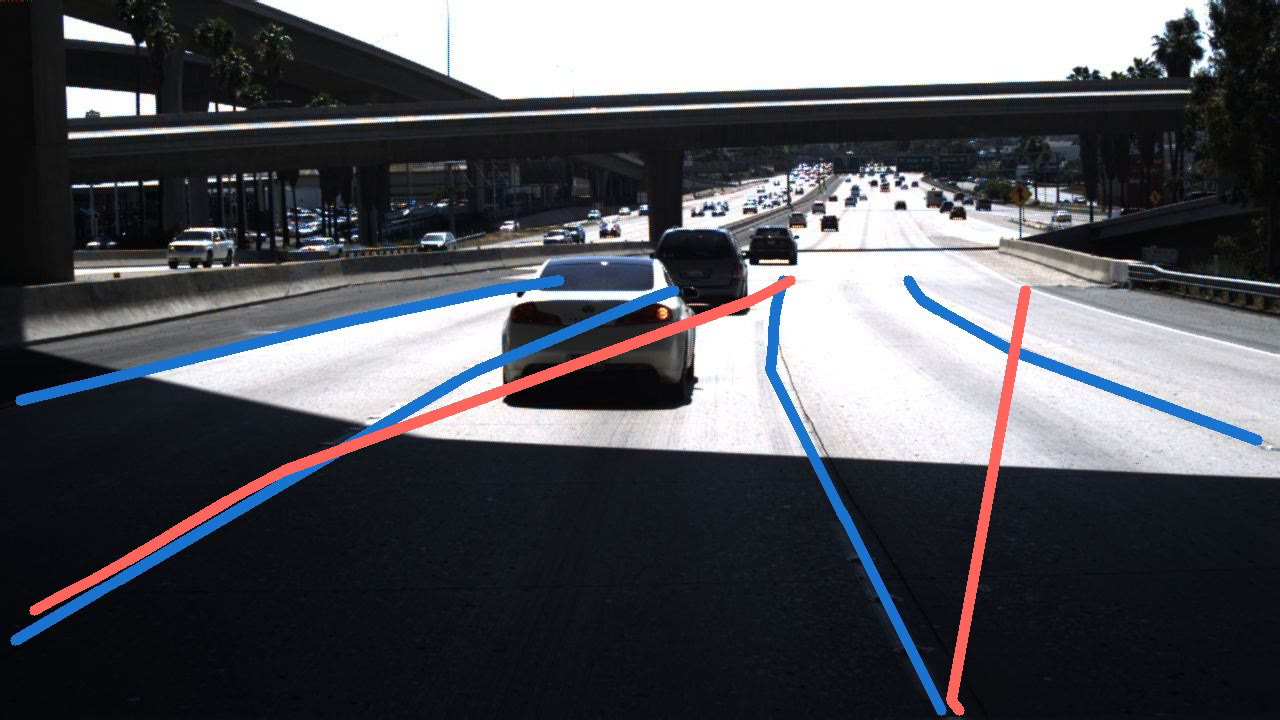}\\
		%
		\textbf{DANN} & 
		\includegraphics[width=.18\linewidth,valign=m]{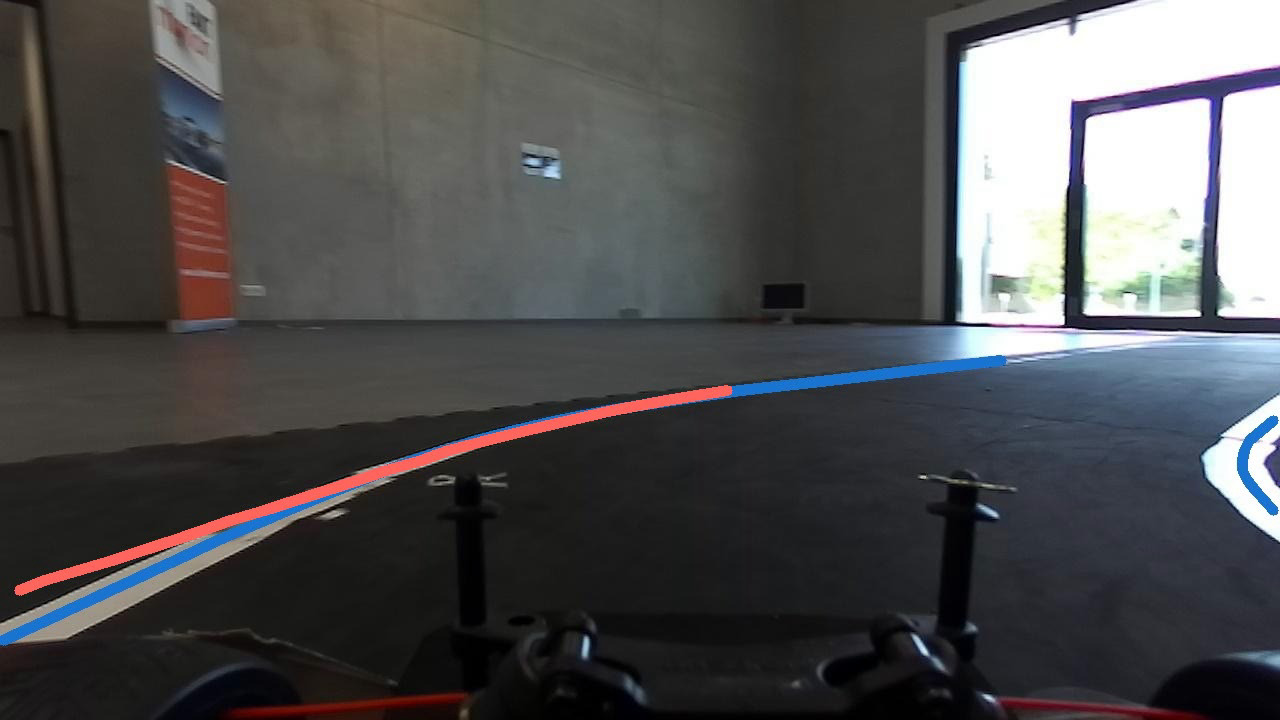} & 
		\includegraphics[width=.18\linewidth,valign=m]{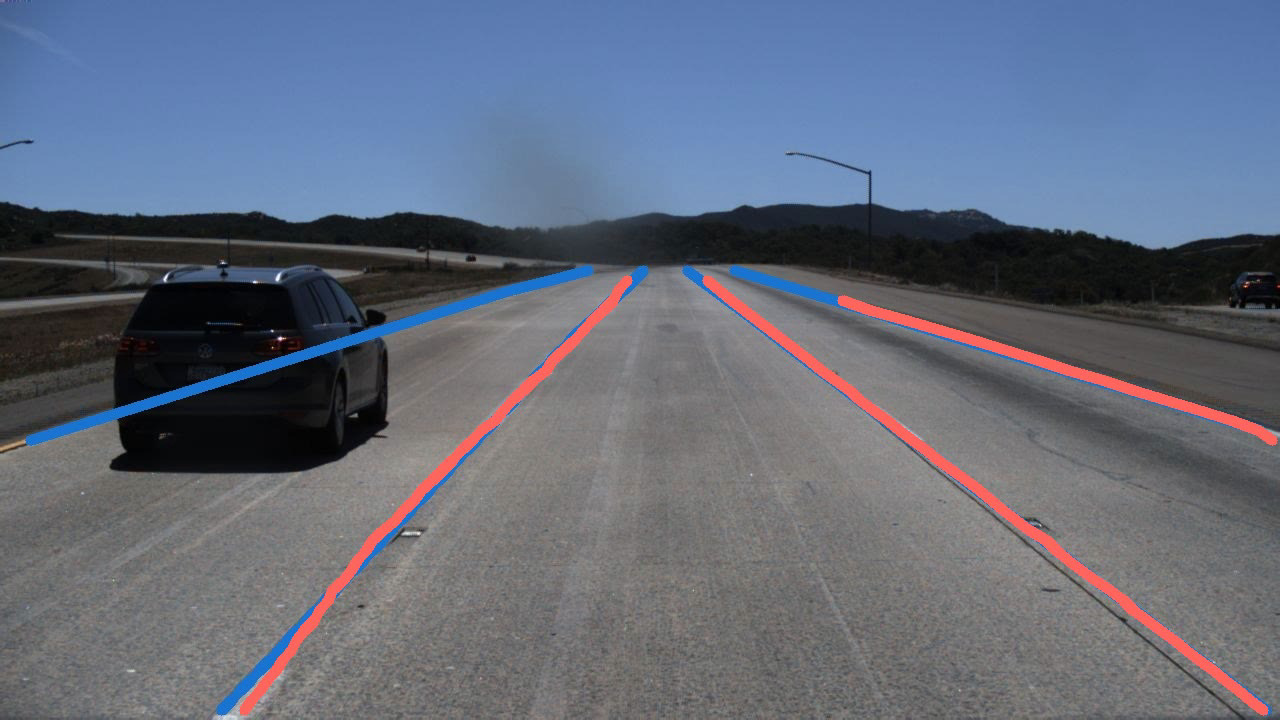} &
		\includegraphics[width=.18\linewidth,valign=m]{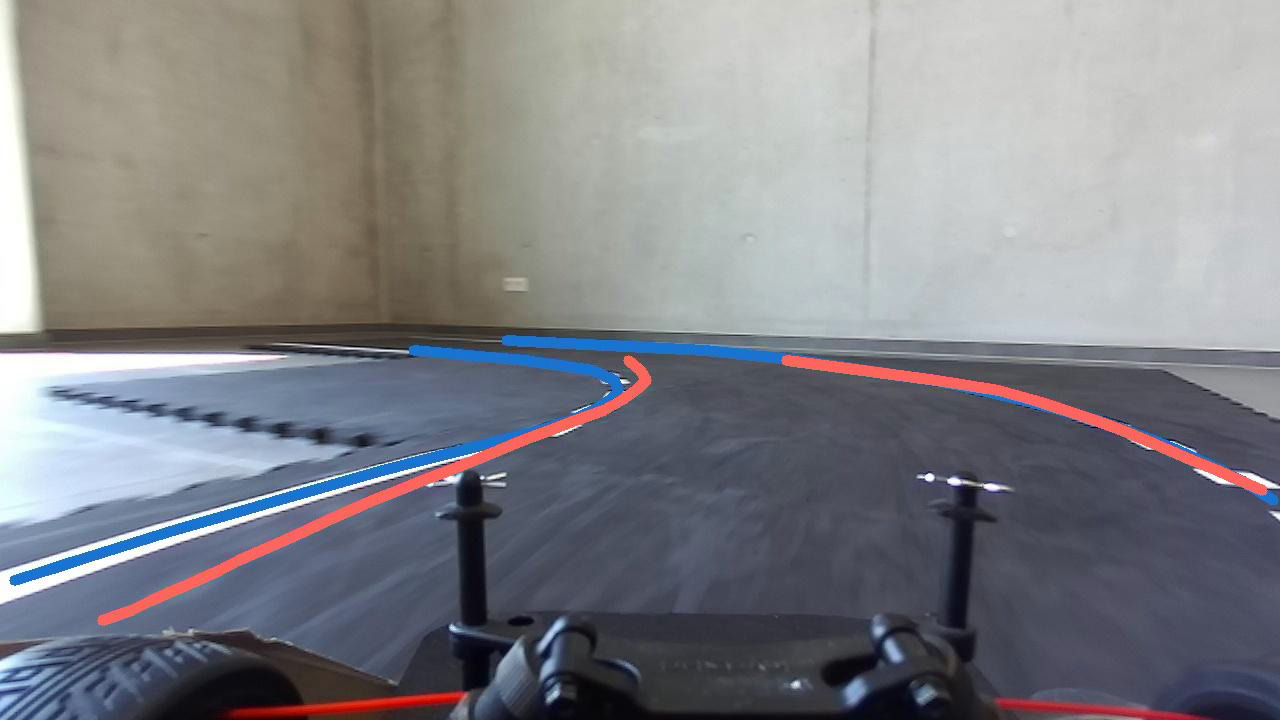} & \includegraphics[width=.18\linewidth,valign=m]{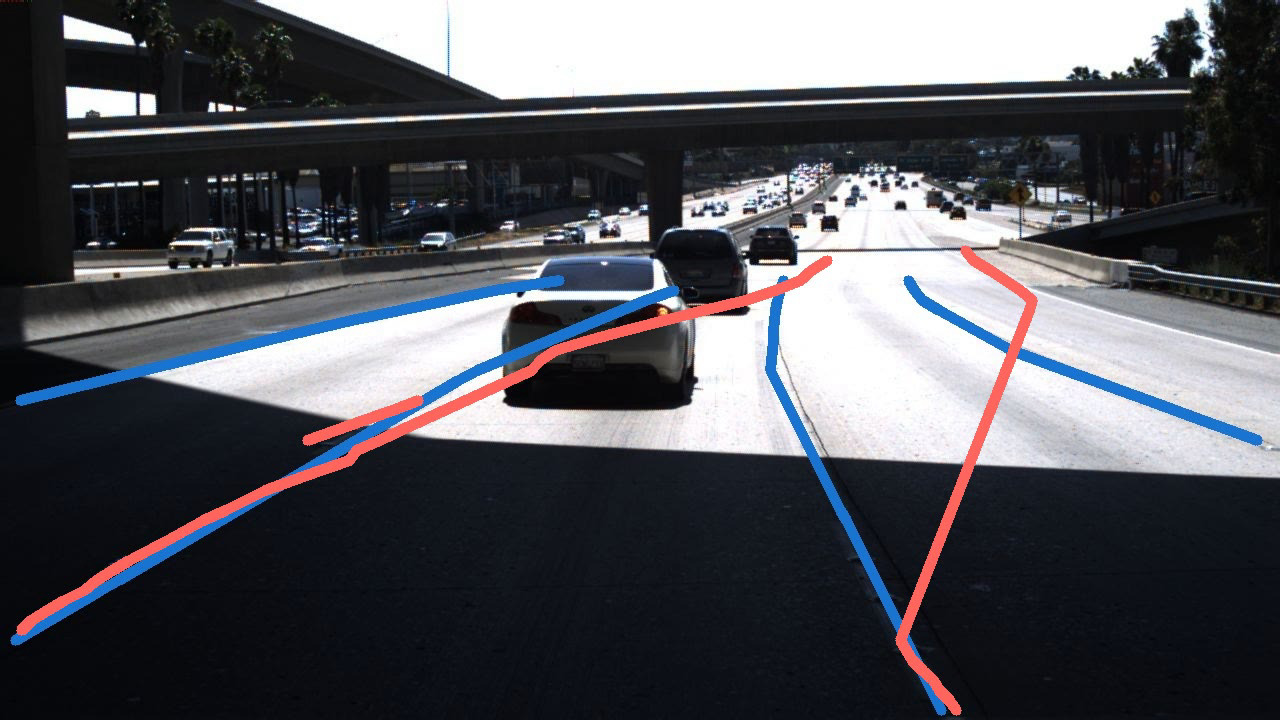}\\
		%
		\textbf{ADDA} & 
		\includegraphics[width=.18\linewidth,valign=m]{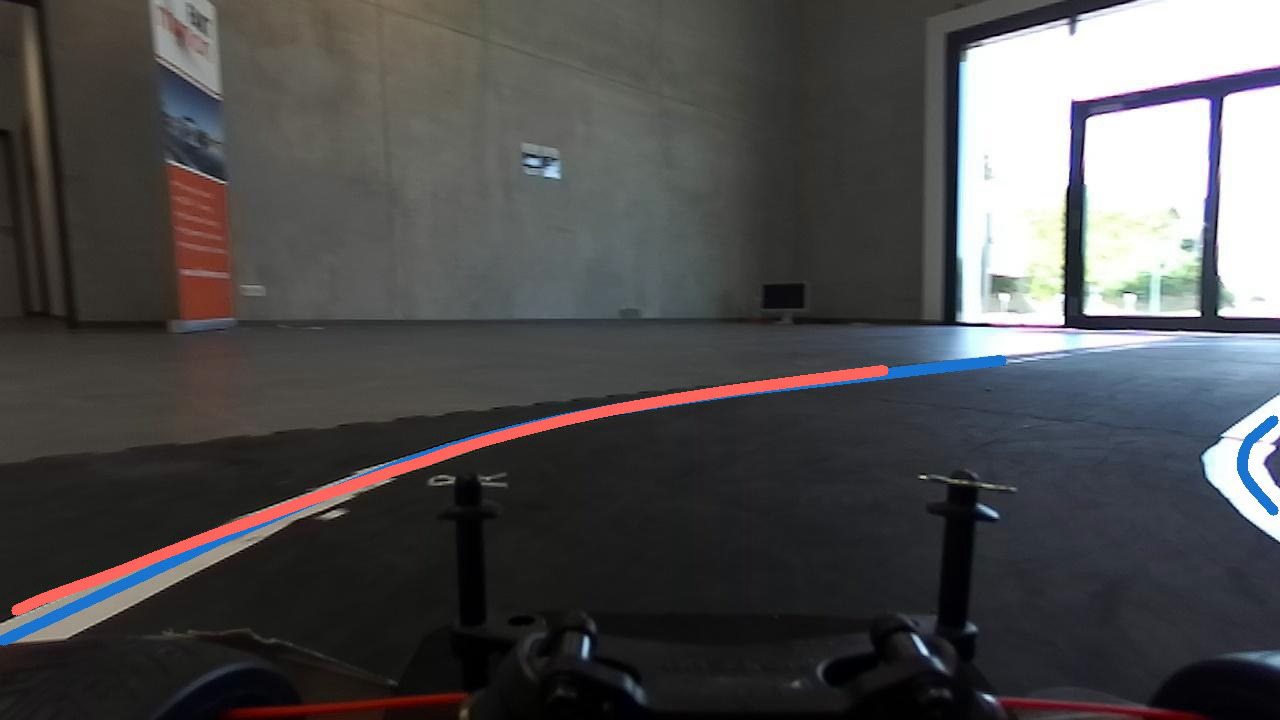} & 
		\includegraphics[width=.18\linewidth,valign=m]{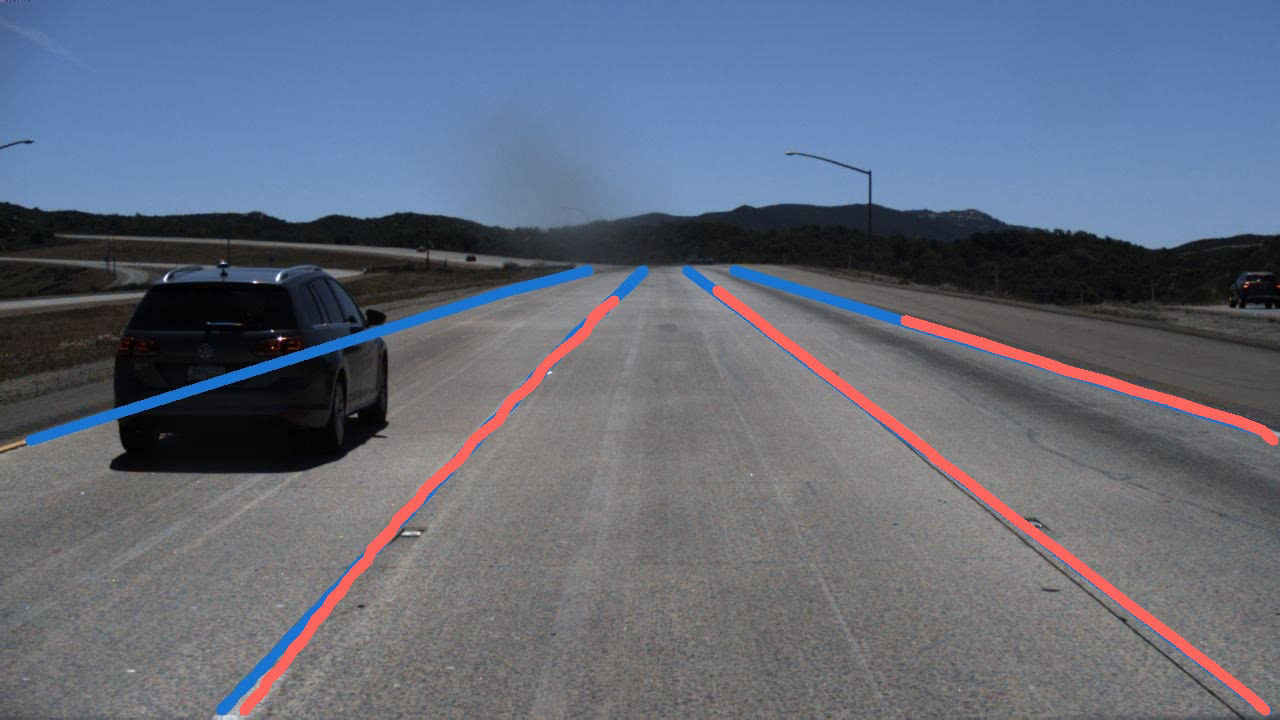} &
		\includegraphics[width=.18\linewidth,valign=m]{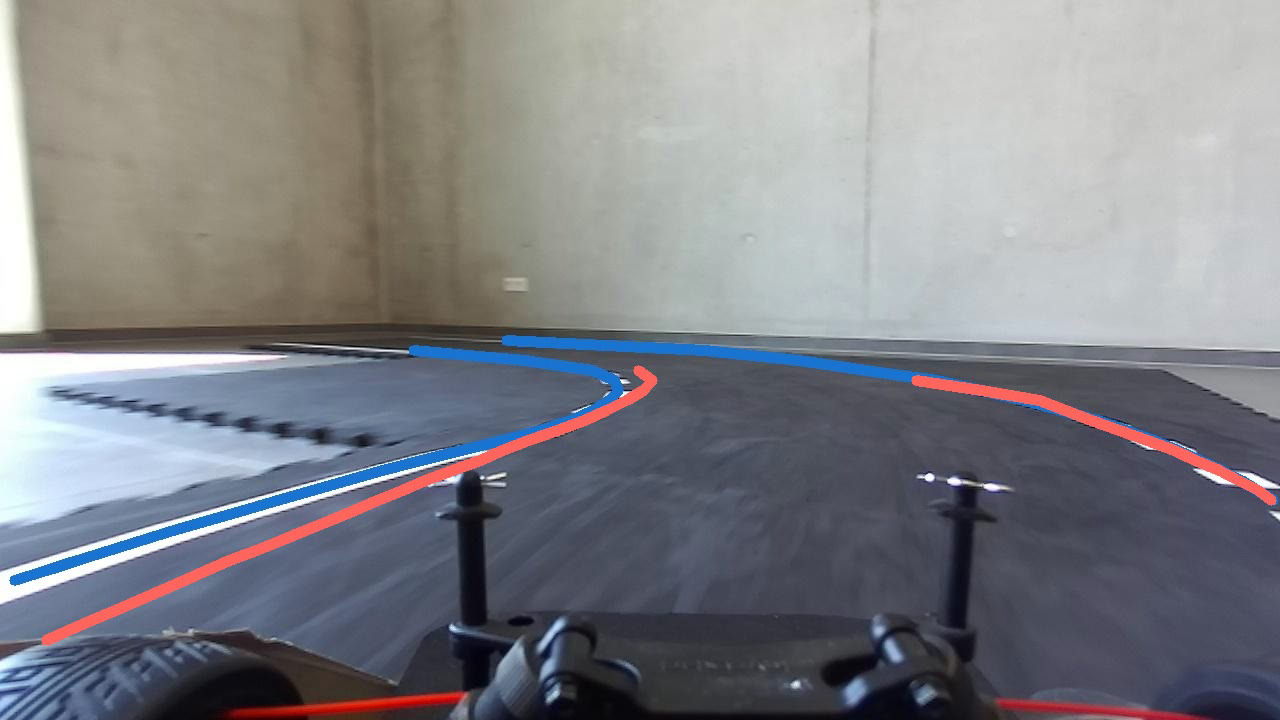} & \includegraphics[width=.18\linewidth,valign=m]{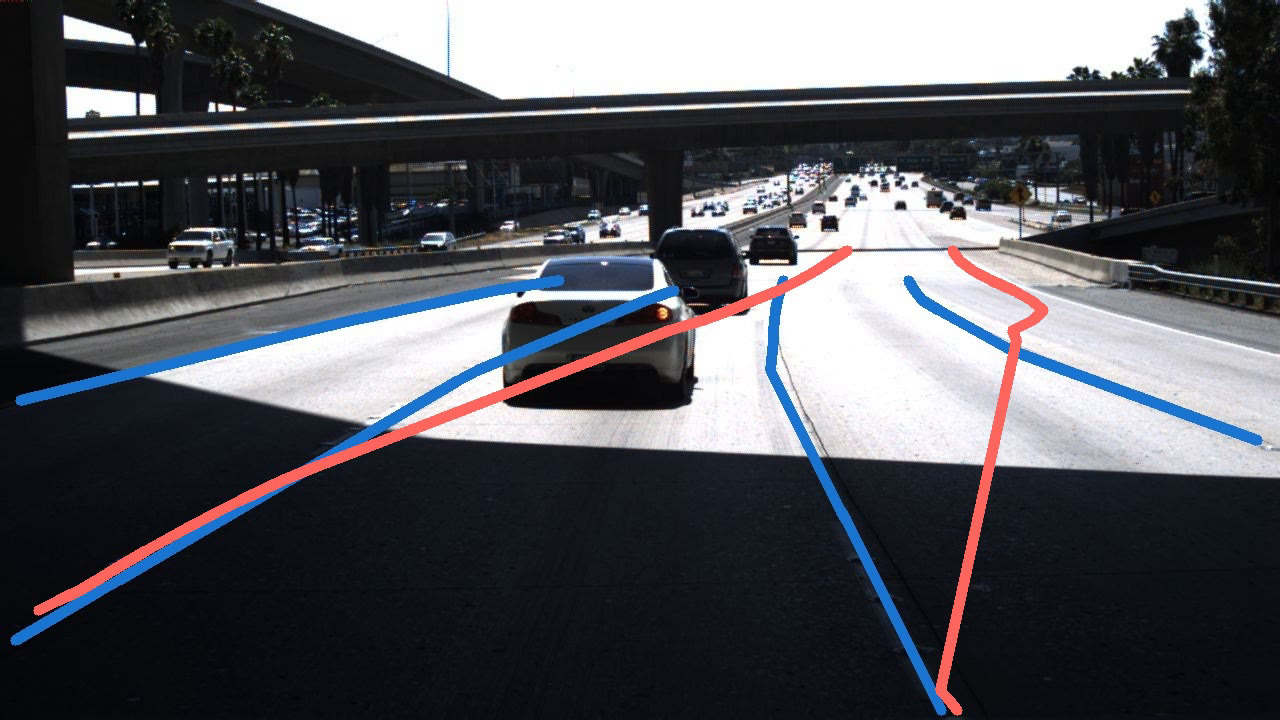}\\
		%
		\textbf{SGADA} & 
		\includegraphics[width=.18\linewidth,valign=m]{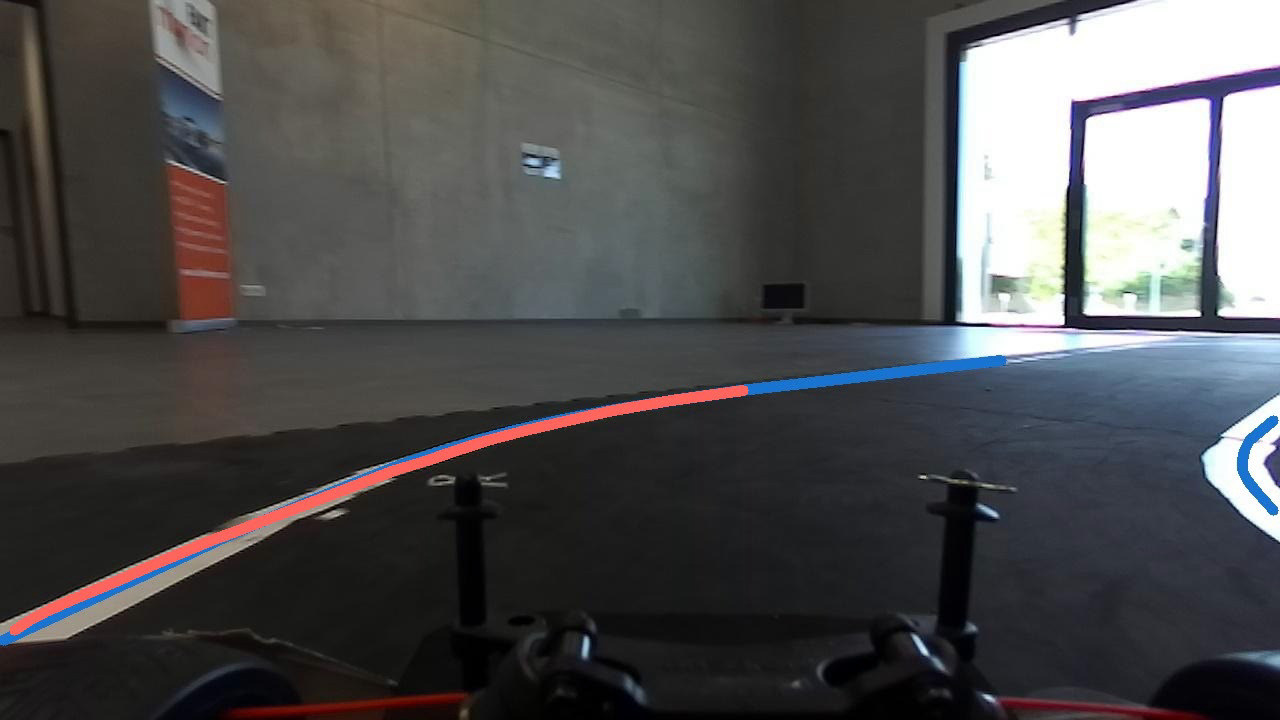} & 
		\includegraphics[width=.18\linewidth,valign=m]{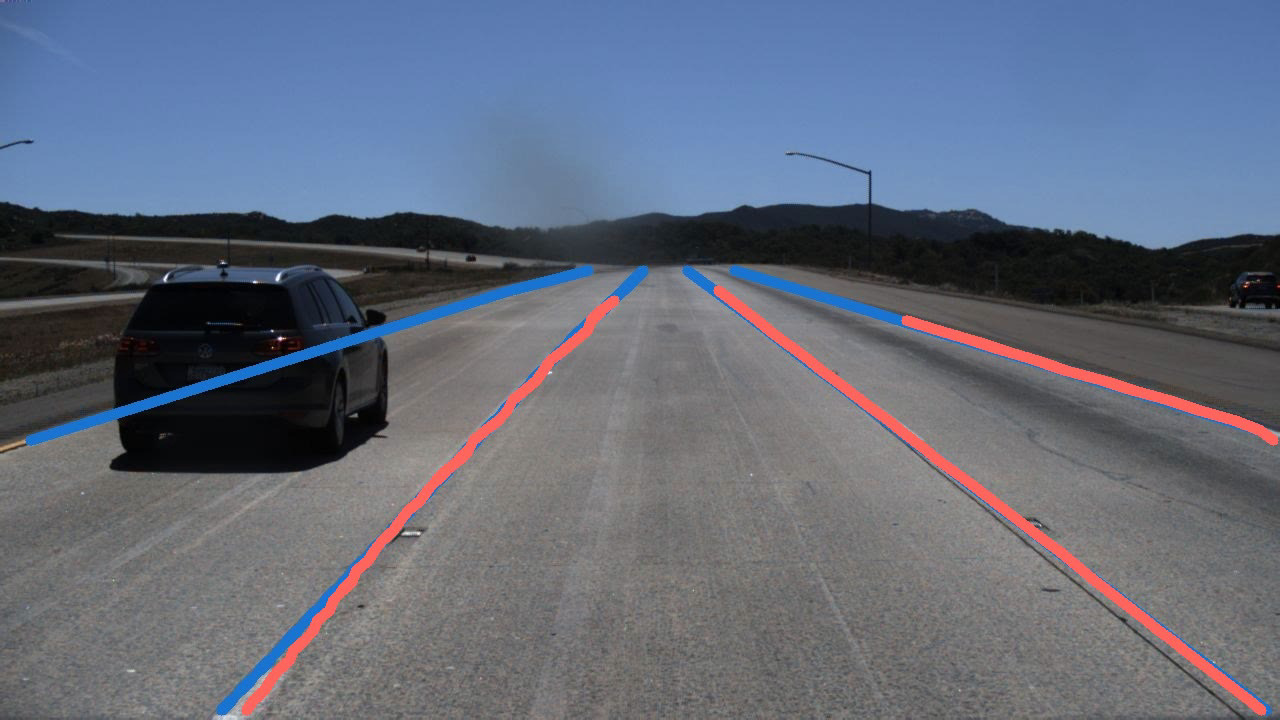} &
		\includegraphics[width=.18\linewidth,valign=m]{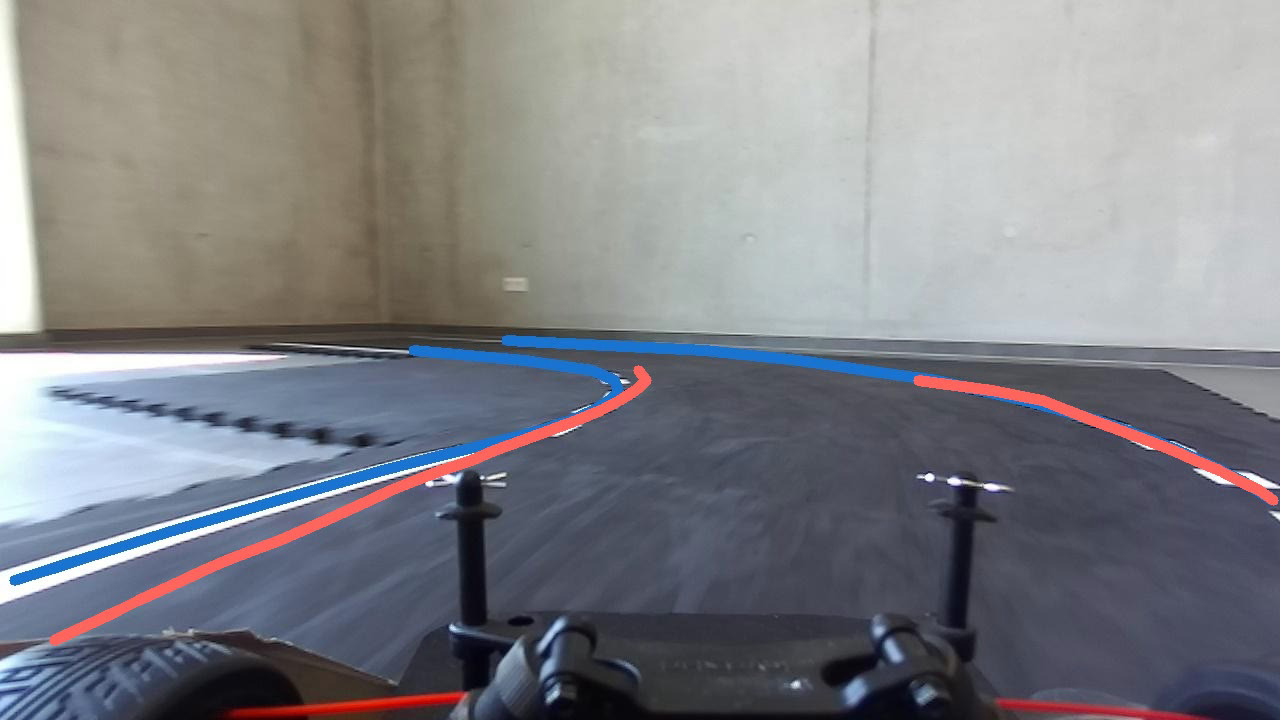} & \includegraphics[width=.18\linewidth,valign=m]{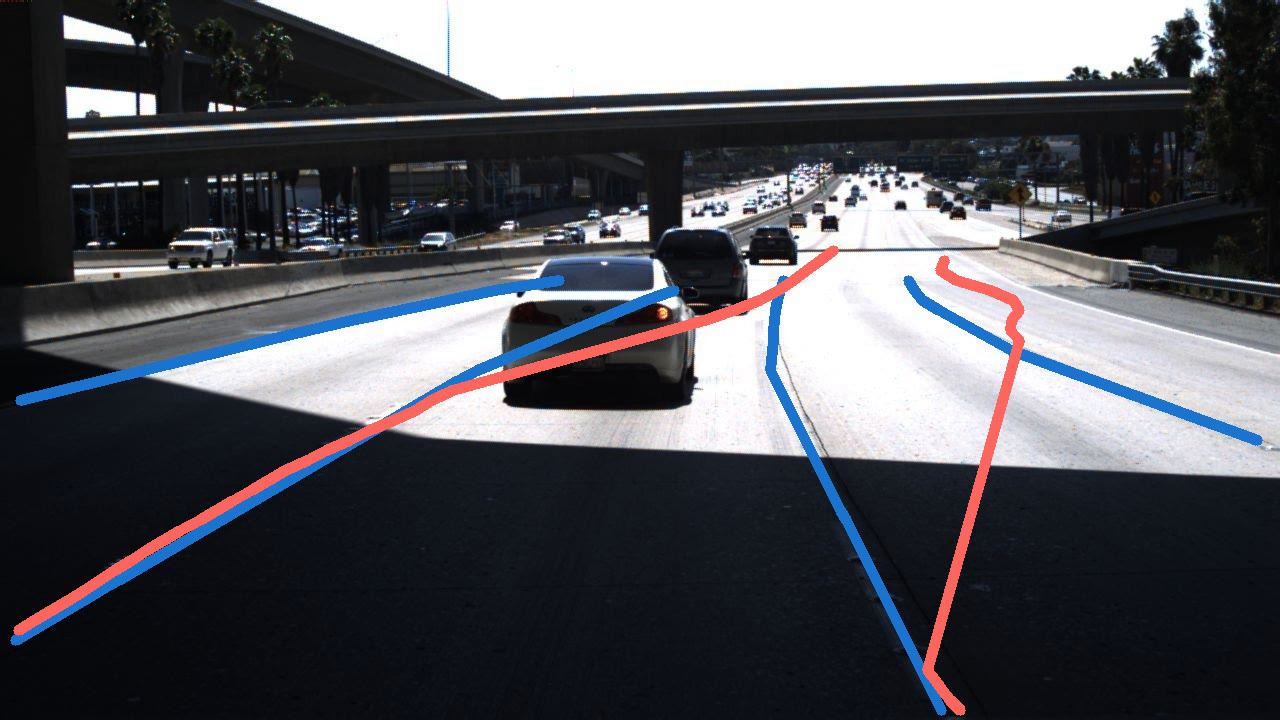}\\
		%
		\textbf{SGPCS} & 
		\includegraphics[width=.18\linewidth,valign=m]{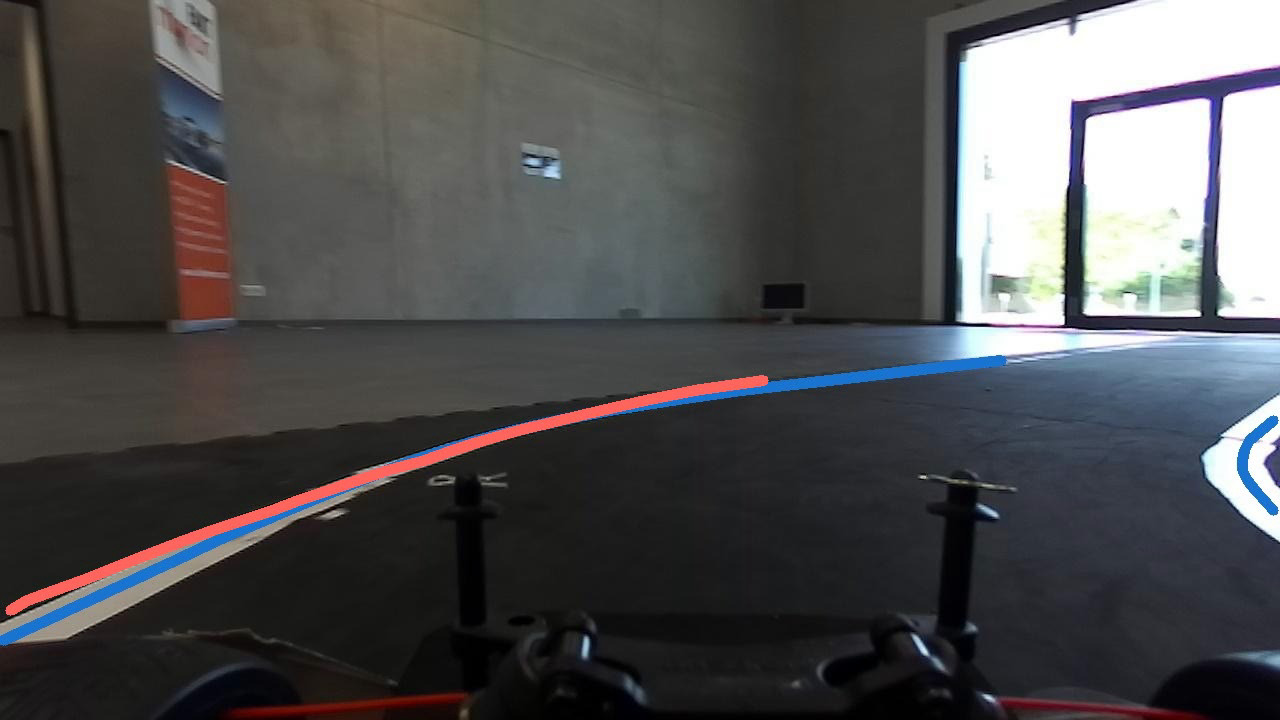} & \includegraphics[width=.18\linewidth,valign=m]{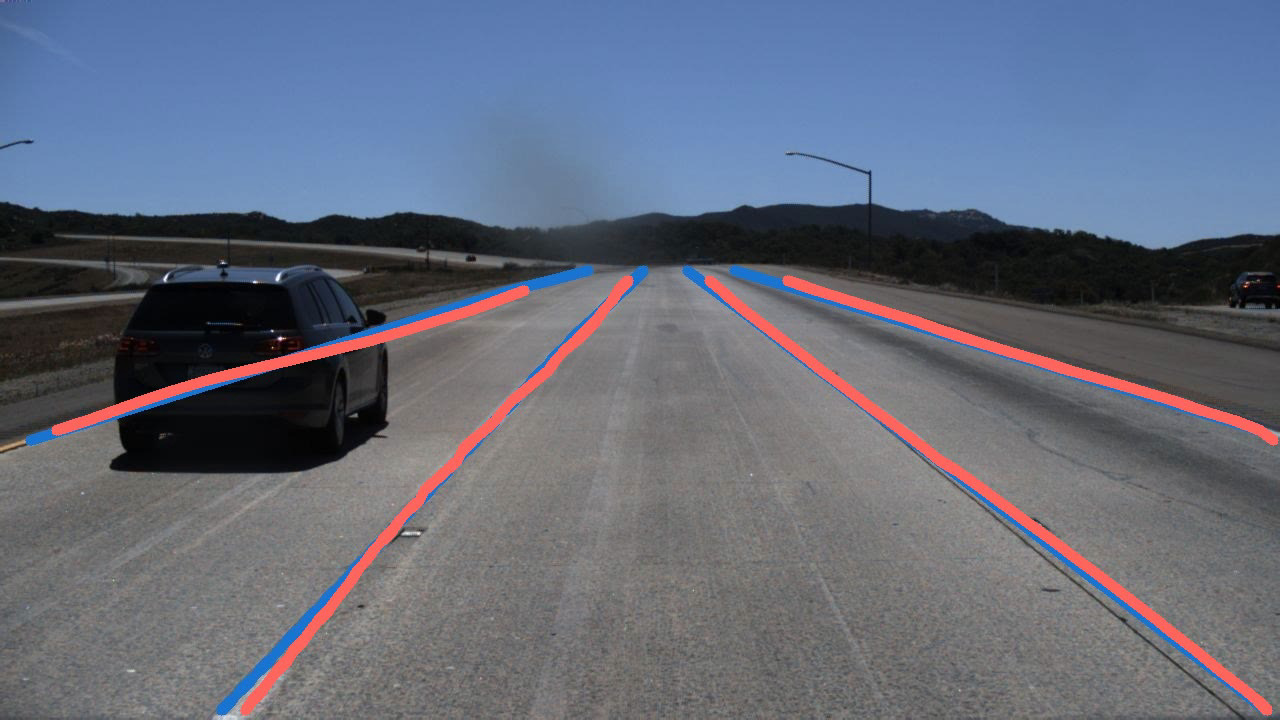} &
		\includegraphics[width=.18\linewidth,valign=m]{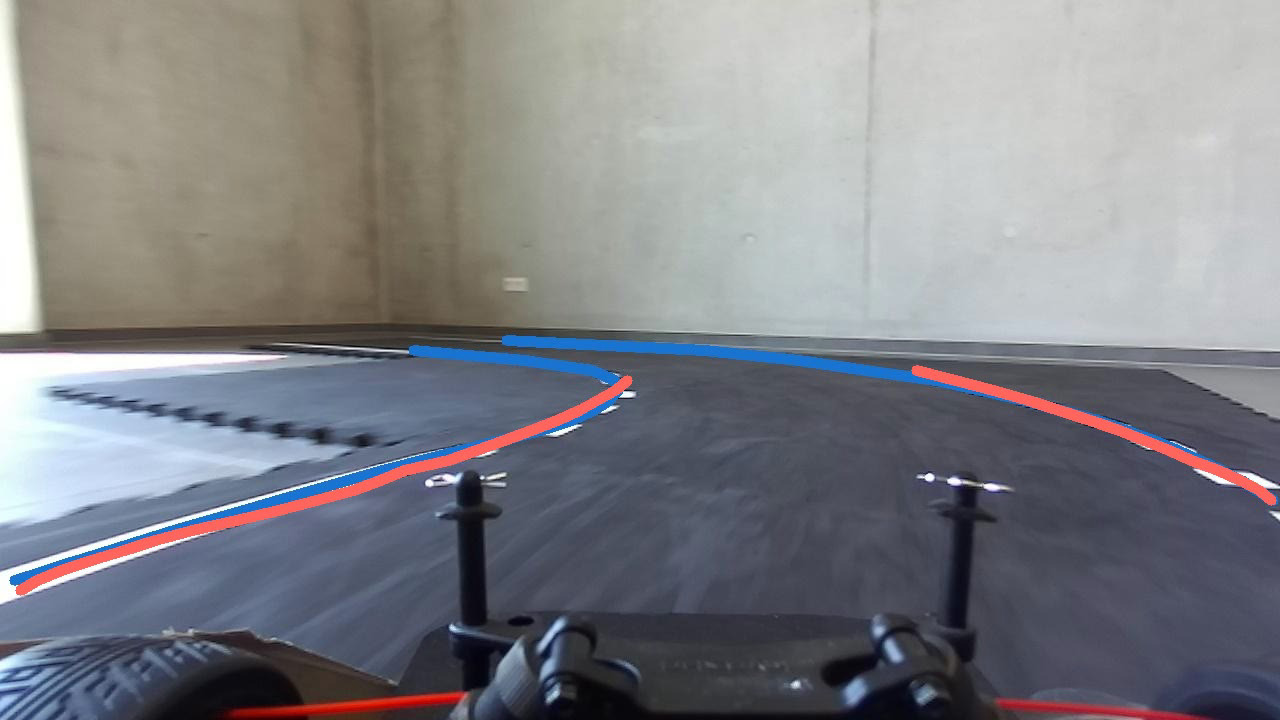} & \includegraphics[width=.18\linewidth,valign=m]{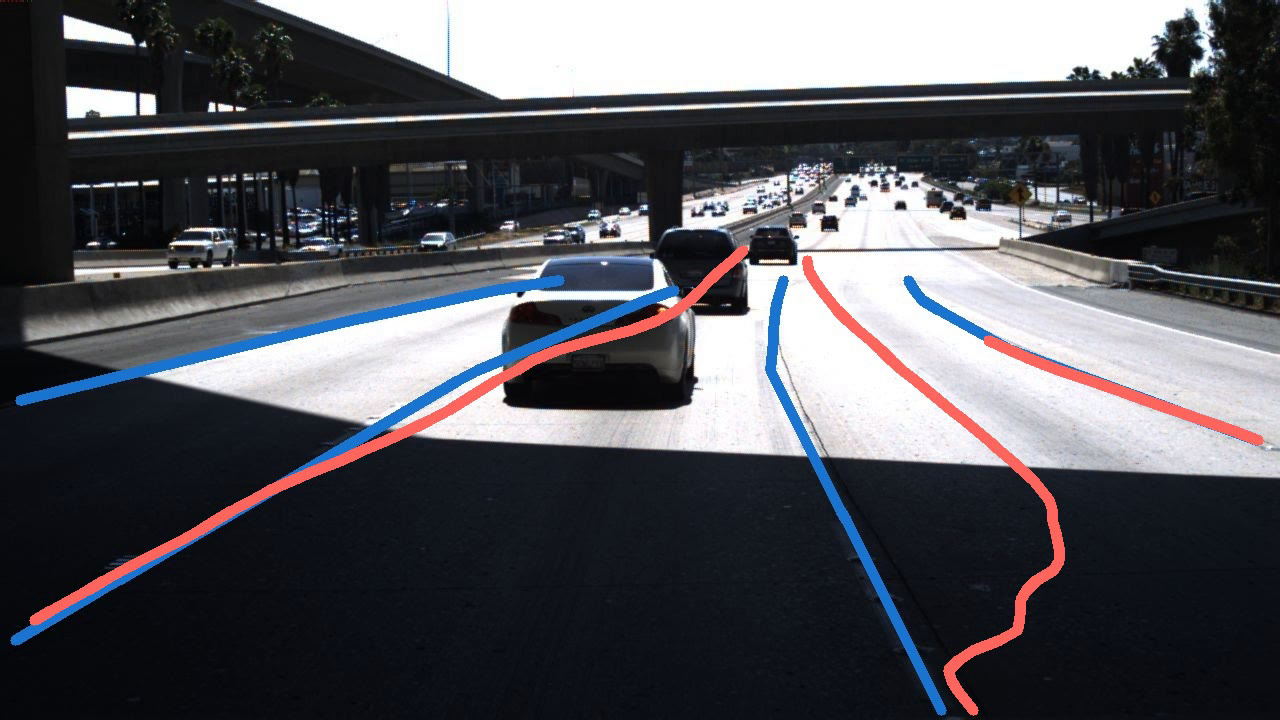}\\
		%
		\textbf{UFLD-TO} & 
		\includegraphics[width=.18\linewidth,valign=m]{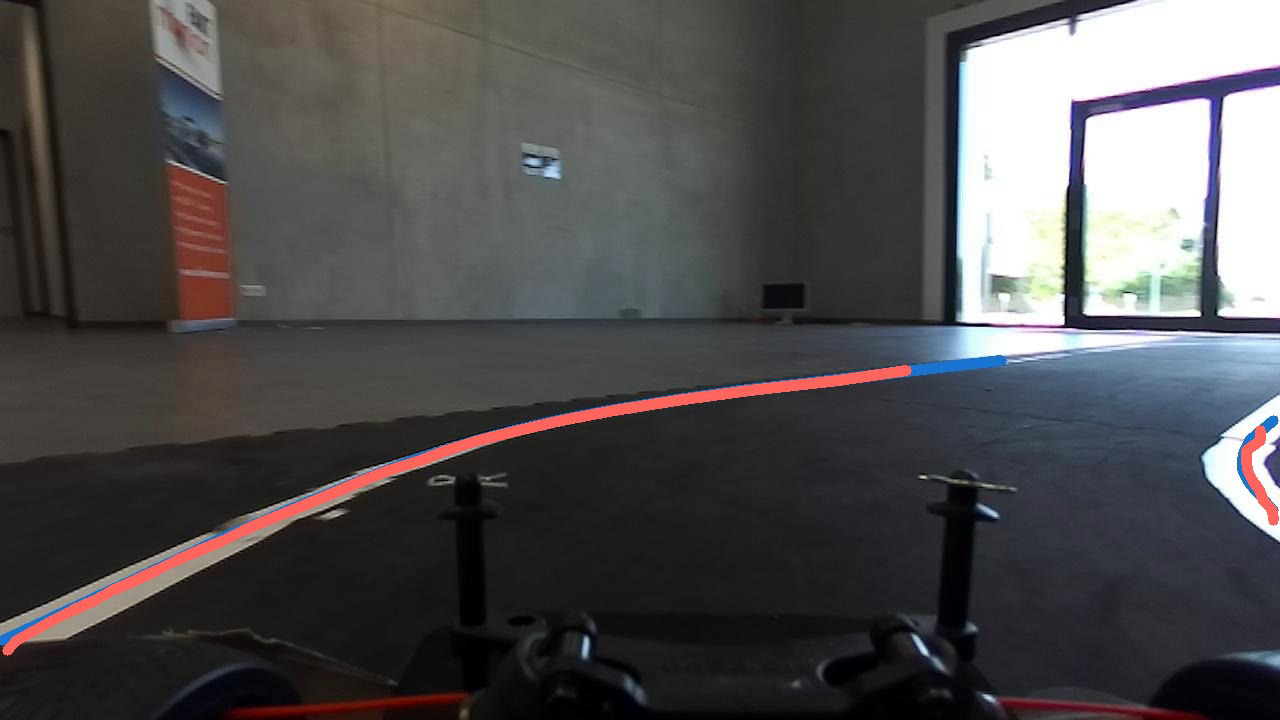} & \includegraphics[width=.18\linewidth,valign=m]{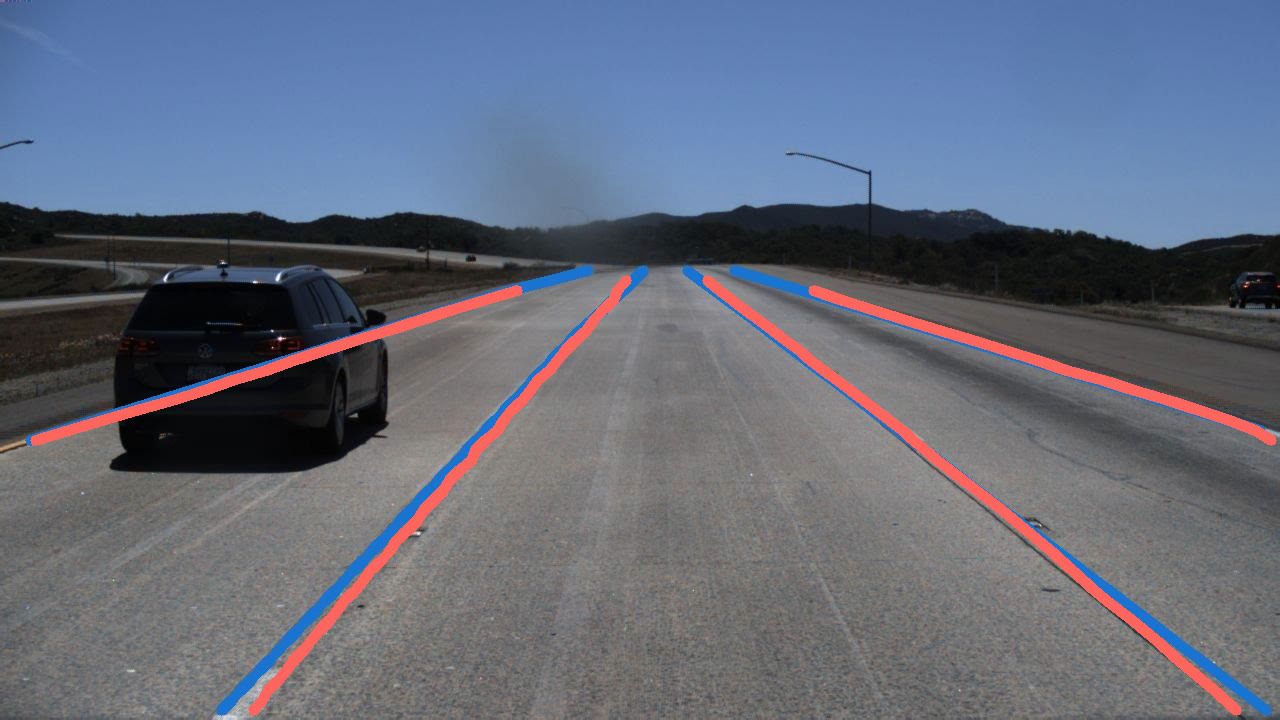} &
		\includegraphics[width=.18\linewidth,valign=m]{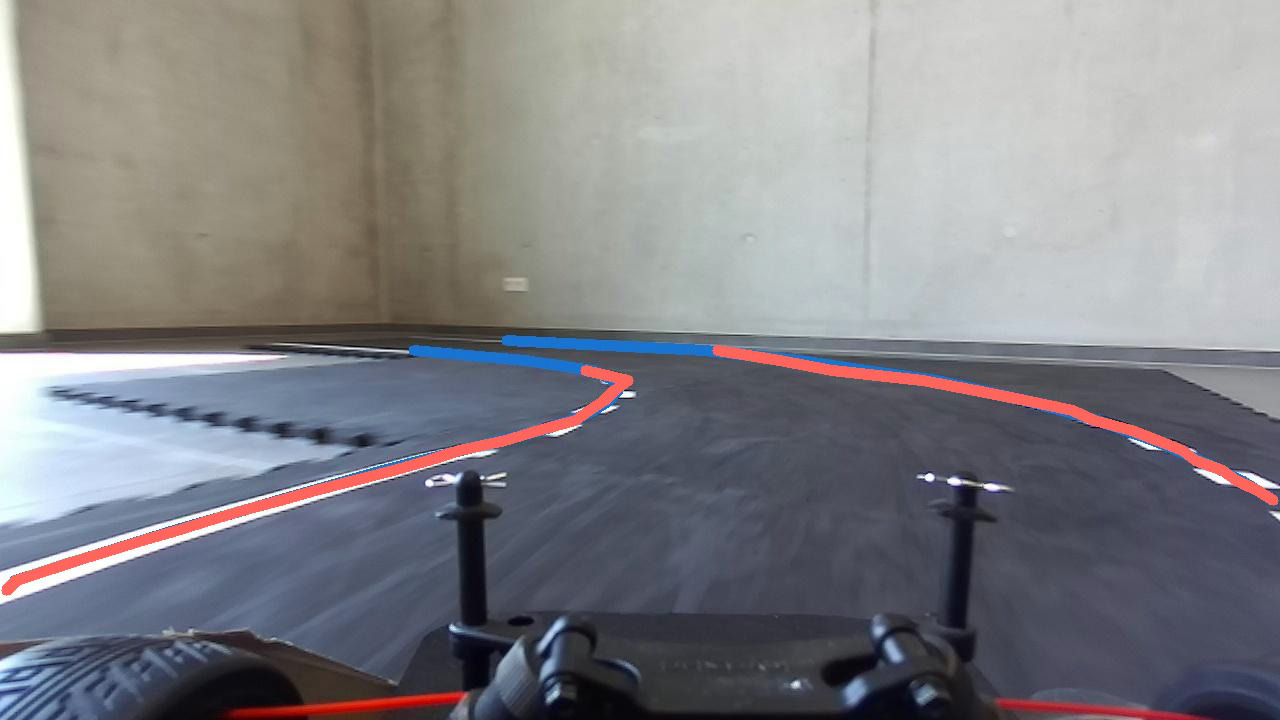} & \includegraphics[width=.18\linewidth,valign=m]{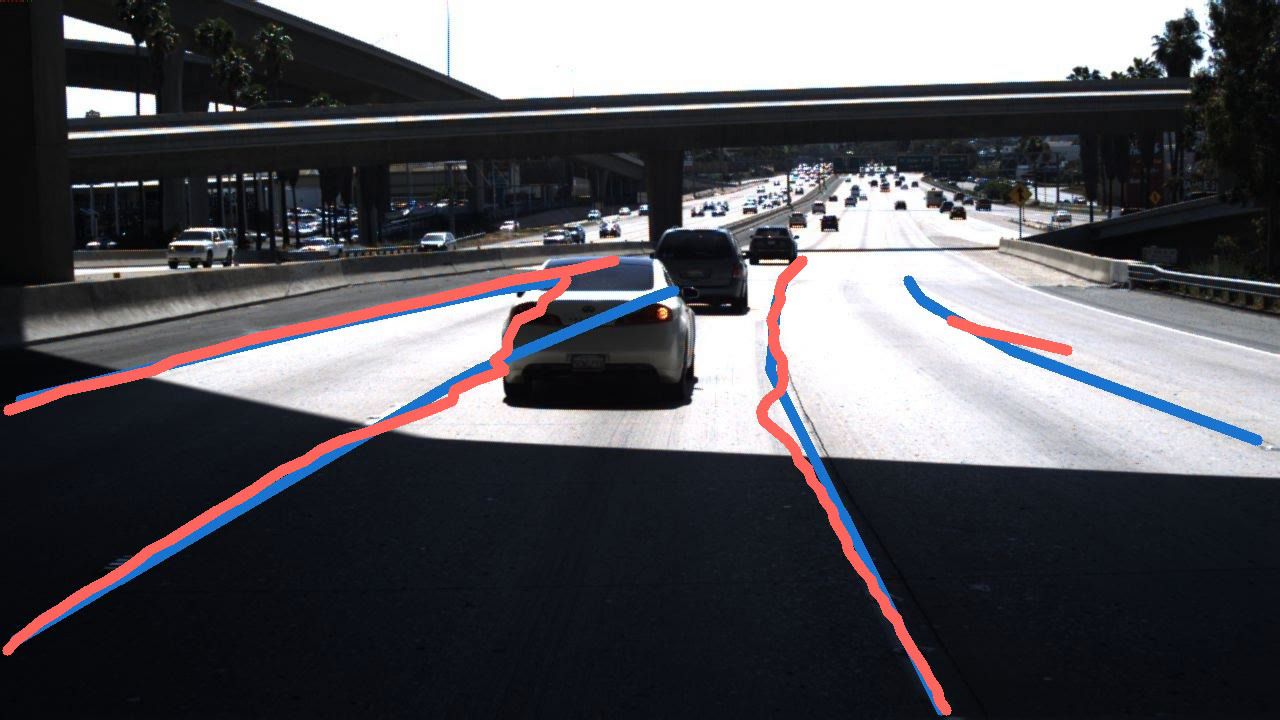}\\
	\end{tabular}
	\caption{Qualitative results of target domain predictions. Images are randomly sampled. Ground truth lane annotations are marked in blue, predictions in red.}
	\label{fig:appendix_inference_samples_1}
\end{figure}

\begin{figure}
	\centering
	\small
	\begin{tabular}{rc@{}c@{}c@{}c}
		~ & \textbf{MoLane} & \textbf{TuLane} & \multicolumn{2}{c}{\textbf{MuLane}} \\
		%
		\textbf{UFLD-SO} & 
		\includegraphics[width=.18\linewidth,valign=m]{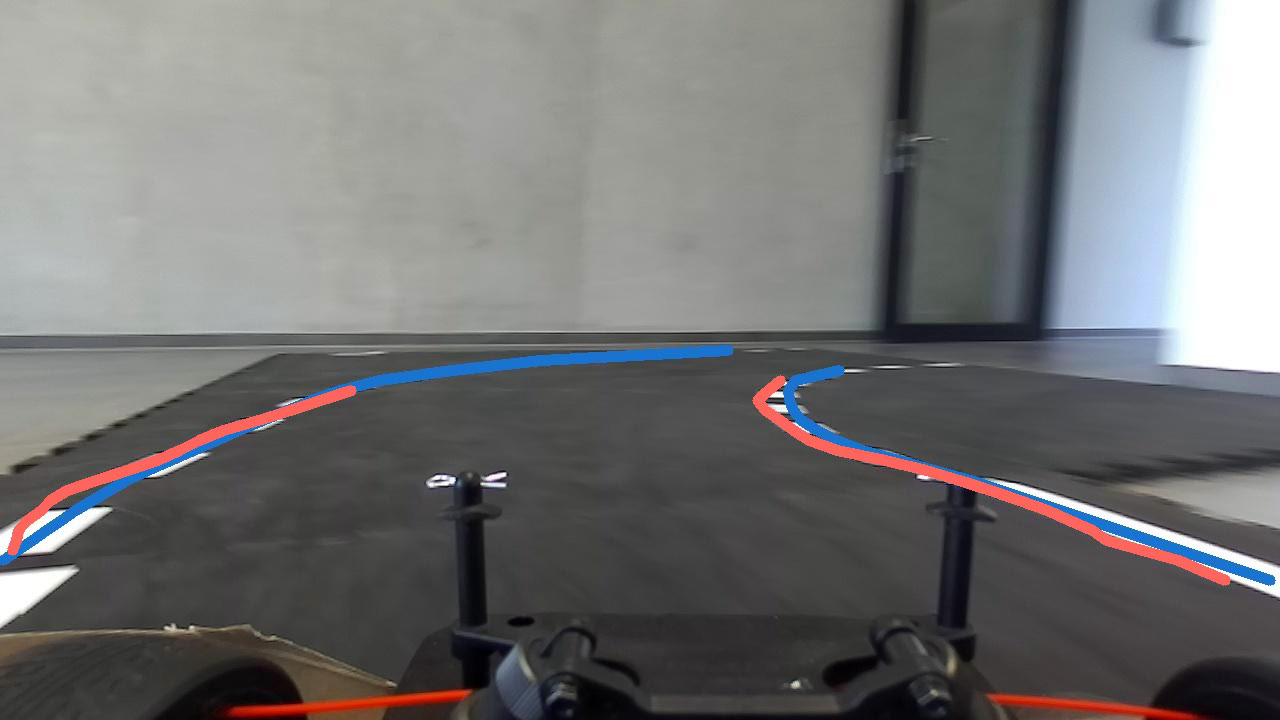} & \includegraphics[width=.18\linewidth,valign=m]{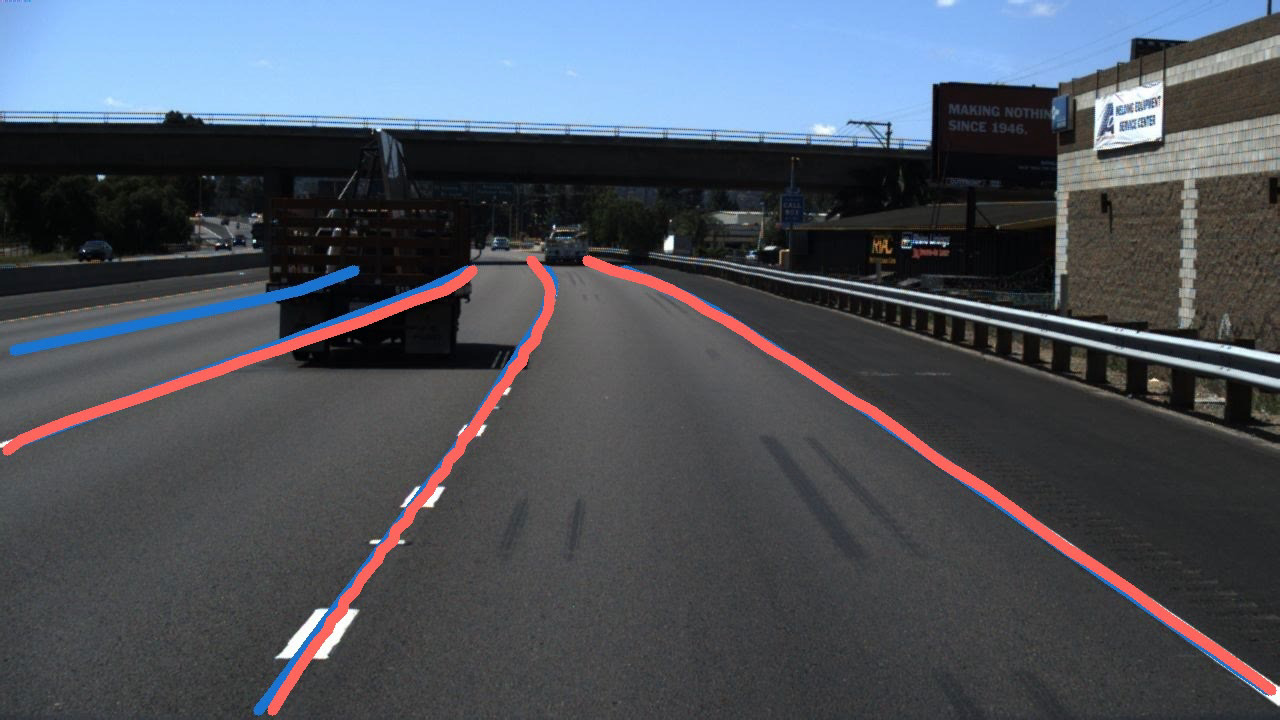} &
		\includegraphics[width=.18\linewidth,valign=m]{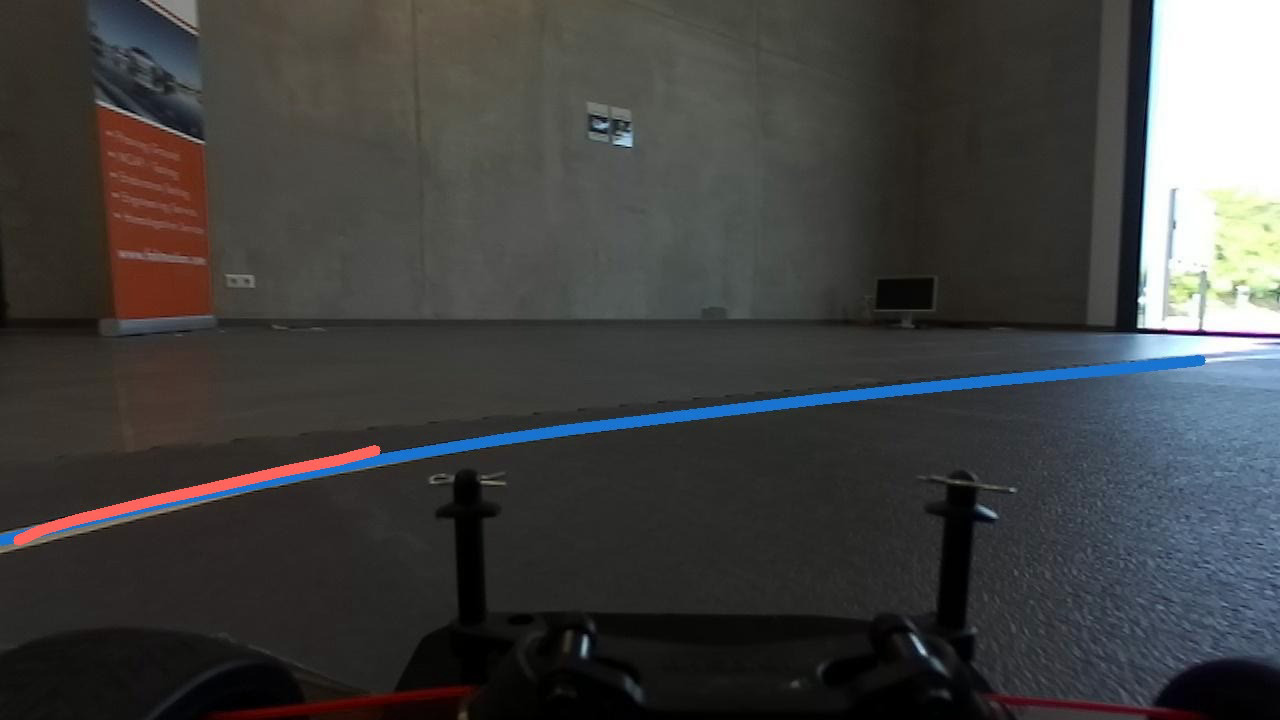} & \includegraphics[width=.18\linewidth,valign=m]{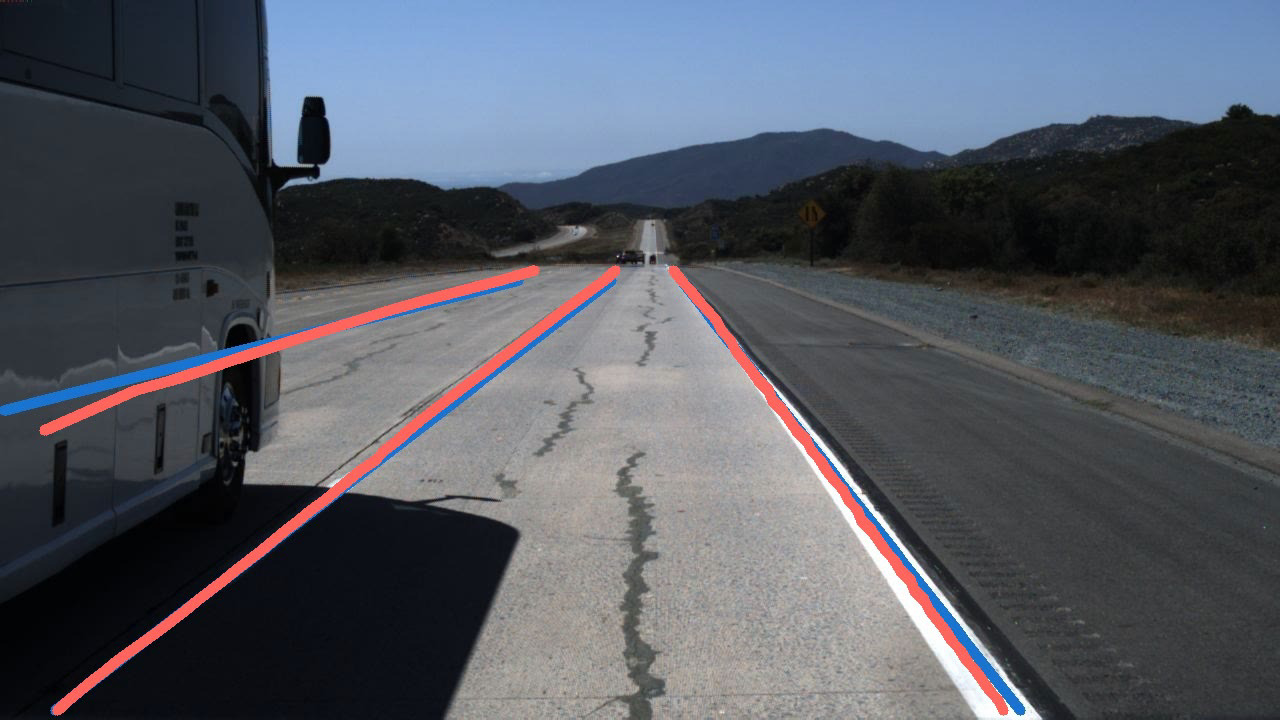}\\
		%
		\textbf{DANN} & 
		\includegraphics[width=.18\linewidth,valign=m]{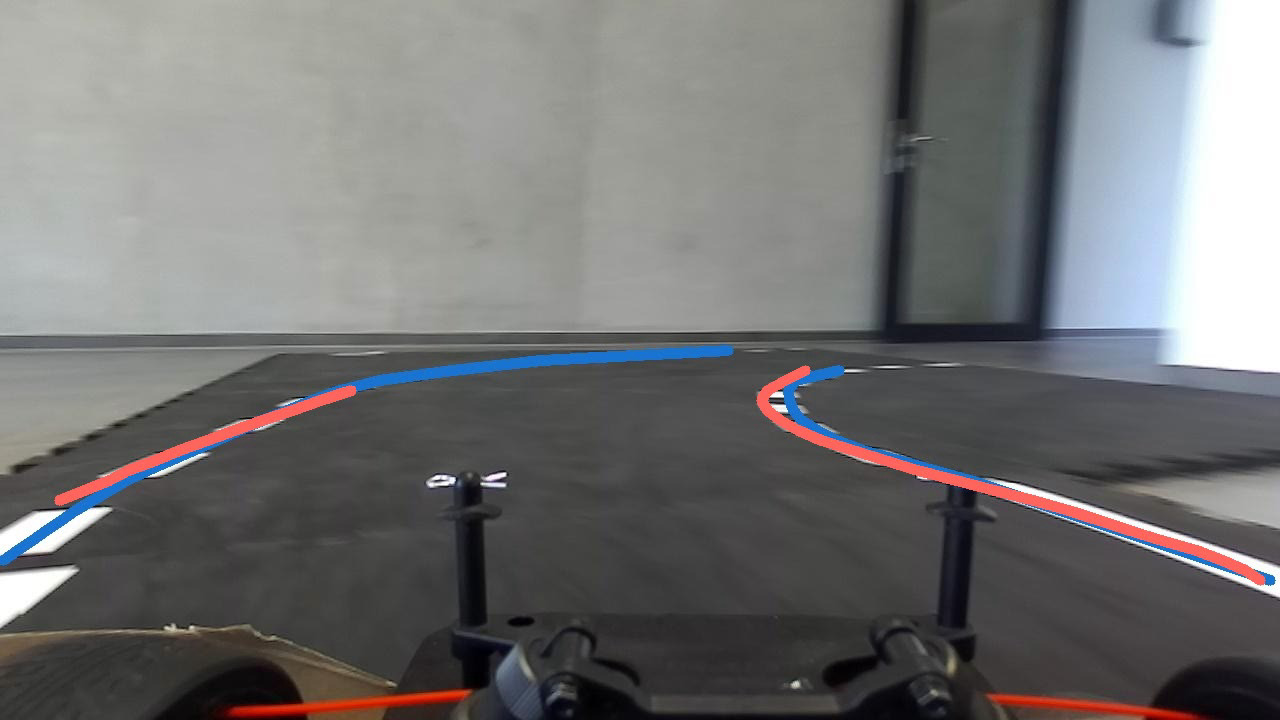} & 
		\includegraphics[width=.18\linewidth,valign=m]{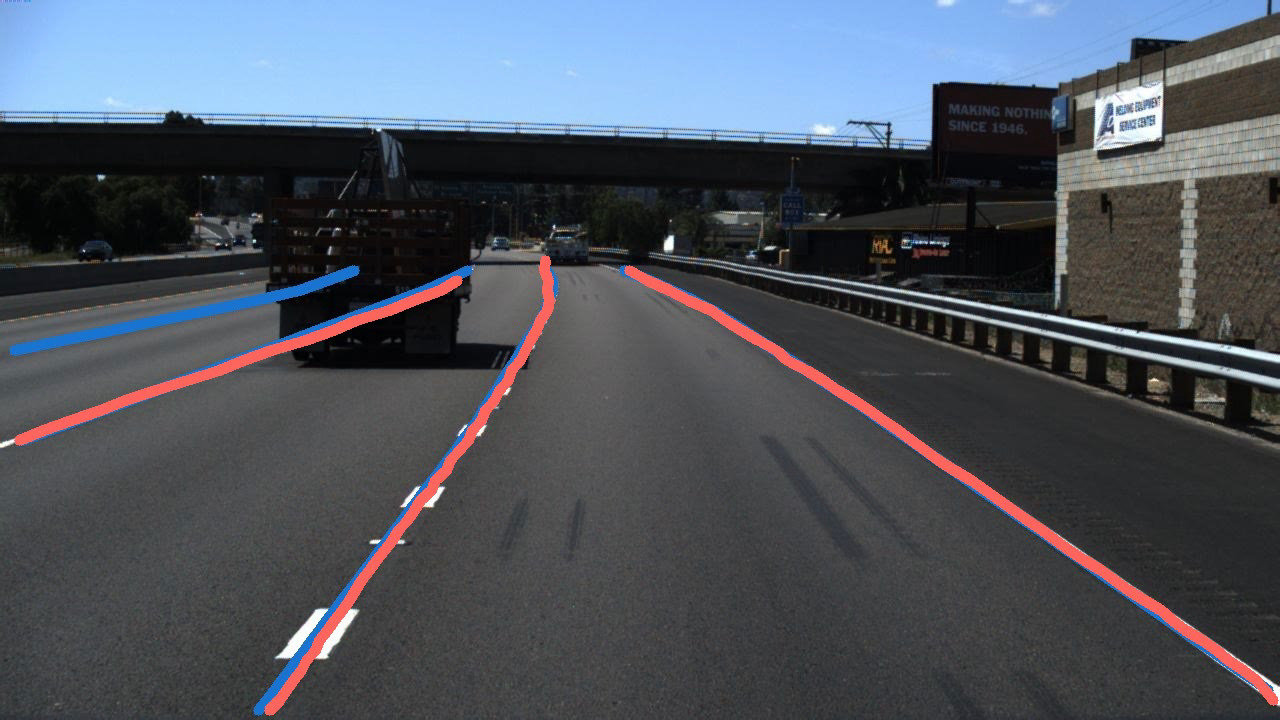} &
		\includegraphics[width=.18\linewidth,valign=m]{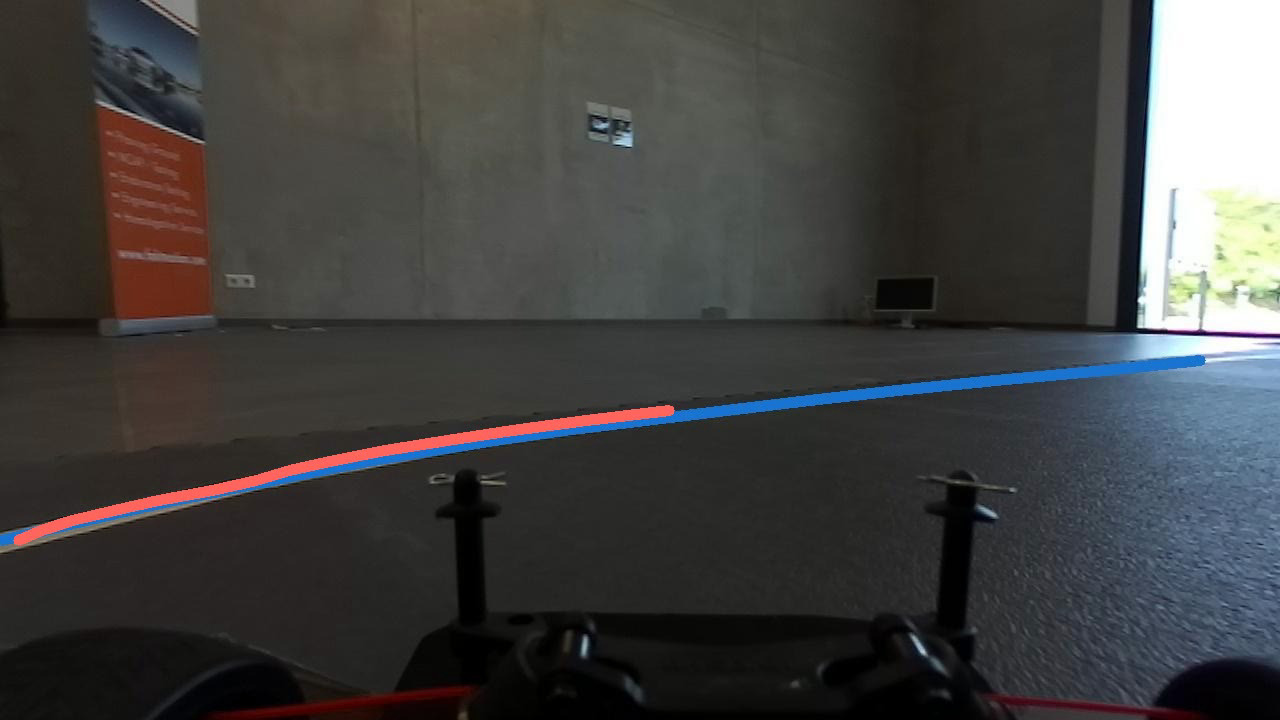} & \includegraphics[width=.18\linewidth,valign=m]{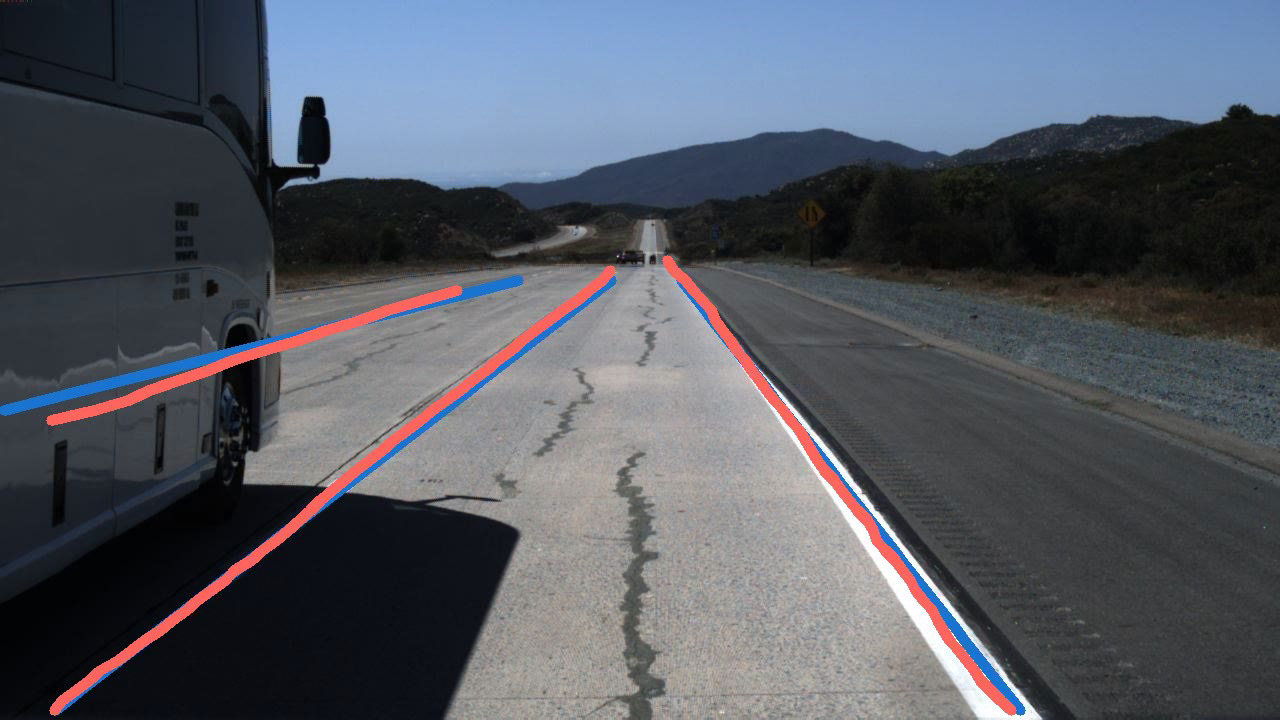}\\
		%
		\textbf{ADDA} & 
		\includegraphics[width=.18\linewidth,valign=m]{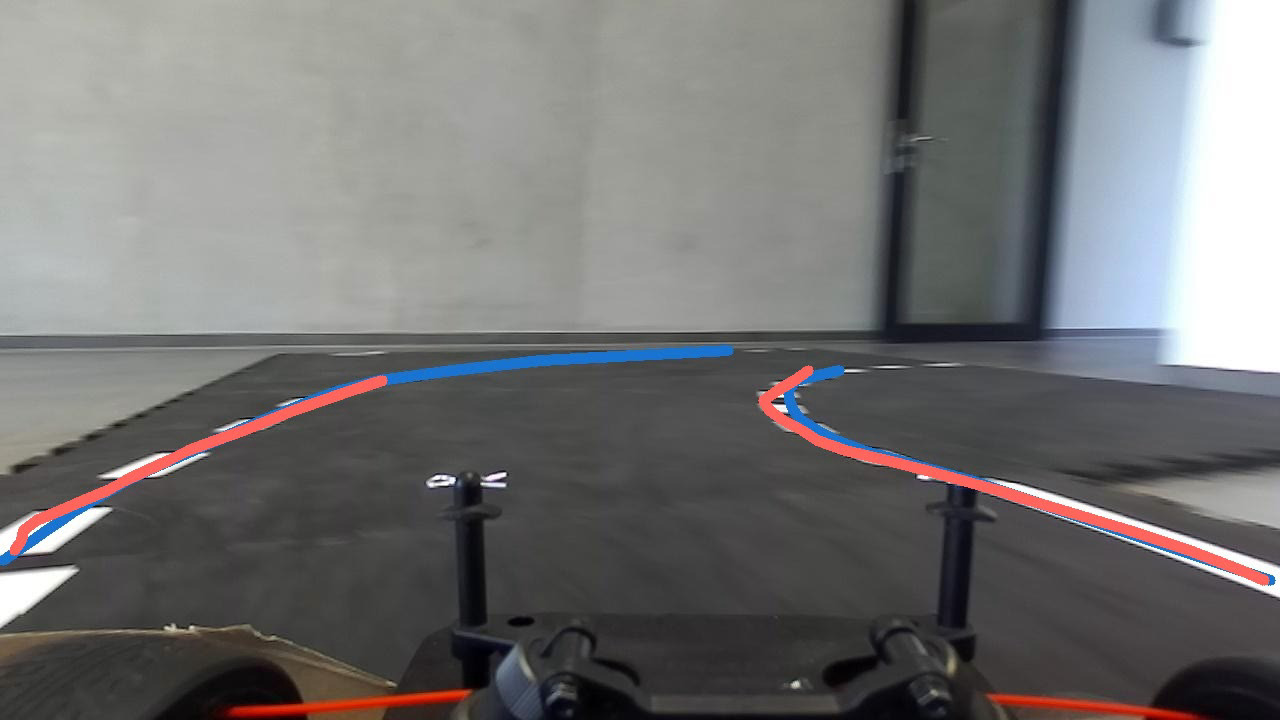} & 
		\includegraphics[width=.18\linewidth,valign=m]{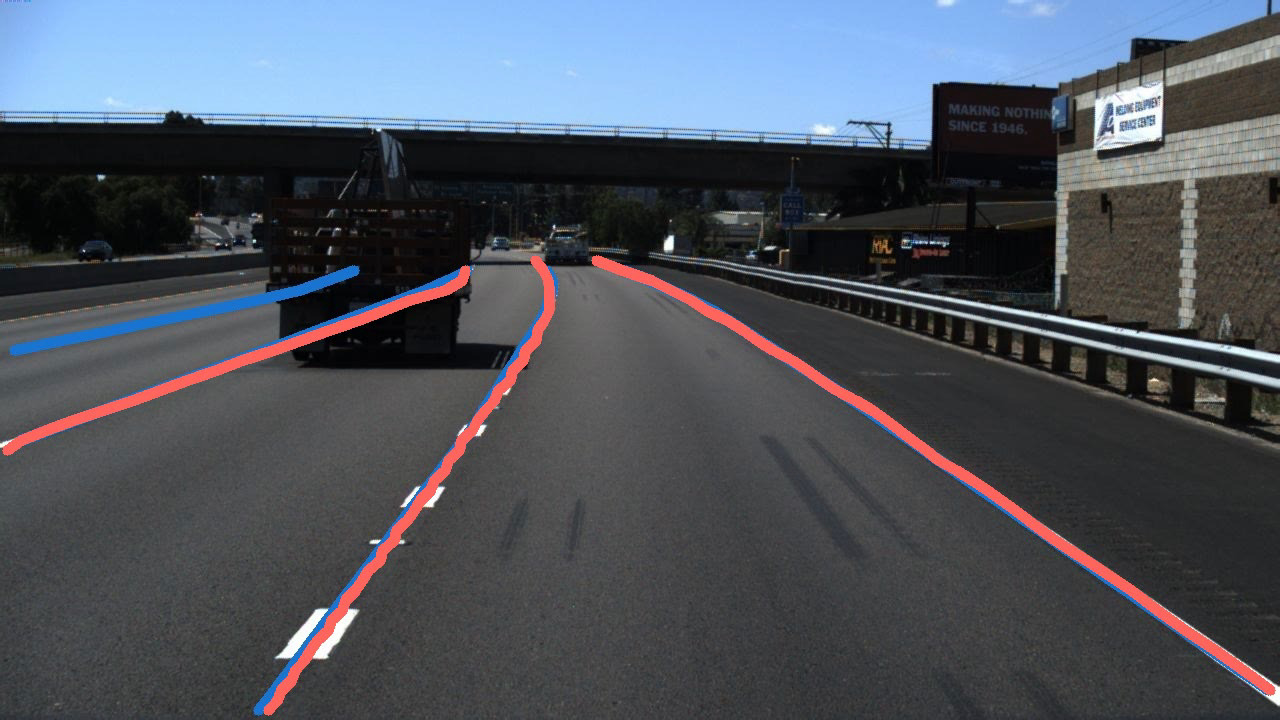} &
		\includegraphics[width=.18\linewidth,valign=m]{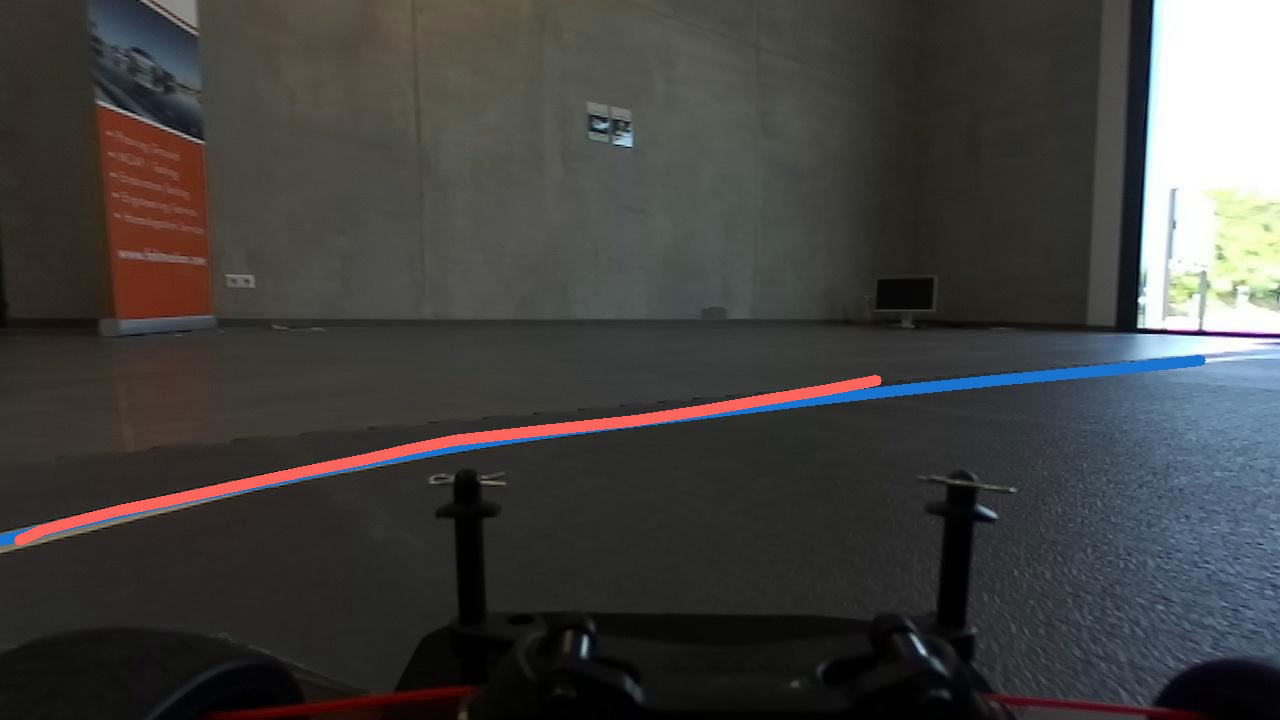} & \includegraphics[width=.18\linewidth,valign=m]{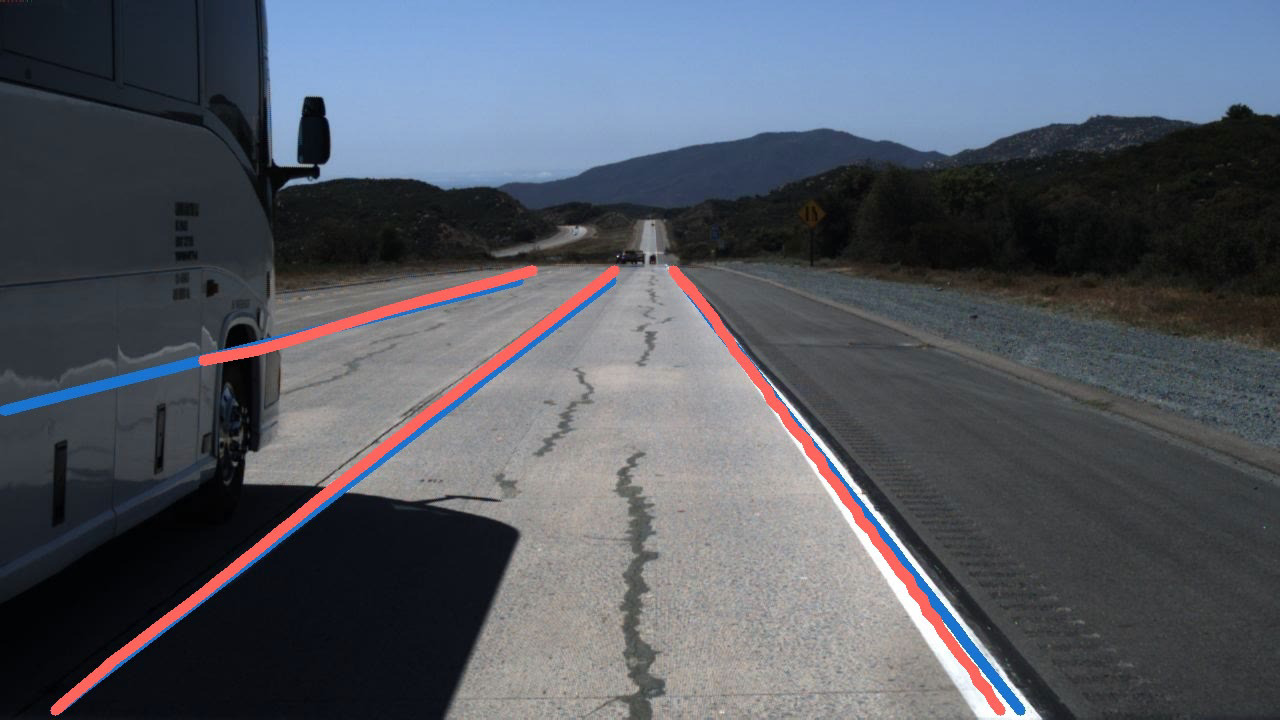}\\
		%
		\textbf{SGADA} & 
		\includegraphics[width=.18\linewidth,valign=m]{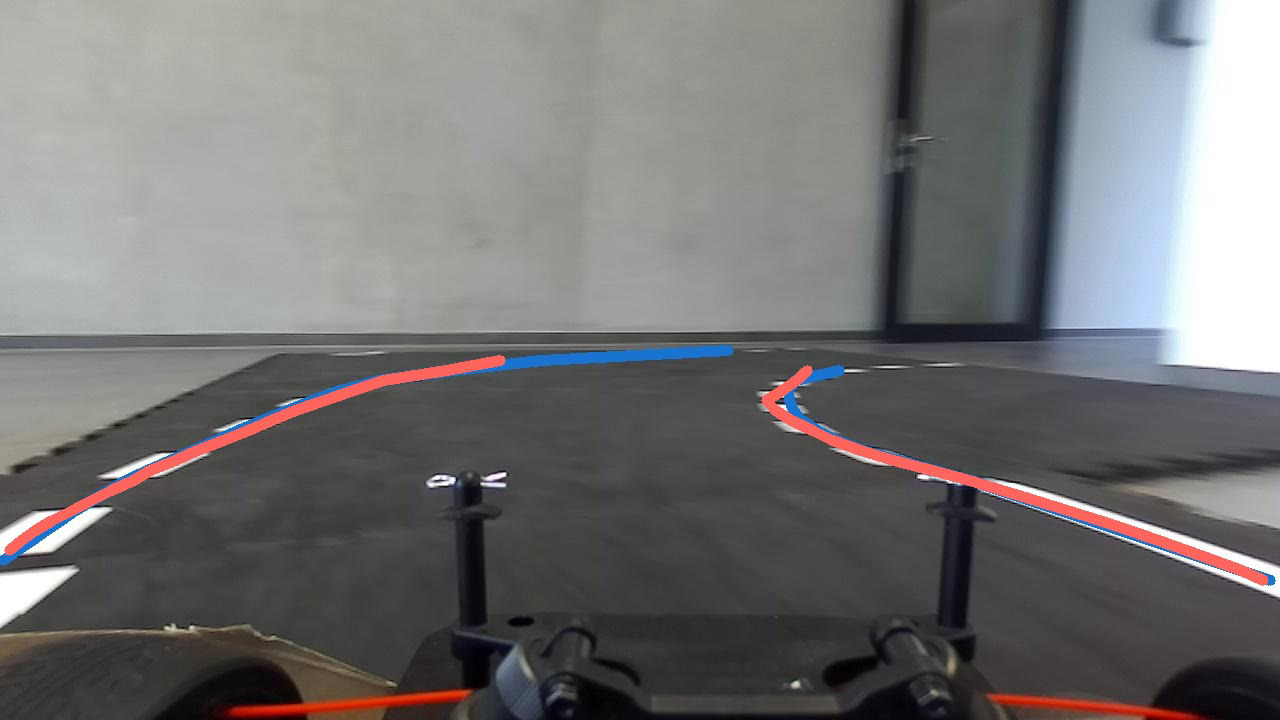} & 
		\includegraphics[width=.18\linewidth,valign=m]{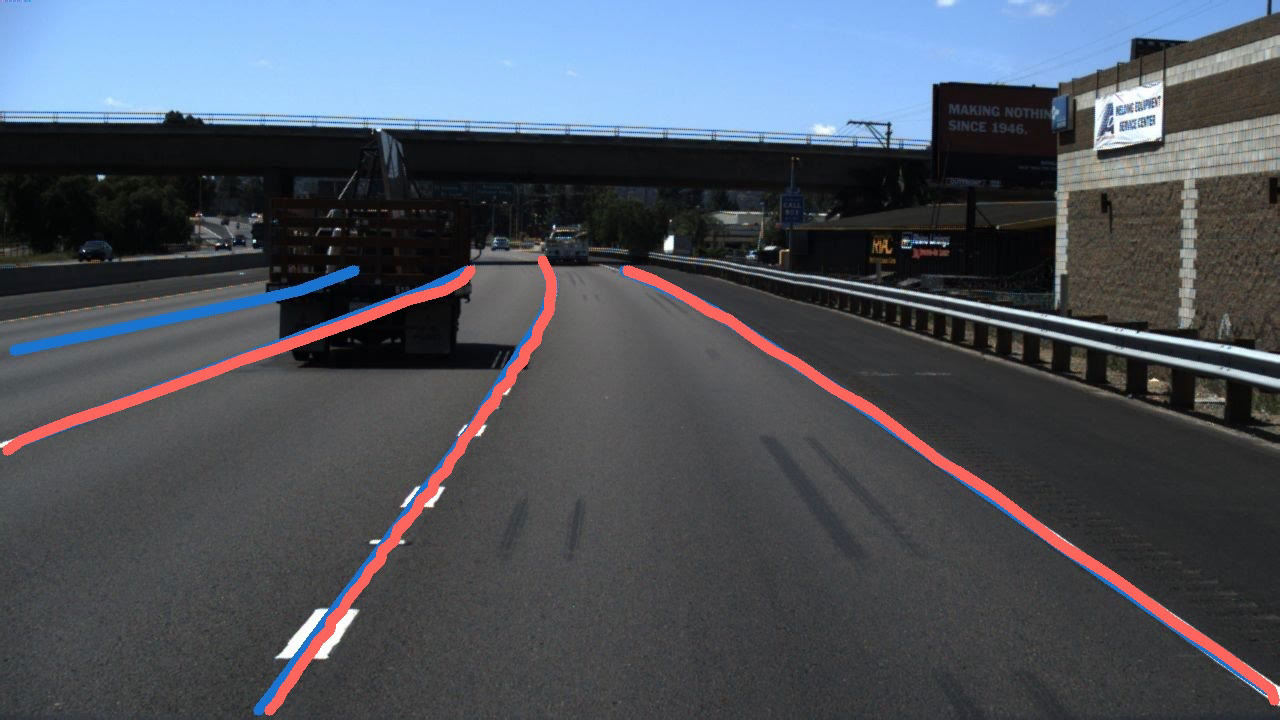} &
		\includegraphics[width=.18\linewidth,valign=m]{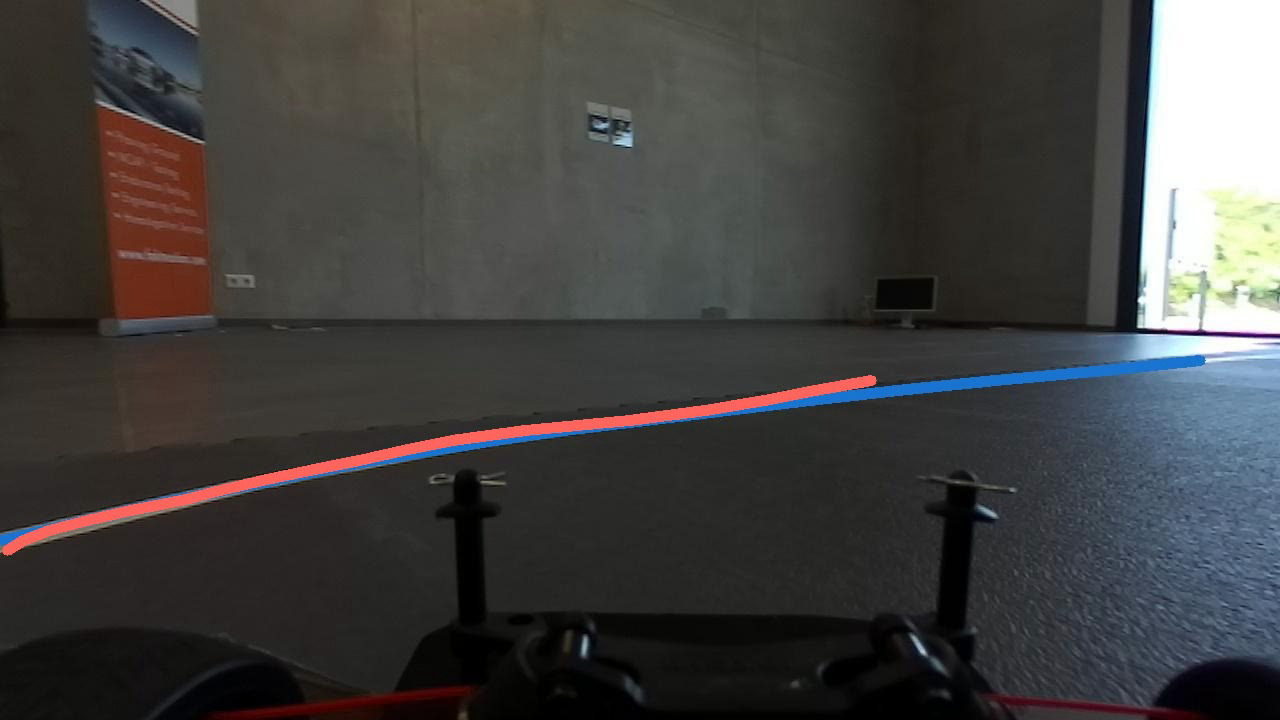} & \includegraphics[width=.18\linewidth,valign=m]{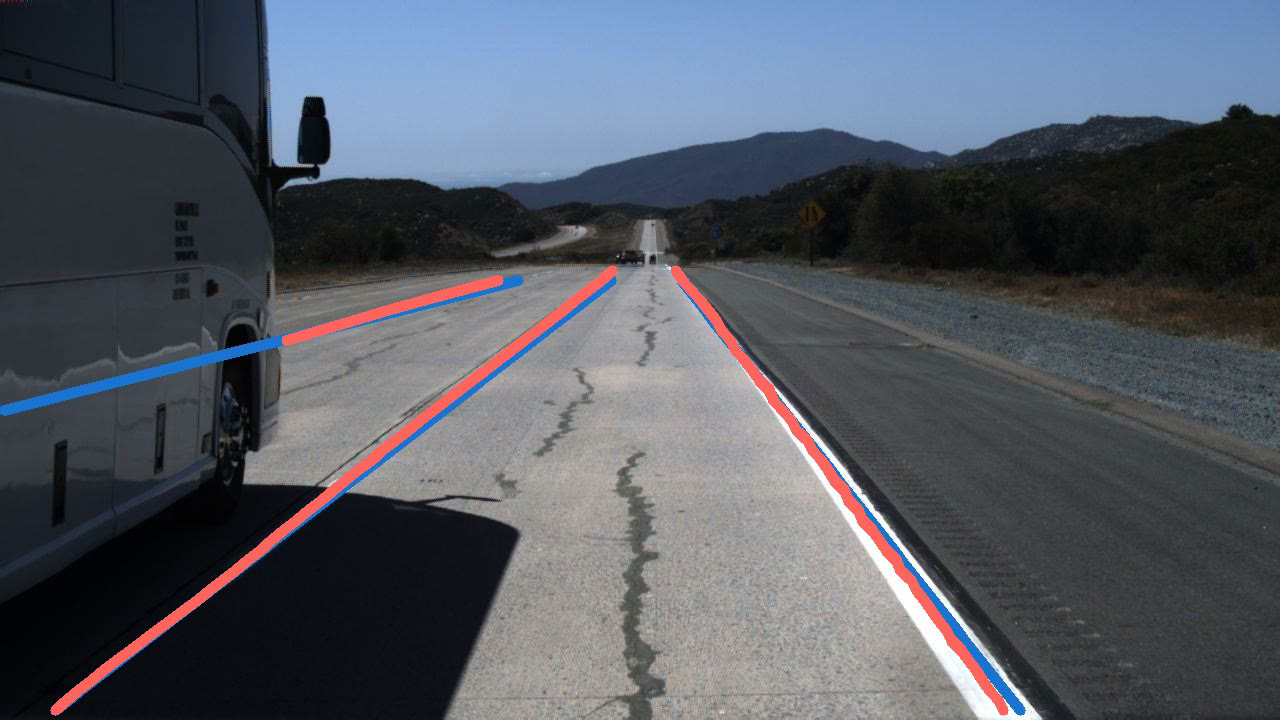}\\
		%
		\textbf{SGPCS} & 
		\includegraphics[width=.18\linewidth,valign=m]{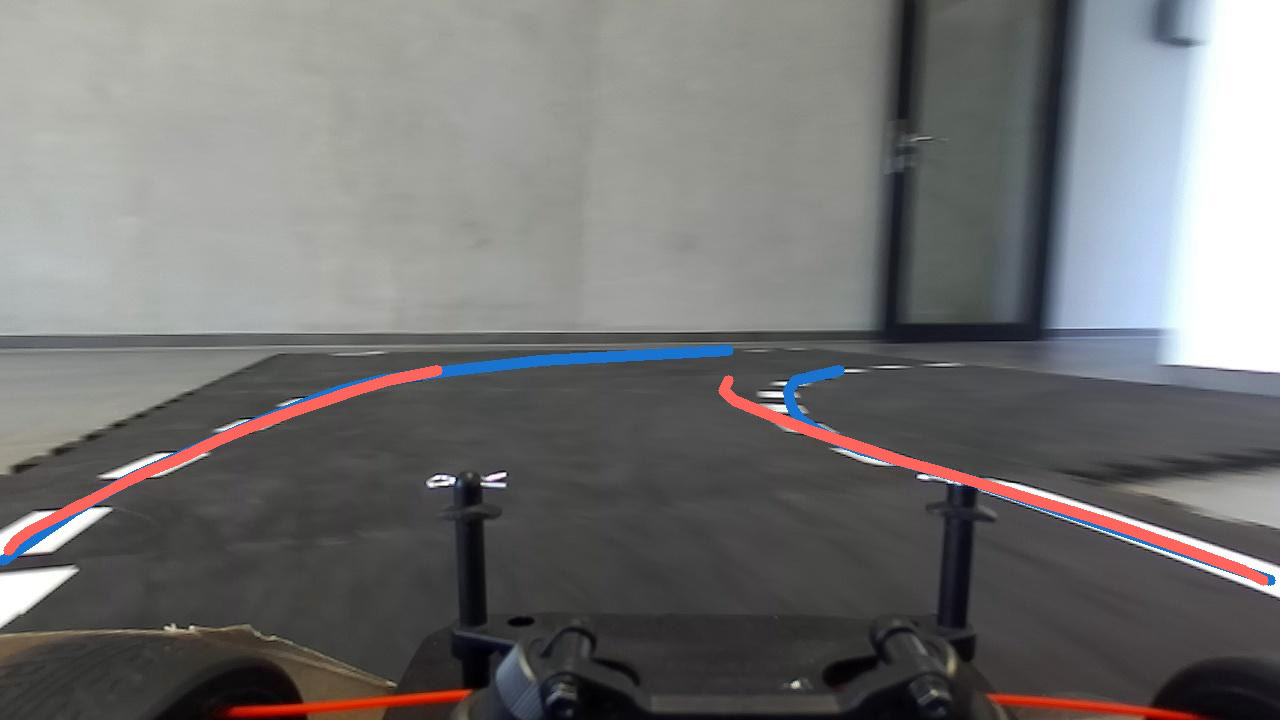} & \includegraphics[width=.18\linewidth,valign=m]{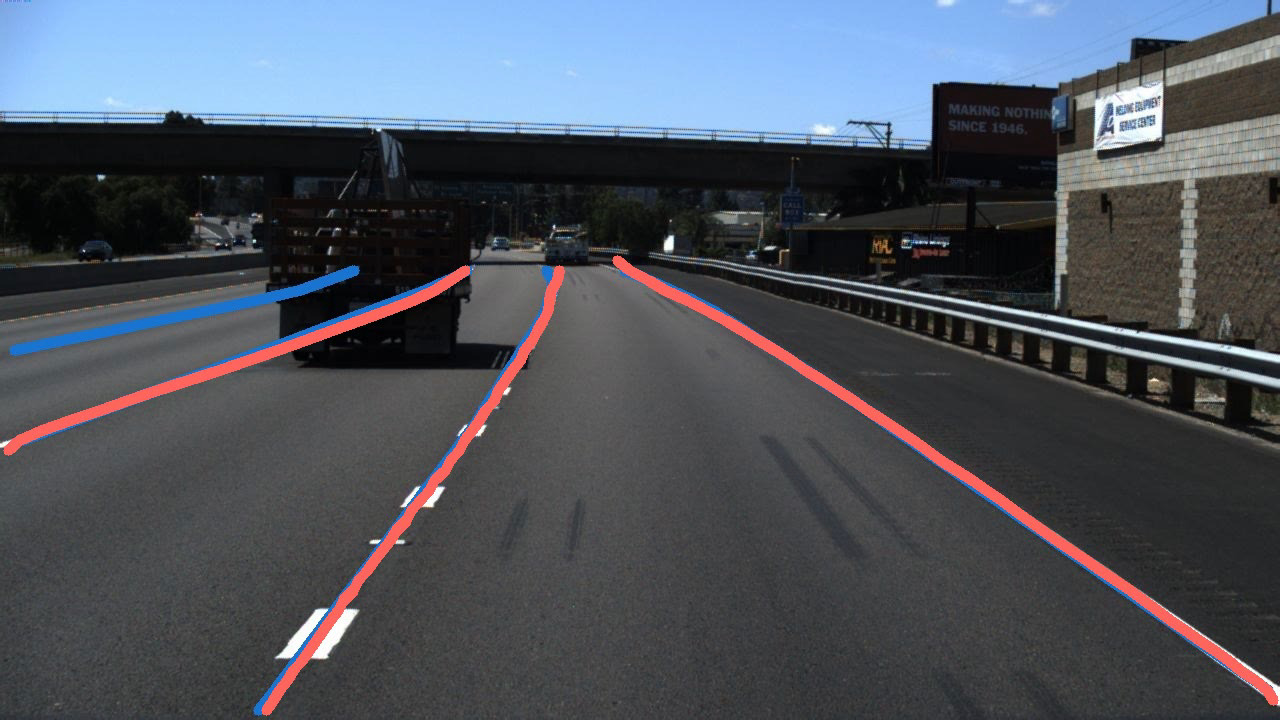} &
		\includegraphics[width=.18\linewidth,valign=m]{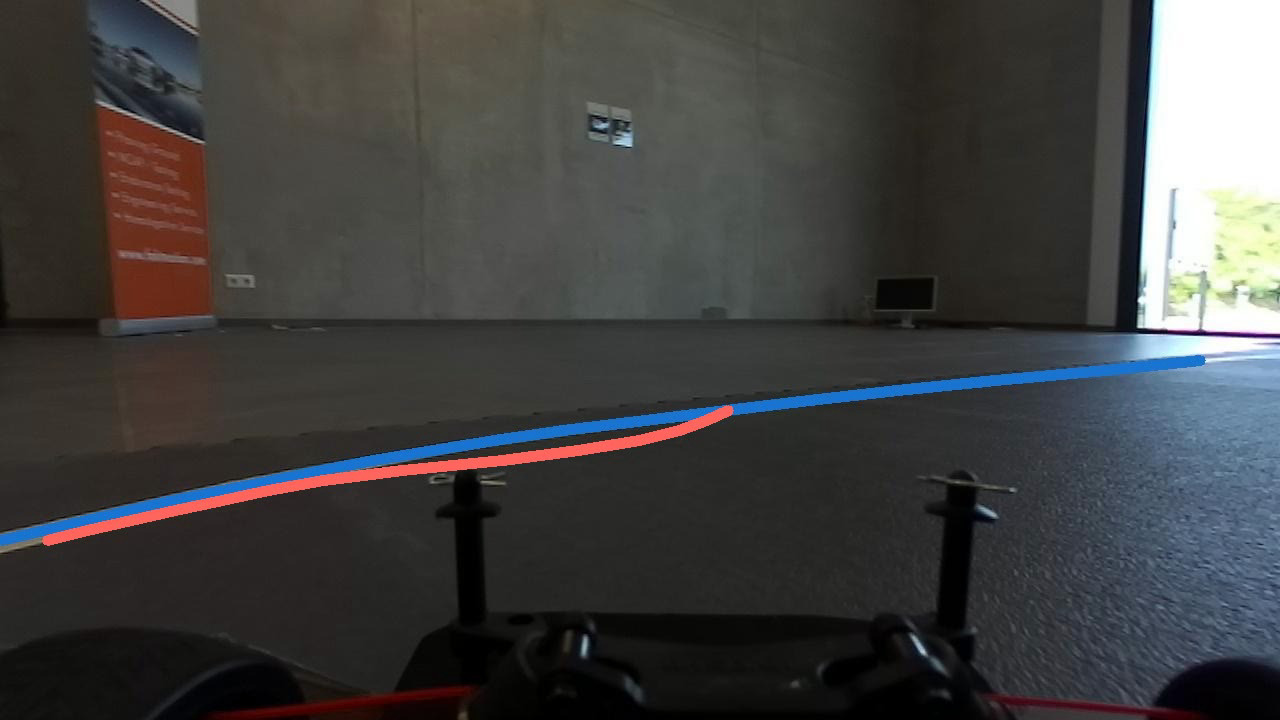} & \includegraphics[width=.18\linewidth,valign=m]{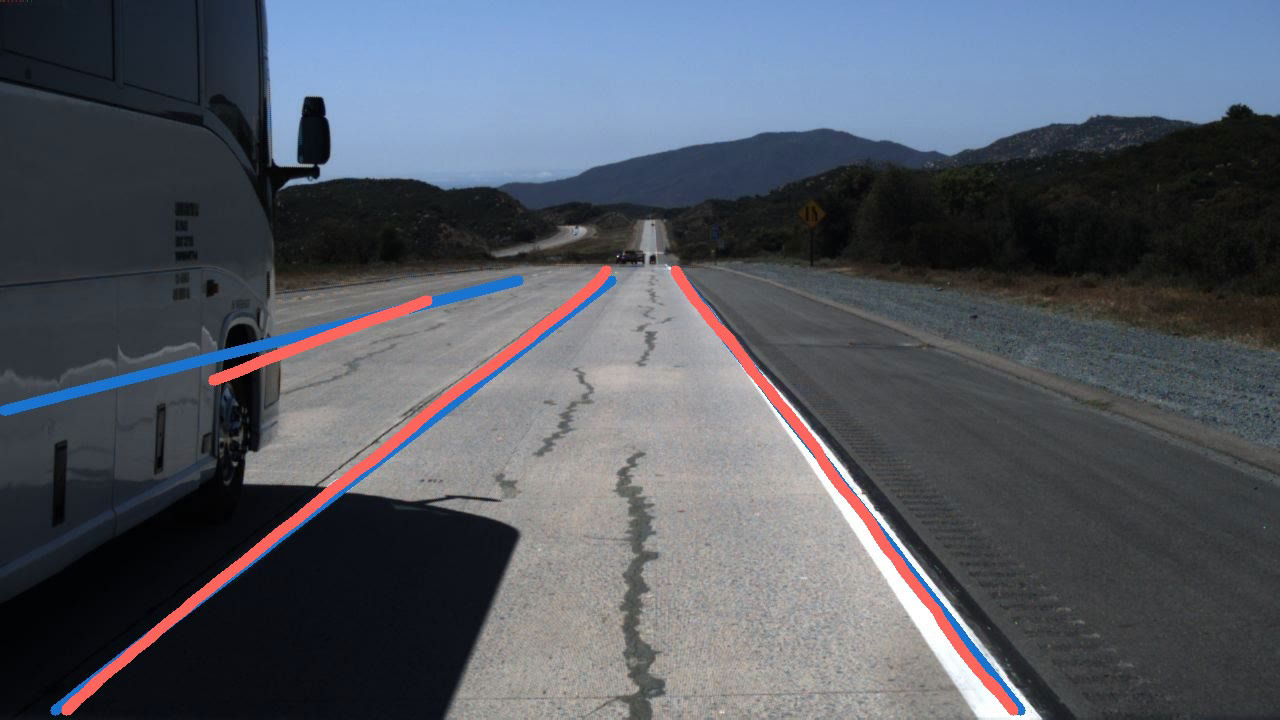}\\
		%
		\textbf{UFLD-TO} & 
		\includegraphics[width=.18\linewidth,valign=m]{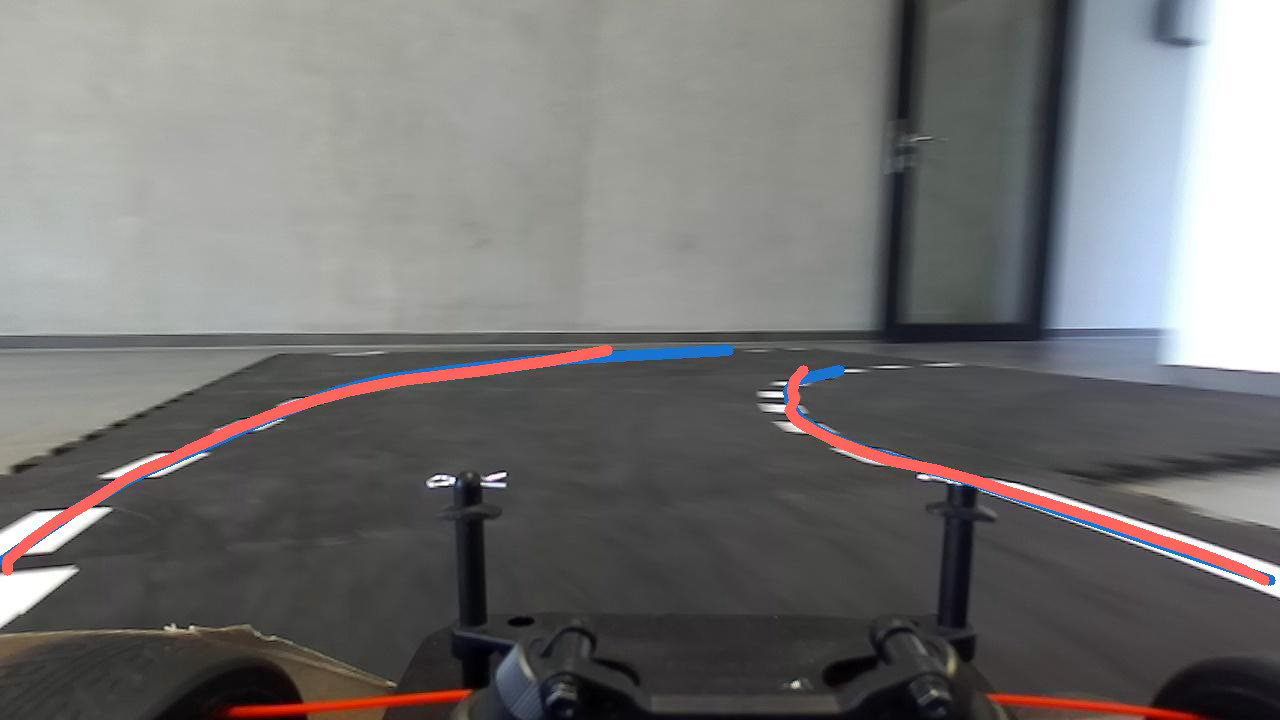} & \includegraphics[width=.18\linewidth,valign=m]{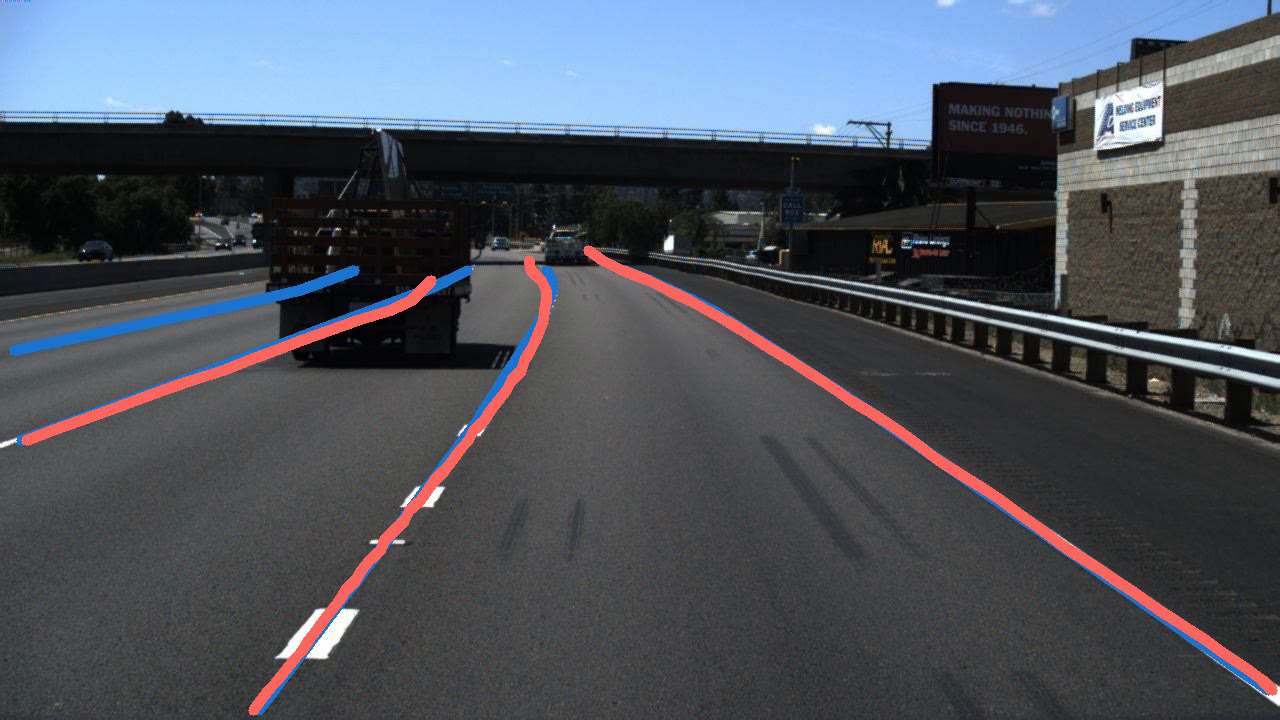} &
		\includegraphics[width=.18\linewidth,valign=m]{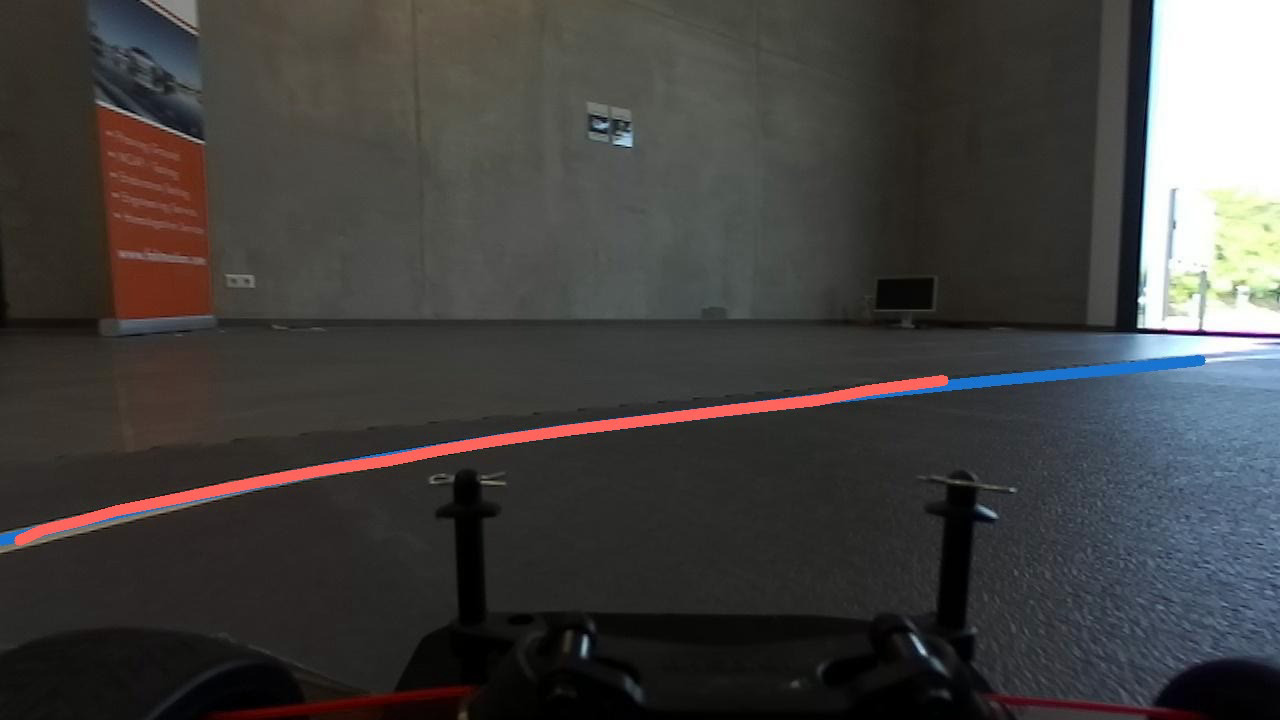} & \includegraphics[width=.18\linewidth,valign=m]{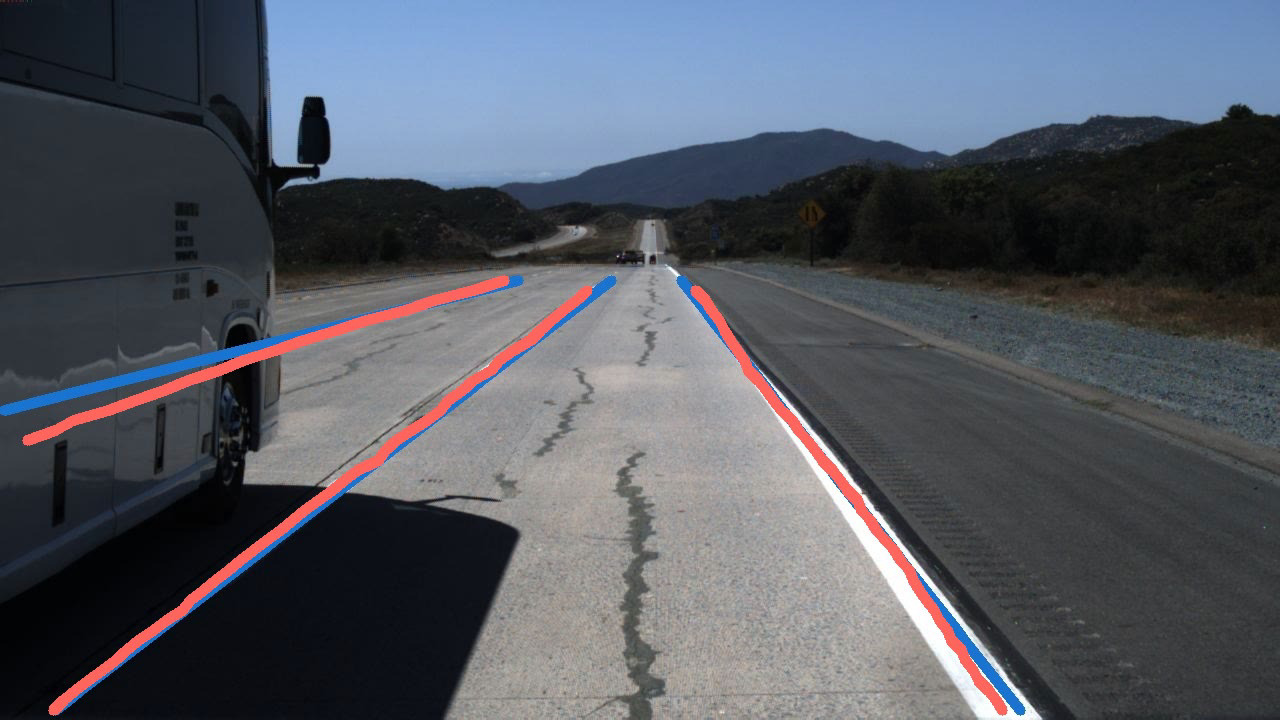}\\
	\end{tabular}
	\caption{Qualitative results of target domain predictions. Images are randomly sampled. Ground truth lane annotations are marked in blue, predictions in red.}
	\label{fig:appendix_inference_samples_2}
\end{figure}

\begin{figure}
	\centering
	\small
	\begin{tabular}{rc@{}c@{}c@{}c}
		~ & \textbf{MoLane} & \textbf{TuLane} & \multicolumn{2}{c}{\textbf{MuLane}} \\
		%
		\textbf{UFLD-SO} & 
		\includegraphics[width=.18\linewidth,valign=m]{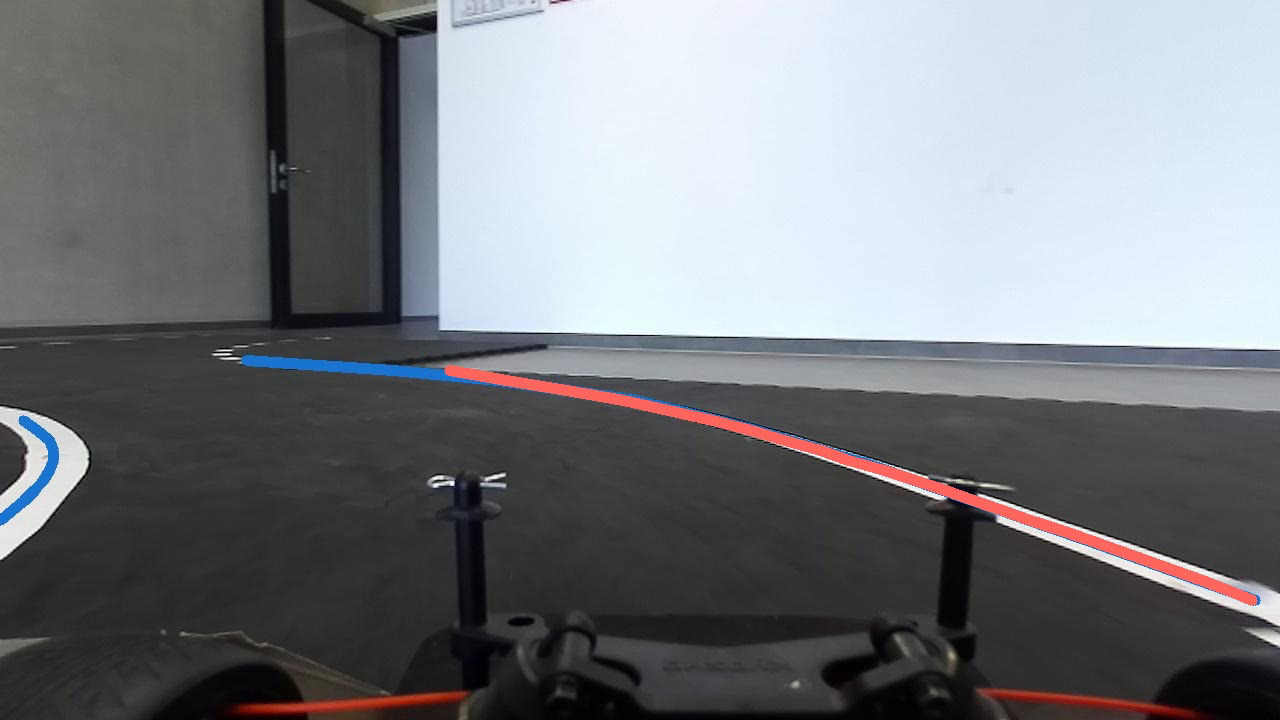} & \includegraphics[width=.18\linewidth,valign=m]{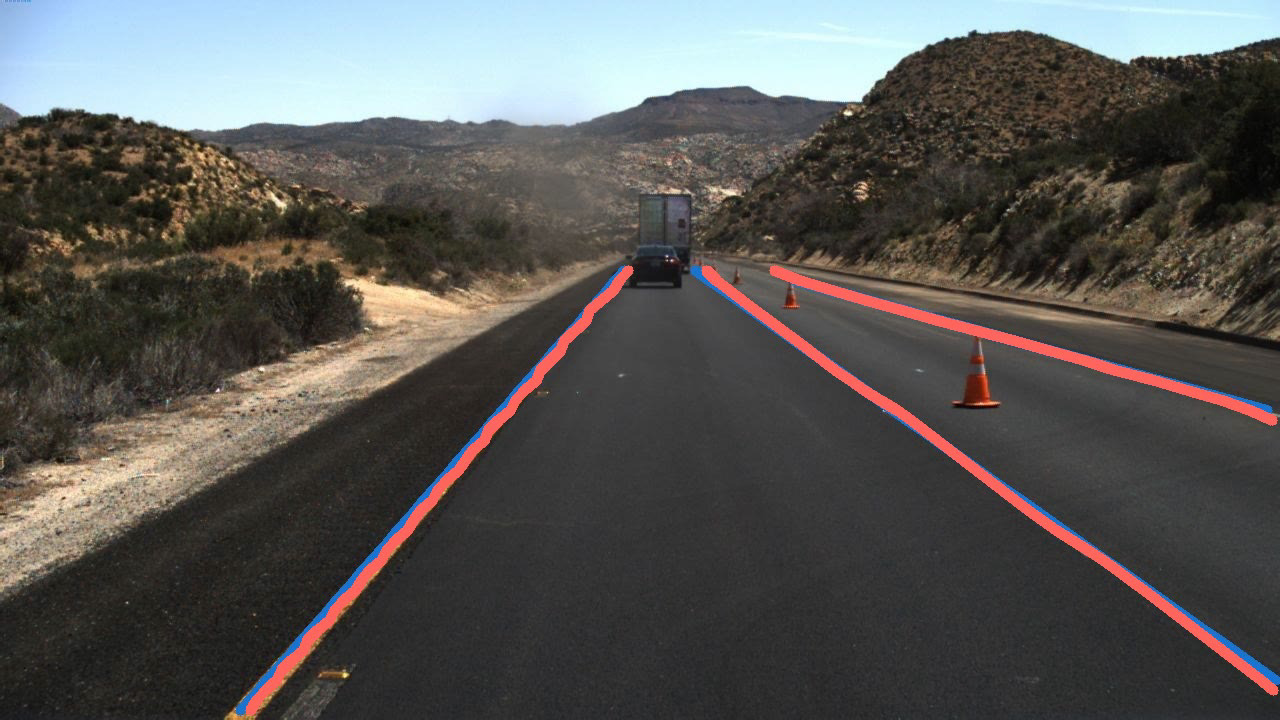} &
		\includegraphics[width=.18\linewidth,valign=m]{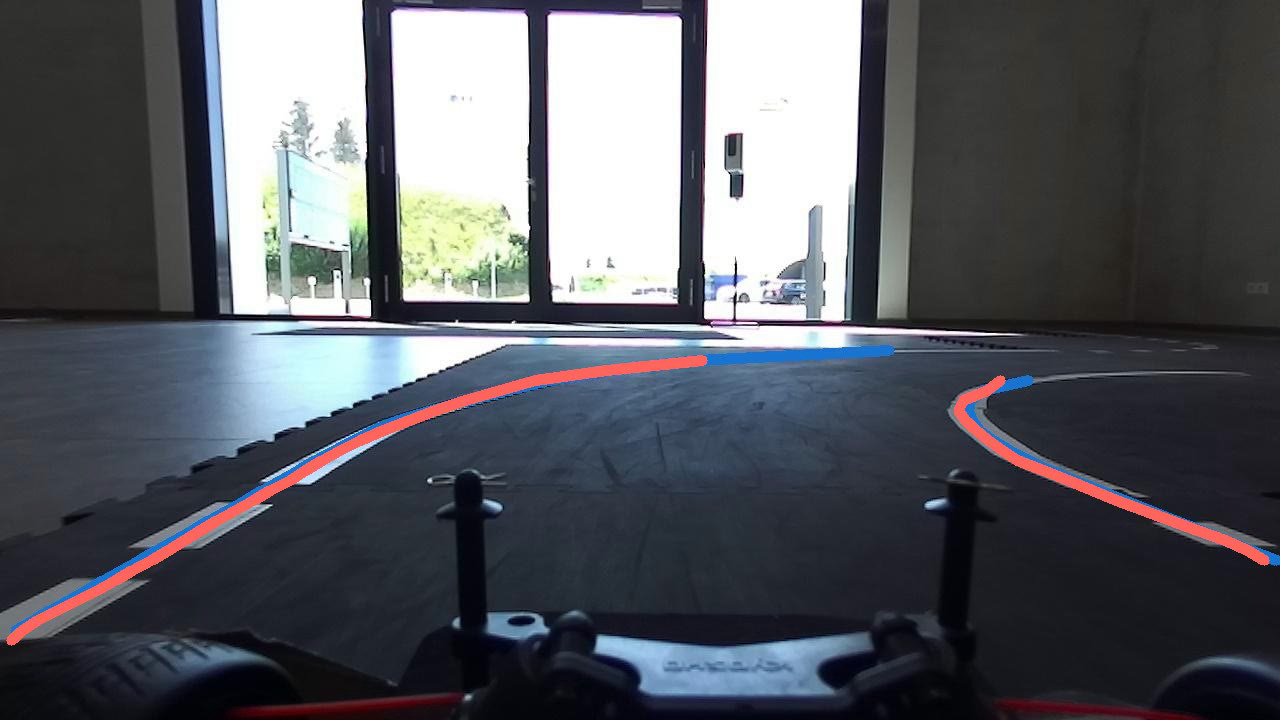} & \includegraphics[width=.18\linewidth,valign=m]{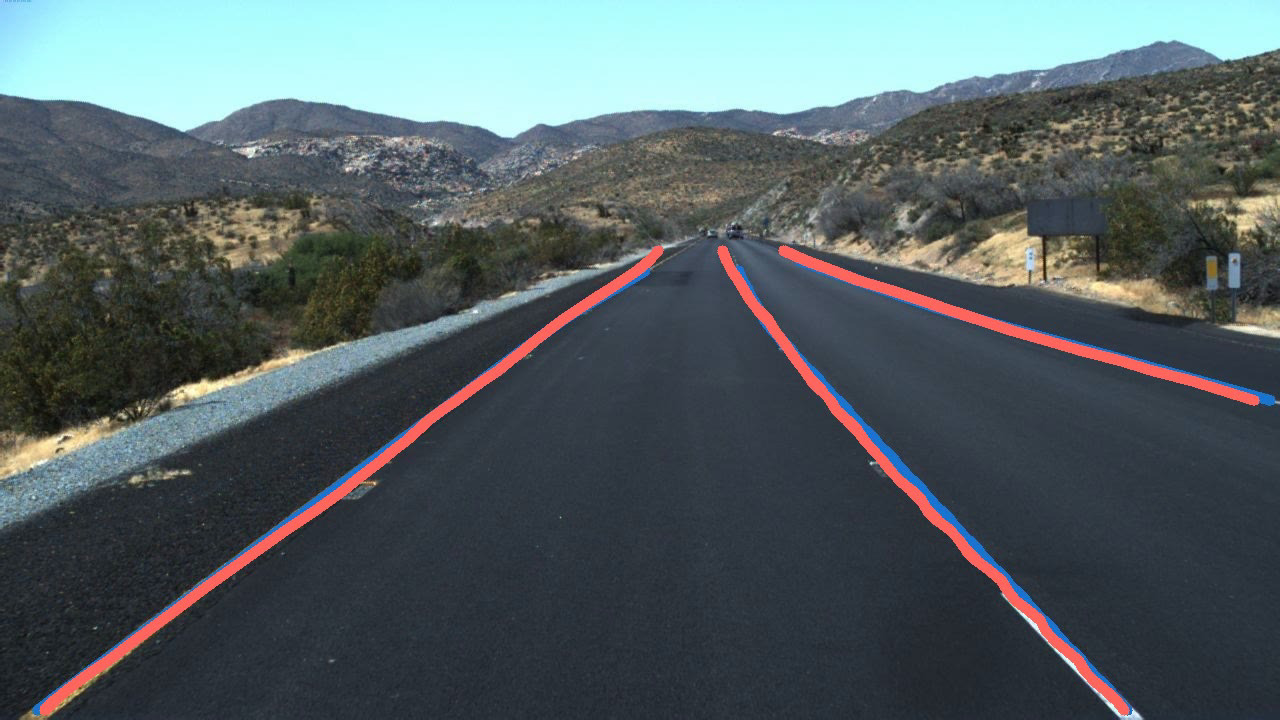}\\
		%
		\textbf{DANN} & 
		\includegraphics[width=.18\linewidth,valign=m]{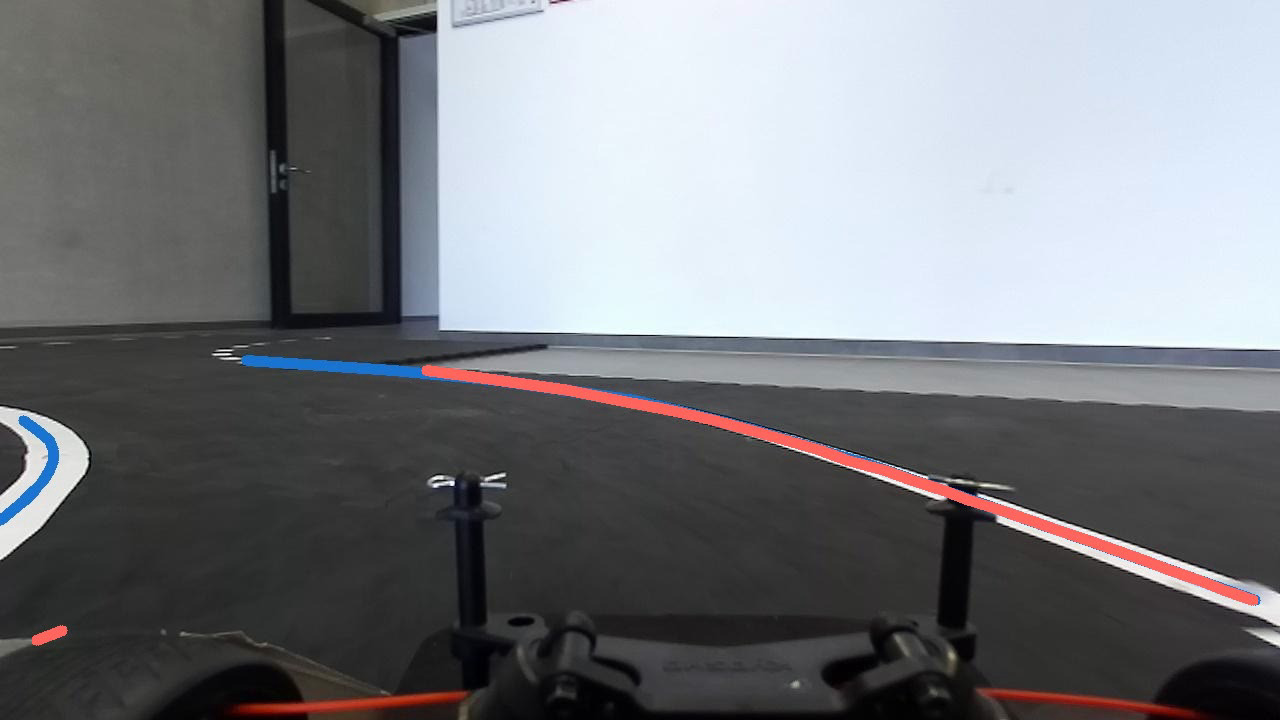} & 
		\includegraphics[width=.18\linewidth,valign=m]{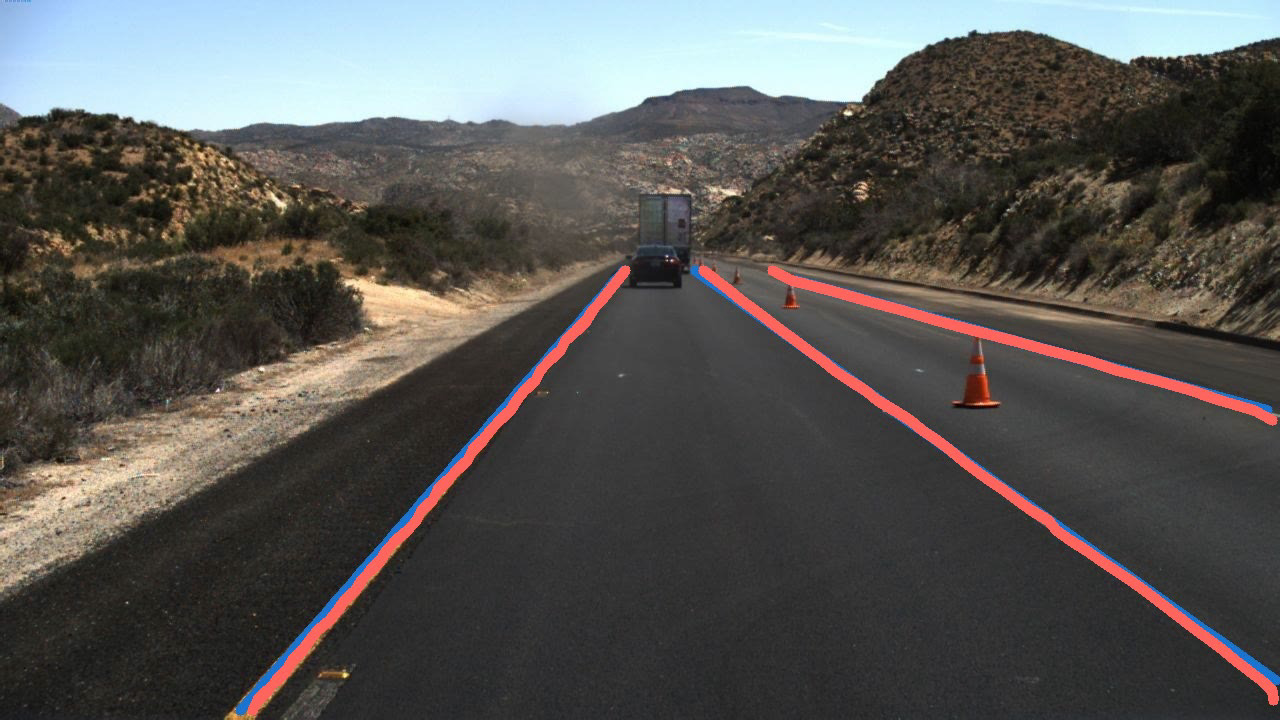} &
		\includegraphics[width=.18\linewidth,valign=m]{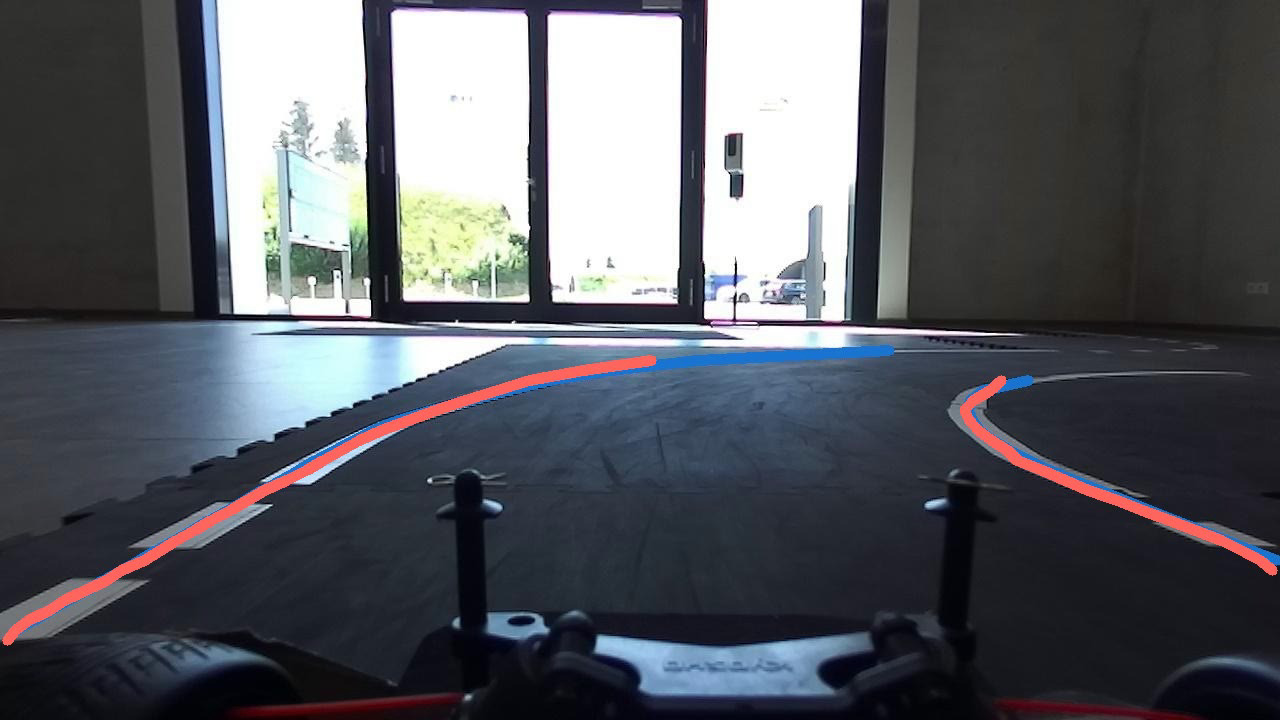} & \includegraphics[width=.18\linewidth,valign=m]{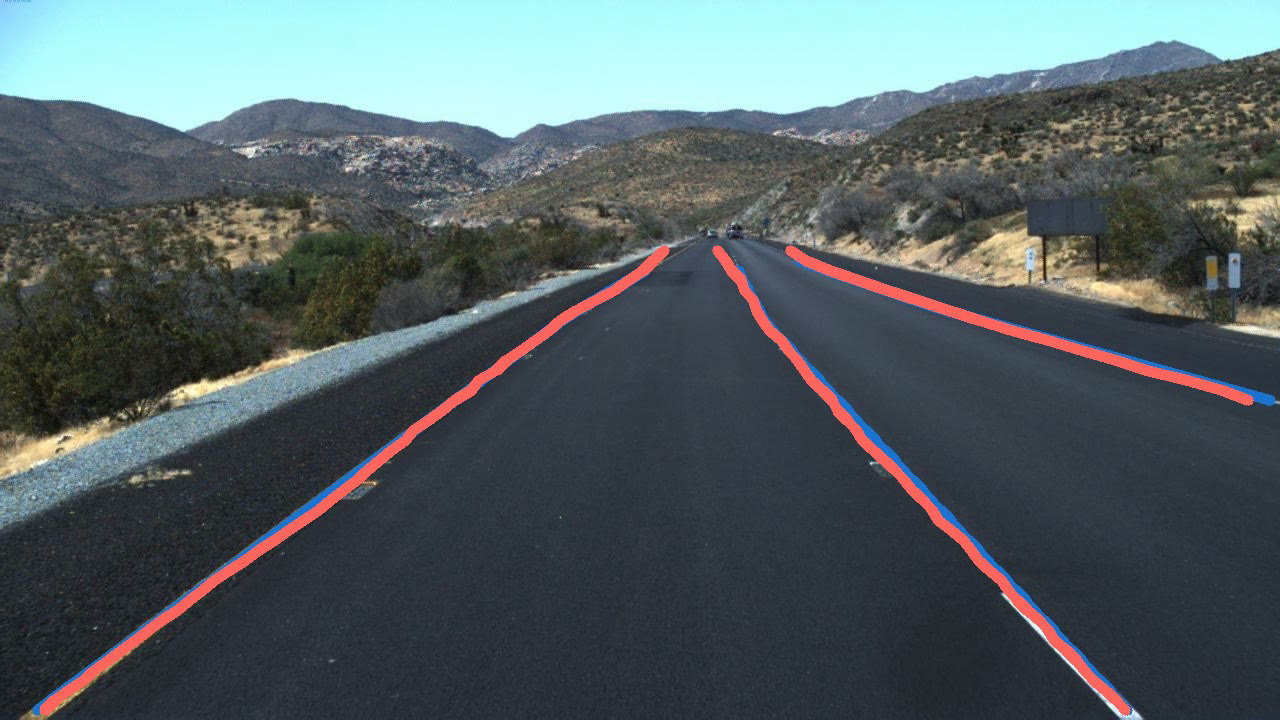}\\
		%
		\textbf{ADDA} & 
		\includegraphics[width=.18\linewidth,valign=m]{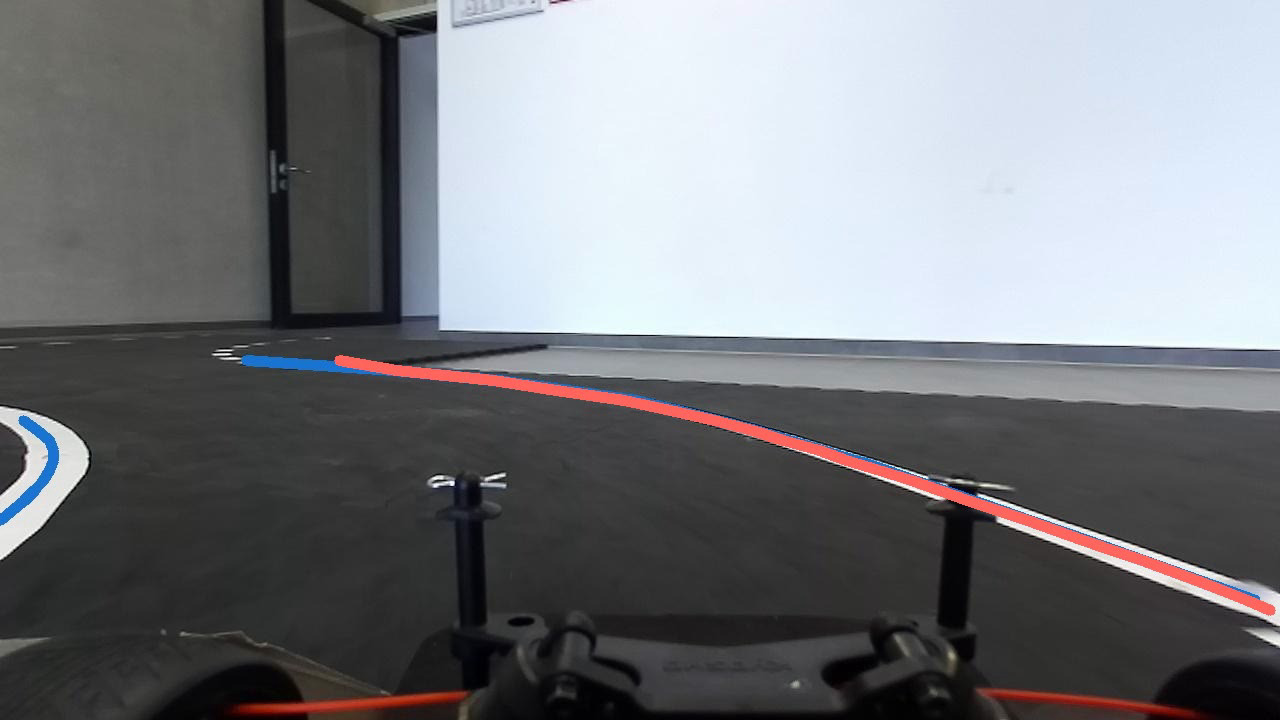} & 
		\includegraphics[width=.18\linewidth,valign=m]{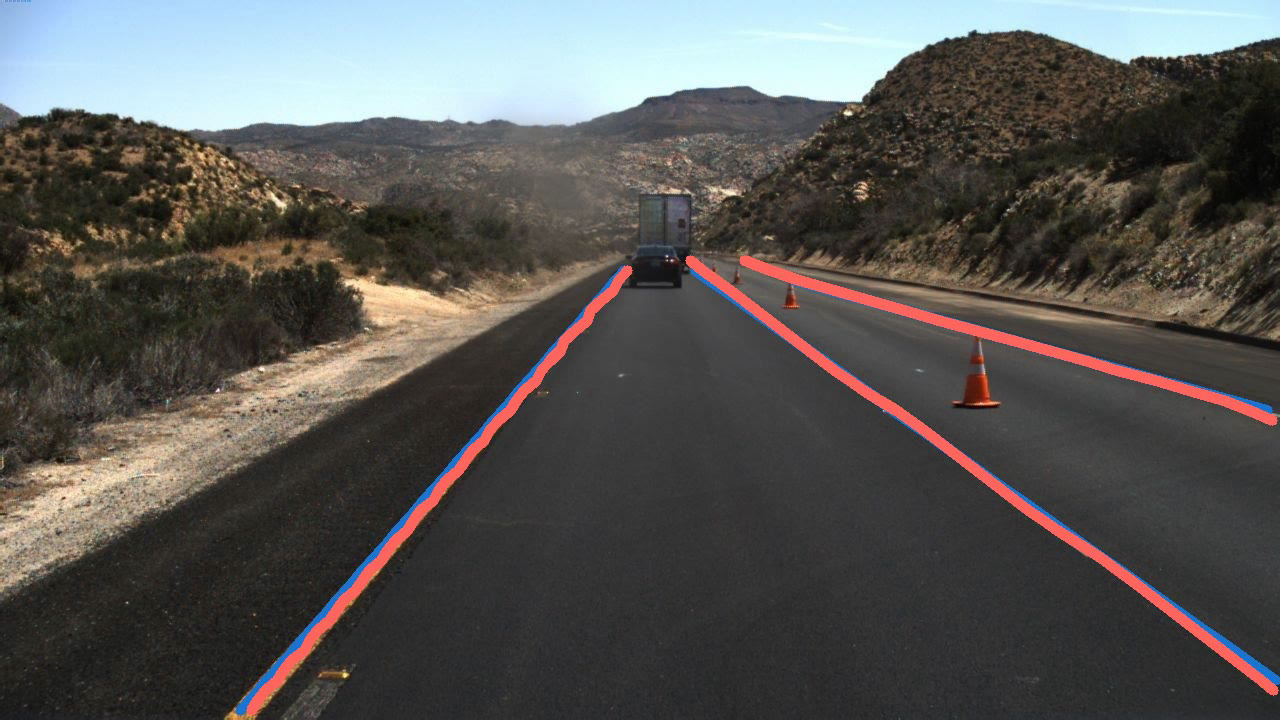} &
		\includegraphics[width=.18\linewidth,valign=m]{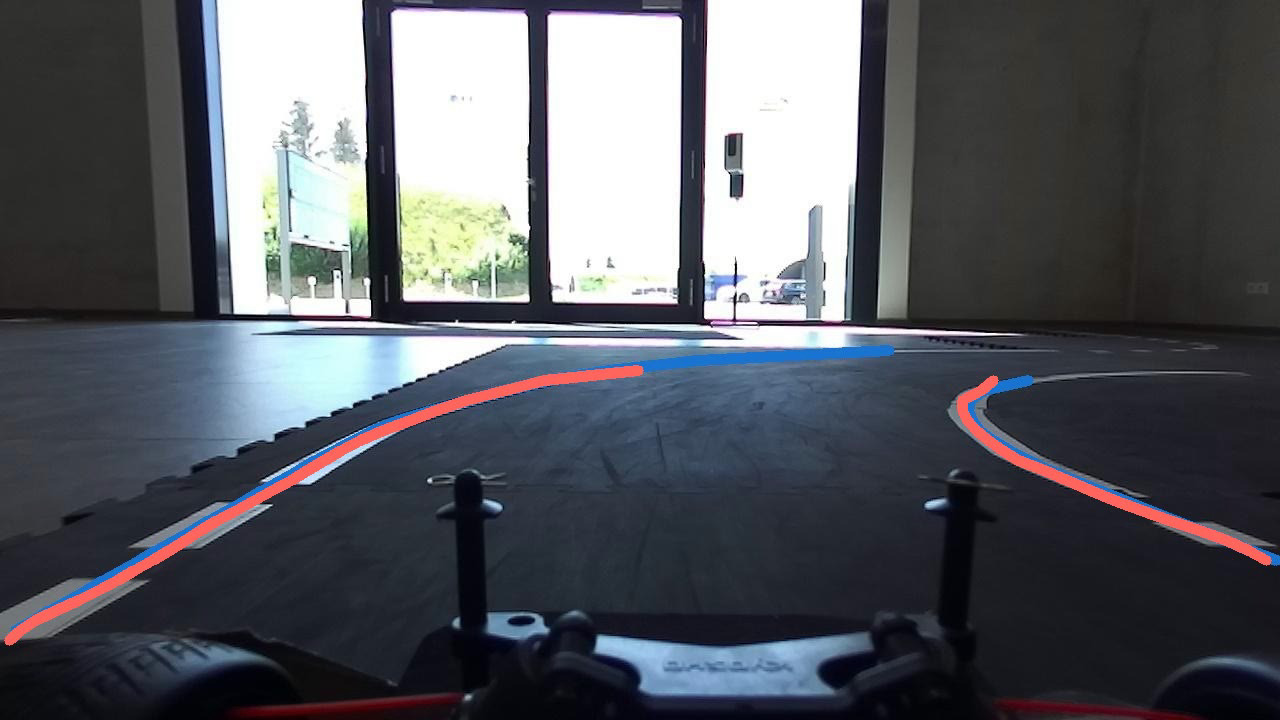} & \includegraphics[width=.18\linewidth,valign=m]{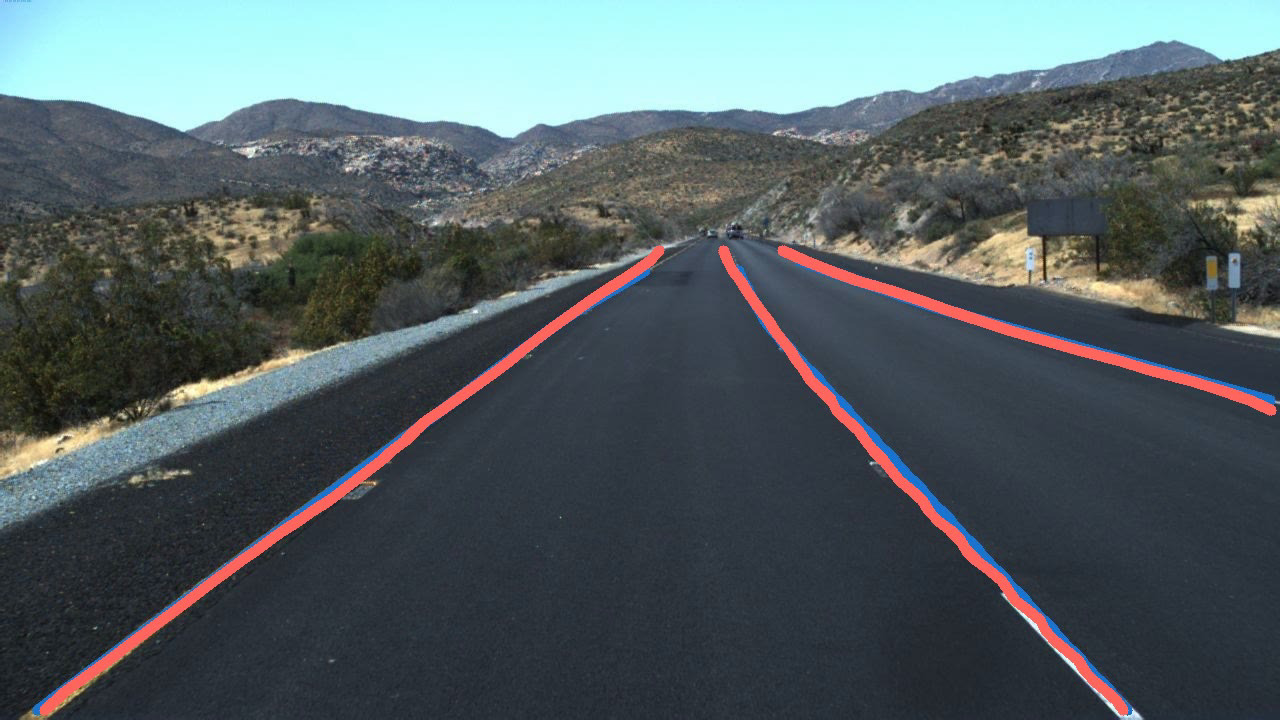}\\
		%
		\textbf{SGADA} & 
		\includegraphics[width=.18\linewidth,valign=m]{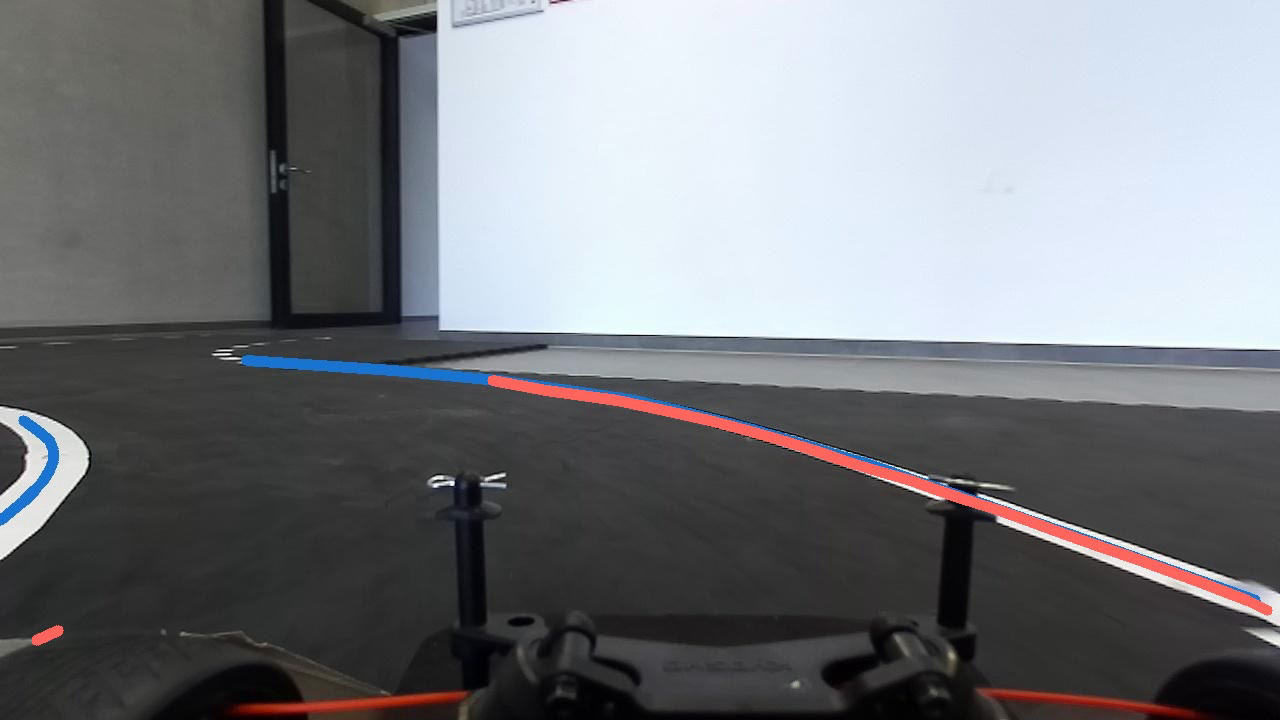} & 
		\includegraphics[width=.18\linewidth,valign=m]{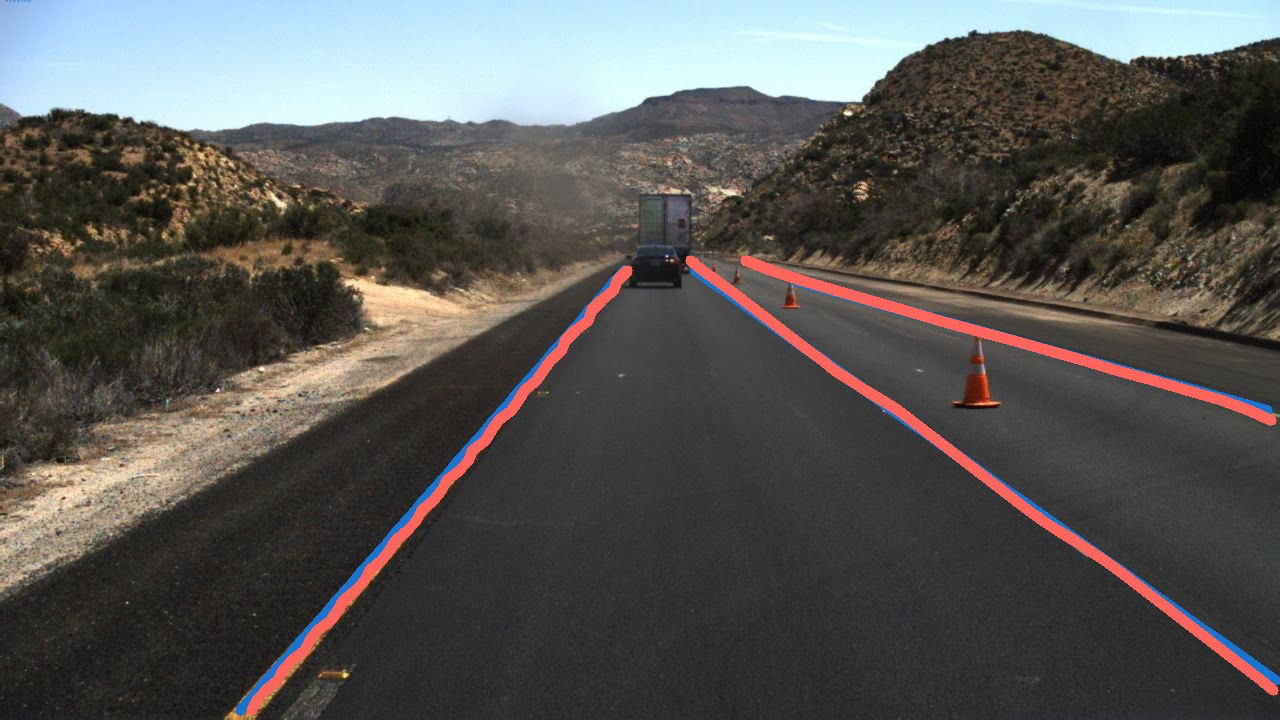} &
		\includegraphics[width=.18\linewidth,valign=m]{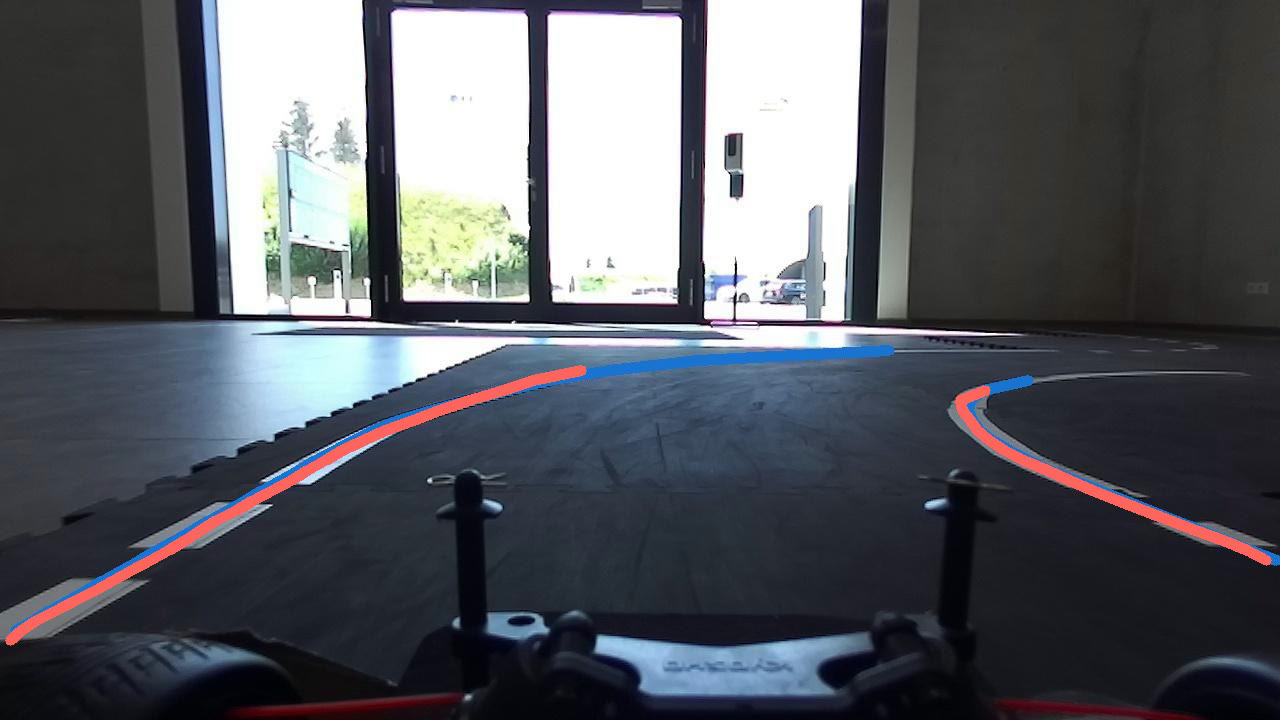} & \includegraphics[width=.18\linewidth,valign=m]{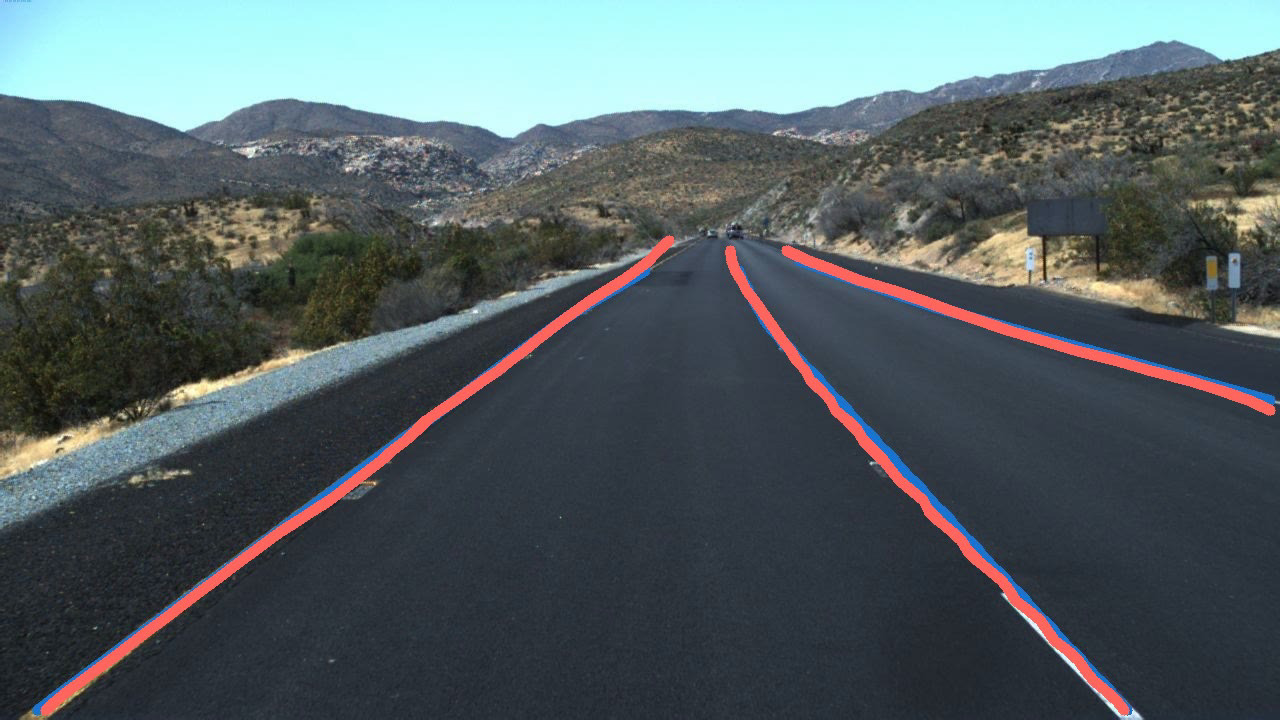}\\
		%
		\textbf{SGPCS} & 
		\includegraphics[width=.18\linewidth,valign=m]{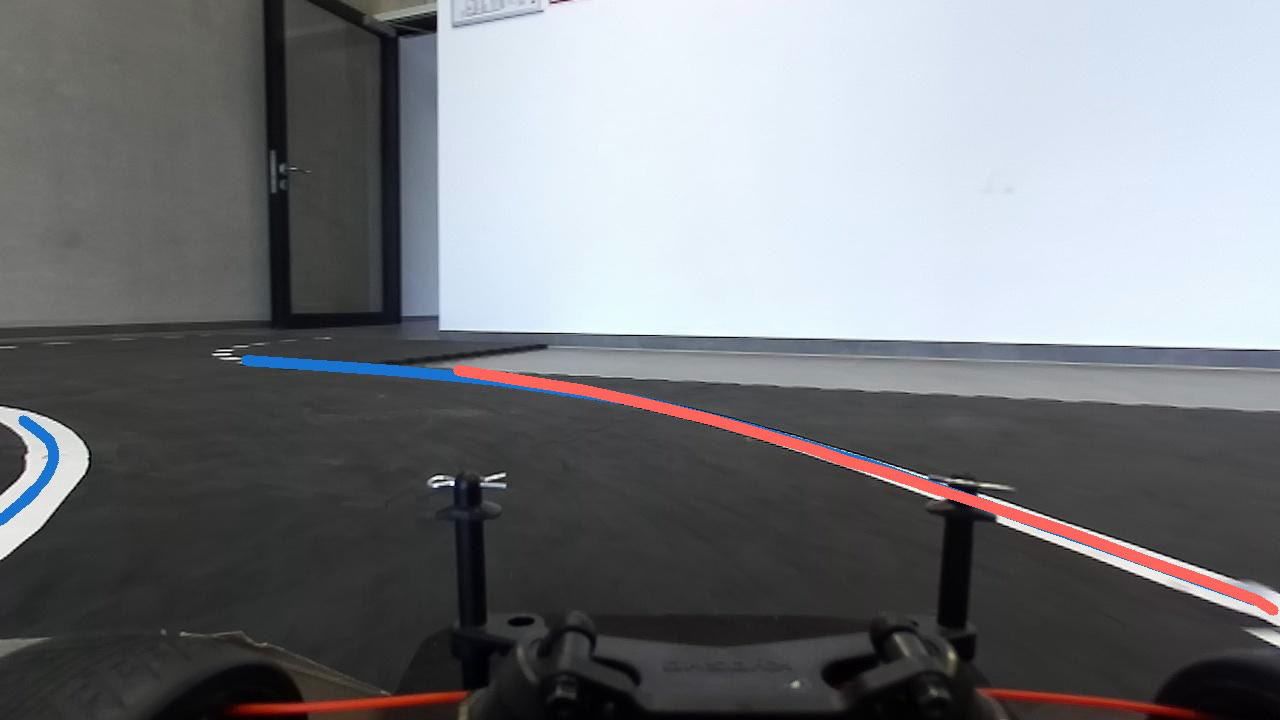} & \includegraphics[width=.18\linewidth,valign=m]{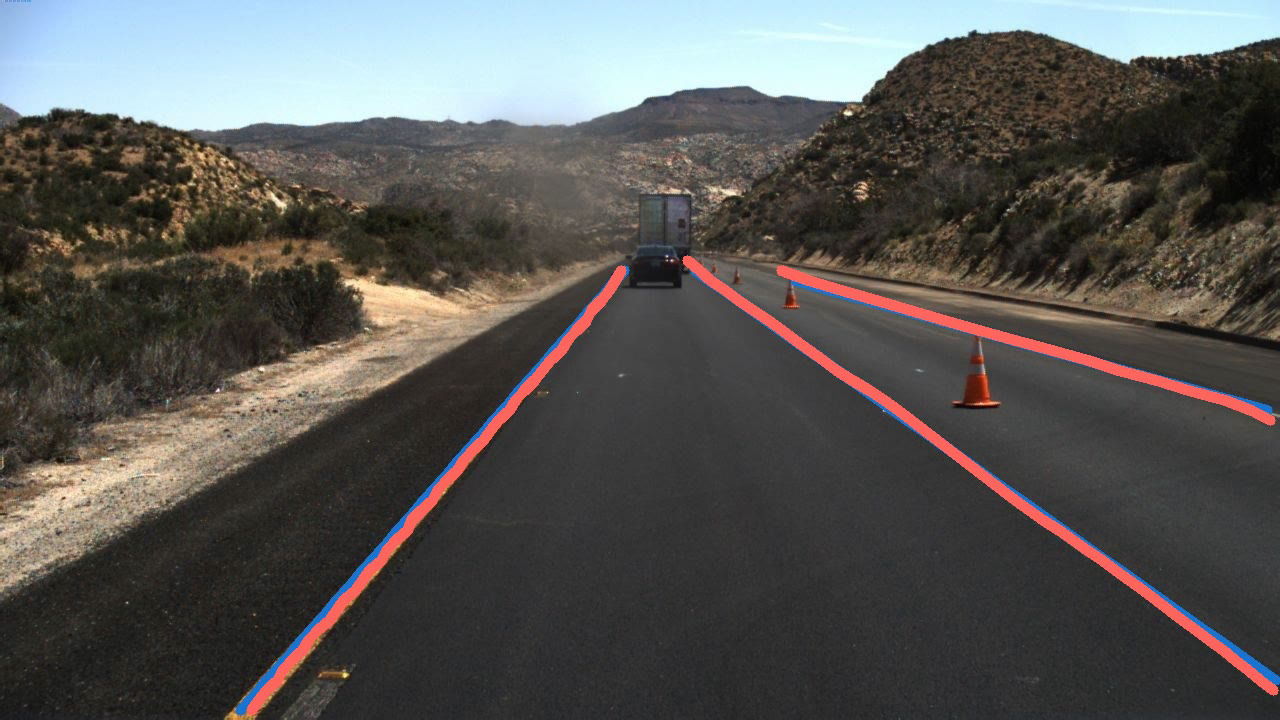} &
		\includegraphics[width=.18\linewidth,valign=m]{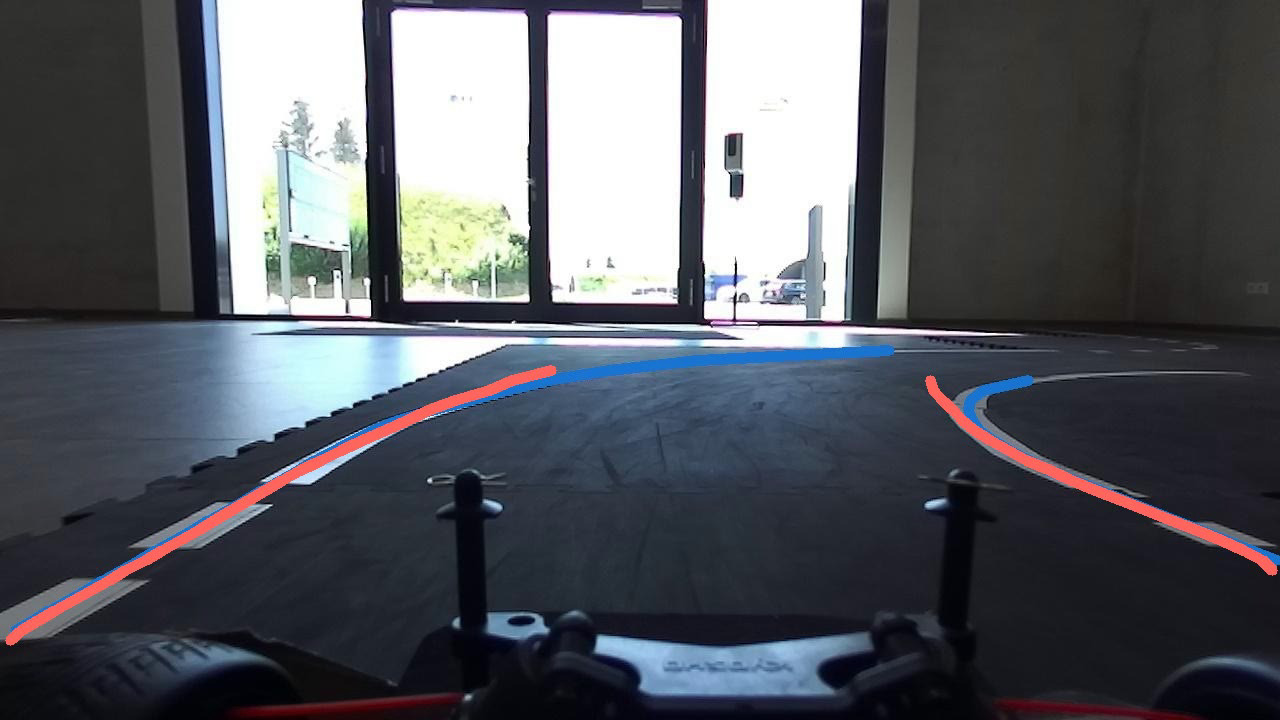} & \includegraphics[width=.18\linewidth,valign=m]{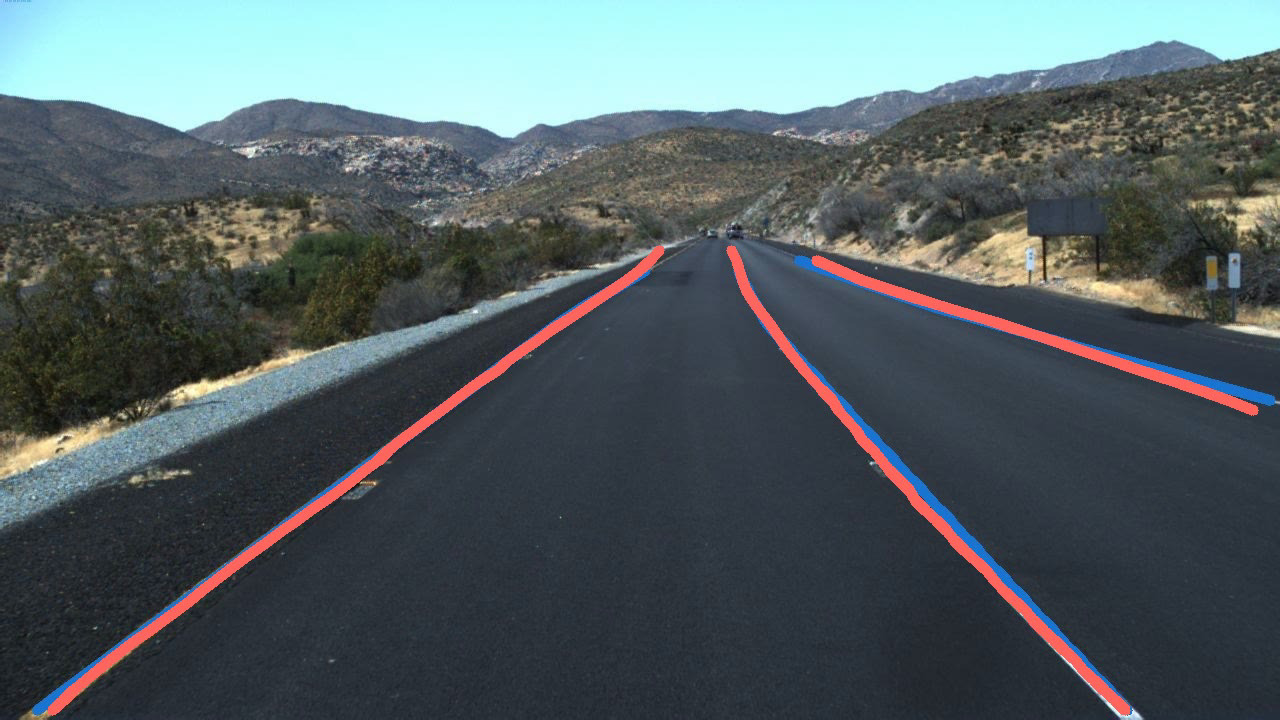}\\
		%
		\textbf{UFLD-TO} & 
		\includegraphics[width=.18\linewidth,valign=m]{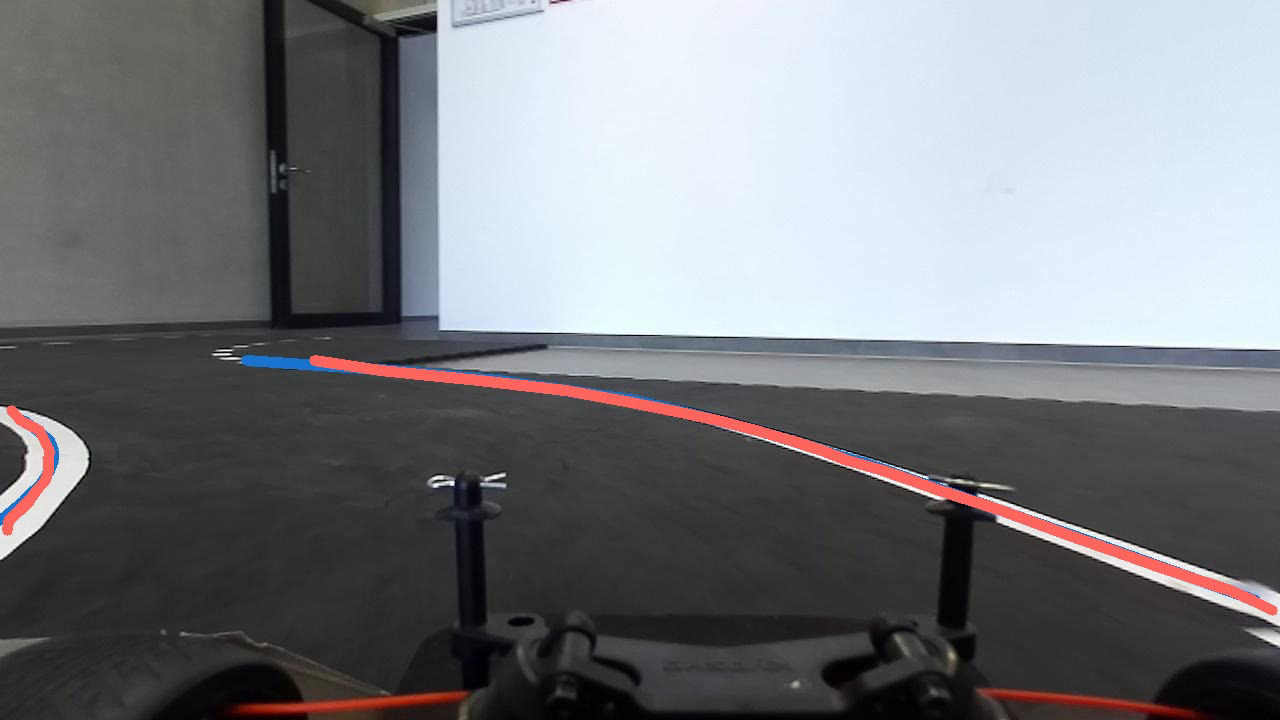} & \includegraphics[width=.18\linewidth,valign=m]{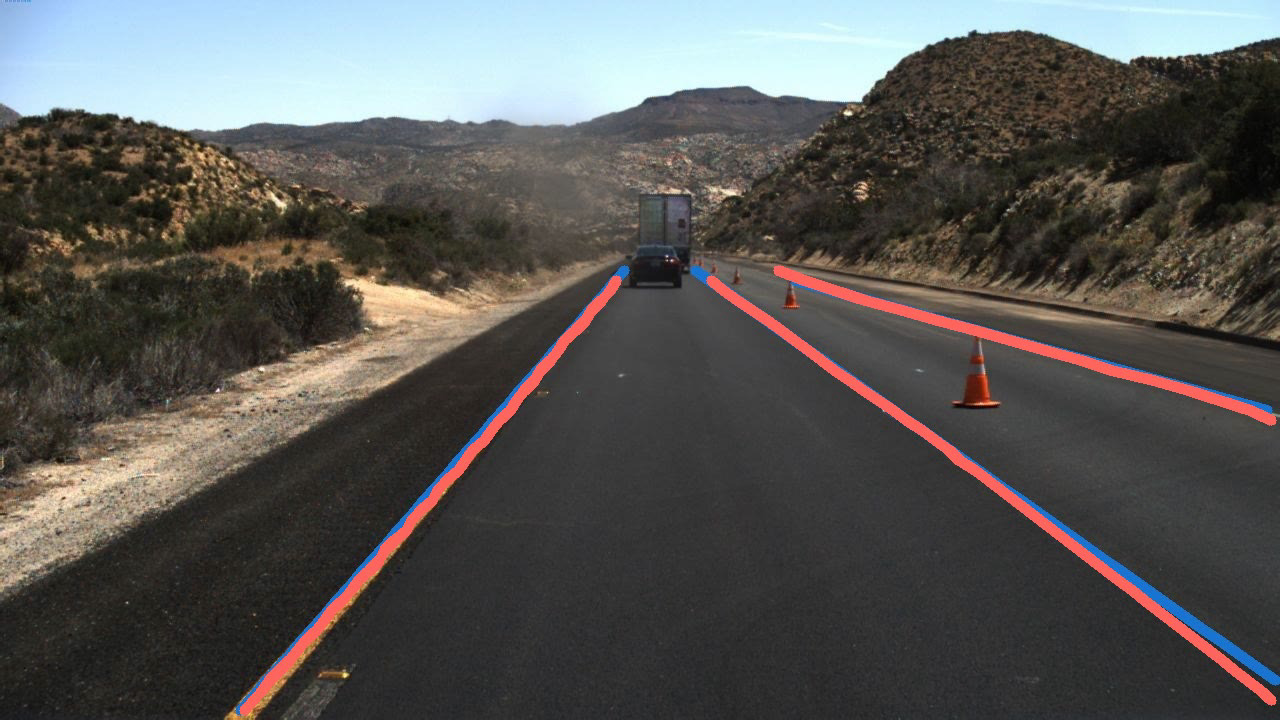} &
		\includegraphics[width=.18\linewidth,valign=m]{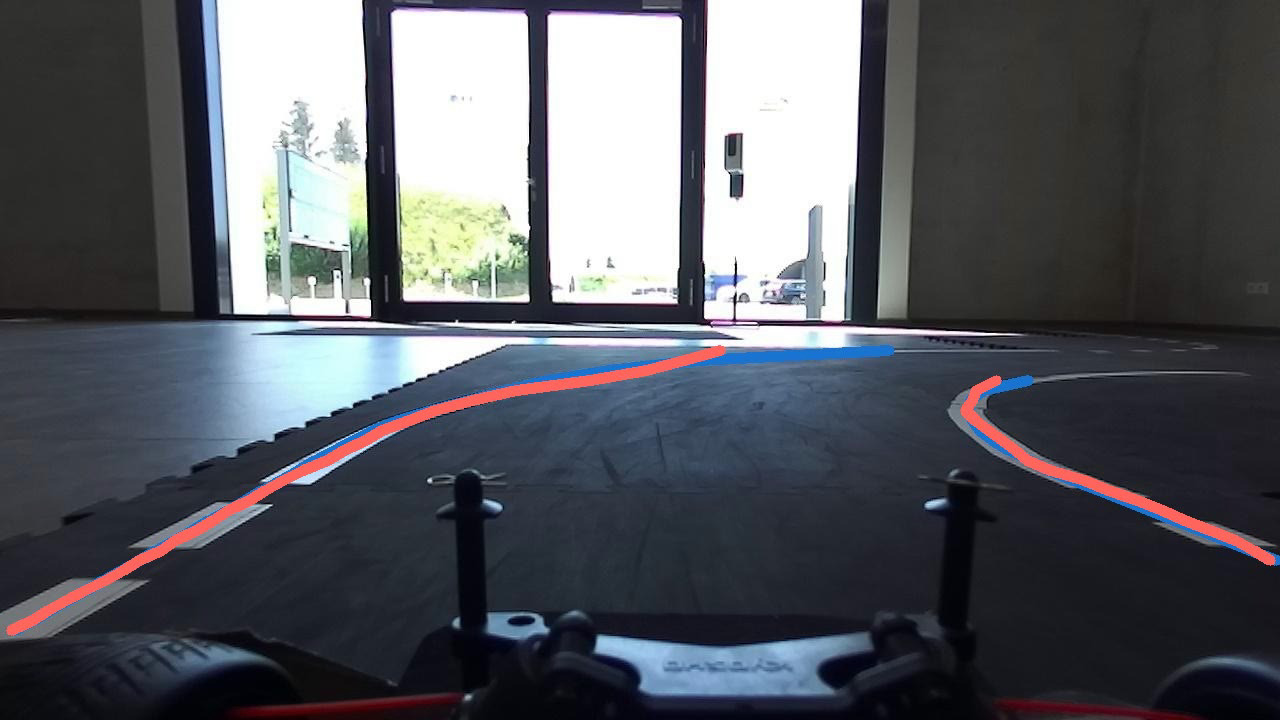} & \includegraphics[width=.18\linewidth,valign=m]{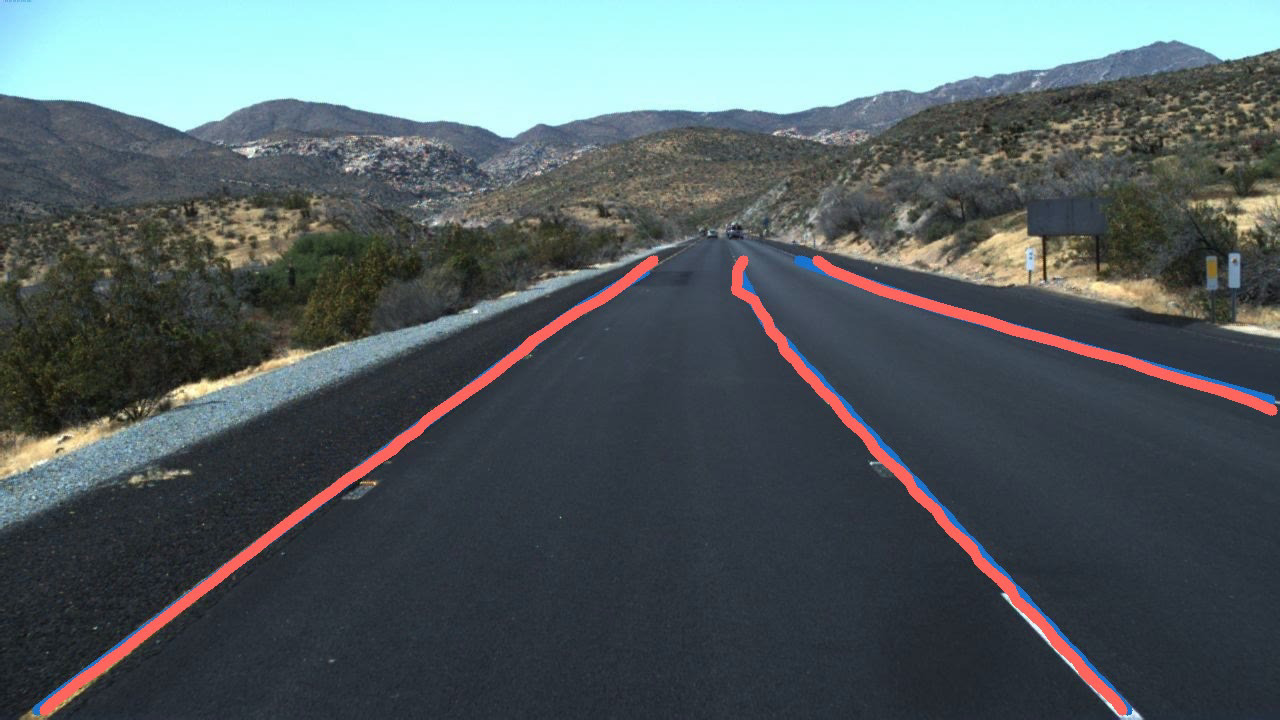}\\
	\end{tabular}
	\caption{Qualitative results of target domain predictions. Images are randomly sampled. Ground truth lane annotations are marked in blue, predictions in red.}
	\label{fig:appendix_inference_samples_3}
\end{figure}

\textbf{Qualitative results.} We randomly sample results from our baselines and show them in Figures \ref{fig:appendix_inference_samples_1}, \ref{fig:appendix_inference_samples_2}, and \ref{fig:appendix_inference_samples_3}. Compared to UFLD-SO, the UDA baselines ADDA, SGADA, and SGPCS increase performance consistently. UFLD-TO samples show the best results on the target domain. 

\subsection{Comparison to Related Work}

\begin{table*}[ht]
	\small
	\caption{Comparison of CARLANE (ours) with datasets created by related work.}
	\centering
	\begin{tabular}{cccccccc}
		\toprule
		\textbf{Dataset} & \textbf{Year}  & \textbf{\begin{tabular}[c]{@{}c@{}}Publicly\\ Available\end{tabular}} &  \textbf{Domains} & \textbf{Simulation} & \textbf{Resolution} & \textbf{\begin{tabular}[c]{@{}c@{}}Total\\ Images\end{tabular}} &  \textbf{Annotations}   \\ 
		\midrule
		\cite{Garnett2019}                    & 2019 & \xmark & sim, real  & blender & $480 \times 360$ & $391$K    & 3D  \\
		\cite{Garnett2020}      & 2020 & \xmark & sim, real  & blender & $480 \times 360$  & $586$K    & 3D \\
		\cite{SimuLanes2022}      & 2022 & \xmark & sim, real  & Carla & $1280 \times 720$  & $23$K    & 2D \\
		ours & 2022 & \cmark & sim, real, scaled  & Carla & $1280 \times 720$ & $163$K  & 2D \\
		\bottomrule
	\end{tabular}
	\label{tab:comparison_data}
\end{table*}

\begin{figure}[ht]
	\centering
	\small
	\begin{tabular}{rc@{}c@{}c}
		%
		\cite{Garnett2019} & 
		\includegraphics[width=.27\linewidth,valign=m]{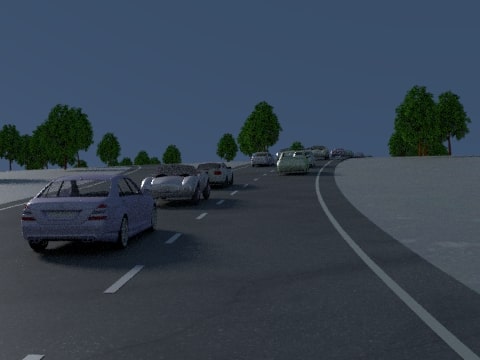} & 
		\includegraphics[width=.27\linewidth,valign=m]{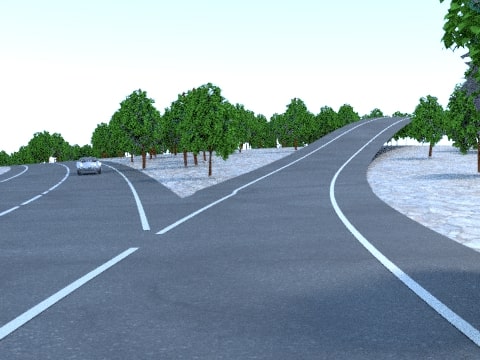} &
		\includegraphics[width=.27\linewidth,valign=m]{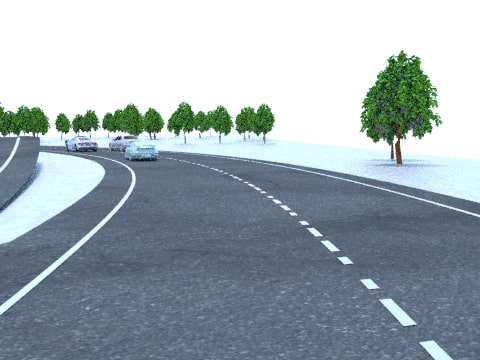}\\
		%
		ours & 
		\includegraphics[width=.27\linewidth,valign=m]{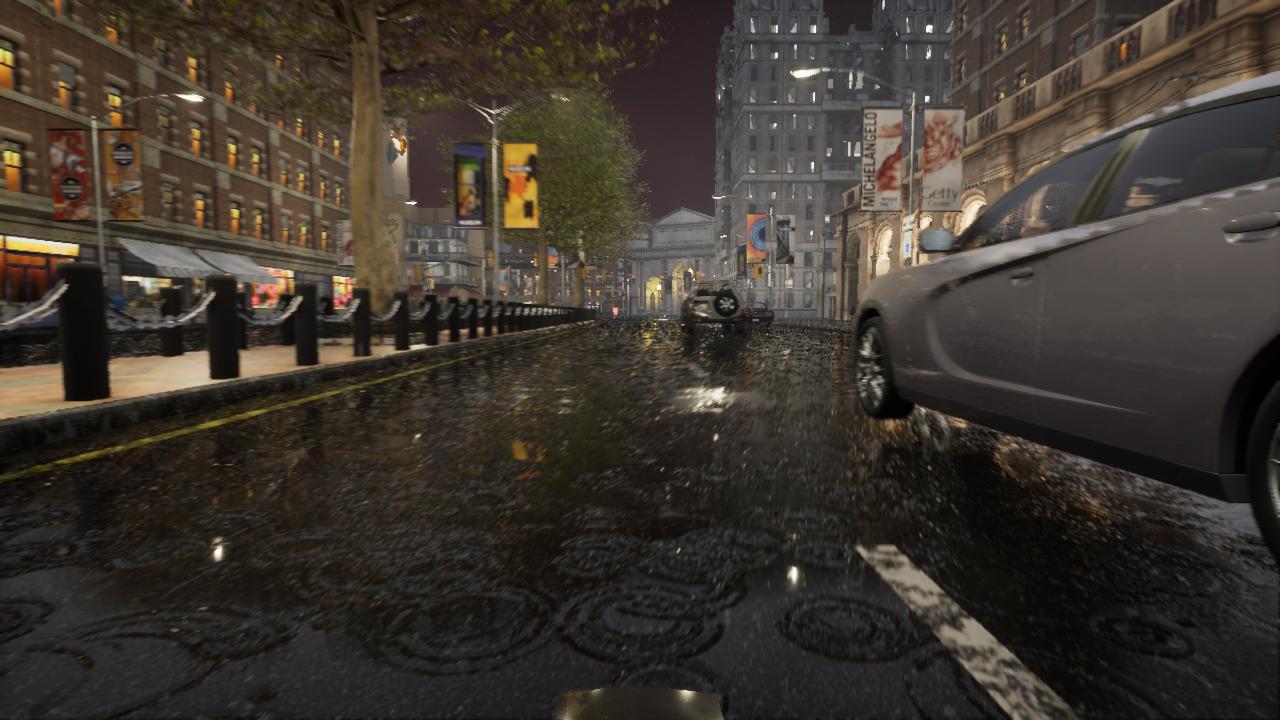} & 
		\includegraphics[width=.27\linewidth,valign=m]{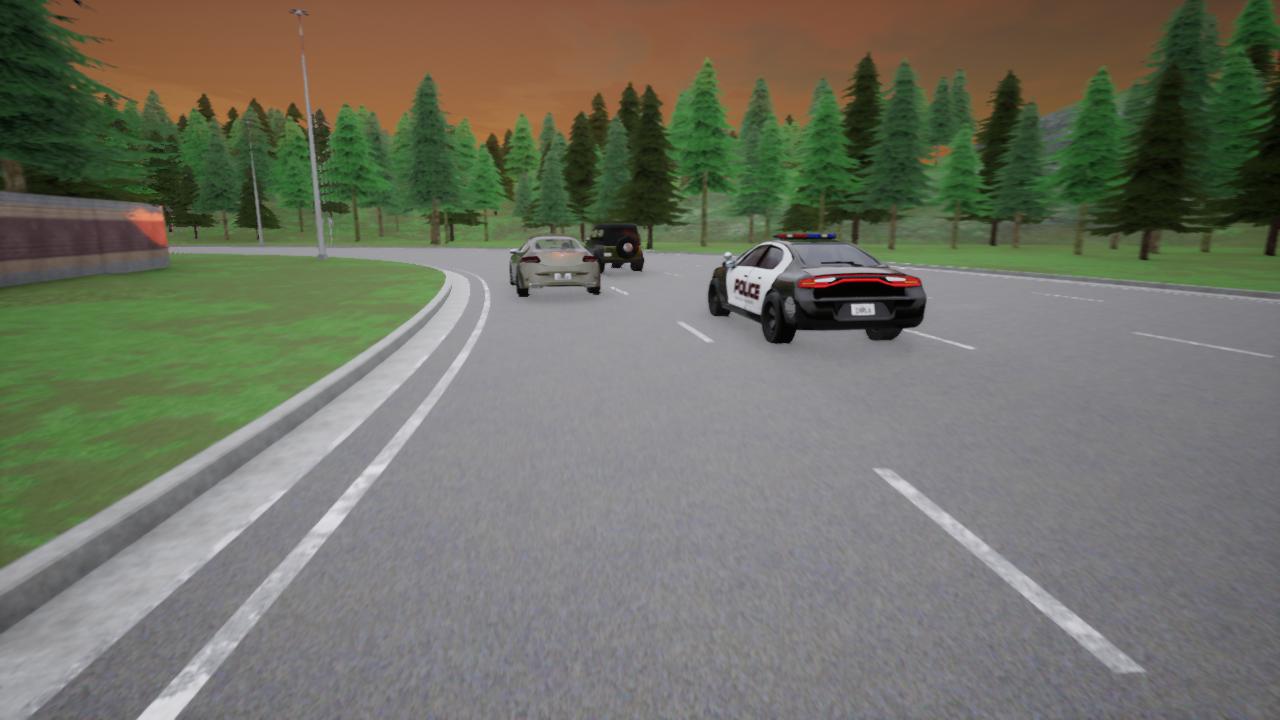} & 
		\includegraphics[width=.27\linewidth,valign=m]{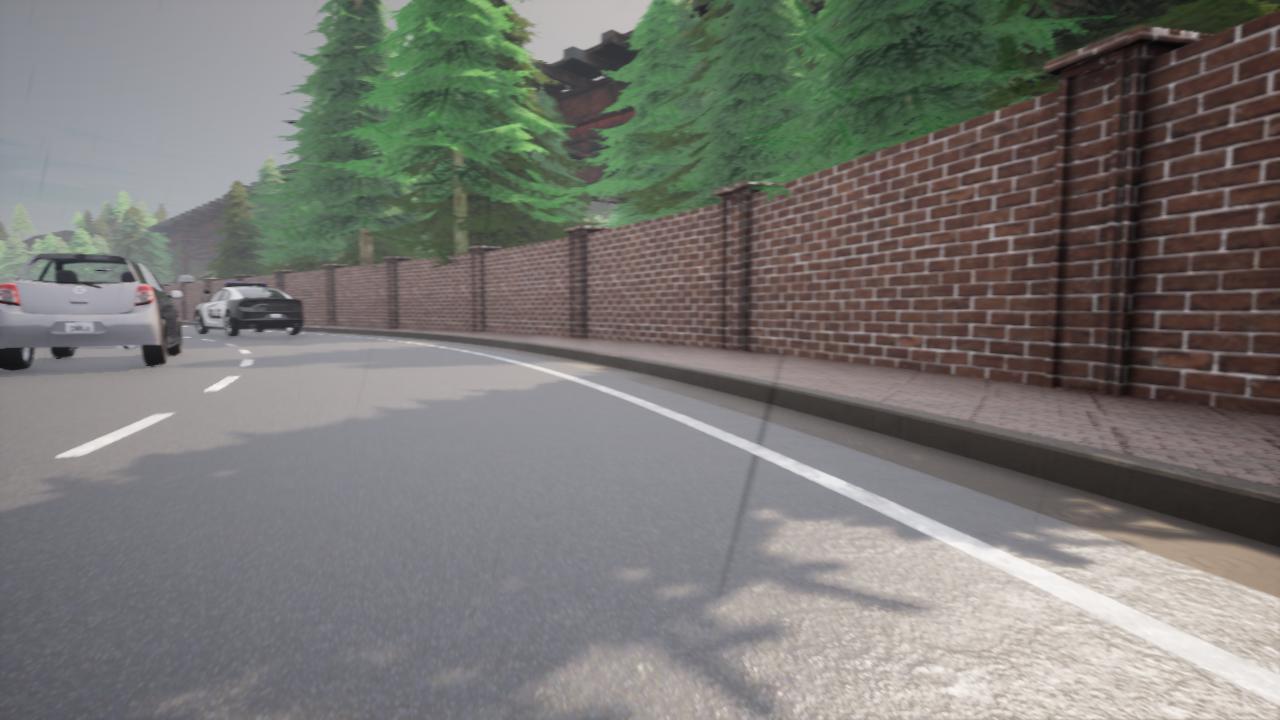} \\
	\end{tabular}
	\caption{Visual comparison of simulation images from the custom blender simulation used in \cite{Garnett2019, Garnett2020} and the Carla simulation used by \cite{SimuLanes2022} and our work. We observe that scenes captured in Carla are more detailed and realistic.}
	\label{fig:compare_sim_images}
\end{figure}

\begin{table*}[ht]
	\small
	\caption{Comparison of applied variations for the collection of the synthetic datasets.}
	\centering
	\begin{tabular}{ccccccccccccccc}
		\toprule
		\textbf{Dataset} & \head{Ego Vehicle}  & \head{Camera Position} &  \head{Lane Deviation} & \head{Traffic} & \head{Pedestrians} &  \head{World Objects}  & \head{Daytime} & \head{Weather} & \head{City} & \head{Rural} & \head{Highway} & \head{Terrain} & \head{Lane Topology} & \head{Road Appearance}\\ 
		\midrule
		\cite{Garnett2019}              & \xmark & \cmark & \cmark & \cmark & \xmark & \cmark & \cmark & \xmark & \xmark & \cmark & \xmark & \cmark & \cmark & \cmark \\
		\cite{Garnett2020}  & \xmark & \cmark & \cmark & \cmark & \xmark & \cmark & \cmark & \xmark & \xmark & \cmark & \xmark & \cmark & \cmark & \cmark \\
		\cite{SimuLanes2022}                  & \xmark & \xmark & \cmark & \cmark & \cmark & \xmark & \cmark & \cmark & \cmark & \cmark & \cmark & \cmark & \cmark & \cmark \\
		ours                            & \cmark & \cmark & \cmark & \cmark & \xmark & \cmark & \cmark & \cmark & \cmark & \cmark & \cmark & \cmark & \cmark & \cmark \\
		\bottomrule
	\end{tabular}
	\label{tab:comparison_varity}
\end{table*}
In \autoref{tab:comparison_data}, we compare CARLANE with the datasets created by related work. The main differentiators are that our dataset contains three distinct domains, including a scaled model vehicle, and is publicly available. To further compare our synthetic datasets with related work, the applied variations during the data collection process are summarized in \autoref{tab:comparison_varity}. Additionally, we highlight noticeable differences in the visual quality of the simulation engines in \autoref{fig:compare_sim_images}. Scenes captured in Carla are more realistic and detailed.

\newpage
\twocolumn

\section{Datasheet for the CARLANE Benchmark}
	\datasheetsection{Motivation}
	\begin{datasheetitem}{For what purpose was the dataset created? \normalfont Was there a specific task in mind? Was there a specific gap that needed to be filled? Please provide a description.}
		CARLANE was created to be the first publicly available single- and multi-target Unsupervised Domain Adaptation (UDA) benchmark for lane detection to facilitate future research in these directions. However, in a broader sense, the datasets of CARLANE were also created for unsupervised and semi-supervised learning and partially for supervised learning. Furthermore, a real-to-real transfer can be performed between the target domains of our datasets.
	\end{datasheetitem}
	\begin{datasheetitem}{Who created the dataset (e.g., which team, research group) and on behalf of which entity (e.g., company, institution, organization)?}
		As released on June 17, 2022, the initial version of CARLANE was created by Julian Gebele, Bonifaz Stuhr, and Johann Haselberger from the Institute for Driver Assistance Systems and Connected Mobility (IFM). The IFM is a part of the University of Applied Sciences Kempten. Furthermore, CARLANE was created by Bonifaz Stuhr as part of his Ph.D. at the Autonomous University of Barcelona (UAB) and by Johann Haselberger as part of his Ph.D. at the Technische Universität Berlin (TU Berlin).
	\end{datasheetitem}
	\begin{datasheetitem}{Who funded the creation of the dataset? \normalfont If there is an associated grant, please provide the name of the grantor and the grant name and number.}
		There is no specific grant for the creation of the CARLANE Benchmark. The datasets were created as part of the work at the IFM and the University of Applied Sciences Kempten.
	\end{datasheetitem}
	\datasheetsection{Composition}	
	\begin{datasheetitem}{What do the instances that comprise the dataset represent (e.g., documents, photos, people, countries)? \normalfont Are there multiple types of instances (e.g., movies, users, and ratings; people and interactions between them; nodes and edges)? Please provide a description.}
		The instances are drives on diverse roads in simulation, in an abstract 1/8th real world, and in full-scale real-world scenarios, along with lane annotations of the up to four nearest lanes to the vehicle.
	\end{datasheetitem}
	\begin{datasheetitem}{How many instances are there in total (of each type, if appropriate)?}
		\begin{table*}
			\caption{Dataset overview. Unlabeled images denoted by *, partially labeled images denoted by **.}
			\label{table:appendix_dataset_overview}
			\small
			\centering
			\begin{tabular}{lcccccc}
				\toprule
				Dataset                  & domain             & total images & train   & validation  & test  & lanes       \\ \midrule
				\multirow{2}{*}{MoLane}  & CARLA simulation   & 84,000       & 80,000  & 4,000       & -     & \(\leq\) 2  \\ 
				& model vehicle      & 46,843       & 43,843* & 2,000       & 1,000 & \(\leq\) 2  \\ \midrule
				\multirow{2}{*}{TuLane}  & CARLA simulation   & 26,400       & 24,000  & 2,400       & -     & \(\leq\) 4  \\ 
				& TuSimple & 6,408        & 3,268   & 358         & 2,782 & \(\leq\) 4  \\ \midrule
				\multirow{2}{*}{MuLane}  & CARLA simulation   & 52,800       & 48,000  & 4,800       & -     & \(\leq\) 4  \\ 
				& model vehicle + TuSimple & 12,536      & 6,536** & 4,000       & 2,000 & \(\leq\) 4  \\ 
				\bottomrule
			\end{tabular}
		\end{table*}
		\autoref{table:appendix_dataset_overview} shows the per-domain and per-subset breakdown of CARLANE instances. TuSimple is available at  \url{https://github.com/TuSimple/tusimple-benchmark} under the Apache License Version 2.0, January 2004. 
	\end{datasheetitem}
	\begin{datasheetitem}{Does the dataset contain all possible instances or is it a sample (not necessarily random) of instances from a larger set? \normalfont If the dataset is a sample, then what is the larger set? Is the sample representative of the larger set (e.g., geographic coverage)? If so, please describe how this representativeness was validated/verified. If it is not representative of the larger set, please describe why not (e.g., to cover a more diverse range of instances, because instances were withheld or unavailable).}
		The datasets of CARLANE contain samples of driving scenarios and lane annotations encountered in simulation and the real world. The datasets are not representative of all these driving scenarios, as the distribution of the latter is highly dynamic and diverse. Instead, the motivation was to resemble the variety and shifts of different domains in which such scenarios occur to strengthen the systematic study of UDA methods for lane detection. Therefore, CARLANE should be considered as an UDA benchmark with step-by-step testing possibility across three domains, which makes it possible to include an additional safety mechanism for real-world testing.
	\end{datasheetitem}
	\begin{datasheetitem}{What data does each instance consist of? \normalfont “Raw” data (e.g., unprocessed text or images) or features? In either case, please provide a description.}
		Each labeled instance consists of the following components: \vspace{-2pt}
		
		\textit{(1)} A single $1280\times720$ image from a driving scenario. \vspace{-2pt}
		
		\textit{(2)} A .json file entry for the corresponding subset containing lane annotations following TuSimple. The lanes’ y-values discretized by 56 raw anchors, the lanes’ x-values to 101 gridding cells, with the last gridding cell representing the absence of a lane. The file path to the corresponding image is also stored in the .json file. \vspace{-2pt}
		
		\textit{(3)} A .png file containing lane segmentations following UFLD (\url{https://github.com/cfzd/Ultra-Fast-Lane-Detection}), where each of the four lanes has a different label. \vspace{-2pt}
		
		\textit{(4)} A .txt file entry for the corresponding subset containing the linkage between the raw image and its segmentation as well as the presence and absence of a lane. \vspace{-2pt}
		
		Each unlabeled instance consists of an $1280\times720$ image from a driving scenario and a .txt file entry for the corresponding subset.
	\end{datasheetitem}
	\begin{datasheetitem}{Is there a label or target associated with each instance? \normalfont If so, please provide a description.}
		As described above, the labels per instance are discretized lane annotations and lane segmentations.
	\end{datasheetitem}
	\begin{datasheetitem}{Is any information missing from individual instances? \normalfont If so, please provide a description, explaining why this information is missing (e.g., because it was unavailable). This does not include intentionally removed information, but might include, e.g., redacted text.}
		Everything is included. No data is missing.
	\end{datasheetitem}
	\begin{datasheetitem}{Are relationships between individual instances made explicit (e.g., users’ movie ratings, social network links)? \normalfont If so, please describe how these relationships are made explicit.}
		There are no relationships made explicit between instances. However, some instances are part of the same drive and therefore have an implicit relationship.
	\end{datasheetitem}
	\begin{datasheetitem}{Are there recommended data splits(e.g., training, development/validation, testing)? \normalfont If so, please provide a description of these splits, explaining the rationale behind them.}
		Each domain is split into training and validation subsets. Details are shown in \autoref{table:appendix_dataset_overview}. The target domains for UDA additionally include test sets, which were recorded from separate tracks (model vehicle) or driving scenarios (TuSimple). Since UDA aims to adapt models to target domains, only the target domains include a test set.
	\end{datasheetitem}
	\begin{datasheetitem}{Are there any errors, sources of noise, or redundancies in the dataset? \normalfont If so, please provide a description.}
		CARLANE was recorded from different drives through simulation and real-world domains. Therefore there are images captured from the same drive, which result in similar scenarios for consecutive images. Target domain samples were annotated by hand and may include human labeling errors. However, we double-checked labels and cleaned TuSimple's test set with our labeling tool.  
	\end{datasheetitem}
	\begin{datasheetitem}{Is the dataset self-contained, or does it link to or otherwise rely on external resources (e.g., websites, tweets, other datasets)? \normalfont If it links to or relies on external resources, a) are there guarantees that they will exist, and remain constant, over time; b) are there official archival versions of the complete dataset (i.e., including the external resources as they existed at the time the dataset was created); c) are there any restrictions (e.g., licenses, fees) associated with any of the external resources that might apply to a dataset consumer? Please provide descriptions of all external resources and any restrictions associated with them, as well as links or other access points, as appropriate.}
		CARLANE is entirely self-contained.
	\end{datasheetitem}
	\begin{datasheetitem}{Does the dataset contain data that might be considered confidential (e.g., data that is protected by legal privilege or by doctor-patient confidentiality, data that includes the content of individuals’ non-public communications)? \normalfont If so, please provide a description.}
		The full-scale real-world target domain contains open-source images with unblurred license plates and people from the TuSimple dataset. This data should be treated with respect and in accordance with privacy policies. The other domains do not contain data that might be considered confidential since there where recorded in simulations or a controlled 1/8th real-world environment.
	\end{datasheetitem}
	\begin{datasheetitem}{Does the dataset contain data that, if viewed directly, might be offensive, insulting, threatening, or might otherwise cause anxiety? \normalfont If so, please describe why.}
		CARLANE includes driving scenarios; therefore, its datasets could cause anxiety in people with driving anxiety.
	\end{datasheetitem}
	\begin{datasheetitem}{Does the dataset identify any subpopulations (e.g., by age, gender)? \normalfont If so, please describe how these subpopulations are identified and provide a description of their respective distributions within the dataset.}
		No.
	\end{datasheetitem}
	\begin{datasheetitem}{Is it possible to identify individuals (i.e., one or more natural persons), either directly or indirectly (i.e., in combination with other data) from the dataset? \normalfont If so, please describe how.}
		Yes, individuals could be identified in the full-scale real-world target domain from TuSimple, since it contains unblurred license plates and people. However, the remaining domains do not contain identifiable individuals. 
	\end{datasheetitem}
	\begin{datasheetitem}{Does the dataset contain data that might be considered sensitive in anyway(e.g., data that reveals race or ethnic origins, sexual orientations, religious beliefs, political opinions or union memberships, or locations; financial or health data; biometric or genetic data; forms of government identification, such as social security numbers; criminal history)? \normalfont If so, please provide a description.}
		The full-scale real-world target domain from TuSimple could implicitly reveal sensitive information printed or put on the vehicles or people's wearings.
	\end{datasheetitem}
	\datasheetsection{Collection Process}	
	\begin{datasheetitem}{How was the data associated with each instance acquired? \normalfont Was the data directly observable (e.g., raw text, movie ratings), reported by subjects (e.g., survey responses), or indirectly inferred/derived from other data (e.g., part-of-speech tags, model-based guesses for age or language)? If the data was reported by subjects or indirectly inferred/derived from other data, was the data validated/verified? If so, please describe how.}
		The source domain images of driving scenarios and the corresponding lane annotations were directly recorded from the simulation. Lanes were manually labeled for the directly recorded real-world images. For the images collected from the model vehicle, the authors annotated the data with a labeling tool created for this task. The labeling tool is publicly available at \url{https://carlanebenchmark.github.io}. The labeling tool is utilized to clean up the annotations of the test set in the real-world domain. The authors do not have information about the labeling process of the full-scale target domain since its data is derived from the TuSimple dataset.
	\end{datasheetitem}
	\begin{datasheetitem}{What mechanisms or procedures were used to collect the data (e.g., hardware apparatuses or sensors, manual human curation, software programs, software APIs)? \normalfont How were these mechanisms or procedures validated?}
		The source domain data was collected using the CARLA simulator and its APIs with a resolution of $1280\times720$ pixels. The real-world 1/8th target domain was collected with a Stereolabs ZEDM camera with 30 FPS and a resolution of $1280\times720$ pixels. The lane distributions were additionally balanced with a bagging approach, and lanes were annotated with a labeling tool. More information can be found in the corresponding paper and the implementation. The implementation and all used tools are publicly available at \url{https://carlanebenchmark.github.io}.
	\end{datasheetitem}
	\begin{datasheetitem}{If the dataset is a sample from a larger set, what was the sampling strategy (e.g., deterministic, probabilistic with specific sampling probabilities)?}
		Source domain dataset entries are sampled based on the relative angle $\beta$ of the agent to the center lane. For MoLane, five lane classes are defined for the bagging approach: strong left curve ($\beta\leq$\ang{-45}), soft left curve (\ang{-45} $ < \beta \leq $ \ang{-15}), straight (\ang{-15} $ < \beta <$ \ang{15}), soft right curve (\ang{15} $ \leq \beta < $ \ang{45}) and strong right curve (\ang{45}$\leq \beta$). 
		
		For TuLane, three lane classes are defined for the bagging approach: left curve (\ang{-12} $ < \beta \leq$ \ang{5}), straight (\ang{-5} $ < \beta <$ \ang{5}) and right curve (\ang{5} $ \leq \beta < $ \ang{12}). 
	\end{datasheetitem}
	\begin{datasheetitem}{Who was involved in the data collection process (e.g., students, crowdworkers, contractors) and how were they compensated (e.g., how much were crowdworkers paid)?}
		Only the authors were involved in the collection process. The authors do not have information about the people involved in collecting the TuSimple dataset.
	\end{datasheetitem}
	\begin{datasheetitem}{Over what timeframe was the data collected? \normalfont Does this timeframe match the creation timeframe of the data associated with the instances (e.g., recent crawl of old news articles)? If not, please describe the timeframe in which the data associated with the instances was created.}
		MoLane's data was collected and annotated from June 2021 to August 2021. Data for TuLane's source domain was collected in February 2022.
	\end{datasheetitem}
	\begin{datasheetitem}{Were any ethical review processes conducted (e.g., by an institutional review board)? \normalfont If so, please provide a description of these review processes, including the outcomes, as well as a link or other access point to any supporting documentation.}
		No ethical reviews have been conducted to date. However, an ethical review may be conducted as part of the paper review process.
	\end{datasheetitem}
	\datasheetsection{Preprocessing/cleaning/labeling}	
	\begin{datasheetitem}{Was any preprocessing/cleaning/labeling of the data done (e.g., discretization or bucketing, tokenization, part-of-speech tagging, SIFT feature extraction, removal of instances, processing of missing values)? \normalfont If so, please provide a description. If not, you may skip the remaining questions in this section.}
		As described above, lane annotations were labeled or cleaned using a labeling tool and sampled based on the relative angle $\beta$ of the agent to the center lane.
	\end{datasheetitem}
	\begin{datasheetitem}{Was the “raw” data saved in addition to the preprocessed/cleaned/labeled data (e.g., to support unanticipated future uses)? \normalfont If so, please provide a link or other access point to the “raw” data.}
		No.
	\end{datasheetitem}
	\begin{datasheetitem}{Is the software that was used to preprocess/clean/label the data available? \normalfont If so, please provide a link or other access point.}
		Yes, the software is available at \url{https://carlanebenchmark.github.io}.
	\end{datasheetitem}
	\datasheetsection{Uses}	
	\begin{datasheetitem}{Has the dataset been used for any tasks already? \normalfont If so, please provide a description.}
		The datasets were used to create UDA baselines for the corresponding paper presenting the CARLANE Benchmark.
	\end{datasheetitem}
	\begin{datasheetitem}{Is there a repository that links to any or all papers or systems that use the dataset? \normalfont If so, please provide a link or other access point.}
		Yes, the baselines presented in the corresponding paper are available at \url{https://carlanebenchmark.github.io}.
	\end{datasheetitem}
	\begin{datasheetitem}{What(other) tasks could the dataset be used for?}
		In a broader sense, the datasets of CARLANE can also be used for unsupervised and semi-supervised learning and partially for supervised learning.
	\end{datasheetitem}
	\begin{datasheetitem}{Is there anything about the composition of the dataset or the way it was collected and preprocessed/cleaned/labeled that might impact future uses? \normalfont For example, is there anything that a dataset consumer might need to know to avoid uses that could result in unfair treatment of individuals or groups (e.g., stereotyping, quality of service issues) or other risks or harms (e.g., legal risks, financial harms)? If so, please provide a description. Is there anything a dataset consumer could do to mitigate these risks or harms?}
		Yes, TuLane and MuLane contain open-source images with unblurred license plates and people. This data should be treated with respect and in accordance with privacy policies. In general, CARLANE contributes to the research in the field of autonomous driving, in which many unresolved ethical and legal questions are still being discussed. The step-by-step testing possibility across three domains makes it possible for our benchmark to include an additional safety mechanism for real-world testing. This can help the consumer to mitigate the risks and harms to some extent.
	\end{datasheetitem}
	\begin{datasheetitem}{Are there tasks for which the dataset should not be used? \normalfont If so, please provide a description.}
		Since CARLANE focuses on UDA for lane detection and spans a limited number of driving scenarios, consumers should not solely really on this dataset to train models for fully autonomous driving. 
	\end{datasheetitem}
	\datasheetsection{Distribution}	
	\begin{datasheetitem}{Will the dataset be distributed to third parties outside of the entity (e.g., company, institution, organization) on behalf of which the dataset was created? \normalfont If so, please provide a description.}
		Yes, CARLANE is publicly available on the internet for anyone interested in using it.
	\end{datasheetitem}
	\begin{datasheetitem}{How will the dataset will be distributed (e.g., tarball on website, API, GitHub)? \normalfont Does the dataset have a digital object identifier (DOI)?}
		CARLANE is distributed through kaggle at \url{https://www.kaggle.com/datasets/carlanebenchmark/carlane-benchmark}\\ 
		
		DOI: 10.34740/kaggle/dsv/3798459
	\end{datasheetitem}
	\begin{datasheetitem}{When will the dataset be distributed?}
		The datasets have been available on kaggle since June 17, 2022.
	\end{datasheetitem}
	\begin{datasheetitem}{Will the dataset be distributed under a copyright or other intellectual property (IP) license, and/or under applicable terms o fuse (ToU)? \normalfont If so, please describe this license and/or ToU, and provide a link or other access point to, or otherwise reproduce, any relevant licensing terms or ToU, as well as any fees associated with these restrictions.}
		CARLANE is licensed under the Apache License Version 2.0, January 2004.
	\end{datasheetitem}
	\begin{datasheetitem}{Have any third parties imposed IP-based or other restrictions on the data associated with the instances? \normalfont If so, please describe these restrictions, and provide a link or other access point to, or otherwise reproduce, any relevant licensing terms, as well as any fees associated with these restrictions.}
		TuSimple, which is used for TuLanes and MuLanes target domains, is licensed under the Apache License Version 2.0, January 2004.
	\end{datasheetitem}
	\begin{datasheetitem}{Do any export controls or other regulatory restrictions apply to the dataset or to individual instances? \normalfont If so, please describe these restrictions, and provide a link or other access point to, or otherwise reproduce, any supporting documentation.}
		Unknown to authors of the datasheet.
	\end{datasheetitem}
	\datasheetsection{Maintenance}	
	\begin{datasheetitem}{Who will be supporting/hosting/maintaining the dataset?}
		CARLANE is hosted on kaggle and supported and maintained by the authors.
	\end{datasheetitem}
	\begin{datasheetitem}{How can the owner/curator/manager of the dataset be contacted (e.g., email address)?}
		The curators of the datasets can be contacted under carlane.benchmark@gmail.com.
	\end{datasheetitem}
	\begin{datasheetitem}{Is there an erratum? \normalfont If so, please provide a link or other access point.}
		No. 
	\end{datasheetitem}
	\begin{datasheetitem}{Will the dataset be updated (e.g., to correct labeling errors, add new instances, delete instances)? \normalfont If so, please describe how often, by whom, and how updates will be communicated to dataset consumers (e.g., mailing list, GitHub)?}
		New versions of CARLANE's datasets will be shared and announced on our homepage (\url{https://carlanebenchmark.github.io}) and at kaggle if corrections are necessary.
	\end{datasheetitem}
	\begin{datasheetitem}{Will older versions of the dataset continue to be supported/hosted/maintained? \normalfont If so, please describe how. If not, please describe how its obsolescence will be communicated to dataset consumers.}
		Yes, we plan to support versioning of the datasets so that all the versions are available to potential users. We maintain the history of versions via our homepage (\url{https://carlanebenchmark.github.io}) and at kaggle. Each version will have a unique DOI assigned.
	\end{datasheetitem}
	\begin{datasheetitem}{If others want to extend/augment/build on/contribute to the dataset, is there a mechanism for them to do so? \normalfont If so, please provide a description. Will these contributions be validated/verified? If so, please describe how. If not, why not? Is there a process for communicating/distributing these contributions to dataset consumers? If so, please provide a description.}
		Others can extend/augment/build on CARLANE with the support of the open-source tools provided on our homepage. Besides these tools, there will be no mechanism to validate or verify the extended datasets. However, others are free to release their extension of the CARLANE Benchmark or its datasets under the Apache License Version 2.0.
	\end{datasheetitem}
\end{document}